\documentclass[10pt,journal,compsoc,vgtc]{IEEEtran}
\usepackage{booktabs,amsmath} 
\usepackage{pdfsync}

\ifCLASSOPTIONcompsoc
  \usepackage[nocompress]{cite}
\else
  \usepackage{cite}
\fi
\ifCLASSINFOpdf
   \usepackage[pdftex]{graphicx}
\else
\fi

\usepackage{xcolor}
\usepackage{subcaption}
\usepackage{stfloats}
\usepackage[export]{adjustbox}
\usepackage{mdframed}
\usepackage{array}
\usepackage{xcolor,colortbl}
\usepackage{cancel}

\newcommand{\figref}[1]{Fig.~\ref{#1}}
\newcommand{\tabref}[1]{Tab.~\ref{#1}}

\newcommand{\eqnref}[1]{Eq.~\ref{#1}}

\def\sec#1{Sec.~\ref{#1}}
\newcommand{\tightcolorbox}[4][0pt]{%
  \begingroup
  \setlength{\fboxsep}{#1}
  \colorbox{#2}{\color{#3} #4}%
  \endgroup
}

\definecolor{corr_l0}{rgb}{1.0, 1.0, 1.0}
\definecolor{corr_l1}{rgb}{1.0, 0.851, 0.851}
\definecolor{corr_l2}{rgb}{1.0, 0.602, 0.602}
\definecolor{corr_l3}{rgb}{1.0, 0.300, 0.302}

\begin{document}

\title{Learning Photometric Feature Transform for Free-form Object Scan}

\author{Xiang~Feng, Kaizhang~Kang, Fan~Pei, Huakeng~Ding, Jinjiang~You, \protect\\Ping Tan, Kun Zhou~\IEEEmembership{Fellow,~IEEE} and Hongzhi Wu~\IEEEmembership{Member, IEEE} 
\IEEEcompsocitemizethanks{\IEEEcompsocthanksitem K. Zhou and H. Wu are the corresponding authors. All authors are with the State Key Lab of CAD~\&~CG, Zhejiang University, Hangzhou 310058, China, except that P. Tan is with Department of Electronic and Computer Engineering, The Hong Kong University of Science and Technology, Hong Kong SAR, China. K. Kang is additionally affiliated with the Visual Computing Center, King Abdullah University of Science and Technology, Thuwal 23955-6900, Saudi Arabia.
\protect\\
E-mail: \{kunzhou,hwu\}@acm.org}
}

\markboth{}%
{Feng \MakeLowercase{\textit{et al.}}: Learning Photometric Feature Transform for Free-form Object Scan}

\IEEEtitleabstractindextext{%
\begin{abstract}
We propose a novel framework to automatically learn to aggregate and transform photometric measurements from multiple unstructured views into spatially distinctive and view-invariant low-level features, which are subsequently fed to a multi-view stereo pipeline to enhance 3D reconstruction. The illumination conditions during acquisition and the feature transform are jointly trained on a large amount of synthetic data. We further build a system to reconstruct both the geometry and anisotropic reflectance of a variety of challenging objects from hand-held scans. The effectiveness of the system is demonstrated with a lightweight prototype, consisting of a camera and an array of LEDs, as well as an off-the-shelf tablet. Our results are validated against reconstructions from a professional 3D scanner and photographs, and compare favorably with state-of-the-art techniques.
\end{abstract}

\begin{IEEEkeywords}
photometric stereo, illumination multiplexing, feature learning, neural acquisition
\end{IEEEkeywords}}

\maketitle

\IEEEdisplaynontitleabstractindextext
\IEEEpeerreviewmaketitle

\IEEEraisesectionheading{\section{Introduction}\label{sec:introduction}}

\IEEEPARstart{F}{ree-form} scanning of 3D geometry in the presence of complex appearance is an important problem in computer graphics and computer vision. It is useful for various applications including e-commerce, visual effects, 3D printing, and cultural heritage.

Despite extensive research on traditional shape reconstruction over the past decades, this problem remains challenging. At one hand, multi-view stereo\cite{furukawa2015multi} usually assumes a Lambertian-dominant reflectance in computing reliable view-invariant features. Complex appearance variation with view or lighting is not welcome. It may alter the native spatial features on object surface, or specularly reflect the projected pattern from active illumination, either of which may result in correspondence matching errors. On the other hand, photometric stereo\cite{chen2018ps,woodham1980photometric} takes as input the images under varying illumination at the \emph{same} view(s), with the help from additional hardware (i.e., a tripod) for fixing the view. In free-form scanning, however, such images are impractical to acquire, as the camera/view is \emph{constantly changing}.

\begin{figure}[tbp]
    \centering
    \begin{minipage}[t]{\linewidth}
        \centering
        \begin{minipage}[t]{.495\linewidth}
            \centering
            \includegraphics[width=.995\linewidth,frame]{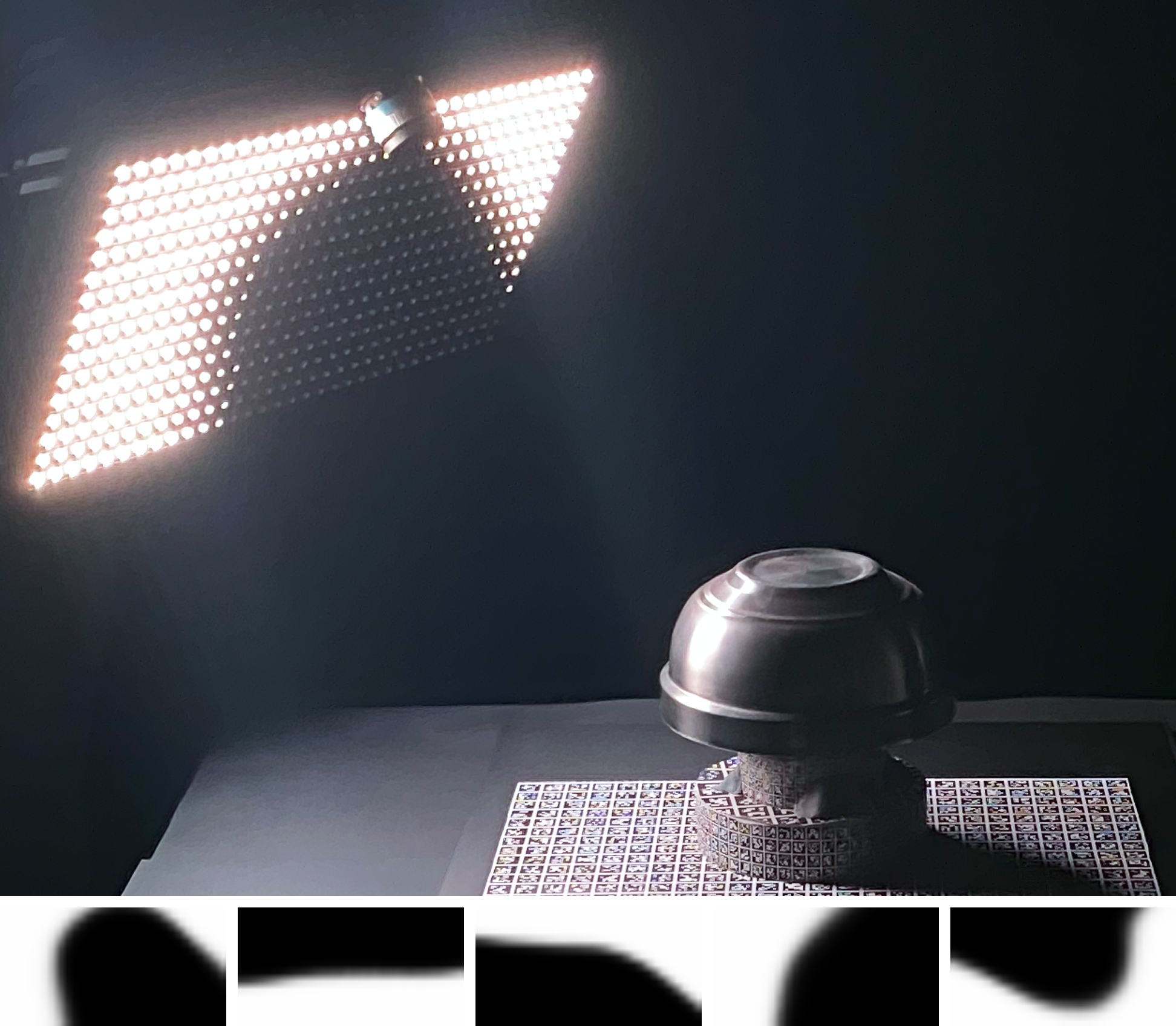}
            \put(-13,99){\scalebox{.9}{\color{white} (a)}}
            \\
            \centering
            \includegraphics[width=\linewidth]{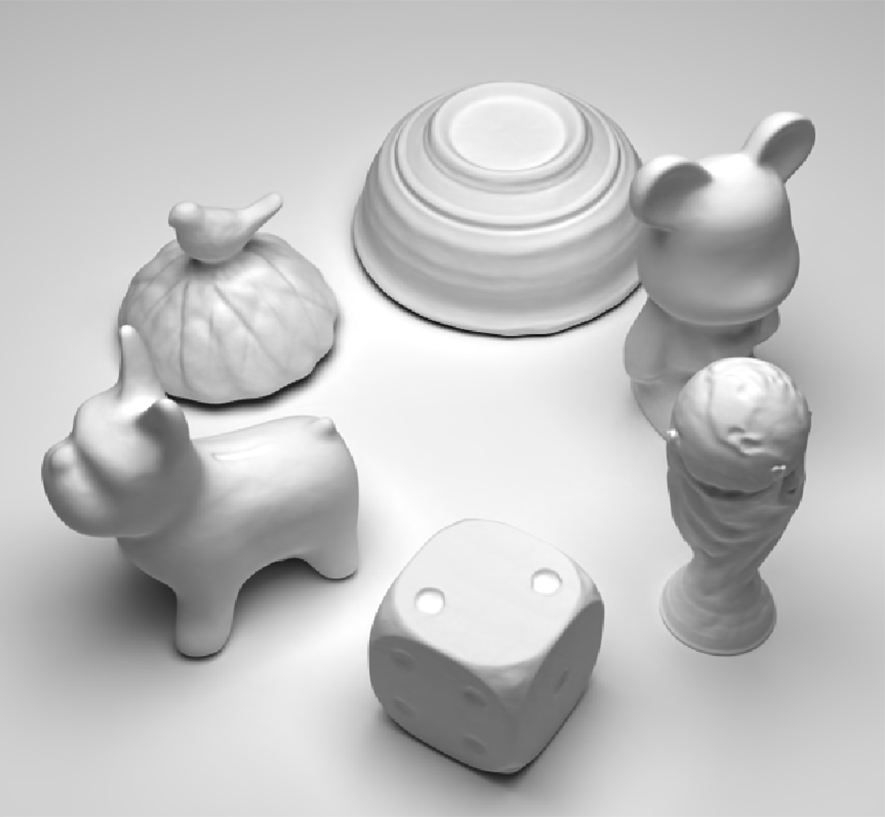}
            \put(-13,105.5){\scalebox{.9}{\color{white} (c)}}
        \end{minipage}
        \begin{minipage}[t]{.495\linewidth}
            \centering
            \includegraphics[width=.995\linewidth,frame]{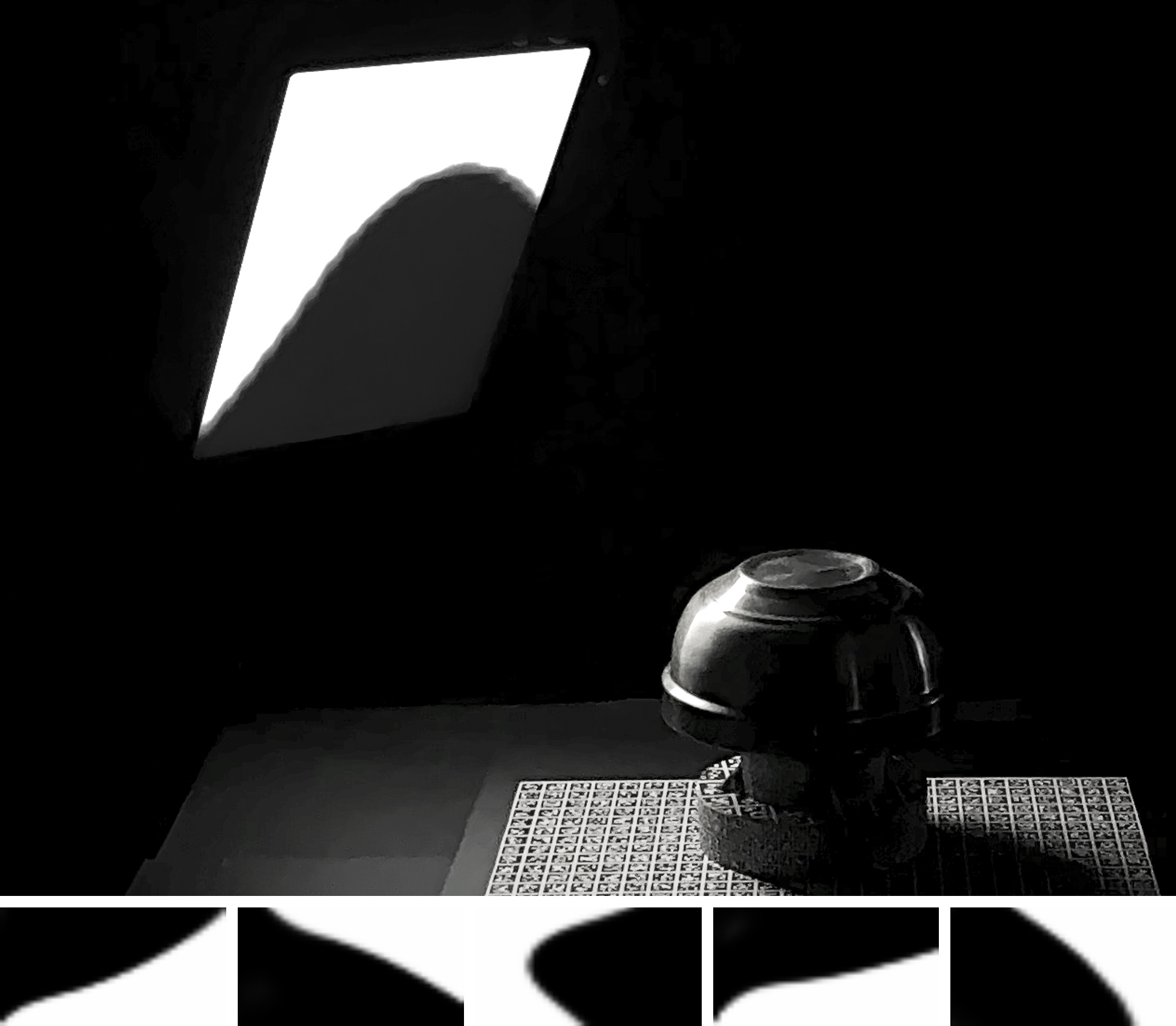}
            \put(-13,99){\scalebox{.9}{\color{white} (b)}}
            \\
            \centering
            \includegraphics[width=\linewidth]{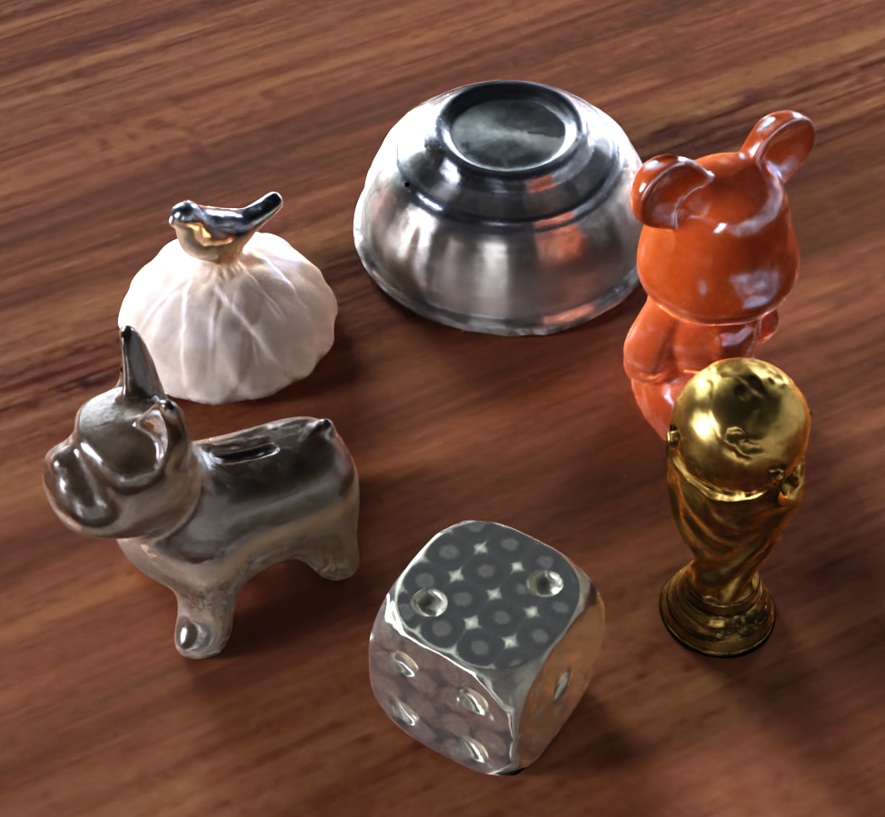}
            \put(-13,105.5){\scalebox{.9}{\color{black} (d)}}
        \end{minipage}
    \end{minipage}
    \captionsetup{type=figure}
    \captionof{figure}{Using an illumination-multiplexing device, such as a lightweight prototype consisting of a single camera and a programmable light array (a) or an off-the-shelf tablet (b), we propose a system that learns to acquire with pre-optimized time-varying lighting patterns (the bottom insets in (a) and (b)) at unstructured views, and reconstruct both the geometry (c) and complex anisotropic reflectance (d) of a number of challenging objects. Please refer to the supplementary video for animated rendering results.}
    \label{fig:teaser}
\end{figure}

Recently, image-driven differentiable optimization makes a significant success in geometric reconstruction\cite{wang2021neus,zeng2023nrhints,li2023neuralangelo}. The geometry and appearance are optimized jointly, with a loss function that encourages the rendering results to approximate corresponding input images in an end-to-end fashion. But for complex appearance such as highly specular or strongly anisotropic reflectance, the result quality is not yet satisfactory, due to the insufficient physical sampling capability (e.g., a flash only produces a single point sample in the illumination domain at a time\cite{zhang2021nerfactor}) and the insufficient fidelity of appearance representation.

\begin{figure}[tbp]
    \centering
    \begin{minipage}{\linewidth}
        \centering
        \begin{minipage}{0.08in}	
            \centering
            \hspace{0.08in}
        \end{minipage}
        \begin{minipage}{.94\linewidth}
            \begin{minipage}{.49\linewidth}
                \centering
                \subcaption*{\small Photograph}
            \end{minipage}
            \begin{minipage}{.49\linewidth}
                \centering
                \subcaption*{\small Feature Map}
            \end{minipage}
        \end{minipage}
    \end{minipage}
    \begin{minipage}{\linewidth}
        \centering
        \begin{minipage}{0.08in}	
            \centering
            \rotatebox{90}{\small \textsc{Bowl}}
        \end{minipage}
        \begin{minipage}{.96\linewidth}
            \begin{minipage}{.24\linewidth}
                \centering
                \includegraphics[width=\linewidth]{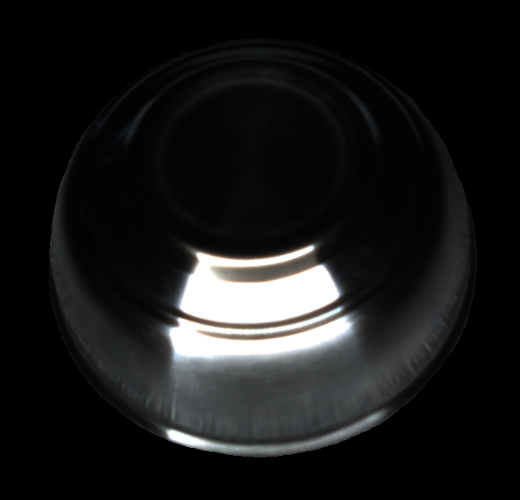}
            \end{minipage}
            \begin{minipage}{.24\linewidth}
                \centering
                \includegraphics[width=\linewidth]{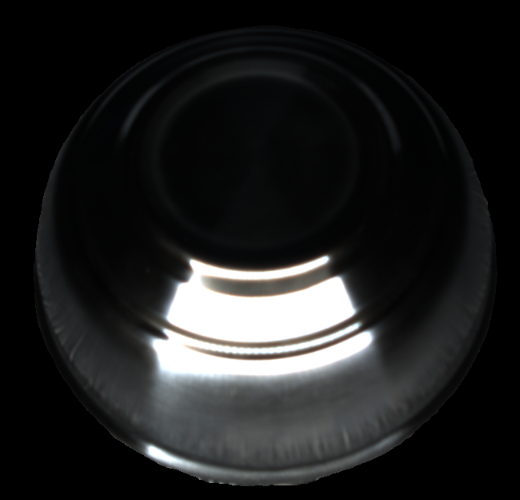}
            \end{minipage}
            \begin{minipage}{.24\linewidth}
                \centering
                \includegraphics[width=\linewidth]{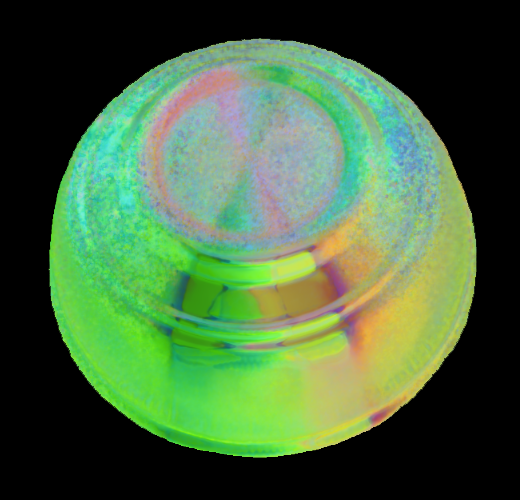}
            \end{minipage}
            \begin{minipage}{.24\linewidth}
                \centering
                \includegraphics[width=\linewidth]{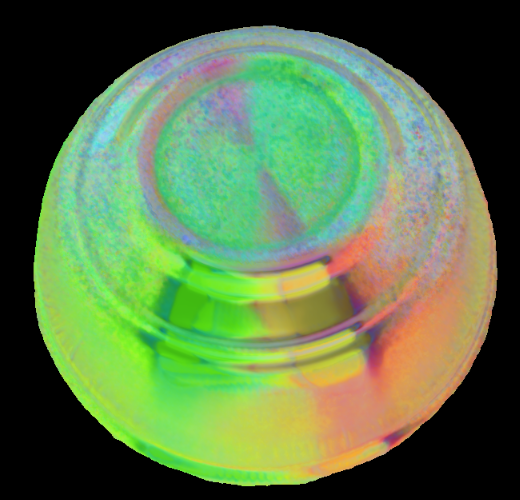}
            \end{minipage}
        \end{minipage}
    \end{minipage}
    \begin{minipage}{\linewidth}
        \centering
        \begin{minipage}{0.08in}	
            \centering
            \rotatebox{90}{\small \textsc{Cup}}
        \end{minipage}
        \begin{minipage}{.96\linewidth}
        \begin{minipage}{.24\linewidth}
            \centering
            \includegraphics[width=\linewidth]{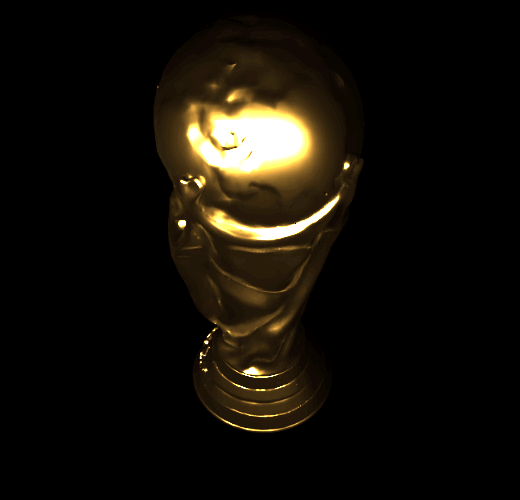}
        \end{minipage}
        \begin{minipage}{.24\linewidth}
            \centering
            \includegraphics[width=\linewidth]{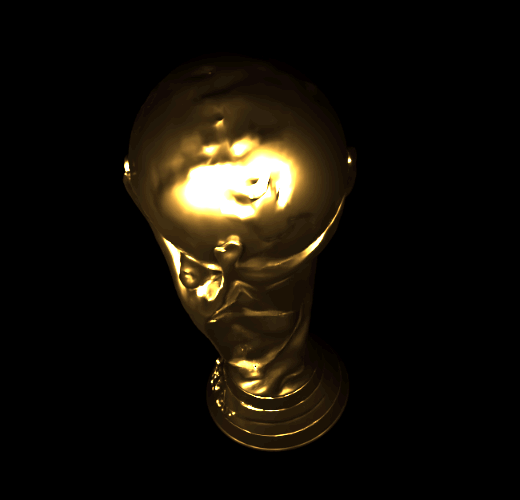}
        \end{minipage}
        \begin{minipage}{.24\linewidth}
            \centering
            \includegraphics[width=\linewidth]{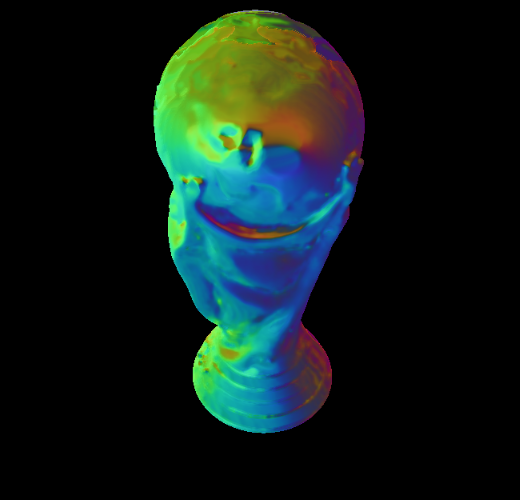}
        \end{minipage}
        \begin{minipage}{.24\linewidth}
            \centering
            \includegraphics[width=\linewidth]{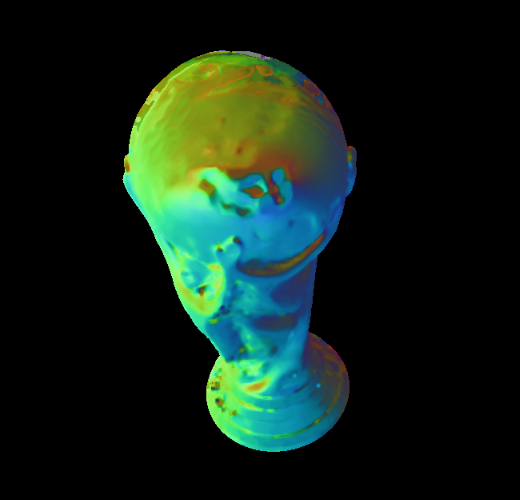}
        \end{minipage}
        \end{minipage}
    \end{minipage}
    \caption{Visualization of our feature maps at 2 views (3rd/4th column) on a scanned object (top row) and a synthetic one (bottom row). The 1st/2nd column are input photographs at the same view as the 3rd/4th column, respectively. Our high-dimensional features are projected to 3D via PCA for visualization.}   
    \label{fig:feature_visualize}
\end{figure}

To tackle the above difficulties, we present a novel framework to automatically learn to aggregate and transform photometric measurements from multiple \emph{unstructured} views into spatially distinctive and view-invariant features~(\figref{fig:feature_visualize}). This low-level transform is modular, and can enhance any multi-view stereo pipeline as a preprocessing step. To encode more angular information for high-quality reconstruction, we employ pre-optimized time-varying illumination multiplexing to physically convolve with a BRDF slice to produce a photometric measurement. To handle varying views, we carefully warp related measurements to preserve their geometric relationships for efficient neural processing. The illumination conditions during acquisition and the feature transform are jointly trained on a large amount of synthetic data. Our data-driven framework is highly flexible and can adapt to various factors, including the physical capabilities/characteristics of different setups, and different types of appearance. Since our photometric measurements reveal useful information about appearance as well, we further build a system to scan and reconstruct both the geometry and reflectance for complete object digitization.

The effectiveness of our system is demonstrated on scanning a number of challenging objects with a wide variation of shape and reflectance (\figref{fig:teaser}). Our framework is not tied to any particular setup (e.g., a lightstage~\cite{wenger2005performance}). We conduct the experiments on a lightweight illumination-multiplexing prototype~\cite{Ma:2021:TRACE}, consisting of a camera and an LED array, as well as an off-the-shelf tablet using its front camera and its screen as a programmable light source. Our shape results are validated against reconstructions from a professional 3D scanner, and our appearance results against photographs. We compare favorably with state-of-the-art techniques both in terms of geometry and reflectance.

\section{Related Work}
Due to the space limit, below we only review previous work that is closely related to our approach. 

\subsection{Photometric Stereo}
\label{sec:ps}
These techniques compute a normal field from appearance variations under different lighting and typically at a fixed view, and then integrate into a depth map~\cite{shi2016benchmark,woodham1980photometric}. Research efforts have been made from the original assumption of a Lambertian reflectance and a calibrated light, to more general appearance~\cite{alldrin2008photometric,goldman2009shape,shi2012elevation} and/or uncalibrated lighting conditions~\cite{alldrin2007resolving,basri2007photometric,lu2013uncalibrated}. Due to indirect measurements, the integrated depth map often suffers from low-frequency distortions. Multi-view photometric stereo leverages photometric cues at multiple views to obtain a complete 3D shape. These methods refine an initial coarse geometry with the estimated normals~\cite{hernandez2008multiview,li2020multi,zhou2013multi}. Vlasic et al.~\cite{vlasic2009dynamic} directly combine multi-view depth maps, each of which is computed from a normal field, to produce the final result. Logothetis et al.~\cite{logothetis2019differential} exploit the relationship between a signed distance field and normals for reconstruction. Bi et al.~\cite{bi2020deep} build a rig of 6 colocated cameras and lights, and estimate normals from sparse multi-view photometric images to refine an initial geometry. Yang et al.~\cite{yang2022ps} jointly estimate geometry, materials and lighting with differentiable rendering. The closest work to ours is EPFT\cite{Kang_2021_ICCV}. They learn to probe the angular information with a set of lighting patterns from a fixed view at a time, and transform the measurements to useful multi-view feature maps for 3D reconstruction.

All above work requires taking multiple photographs at a fixed view; none can be applied to free-form scan, in which the camera is constantly moving relative to the physical sample. In comparison, we propose a data structure and network to efficiently aggregate and transform photometric measurements at unstructured views, which are unique in free-form scanning, to low-level geometric features.

\subsection{Multi-view Stereo}
\label{sec:mvs}
Traditional methods extract low-level features from each image, compute feature correspondences across multiple views, and apply triangulation to obtain 3D information~\cite{furukawa2015multi}. Spatial aggregation is typically required, as the raw measurements at each pixel are not distinctive enough to establish reliable correspondences. While excellent results are achieved on Lambertian-dominant appearance~\cite{galliani2015massively,schoenberger2016mvs}, the reconstruction quality cannot be guaranteed in the presence of complex materials that vary with view or lighting~\cite{levoy2000digital,salvi2004pattern}. Recently, hand-crafted features are replaced with automatically learned ones~\cite{simonyan2014learning,zagoruyko2015learning,luo2016efficient}. However, learning high-quality features in the presence of complex materials remains difficult, due to the lack of corresponding training datasets. 

With the advances in machine learning, considerable progress has been made in developing end-to-end frameworks which take as input photographs and directly predict depth maps. Yao et al.~\cite{yao2018mvsnet} encode camera geometries in the network as differentiable homography, and construct 3D cost volumes to regress per-view depth map. Gu et al.~\cite{gu2020cascade} introduce cascade cost volume to predict per-view depth map in a coarse-to-fine manner. These approaches do not exploit physical appearance information for 3D reconstruction.

Our work is orthogonal to most techniques here, which focus on efficient processing in the spatial domain. In comparison, we learn how to aggregate useful information in the high-dimensional view-illumination domain for enhanced geometric reconstruction. The optimized lighting patterns essentially serve as convolution kernels to actively probe the \textit{angular} domain. Unlike the majority of related work, which tries to exclude photometric information, we exploit such information to efficiently handle highly challenging anisotropic appearance. Moreover, our learned low-level feature transform can be plugged in any existing multi-view stereo pipeline as a preprocessing module.

\subsection{Image-driven Differentiable Optimization}
\label{sec:opt}

Recent research optimizes both shape and appearance, using captured images as a form of self-supervision~\cite{munkberg2022extracting,yariv2020multiview,yariv2021volume,wang2021neus,li2023neuralangelo}. Various neural geometric representations are proposed, along with a shading network to account for appearance variations~\cite{yariv2020multiview,yariv2021volume,wang2021neus,li2023neuralangelo}. While the network considerably improves the reconstruction robustness compared with the Lambertian model, it cannot model challenging appearance like anisotropic materials with high fidelity.

A number of techniques~\cite{bi2020deep,luan2021unified,nam2018practical,iron-2022} focus on recovering detailed geometry and materials using a point light colocated with the camera. These methods struggle with strong specular highlights or anisotropic reflections, due to the extremely low sampling efficiency in the angular domain. Another line of work optimizes geometry and appearance under environment lighting~\cite{zhang2021nerfactor,srinivasan2021nerv,zhang2021physg,boss2021nerd,munkberg2022extracting,liu2023nero}. Fundamentally, the material reconstruction of passive photometric approaches is limited by the frequency distribution of the environment illumination~\cite{wu2015simultaneous}. As a result, the quality of jointly optimized geometry is affected.

Our approach is different mainly from 3 aspects. First, we jointly optimize the illumination condition during acquisition to pack more useful information in the measurements for improved reconstruction. The light array we use also has a substantially higher physical sampling capability compared with a flash. Second, we leverage the existing domain knowledge on appearance (i.e., the GGX BRDF model) in training our network, while such knowledge is entirely learned from measurements on a per-object basis in the majority of related work. Finally, we do not jointly optimize the shape and appearance. This decoupling prevents appearance reconstruction errors from propagating to geometry results~\cite{zeng2023nrhints}.

\section{Acquisition Setup}
\label{sec:setup}

We conduct acquisition experiments on two illumination-multiplexing devices, a lightweight custom-built scanner similar to\cite{Ma:2021:TRACE} (\figref{fig:teaser}-a) and an off-the-shelf tablet (\figref{fig:teaser}-b). The intrinsic/extrinsic parameters of the camera, as well as the positions, orientations, angular intensity and spectral distribution of the light sources, are carefully calibrated.  We acquire 30/10 images per second on the scanner/tablet during scanning, respectively.

\textbf{Prototype Scanner.}
Our prototype consists of a rectangular RGB LED array and a single machine vision camera. The LED array has 32$\times$16=512 RGB LEDs, with a pitch of 1cm and a maximum total power of 40W. The intensity of each LED is independently controlled, and quantized with 8 bits per channel for implementation via Power Width Modulation (PWM). A 5MP Basler acA2440-75uc camera is mounted on the top edge of the LED array. A house-made circuit board is in charge of high-precision synchronization between the camera and the LED array.

\textbf{Tablet.}
Our tablet is a 12.9-inch iPad Pro (4th gen.). We use its screen as a programmable light source, and employ its front-facing 7MP camera to take photographs. Note that the power of the iPad screen is considerably lower than our scanner, which translates to a higher requirement on hand-held stability to avoid blurred images.

\section{Preliminaries}
\label{sec:pre}

The following equation describes the relationship among the image measurement $B$ from a surface point $\mathbf{p}$, the reflectance $f$ and the intensity $I$ of each LED of the scanner/each pixel of the tablet, for a single channel.
\begin{align}
B(I;\mathbf{p}) =  &\sum_{l}  I(l)  \int  \frac{1}{|| \mathbf{x_{l}} - \mathbf{x_{p}} ||^2} \Psi(\mathbf{x_{l}}, -\mathbf{\omega_{i}})V(\mathbf{x_{l}}, \mathbf{x_{p}})   \nonumber \\
 &f(\mathbf{\omega_{i}}; \mathbf{\omega_{o}}, \mathbf{p}) (\mathbf{\omega_{i}} \cdot \mathbf{n_{p}})^{+} (-\mathbf{\omega_{i}} \cdot \mathbf{n_{l}})^{+} d\mathbf{x_{l}}.
\label{eq:render}
\end{align}
Here $l$ is the index of a locally planar light source, and $I(l)$ is its intensity in the range of [0, 1], the collection of which with each possible $l$ is a \textbf{lighting pattern}. Moreover, $\mathbf{x_{p}}$/$\mathbf{n_{p}}$ is the position/normal of $\mathbf{p}$, while $\mathbf{x_{l}}$/$\mathbf{n_{l}}$ is the position/normal of a point on the light with an index of $l$. We denote $\mathbf{\omega_{i}}$/$\mathbf{\omega_{o}}$ as the lighting/view direction. $\Psi(\mathbf{x_{l}}, \cdot)$ represents the angular distribution of the light intensity. $V$ is a binary visibility function between $\mathbf{x_{l}}$ and $\mathbf{x_{p}}$. The operator $( \cdot )^{+}$ computes the dot product between two vectors, and clamps a negative result to zero. Finally, $f$ is a 2D slice of anisotropic GGX BRDF~\cite{Walter:2007:GGX}. 

As $B$ is linear with respect to $I$~(\eqnref{eq:render}), it can be expressed as the dot product between $I$ and a \textbf{lumitexel} $c$:
\begin{align}
B(I;\mathbf{p})&=\sum\limits_l I(l)c(l;\mathbf{p}), \notag\\
c(l;\mathbf{p})&=B(\{I(l)=1,\forall_{j\neq l}I(j)=0\};\mathbf{p}),
\label{eq:lumi}
\end{align}
where $c$ is a function of the light index $l$, defined on the surface point $p$ of the sampled object~\cite{lensch2003image}. 

\begin{figure}[tbp]
    \centering
    \includegraphics[width=\linewidth]{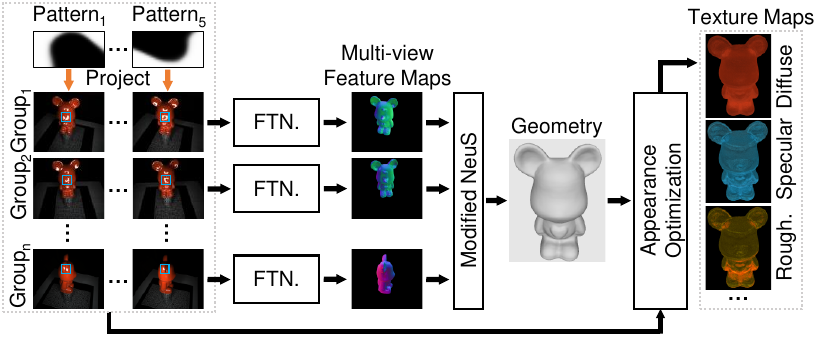}
    \caption{Our runtime pipeline. First, we partition continuously captured images into groups of 5, each acquired under a different lighting pattern. Next, we crop patches from each image in the group, centered at a same pixel location. A network (Feature Transform Network) then transforms these data into a per-pixel high-dimensional feature at that location, the collection of which forms a feature map for the center view. We feed the feature maps from every group into a multi-view stereo pipeline for 3D reconstruction. With the computed shape, the appearance of the object is differentiably optimized with respect to all input images, and then stored as texture maps of GGX BRDF parameters. FTN. $=$ Feature Transform Network.
    }
    \label{fig:runtime_overview}
\end{figure}

\section{Overview}
\label{sec:over}

To scan an object, we move an illumination-multiplexing device around it to take photographs continuously. The light source on the device is programmed to loop over $\#p$ pre-optimized lighting patterns. The camera exposure is synchronized with the pattern projection. We define every $\#p$ consecutively captured images (i.e., views) as a \textbf{group}, with the first image acquired with the first lighting pattern, etc. 

To reconstruct the geometry~(\sec{sec:net}), for each group, we aggregate and transform the multi-view photometric information from all $\#p$ images into a high-dimensional feature map, defined at the \textbf{center view} (i.e., the view of the $\lceil\#p/2\rceil$-th image). The feature maps of all groups are directly sent as input to a multi-view stereo pipeline (e.g., NeuS~\cite{wang2021neus}) for 3D reconstruction. To recover the reflectance~(\sec{sec:app_fitting}), we optimize the GGX BRDF parameters with respect to input photographs, given the previously reconstructed mesh. The results are stored as texture maps, which can be rendered with any standard pipeline under novel view and illumination conditions. Please see~\figref{fig:runtime_overview} for an illustration.

Note that each group is actually a small set of frames similar in view, but different in lighting. This structure generalizes from standard photometric stereo, whose input images are taken with varying illumination at the \emph{same} view. In comparison, the images in one of our groups have slightly \emph{different} views, as it is impossible to maintain a fixed view with a handheld device during acquisition (i.e., one key challenge we address in this paper).

\begin{figure*}
    \centering
    \includegraphics[width=\textwidth]{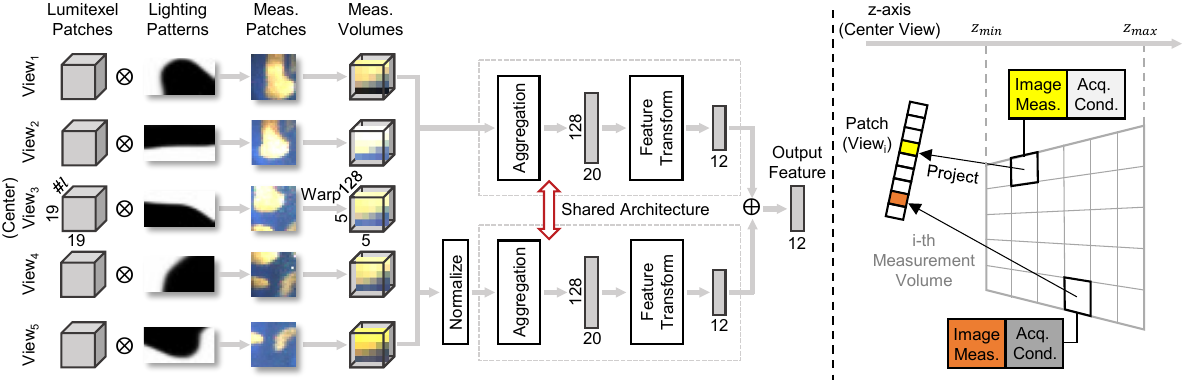}
    \put(-165,5){\scalebox{.8}{\color{black} (a)}}
    \put(-18,5){\scalebox{.8}{\color{black} (b)}}
    \caption{Our network (a) and warping illustration (b). The network (a) takes as input the lumitexels corresponding to all pixels in 5 patches from a group, and encodes them as measurements by simulating lighting pattern projections. The measurements of each view are warped to their respective measurement volume (b), defined at the same center view. A total of 5 volumes are aggregated and transformed to a feature vector, by combining the outputs from an unnormalized and a normalized network branch which share the same architecture. A 2D warping example of a patch at the i-th view is shown in (b). First, a measurement volume is set up with respect to the center view. We then fill each voxel by projecting its center to the patch, and fetching the corresponding image measurement. The acquisition condition, including view and lighting information, is also stored. Meas. = measurement, Acq. = Acquisition, Cond. = condition.}
    \label{fig:pipeline}
\end{figure*}

\section{Feature Transform Network}
\label{sec:net}

It models physical acquisition and computational reconstruction together, to enable the automatic, joint optimization of both processes~(\figref{fig:pipeline}). For training, the network input are $\#p$ patches of lumitexels, representing the spatially-and-angularly varying appearance at $\#p$ views in a group. These patches of lumitexels are encoded by corresponding lighting patterns to simulate the measurement process~(\sec{sec:encoder}). Next, to handle unstructured views and unknown depths, each patch of measurements are warped to a measurement volume defined at the center view~(\sec{sec:warp}). Finally, all warped measurements are aggregated~(\sec{sec:aggregate}), and transformed to a high-dimensional feature vector, corresponding to the center pixel of the patch at the center view~(\sec{sec:ft}). At runtime, for each group, we slide a window over the image domain, crop from all captured images in the group, send the resulting $\#p$ patches of measurements to our network, and assemble the final \textit{per-pixel} feature vectors as a feature map. \figref{fig:architecture} visualizes our network architecture. We use $\#p$=5 in all experiments.

\subsection{Encoding}
\label{sec:encoder}
The first part of our network is an encoder that maps the measurement process. It consists of a linear fully-connected (fc) layer, whose weights correspond to the lighting patterns in acquisition. The output are 5 patches of 19$\times$19 pixels, corresponding to the 5 views in a group. Essentially, each output pixel represents an image measurement under a lighting pattern from a particular view, modeled as the dot product between an input lumitexel and a pattern~(\eqnref{eq:lumi}).

Note that increasing the patch size would increase the computation cost, while decreasing it would reduce the chance of capturing essential information for transforming to a high-quality feature. The current patch size is determined via experiments.

\subsection{Warping}
\label{sec:warp}

To represent the potential geometric relationships between measurements, the second part of our network warps each patch to a \textbf{measurement volume}~(\figref{fig:pipeline}-b). As a result, 5 patches in a group are warpped to 5 different measurement volumes. This step is critical for efficient handling of free-form scanned data with unknown depths.

First, we define a measurement volume as a view frustum at the \textit{same} center view with a resolution of $5\times5\times128$. Specifically, we cast rays from the camera center towards each pixel in the center $5\times5$ window of the patch at the center view. Each ray is discretized into 128 
depth hypotheses, uniformly sampled in the range of $[z_{\rm min}, z_{\rm max}]$.
The depth range $z_{\rm min}$/$z_{\rm max}$ is calculated from a coarse bounding box (or can be manually specified). Ideally, a volume of $1\times1\times128$ is sufficient to model \textit{depth uncertainty} for the center pixel of a patch. However, due to inevitable registration errors in practice, we enlarge the lateral neighborhood to 5$\times$5 to tolerate such inaccuracies.

Next, we convert each patch into its own measurement volume (\figref{fig:pipeline}-b). Specifically, we loop over all voxels in the volume, project each voxel center to the patch, and store the corresponding measurement (black if the projection falls outside the patch) along with the acquisition condition to the current voxel. Our acquisition condition includes (1) one hot-encoding of the lighting pattern index, (2) the center pixel location of the current patch in the image, (3) the depth of the current voxel, (4) the camera transform relative to the center view, and (5) the camera pose of the center view. The above information is stored for each voxel, as we want our network to be aware of the factors that are related to final geometric features. Note that the idea of encoding acquisition conditions for neural processing is proposed in~\cite{Ma:2021:TRACE} for free-form appearance scanning.

In the end, a voxel in the i-th measurement volume contains the image measurements from the i-th view, whose corresponding 3D positions \emph{might} fall within the voxel.

\begin{figure*}[t]
    \centering
    \includegraphics[width=\textwidth]{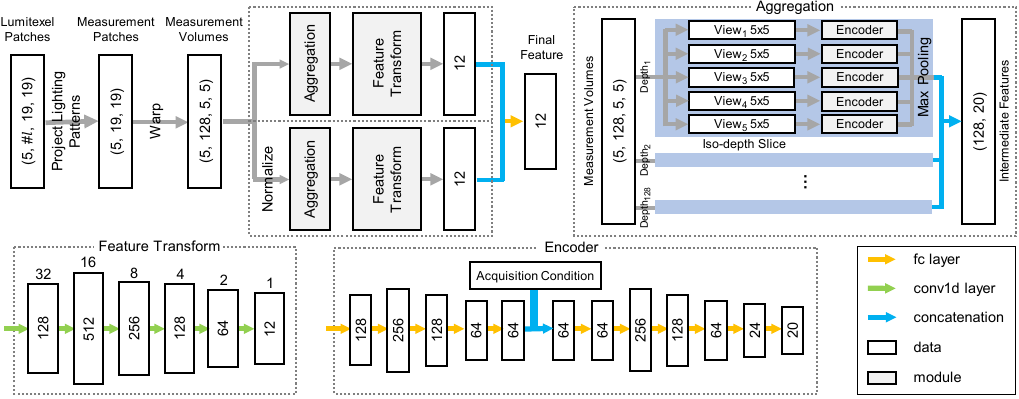}
    \caption{Network architecture. Our network takes as input the lumitexels corresponding to all pixels in 5 patches from a group, and encodes them as measurements by simulating lighting pattern projections. The measurements of each view are warped to a measurement volume. The total of 5 volumes are aggregated and transformed to a feature vector, by combining the outputs from an unnormalized and a normalized branch that shares the same structure. The dimension of data is specified in the corresponding block. In the feature transform module, the dimension of depth is additionally specified on the top of each block. In the aggregation module, we loop over each of 5 iso-depth slices within 5 measurement volumes and transform them into a 20D intermediate feature. In the end, all intermediate features at 128 different depth hypotheses are aggregated to a final 12D feature.}
    \label{fig:architecture}
\end{figure*}

\subsection{Aggregation} 
\label{sec:aggregate}
The third part of the network aggregates the information from all 5 measurement volumes, each of which is computed from a patch.
First, for each iso-depth slice in one volume, we flatten and transform it to a lower dimensional latent vector using an encoder. As a result, each volume is converted to 128 latent vectors. This step can be viewed as an aggregation along the lateral dimensions. Next, for each depth, we aggregate across 5 views by performing max pooling on all related latent vectors, and store the result as an intermediate feature. Note that this step aggregates illumination as well, since each view is associated with a different lighting pattern. The output of this part is 128 intermediate features. Now the only dimension that has not been aggregated is the depth, which is left for the next step.

\subsection{Feature Transform}
\label{sec:ft}
This fourth part produces the final per-pixel geometric feature, the collection of which will be fed to multi-view stereo for 3D reconstruction. We send all 128 intermediate features to a convolutional neural network to extract a 12D feature as output. This allows the network to automatically learn how to "softly" select the most matching depth (among all 5 views) as well as its corresponding feature.

The above architecture~(\sec{sec:aggregate}-\ref{sec:ft}) is repeated to build two branches, as illustrated in~\figref{fig:pipeline}-a~\&~\ref{fig:architecture}. One unnormalized branch works exactly as described above, while the other branch will normalize the volume across multiple views on a per-voxel basis. The idea is to prevent the network from learning stable diffuse albedos only. After normalization, the absolute values of albedos no longer matter, therefore forcing the network to exploit other useful sources of information. Finally, the two 12D output vectors from both branches are combined into a final 12D feature via a linear fc layer, similar to~\cite{Kang_2021_ICCV}.

\subsection{Loss Function}
\label{sec:loss_function}

The function is defined as follows:
\begin{equation}
    L=\lambda_{0}L_0+\lambda_{1}L_1+\lambda_{2}L_2+\lambda_p L_p,
    \label{eq:overall_loss}
\end{equation}
where $L_0,L_1,L_2$ are the similarity loss~\cite{tian2017l2} on the final feature, and the feature from the unnormalized/normalized branch, respectively. The similarity loss is defined as:
\begin{gather}
L_{\rm sim}=\frac{1}{2}(\sum\limits_i{\rm log} s_{ii}^c+\sum\limits_i{\rm log} s_{ii}^r), \notag\\
s_{ij}^c=\frac{{\rm exp}(-d_{ij})}{\sum_m{\rm exp}(-d_{mj})}, s_{ij}^r=\frac{{\rm exp}(-d_{ij})}{\sum_n{\rm exp}(-d_{in})}. \notag
\end{gather}
Here $d_{ij}$ is the Euclidean distance matrix of features in a batch of training samples. In each batch, we consider the features of the \textit{same} 3D point at different views, as well as the features corresponding to \textit{different} points. The e1 loss essentially decreases the distances of the former (view invariance), while increases the distances of the latter (spatial distinctiveness). Finally, $L_p$ is a loss term on the brightness of lighting patterns, as brighter patterns are desired for high SNR acquisition during scanning, defined as: $L_p=\sqrt{\#l / \sum_l I(l)},$
where $\#l$ is the number of independently controlled light sources. We use $\lambda_0=0.33,\lambda_1=0.33,\lambda_2=0.33$ and $\lambda_p=0.05$ in our experiments.

\begin{figure*}
    \centering
    \begin{minipage}{\textwidth}
        \includegraphics[width=\linewidth,height=.12\linewidth]{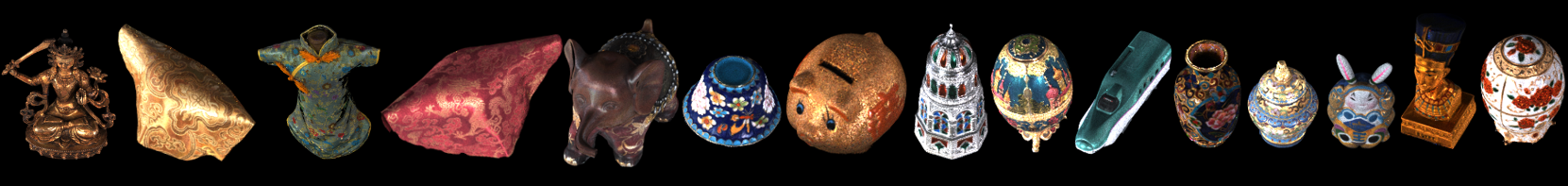}
    \end{minipage}
    \captionsetup{type=figure}
    \caption{Our training dataset of 15 high-quality objects, digitized by a commercial 3D scanner and a professional light stage\cite{Kang:2019:JOINT}.}
    \label{fig:dataset}
\end{figure*}%

\subsection{Training}

We implement our network with PyTorch, using the Adam optimizer with batch size of 32 and a momentum of 0.9. Xavier initialization is applied to all weights in the network. We train 100K iterations with a learning rate of $10^{-4}$, which takes 8 hours to finish.

The training data are synthetically generated using 15 pre-captured objects (\figref{fig:dataset}) and captured scanning motions. Each object consists of a 3D mesh along with texture maps of GGX BRDF parameters. 25 scanning processes are recorded, each of which consists of 2,000 consecutive camera poses. We compute the relative motion of every 5 views with respect to the center view, which will be processed to synthesize the camera motion for training a single group. Note that our approach is \emph{not tied} to the current data since it is entirely data-driven.

Specifically, the synthetic data for a group are generated as follows. We first synthesize the center view. To do so, we randomly sample a 3D point on a random object~(\figref{fig:dataset}) and a visible view direction based on its geometry normal. For the viewing direction, a camera position is randomly generated along the direction within a predefined range of valid distance (16-65cm in our experiments). Next, a random foreground pixel is selected as the projected position of the 3D point. With the projected pixel and the position of the 3D point, a camera pose is initialized, and then we randomly rotate it around the view direction. To produce other views in the group, we random select part of pre-captured continuous scanning motions, and use the relative tranforms between frames to synthesize all other 4 views. Finally, we compute the lumitexels for all pixels in the patch of each view, whose center is the projected pixel, by ray-tracing to the object surface and simulating the light reflections with the associated BRDF parameters, according to~\eqref{eq:lumi}. The patches of lumitexels will be used to synthesize images measurements~(\sec{sec:encoder}). Please also refer to~\cite{Kang_2021_ICCV} for descriptions on a similar process.

\section{Appearance Optimization}
\label{sec:app_fitting}
After geometric reconstruction, we establish a uv-parame-terization over object surfaces, and compute BRDF parameters at each valid texel via differentiable optimization. For a specific texel, we first project the corresponding 3D position to all visible views to gather its image measurements under learned lighting patterns. We reparameterize the GGX BRDF model plus the local frame with a 16D latent code and jointly train a fully-connected network that transforms the latent code to GGX BRDF parameters and the local frame as in~\cite{Xu:2023:StructuredLight}, by minimizing the difference between rendering results~(\eqnref{eq:render}) and the gathered measurements. Finally, we convert the latent code at each texel to anisotropic GGX BRDF parameters and store them in texture maps as the appearance result. Please refer to the supplementary material for more details on appearance modeling.

\section{Implementation Details}
We remove over-blurry images from our sequence to avoid the negative impact over the final results. We calculate the level of blurriness for each image\cite{crete2007blur}, and discard an entire group if the blurriness of any image in the group reaches a threshold. For each remaining image, we perform structure-from-motion with COLMAP~\cite{schoenberger2016sfm} to compute camera poses from the ARTags~\cite{fiala2005artag} placed along with the object (\figref{fig:teaser}-a~\&~b). We apply SAM~\cite{kirillov2023segany} to segment the object from the background for each center-view image. After geometry reconstruction, a uv-parameterization with a texture resolution of 1024$\times$1024 is generated for appearance optimization (\sec{sec:app_fitting}). In terms of lighting patterns for the tablet, we use a resolution of $27\times36$, which is much lower than the native one, to save the otherwise prohibitively expensive computational costs. We modify the original NeuS\cite{wang2021neus} by changing the output dimension of the last fully-connected layer to 12, the same dimension as our feature vector. Similar modification is applied to Neuralangelo\cite{li2023neuralangelo}. Please refer to the supplementary material for more details.

\begin{figure*}
    \begin{minipage}{\textwidth}
        \centering
        \begin{minipage}{0.065in}
            \hspace{0.065in}
        \end{minipage}	
        \begin{minipage}{.985\textwidth}
            \centering
            \begin{minipage}{\textwidth}
                \begin{minipage}{.0832\textwidth}
                    \centering
                    \subcaption*{\scalebox{.75}{\small Photograph}}
                \end{minipage}%
                \begin{minipage}{.0832\textwidth}
                    \centering
                    \subcaption*{\scalebox{.75}{\small Ground-truth}}
                \end{minipage}%
                \begin{minipage}{.0832\textwidth}
                    \centering
                    \subcaption*{\scalebox{.75}{\small Ours+\cite{wang2021neus}}}
                \end{minipage}%
                \begin{minipage}{.0832\textwidth}
                    \centering
                    \subcaption*{\scalebox{.75}{\small Ours+\cite{li2023neuralangelo}}}
                \end{minipage}%
                \begin{minipage}{.0832\textwidth}
                    \centering
                    \subcaption*{\scalebox{.75}{\small Ref-NeuS\cite{ge2023ref}}}
                \end{minipage}%
                \begin{minipage}{.0832\textwidth}
                    \centering
                    \subcaption*{\scalebox{.75}{\small NeRO\cite{liu2023nero}}}
                \end{minipage}%
                \begin{minipage}{.0832\textwidth}
                    \centering
                    \subcaption*{\scalebox{.6}{\small Neuralangelo\cite{li2023neuralangelo}}}
                \end{minipage}%
                \begin{minipage}{.0832\textwidth}
                    \centering
                    \subcaption*{\scalebox{.75}{\small NeuS\cite{wang2021neus}}}
                \end{minipage}%
                \begin{minipage}{.0832\textwidth}
                    \centering
                    \subcaption*{\scalebox{.75}{\small EPFT\cite{Kang_2021_ICCV}}}
                \end{minipage}%
                \begin{minipage}{.0832\textwidth}
                    \centering
                    \subcaption*{\scalebox{.75}{1 pattern~\cite{gu2020cascade}}}
                \end{minipage}%
                \begin{minipage}{.0832\textwidth}
                    \centering
                    \subcaption*{\scalebox{.75}{5 patterns~\cite{gu2020cascade}}}
                \end{minipage}%
                \begin{minipage}{.0832\textwidth}
                    \centering
                    \subcaption*{\scalebox{.7}{\small COLMAP~\cite{schoenberger2016mvs}}}
                \end{minipage}%
            \end{minipage}
        \end{minipage}       
    \end{minipage}
    \begin{minipage}{\textwidth}
        \centering
        \begin{minipage}{0.065in}	
            \centering
            \rotatebox{90}{\footnotesize \textsc{Bowl}}
        \end{minipage}
        \begin{minipage}{.985\textwidth}
            \centering
            \begin{minipage}{.0832\textwidth}
                \includegraphics[width=\textwidth]{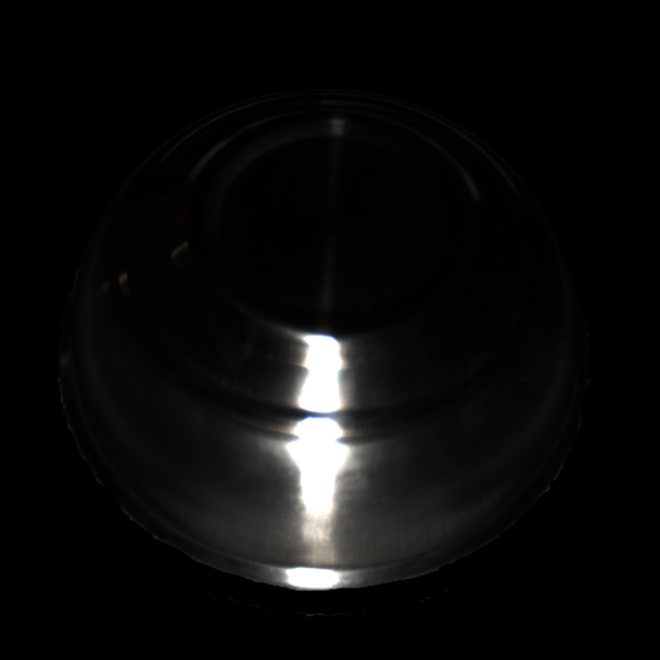}
            \end{minipage}%
            \begin{minipage}{.0832\textwidth}
                \includegraphics[width=\textwidth]{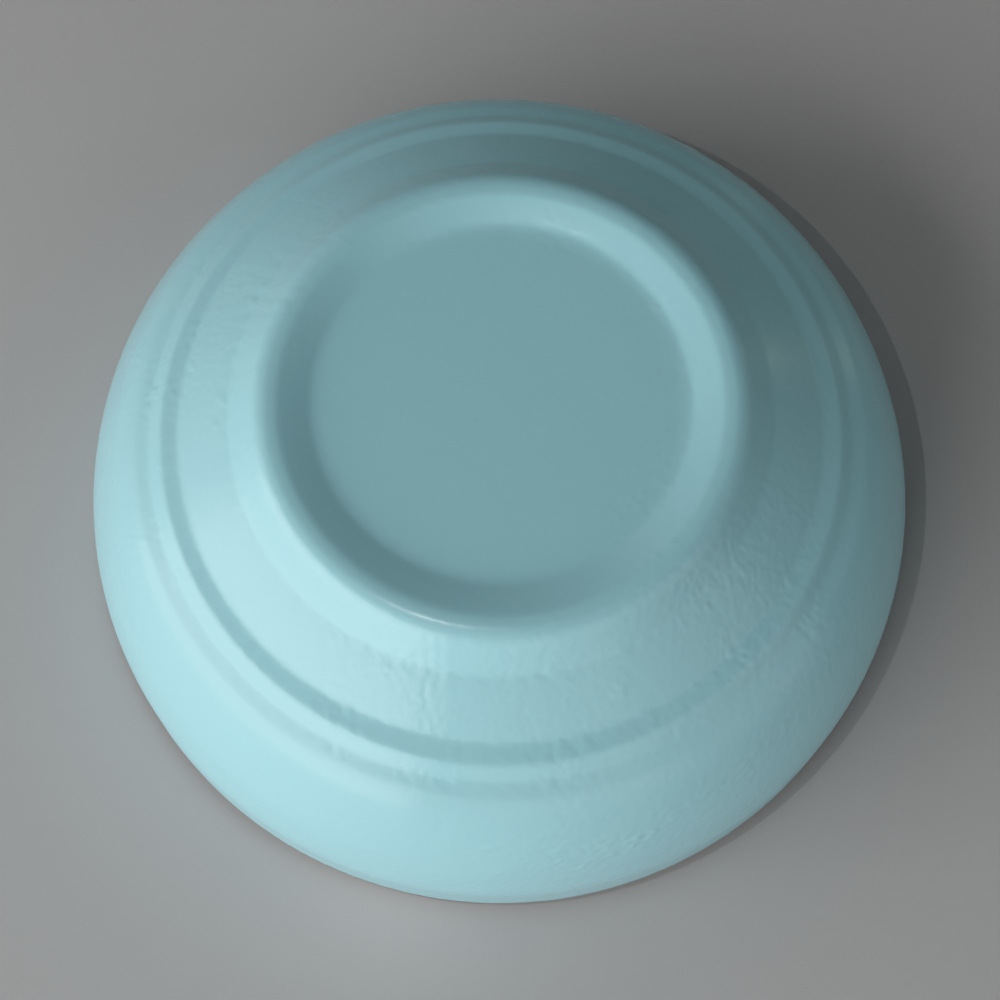}
            \end{minipage}%
            \begin{minipage}{.0832\textwidth}
                \includegraphics[width=\textwidth]{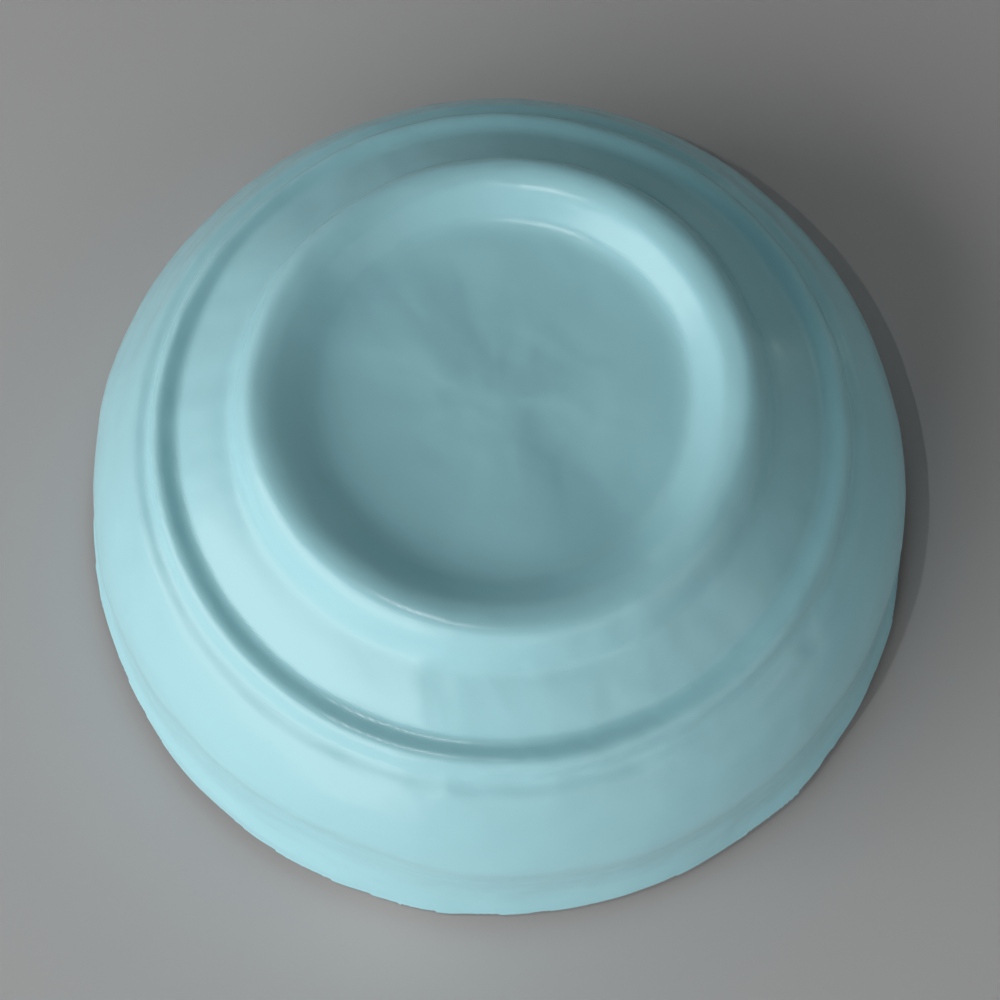}
                \put(-9.0,1.3){\scalebox{0.6}{%
                  \tightcolorbox[1pt]{black!70}{white}{0.5}%
                }}
            \end{minipage}%
            \begin{minipage}{.0832\textwidth}
                \includegraphics[width=\textwidth]{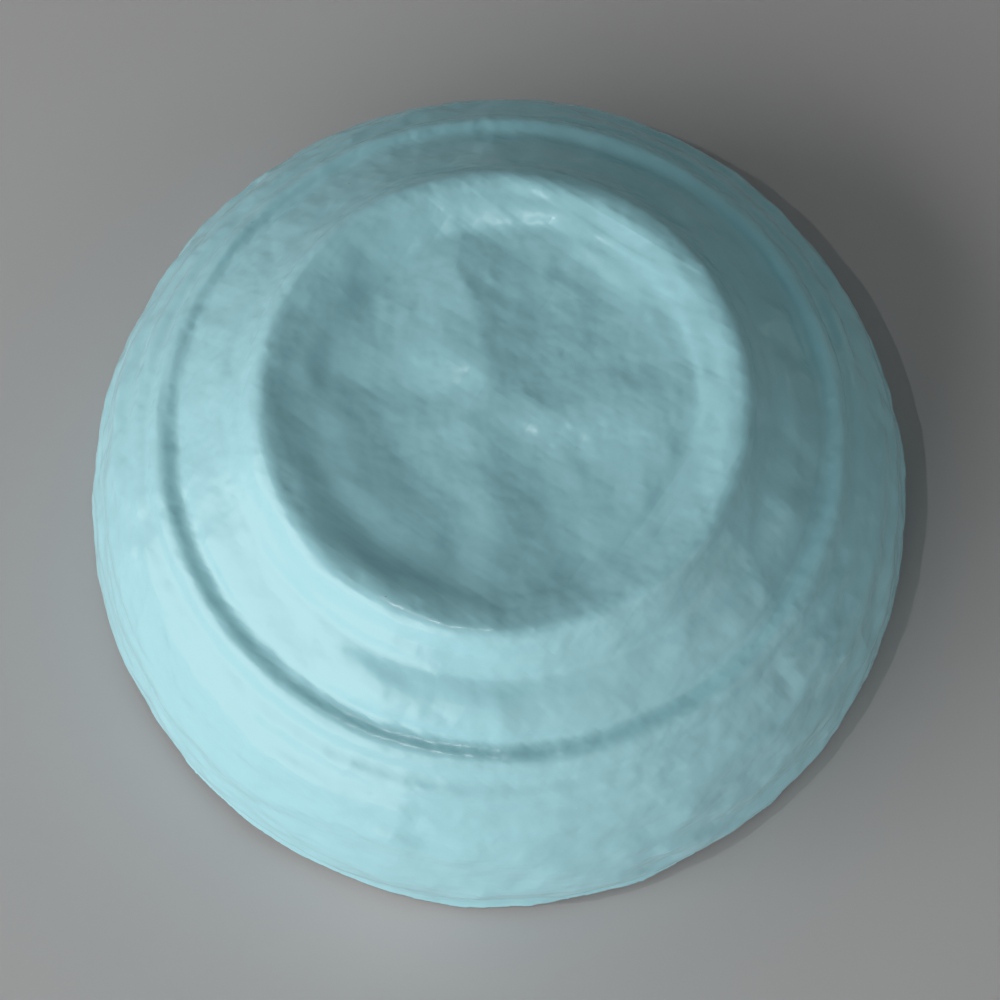}
                \put(-9.0,1.3){\scalebox{0.6}{%
                  \tightcolorbox[1pt]{black!70}{white}{0.3}%
                }}
            \end{minipage}%
            \begin{minipage}{.0832\textwidth}
                \includegraphics[width=\textwidth]{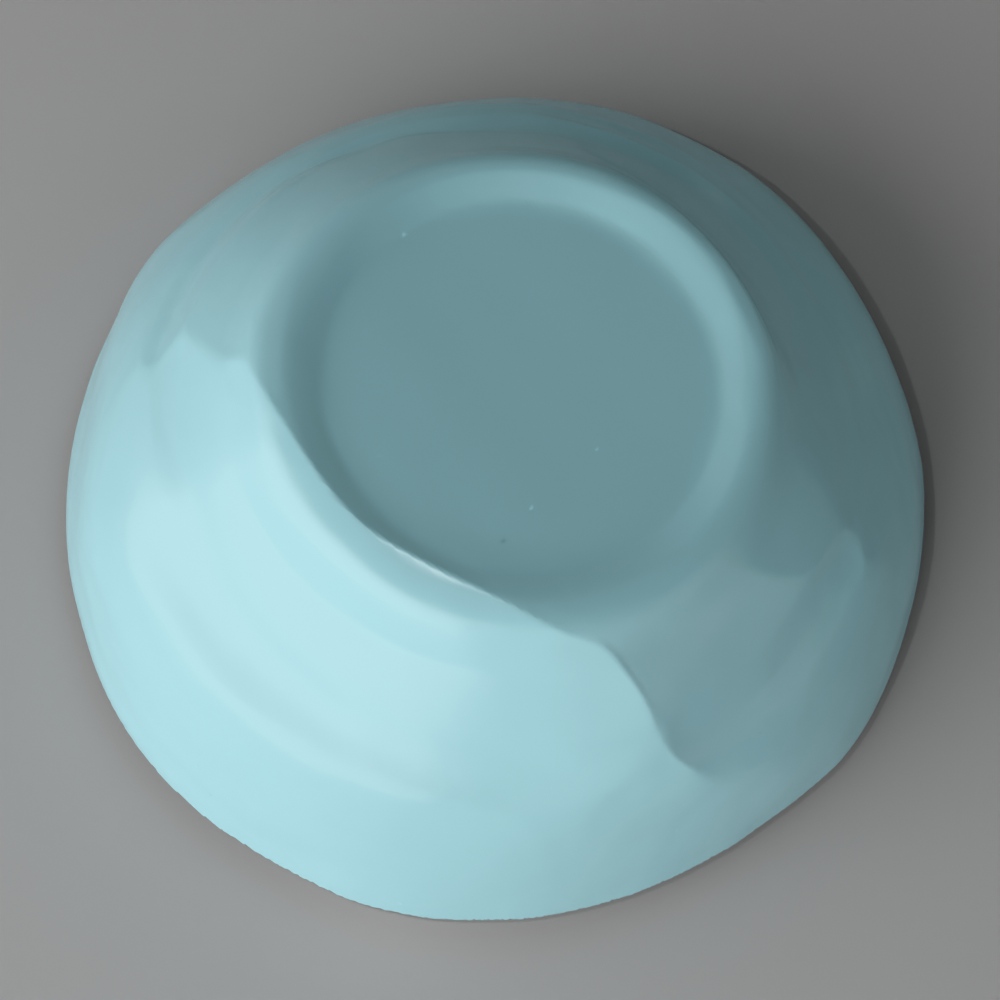}
                \put(-11.7,1.3){\scalebox{0.6}{%
                  \tightcolorbox[1pt]{black!70}{white}{11.9}%
                }}
            \end{minipage}%
            \begin{minipage}{.0832\textwidth}
                \includegraphics[width=\textwidth]{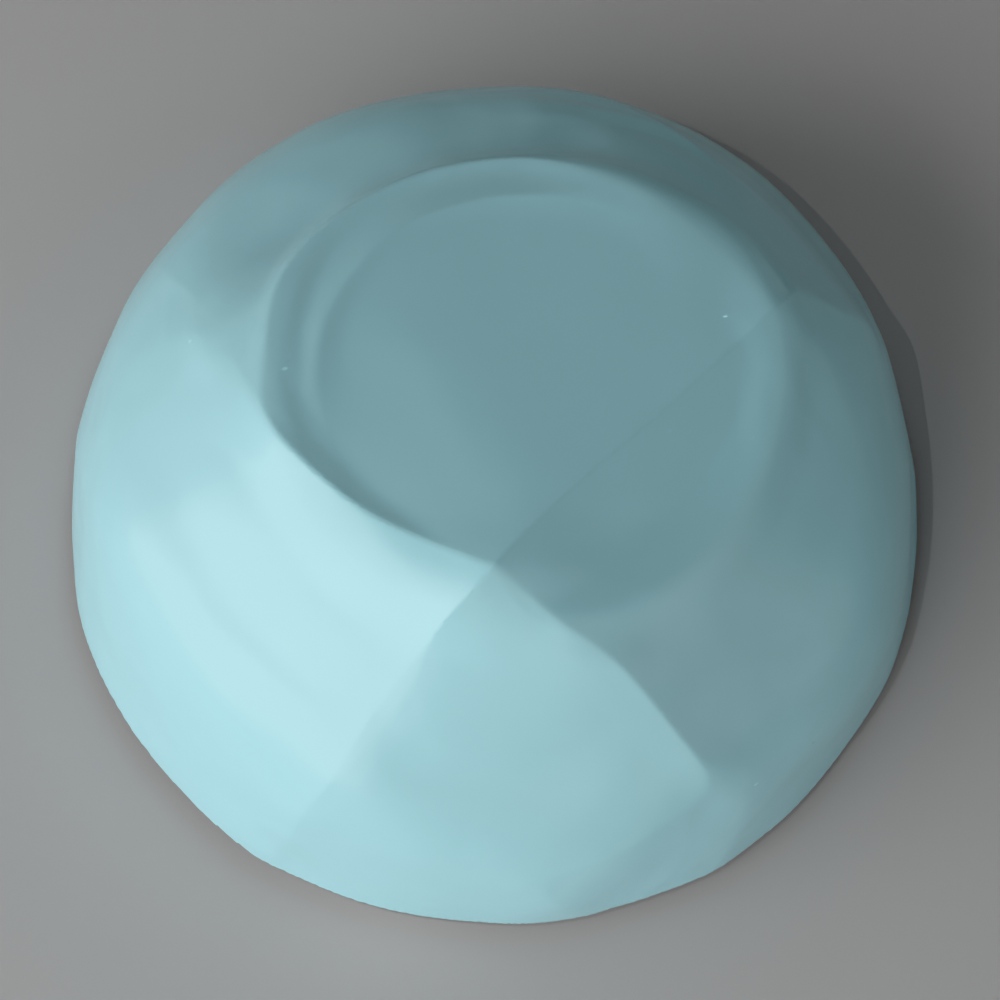}
                \put(-11.7,1.3){\scalebox{0.6}{%
                  \tightcolorbox[1pt]{black!70}{white}{18.0}%
                }}
            \end{minipage}%
            \begin{minipage}{.0832\textwidth}
                \includegraphics[width=\textwidth]{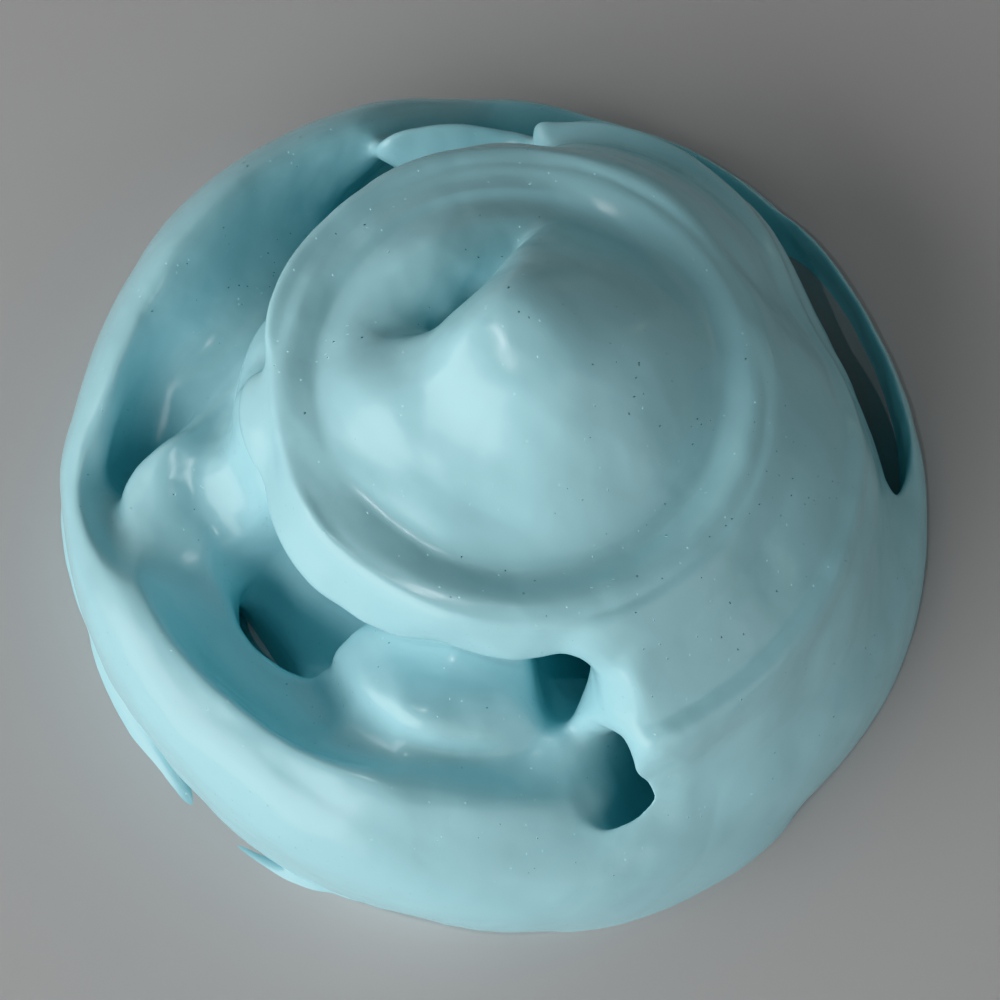}
                \put(-11.7,1.3){\scalebox{0.6}{%
                  \tightcolorbox[1pt]{black!70}{white}{42.4}%
                }}
            \end{minipage}%
            \begin{minipage}{.0832\textwidth}
                \includegraphics[width=\textwidth]{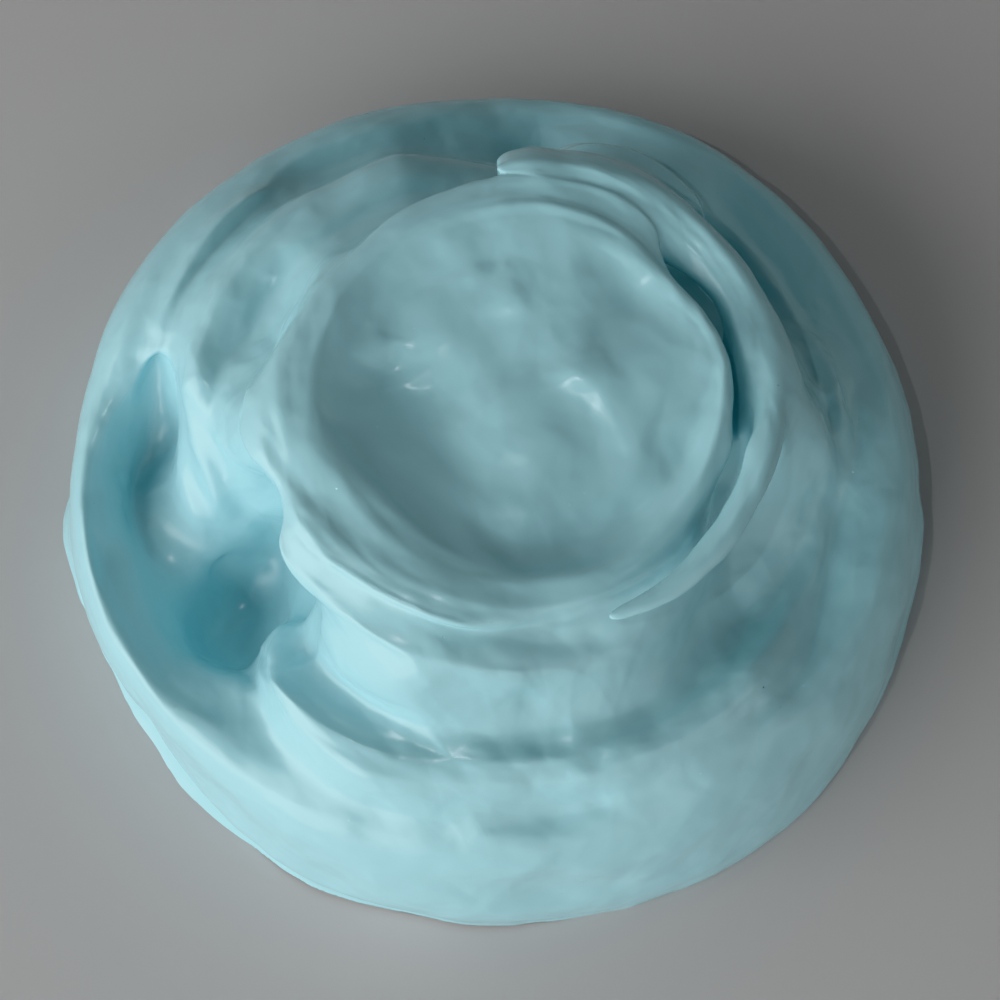}
                \put(-11.7,1.3){\scalebox{0.6}{%
                  \tightcolorbox[1pt]{black!70}{white}{48.7}%
                }}
            \end{minipage}%
            \begin{minipage}{.0832\textwidth}
                \includegraphics[width=\textwidth]{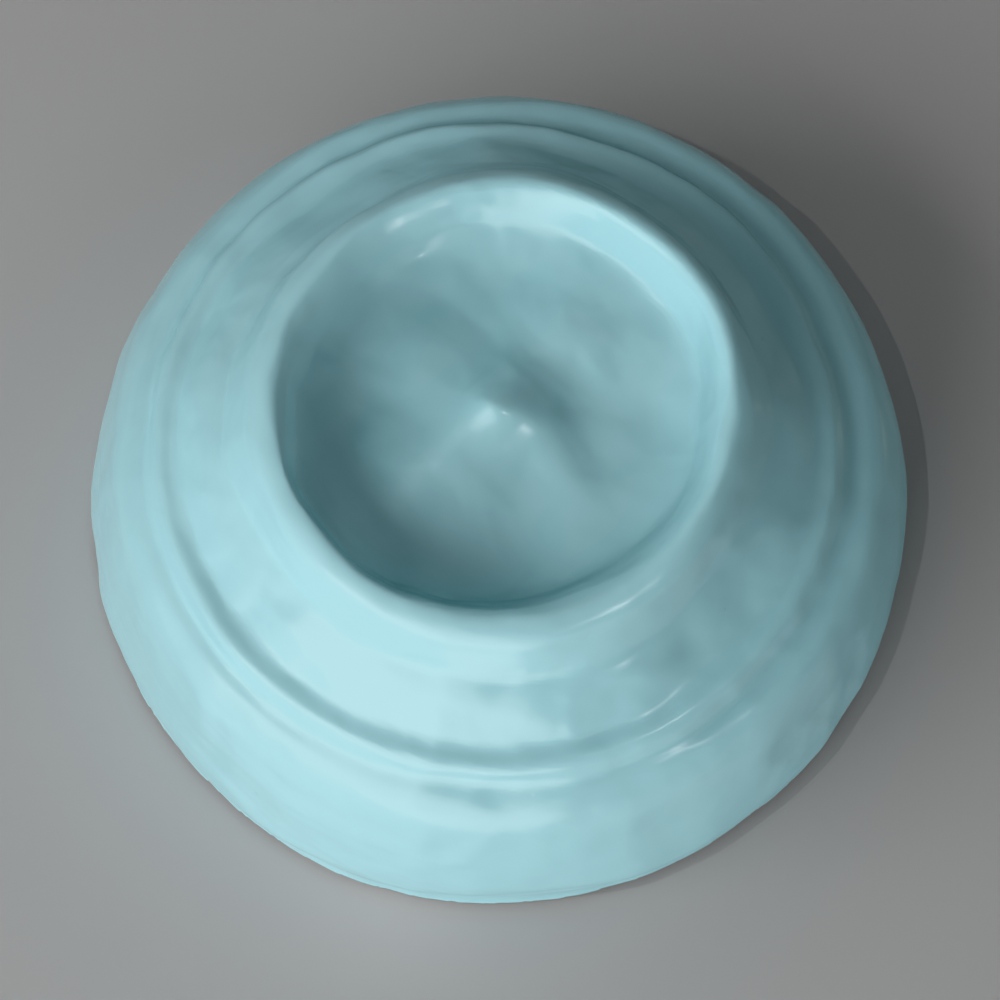}
                \put(-11.7,1.3){\scalebox{0.6}{%
                  \tightcolorbox[1pt]{black!70}{white}{16.4}%
                }}
            \end{minipage}%
            \begin{minipage}{.0832\textwidth}
                \includegraphics[width=\textwidth]{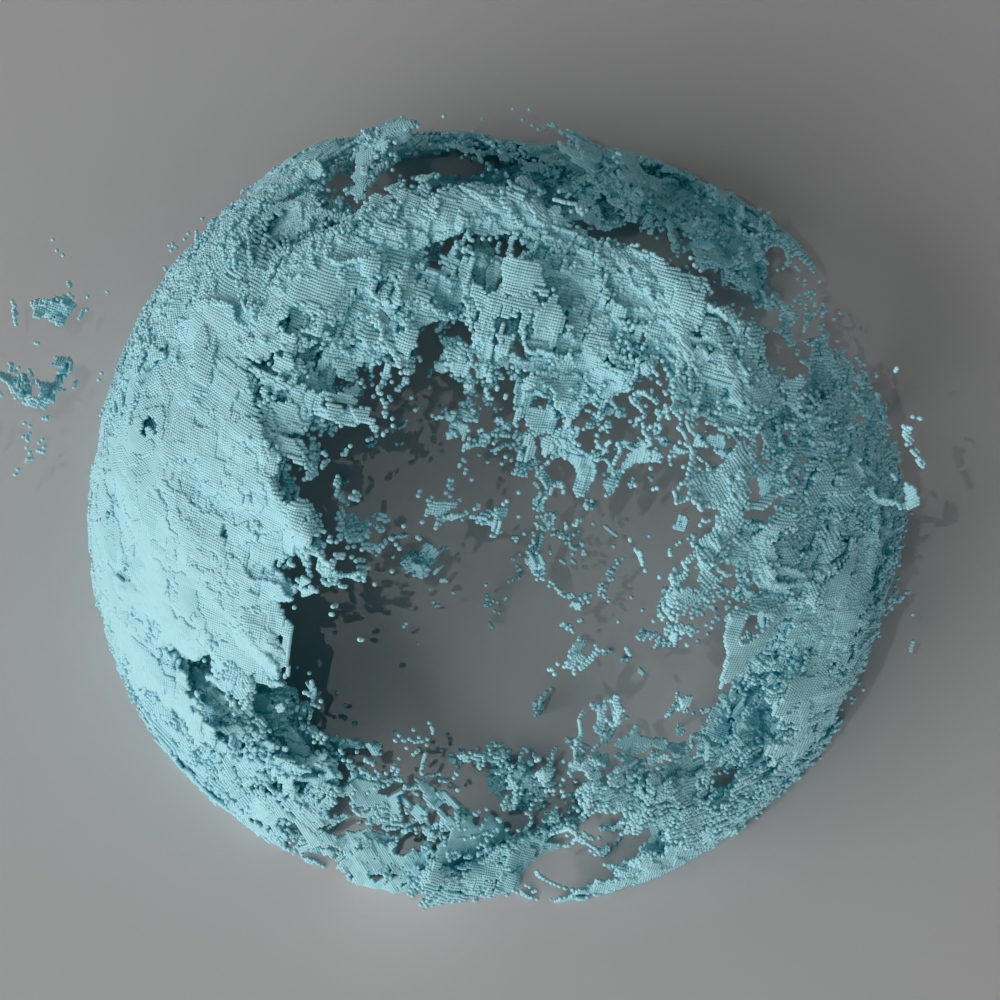}
                \put(-11.7,1.3){\scalebox{0.6}{%
                  \tightcolorbox[1pt]{black!70}{white}{45.7}%
                }}
            \end{minipage}%
            \begin{minipage}{.0832\textwidth}
                \includegraphics[width=\textwidth]{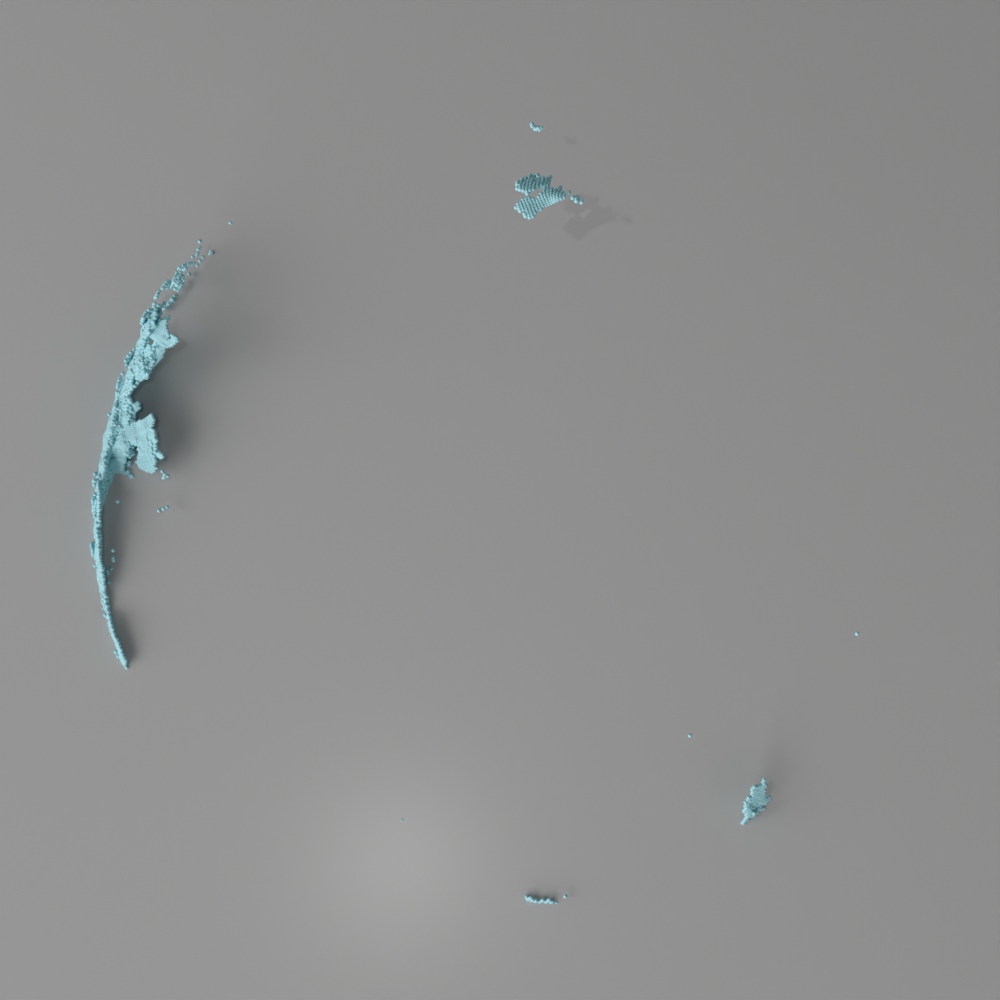}
                \put(-14.5,1.3){\scalebox{0.6}{%
                  \tightcolorbox[1pt]{black!70}{white}{991.5}%
                }}
            \end{minipage}%
            \begin{minipage}{.0832\textwidth}
                \includegraphics[width=\textwidth]{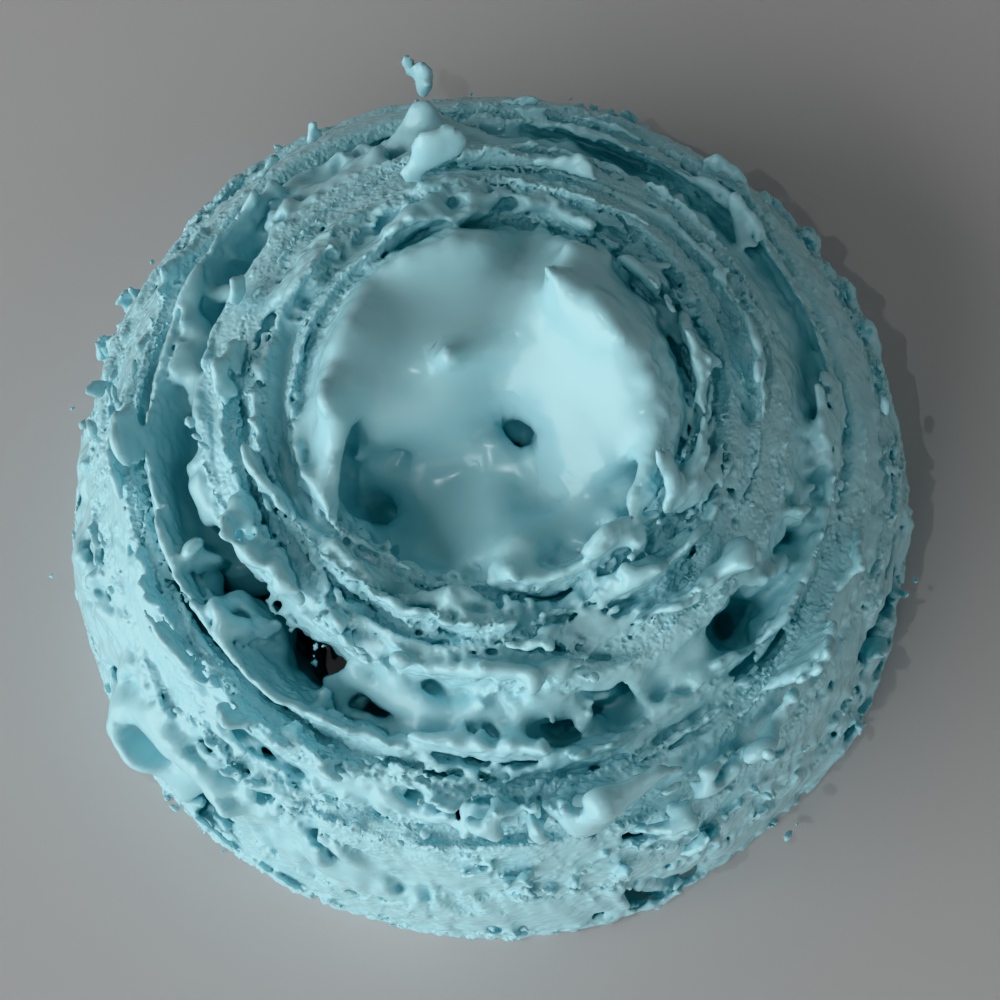}
                \put(-11.7,1.3){\scalebox{0.6}{%
                  \tightcolorbox[1pt]{black!70}{white}{39.6}%
                }}
            \end{minipage}%
        \end{minipage}
    \end{minipage}
    \begin{minipage}{\textwidth}
        \centering
        \begin{minipage}{0.065in}	
            \centering
            \rotatebox{90}{\footnotesize \textsc{Dog}}
        \end{minipage}
        \begin{minipage}{.985\textwidth}
            \centering
            \begin{minipage}{.0832\textwidth}
                \includegraphics[width=\textwidth]{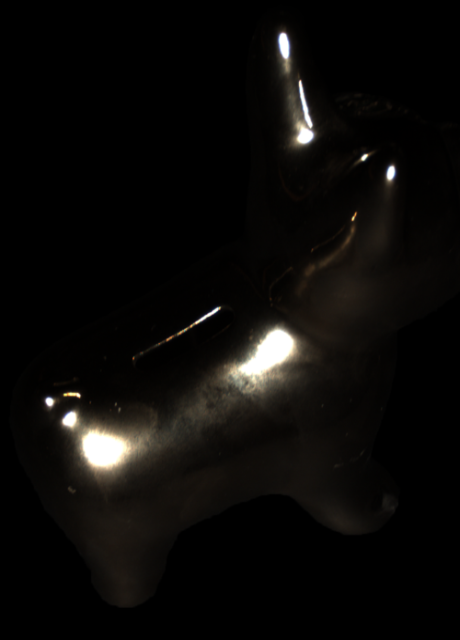}
            \end{minipage}%
            \begin{minipage}{.0832\textwidth}
                \includegraphics[width=\textwidth]{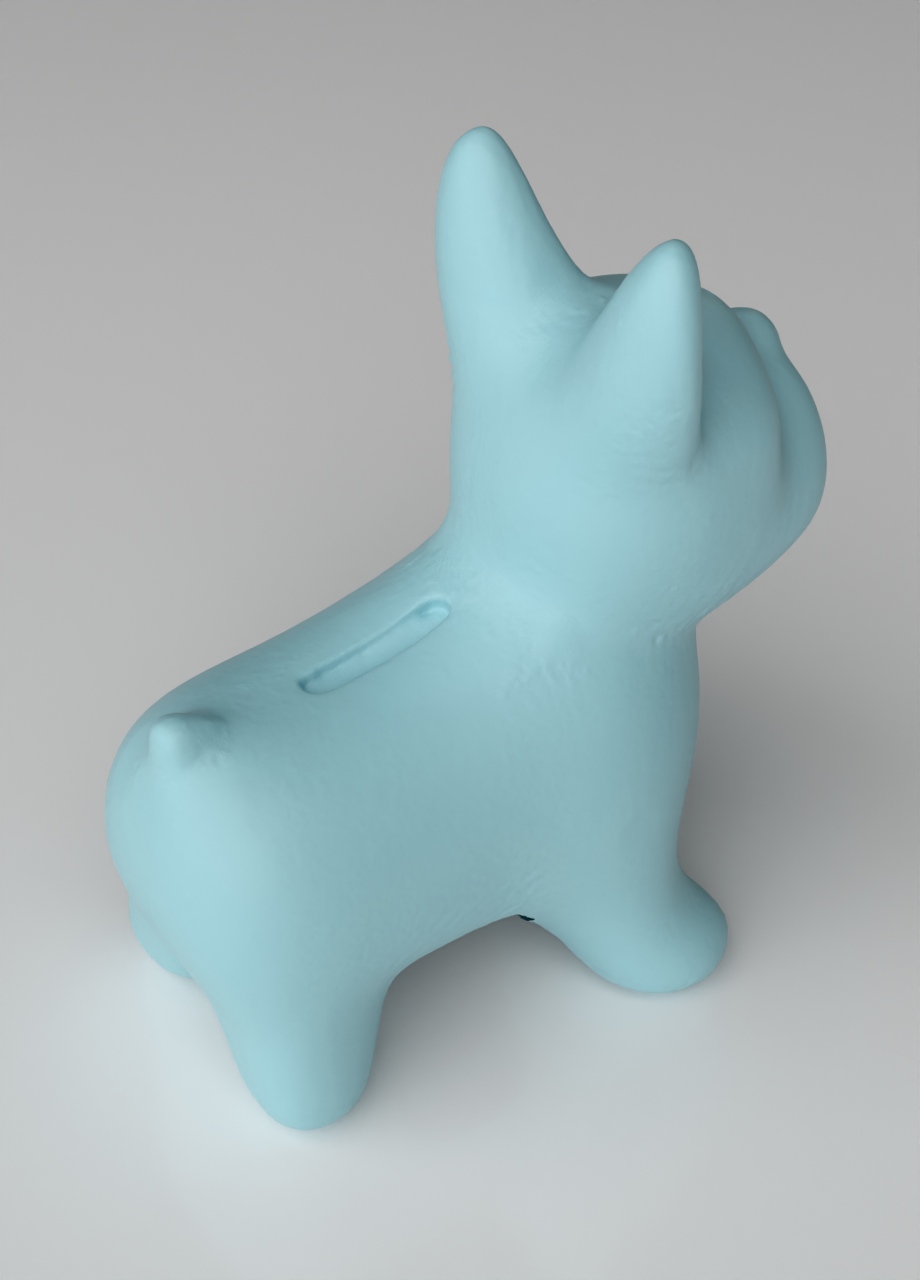}
            \end{minipage}%
            \begin{minipage}{.0832\textwidth}
                \includegraphics[width=\textwidth]{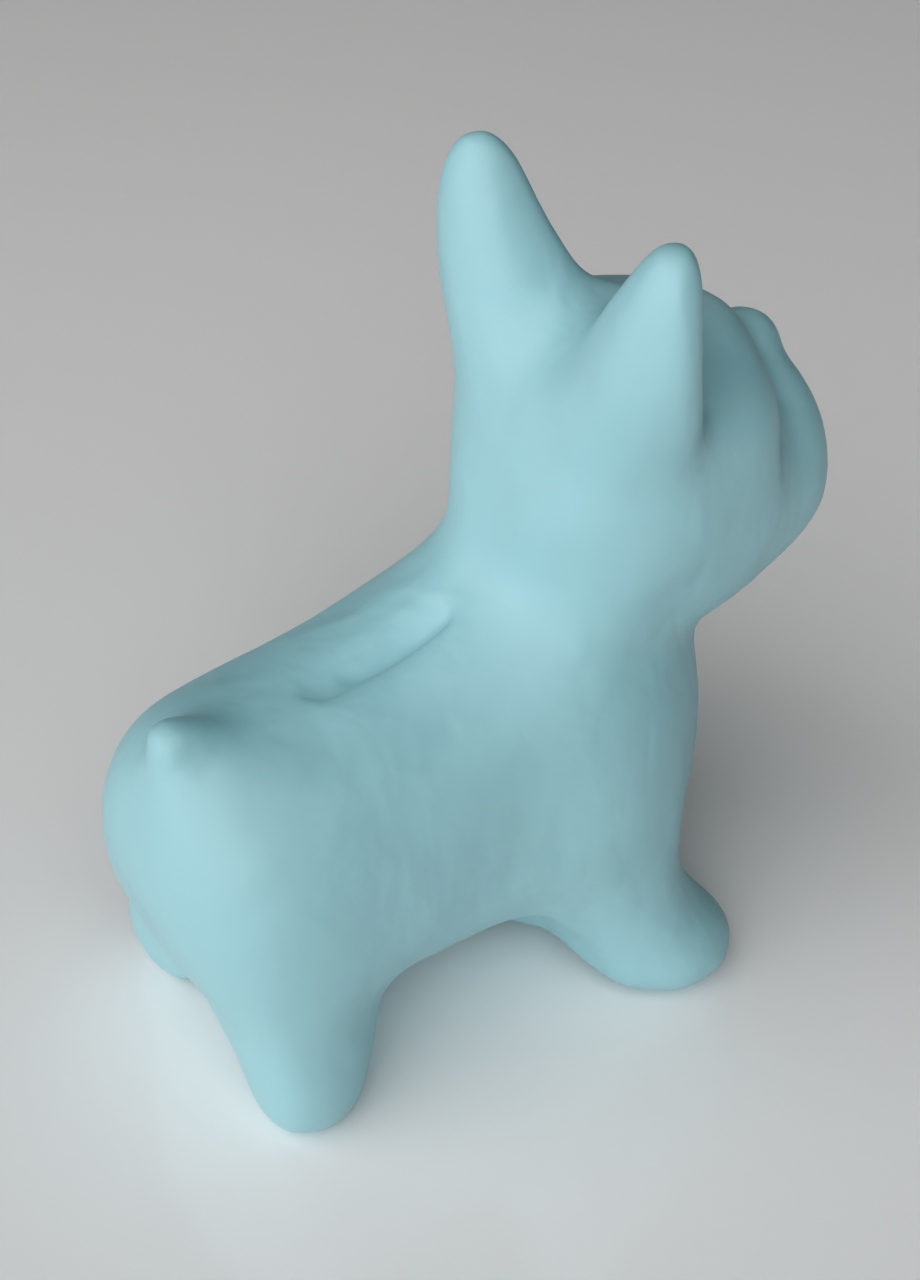}
                \put(-9.0,1.3){\scalebox{0.6}{%
                  \tightcolorbox[1pt]{black!70}{white}{2.9}%
                }}
            \end{minipage}%
            \begin{minipage}{.0832\textwidth}
                \includegraphics[width=\textwidth]{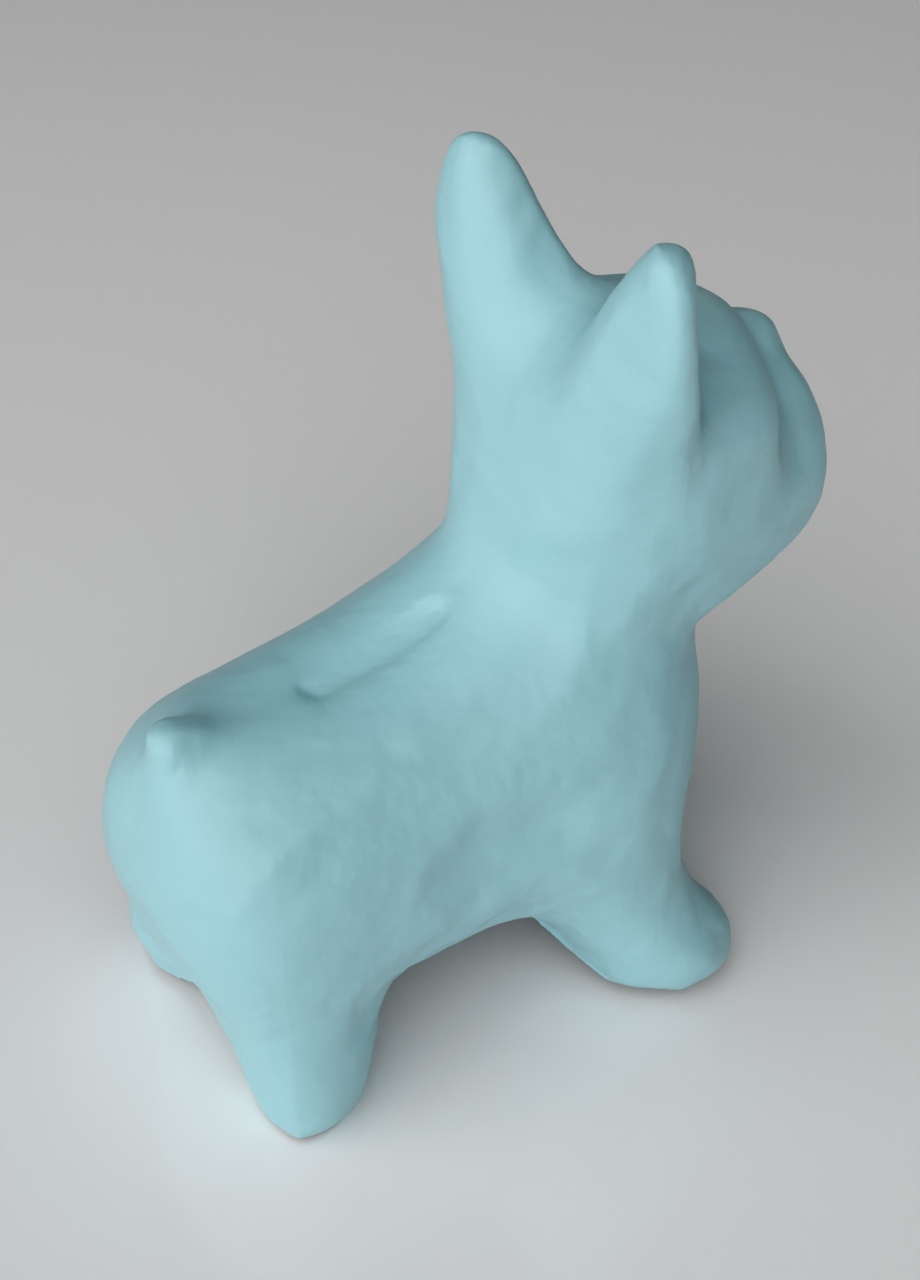}
                \put(-9.0,1.3){\scalebox{0.6}{%
                  \tightcolorbox[1pt]{black!70}{white}{3.6}%
                }}
            \end{minipage}%
            \begin{minipage}{.0832\textwidth}
                \includegraphics[width=\textwidth]{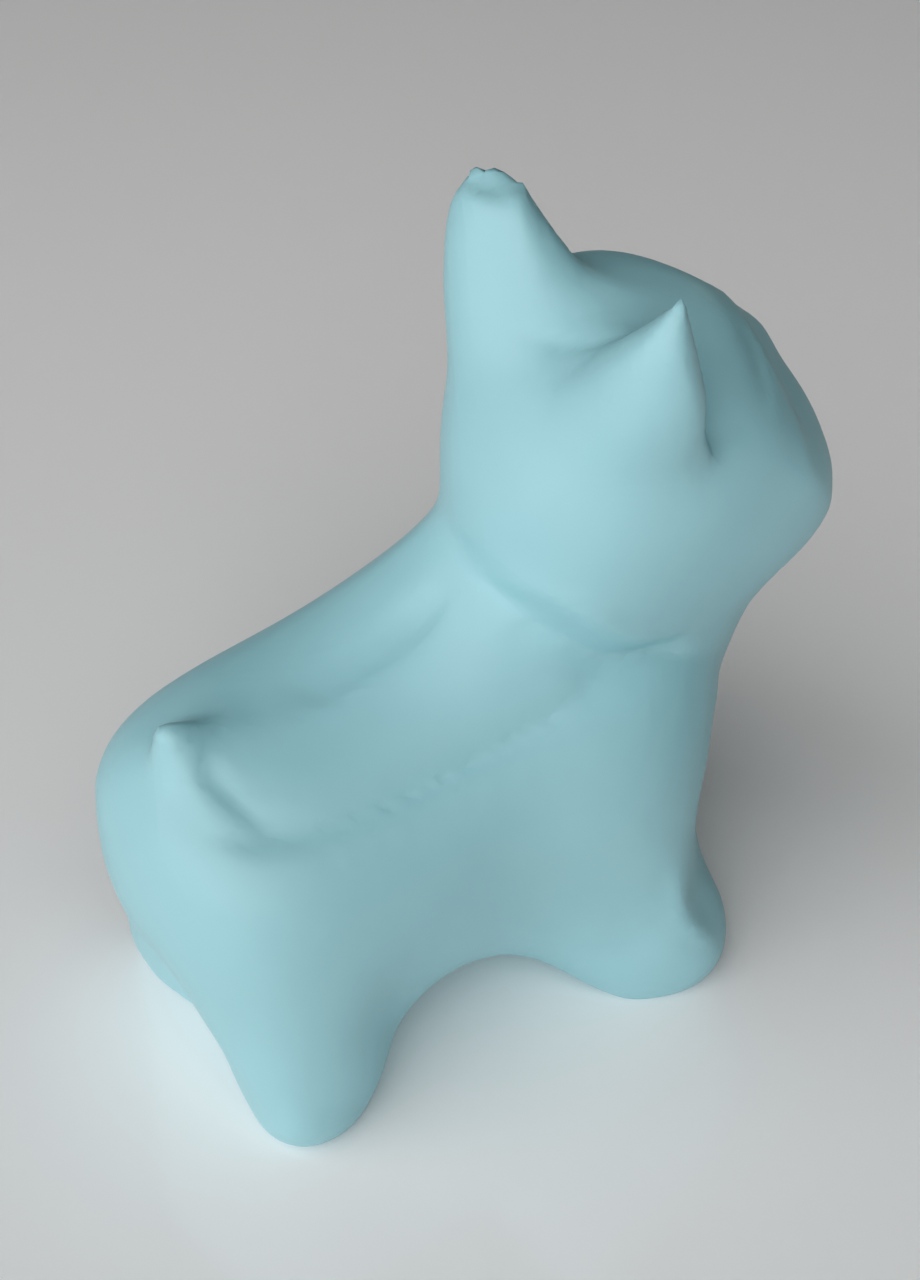}
                \put(-11.7,1.3){\scalebox{0.6}{%
                  \tightcolorbox[1pt]{black!70}{white}{49.7}%
                }}
            \end{minipage}%
            \begin{minipage}{.0832\textwidth}
                \includegraphics[width=\textwidth]{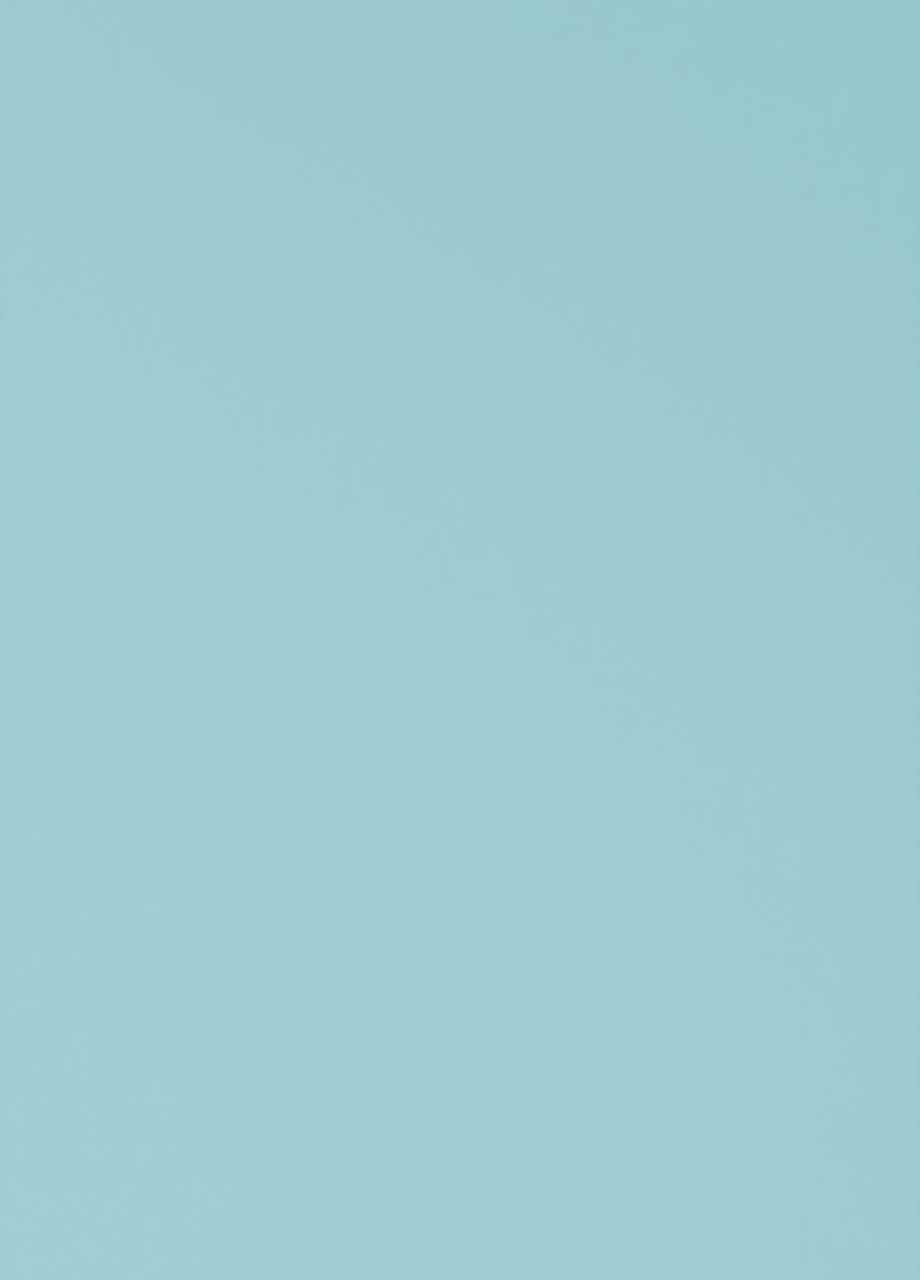}
                \put(-20.4,1.3){\scalebox{0.6}{%
                  \tightcolorbox[1pt]{black!70}{white}{95238.0}%
                }}
            \end{minipage}%
            \begin{minipage}{.0832\textwidth}
                \includegraphics[width=\textwidth]{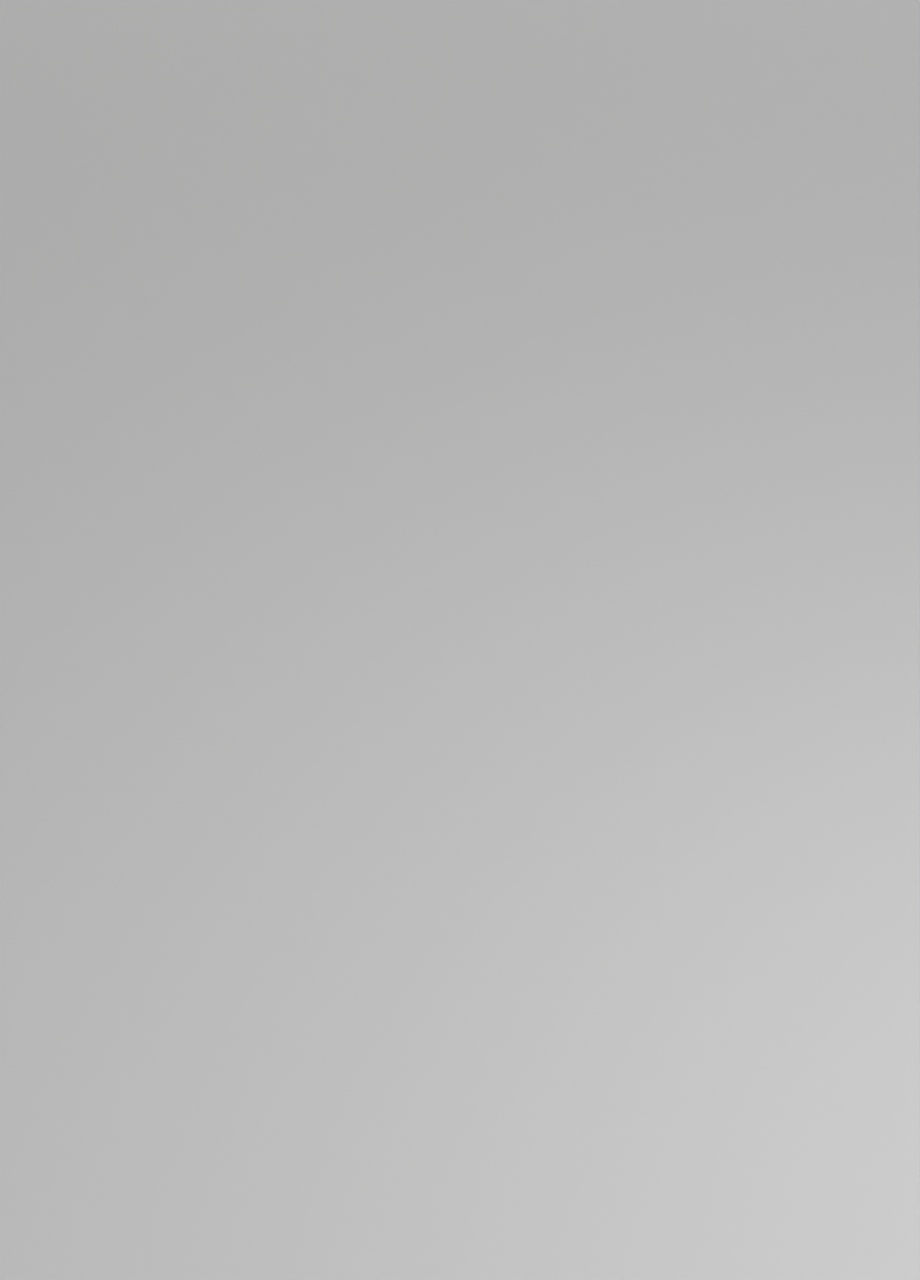}
                \put(-13.3,1.3){\scalebox{0.6}{%
                  \tightcolorbox[0pt]{black!70}{white}{N/A}%
                }}
            \end{minipage}%
            \begin{minipage}{.0832\textwidth}
                \includegraphics[width=\textwidth]{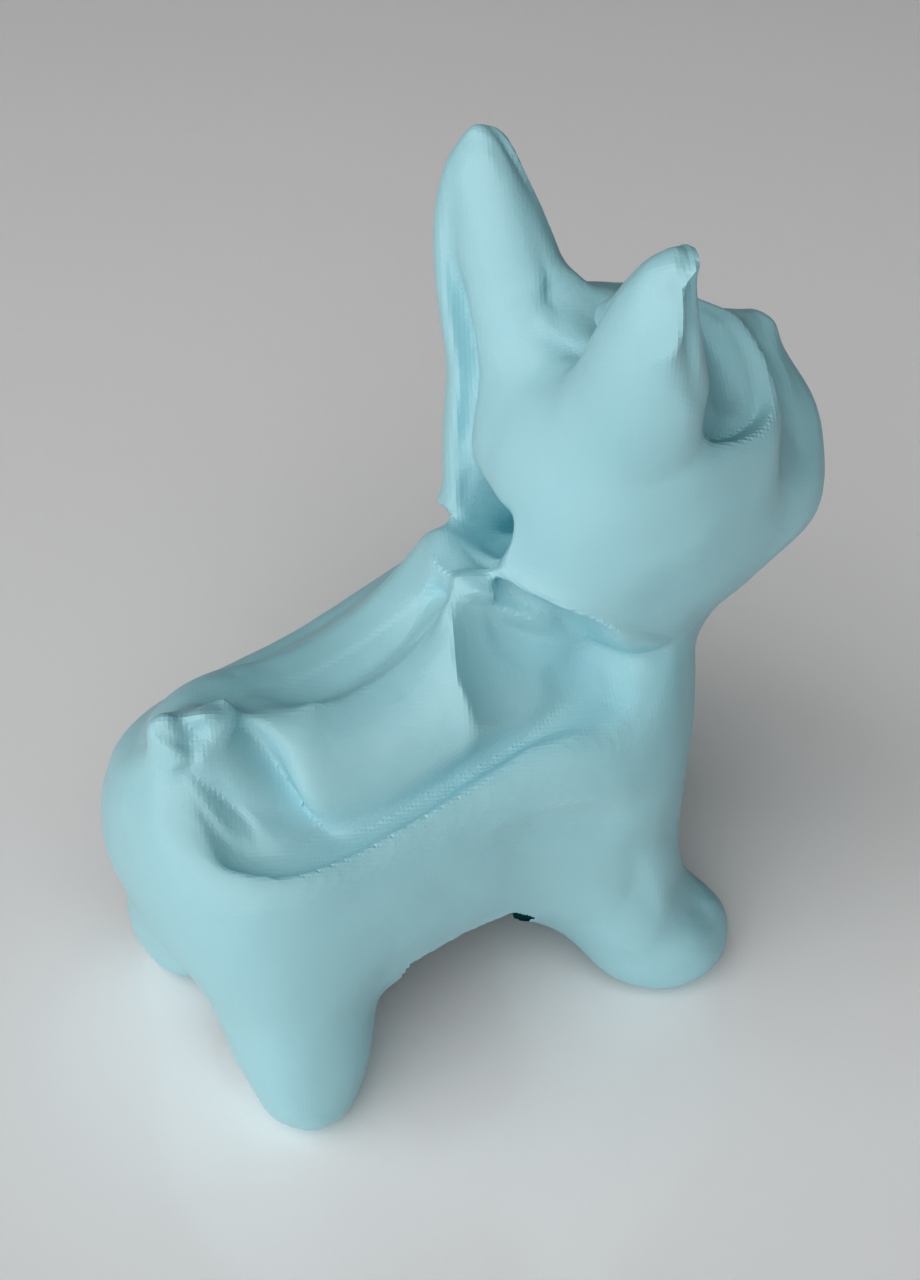}
                \put(-11.7,1.3){\scalebox{0.6}{%
                  \tightcolorbox[1pt]{black!70}{white}{15.4}%
                }}
            \end{minipage}%
            \begin{minipage}{.0832\textwidth}
                \includegraphics[width=\textwidth]{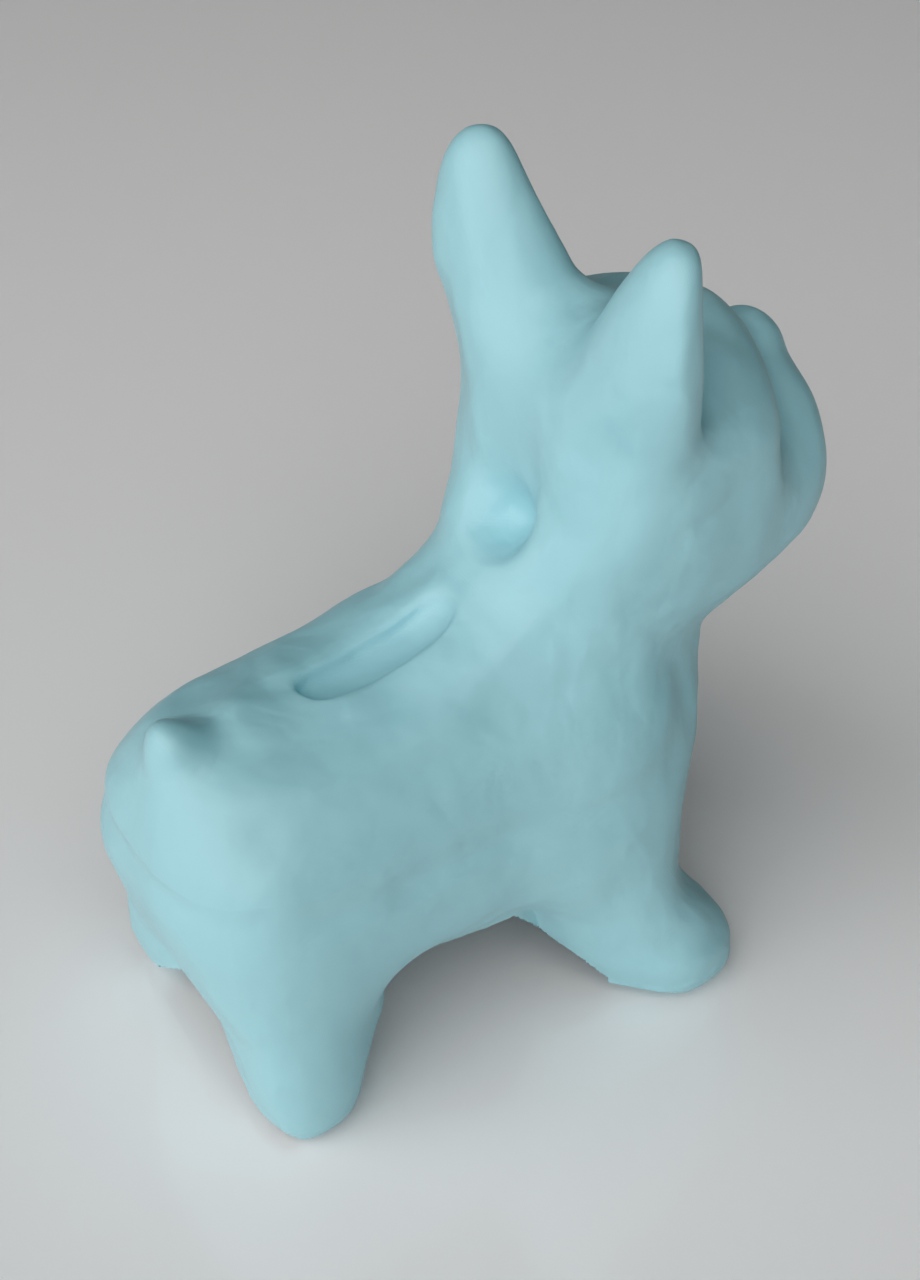}
                \put(-11.7,1.3){\scalebox{0.6}{%
                  \tightcolorbox[1pt]{black!70}{white}{14.2}%
                }}
            \end{minipage}%
            \begin{minipage}{.0832\textwidth}
                \includegraphics[width=\textwidth]{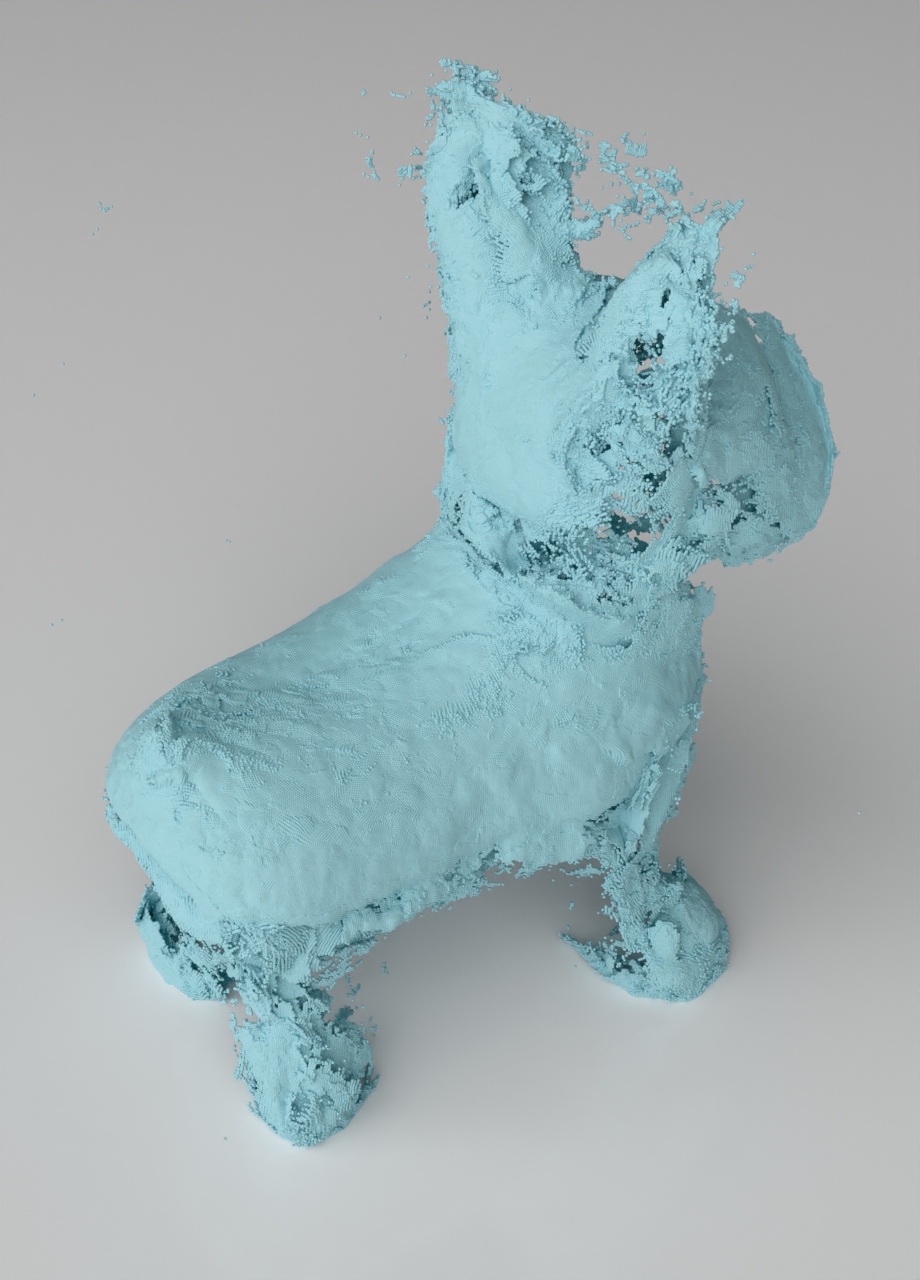}
                \put(-11.7,1.3){\scalebox{0.6}{%
                  \tightcolorbox[1pt]{black!70}{white}{34.0}%
                }}
            \end{minipage}%
            \begin{minipage}{.0832\textwidth}
                \includegraphics[width=\textwidth]{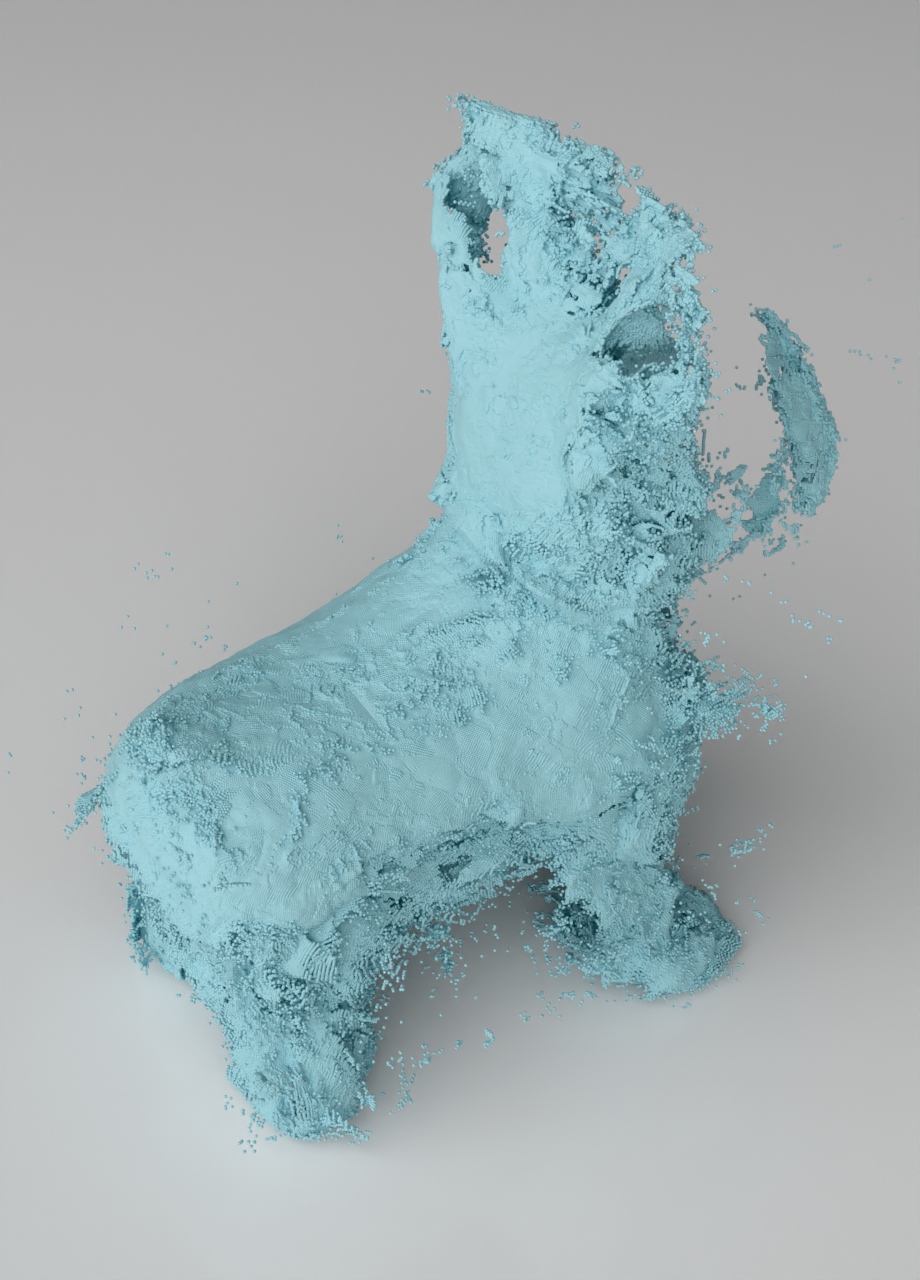}
                \put(-11.7,1.3){\scalebox{0.6}{%
                  \tightcolorbox[1pt]{black!70}{white}{20.0}%
                }}
            \end{minipage}%
            \begin{minipage}{.0832\textwidth}
                \includegraphics[width=\textwidth]{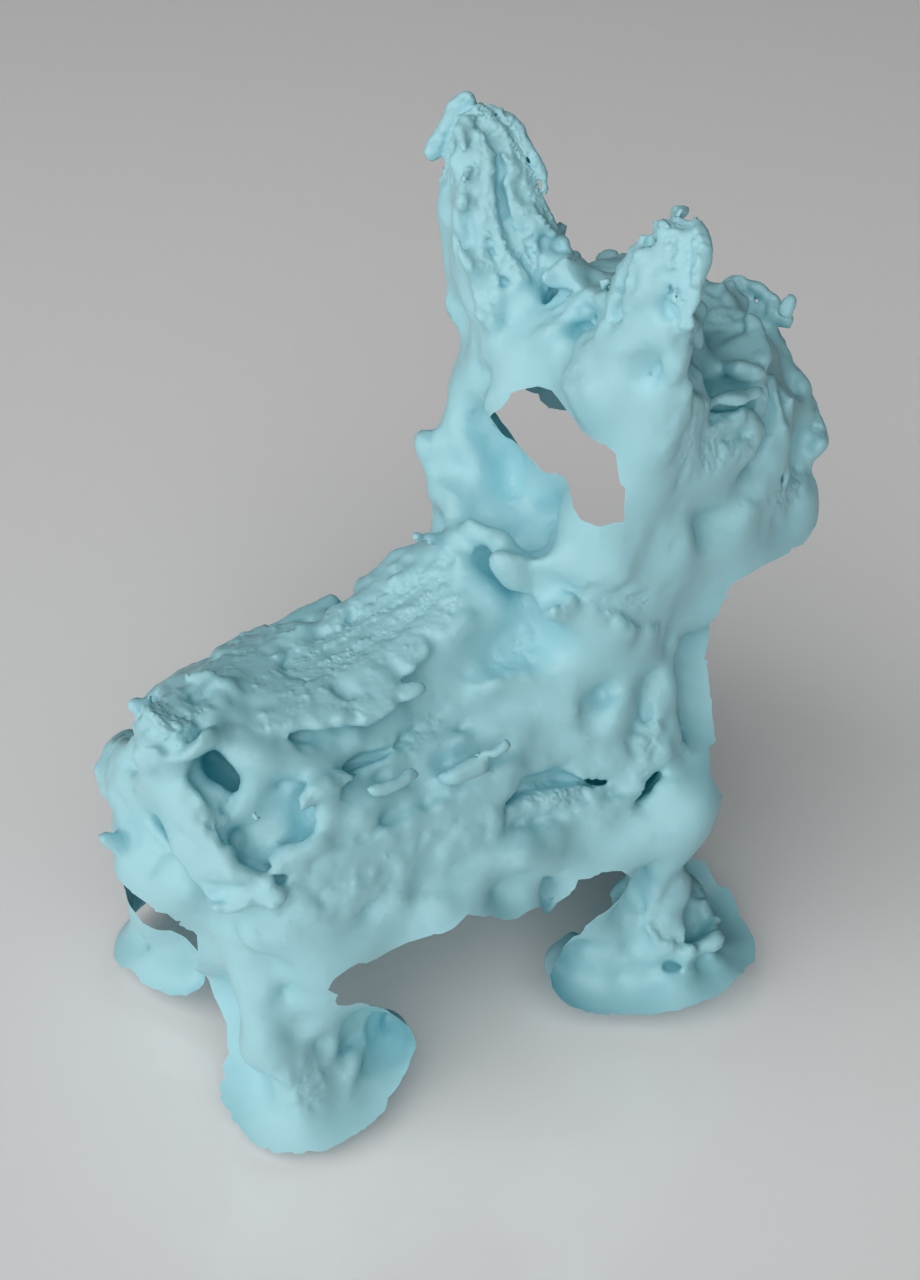}
                \put(-11.7,1.3){\scalebox{0.6}{%
                  \tightcolorbox[1pt]{black!70}{white}{58.5}%
                }}
            \end{minipage}%
        \end{minipage}
    \end{minipage}
    \begin{minipage}{\textwidth}
        \centering
        \begin{minipage}{0.065in}	
            \centering
            \rotatebox{90}{\footnotesize \textsc{Bear}}
        \end{minipage}
        \begin{minipage}{.985\textwidth}
            \centering
            \begin{minipage}{.0832\textwidth}
                \includegraphics[width=\textwidth]{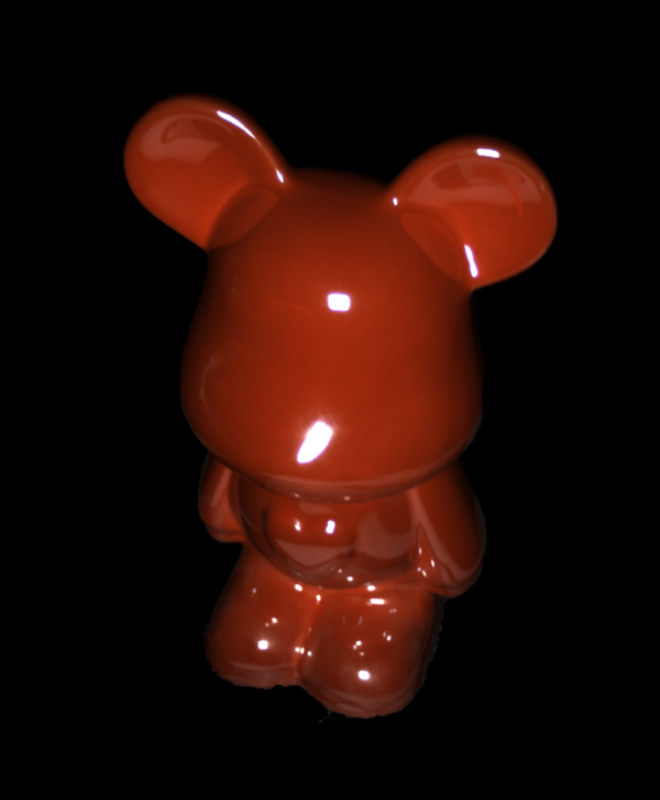}
            \end{minipage}%
            \begin{minipage}{.0832\textwidth}
                \includegraphics[width=\textwidth]{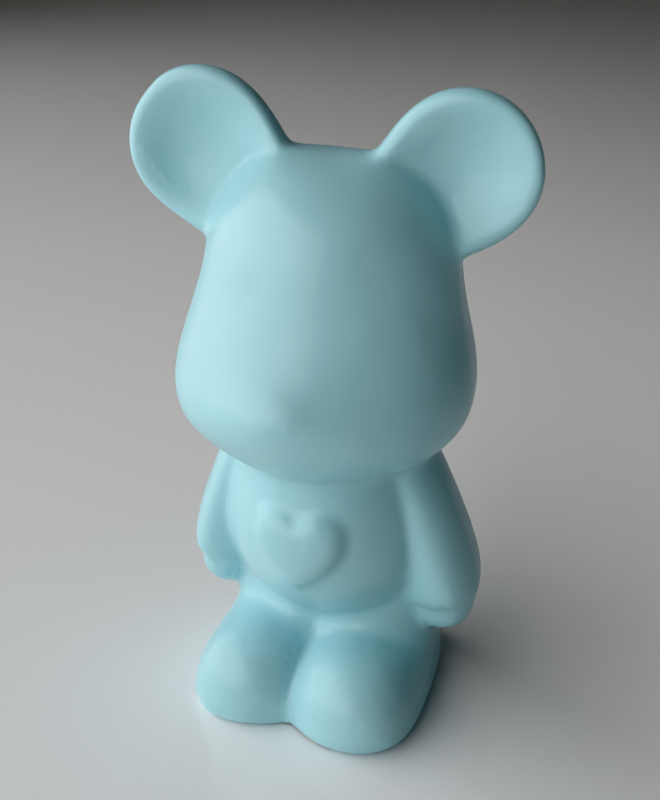}
            \end{minipage}%
            \begin{minipage}{.0832\textwidth}
                \includegraphics[width=\textwidth]{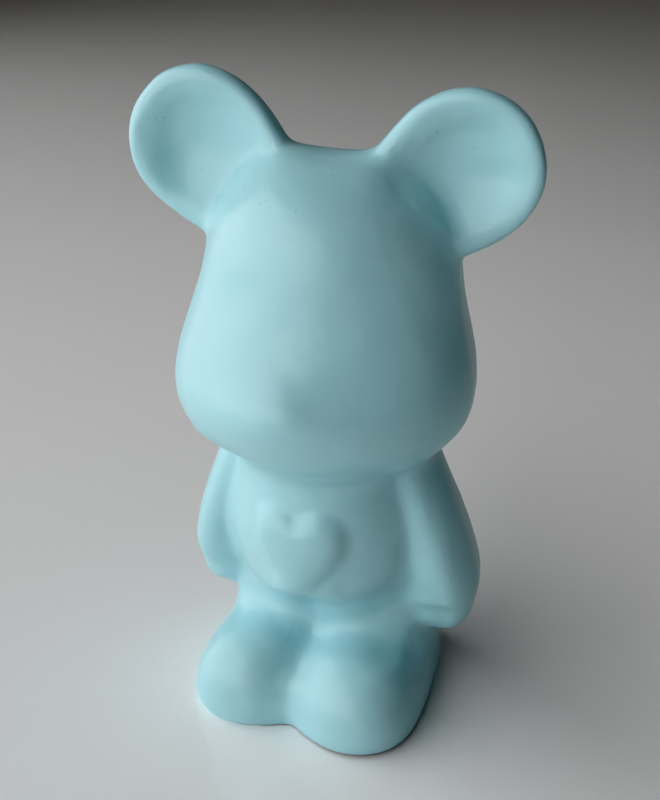}
                \put(-9.0,1.3){\scalebox{0.6}{%
                  \tightcolorbox[1pt]{black!70}{white}{0.4}%
                }}
            \end{minipage}%
            \begin{minipage}{.0832\textwidth}
                \includegraphics[width=\textwidth]{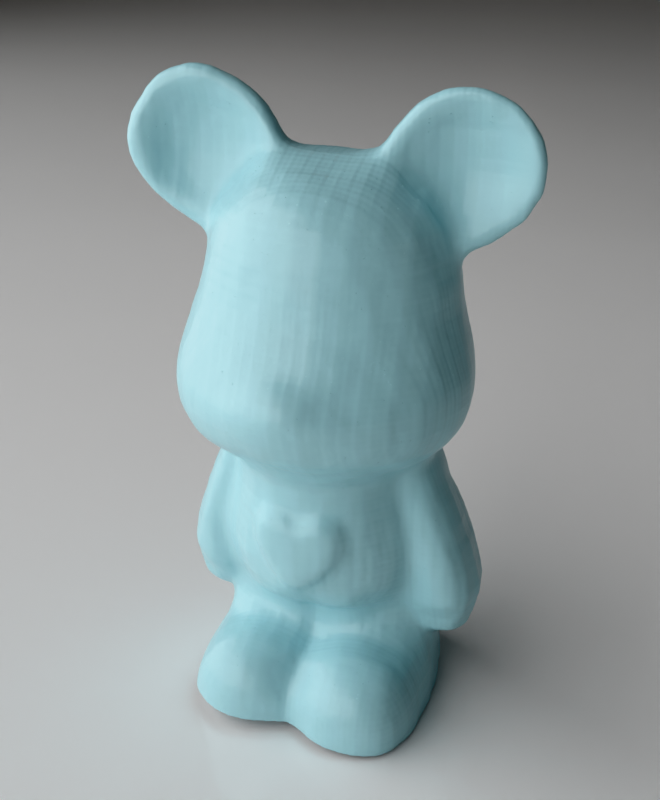}
                \put(-9.0,1.3){\scalebox{0.6}{%
                  \tightcolorbox[1pt]{black!70}{white}{0.8}%
                }}
            \end{minipage}%
            \begin{minipage}{.0832\textwidth}
                \includegraphics[width=\textwidth]{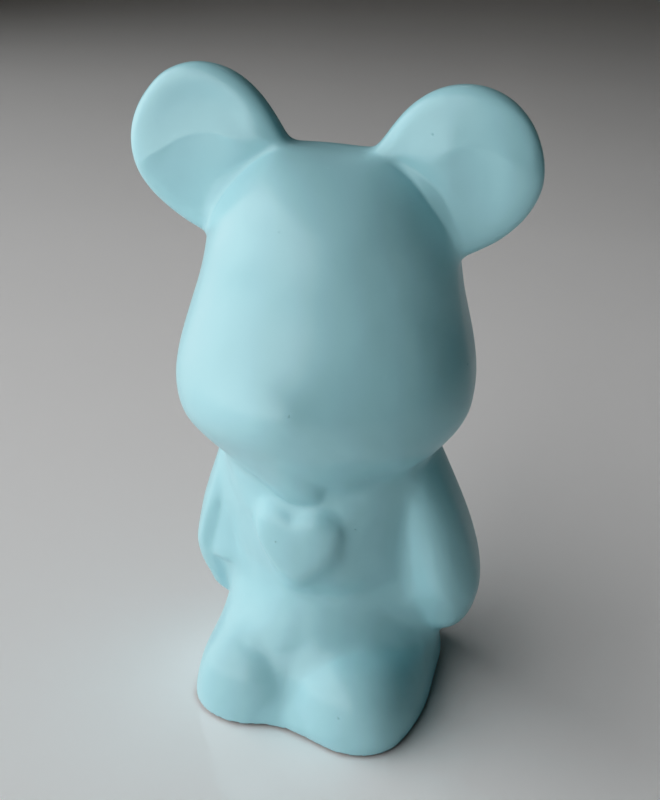}
                \put(-9.0,1.3){\scalebox{0.6}{%
                  \tightcolorbox[1pt]{black!70}{white}{4.5}%
                }}
            \end{minipage}%
            \begin{minipage}{.0832\textwidth}
                \includegraphics[width=\textwidth]{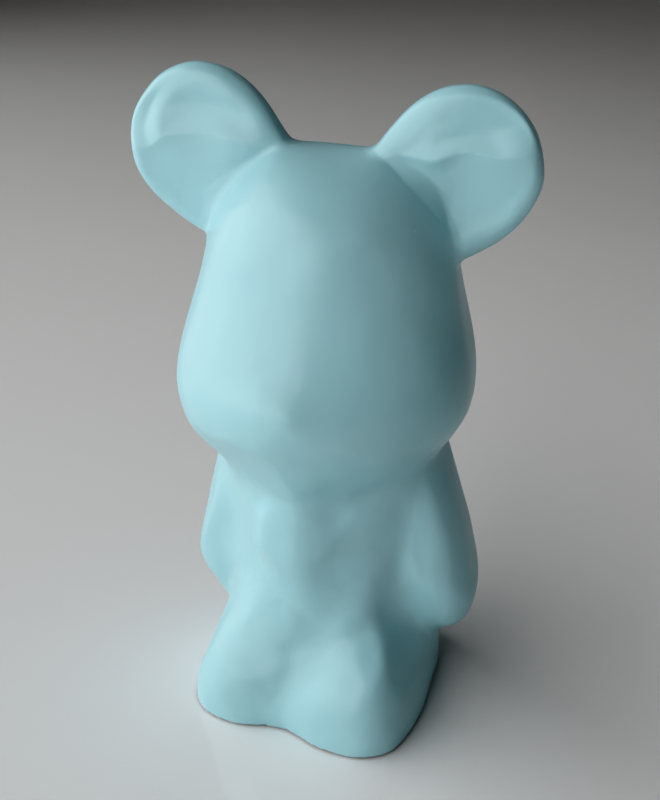}
                \put(-9.0,1.3){\scalebox{0.6}{%
                  \tightcolorbox[1pt]{black!70}{white}{6.9}%
                }}
            \end{minipage}%
            \begin{minipage}{.0832\textwidth}
                \includegraphics[width=\textwidth]{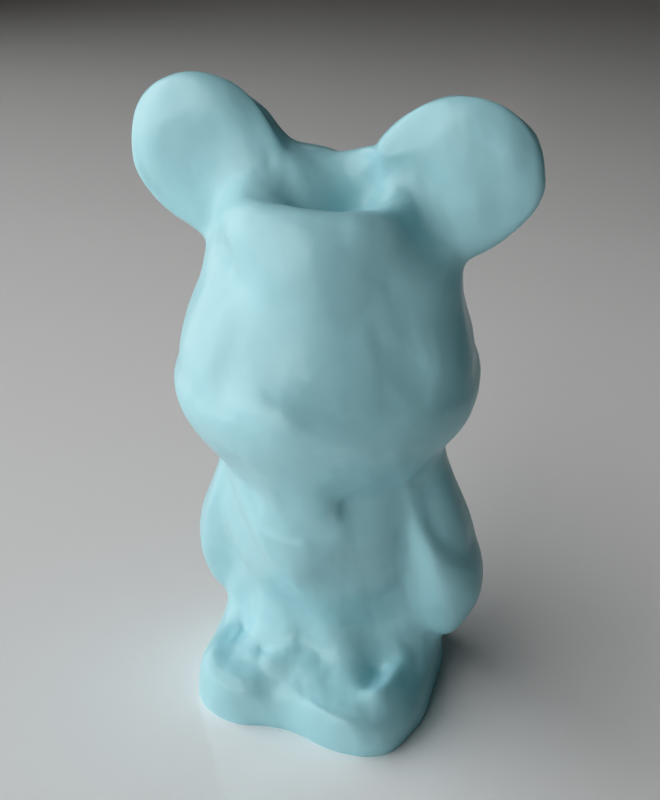}
                \put(-9.0,1.3){\scalebox{0.6}{%
                  \tightcolorbox[1pt]{black!70}{white}{7.5}%
                }}
            \end{minipage}%
            \begin{minipage}{.0832\textwidth}
                \includegraphics[width=\textwidth]{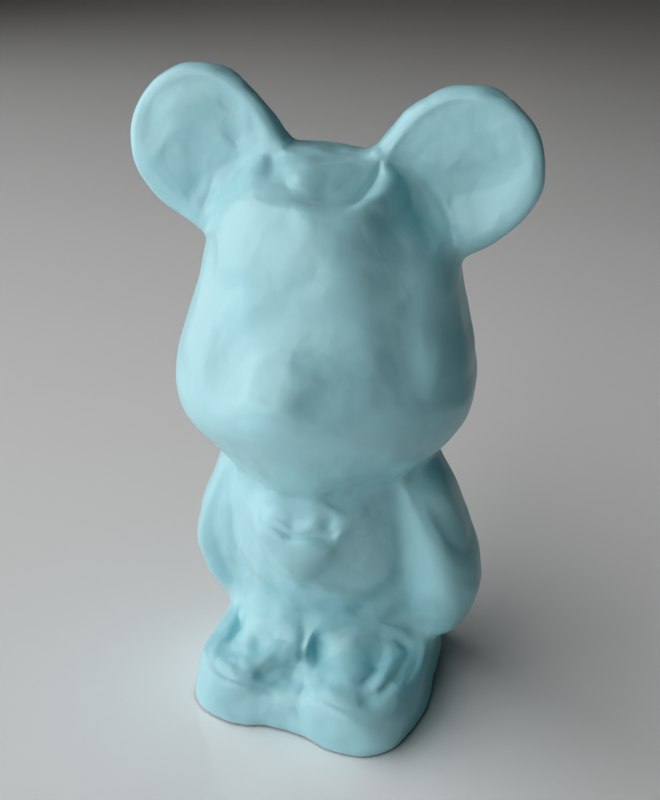}
                \put(-9.0,1.3){\scalebox{0.6}{%
                  \tightcolorbox[1pt]{black!70}{white}{6.8}%
                }}
            \end{minipage}%
            \begin{minipage}{.0832\textwidth}
                \includegraphics[width=\textwidth]{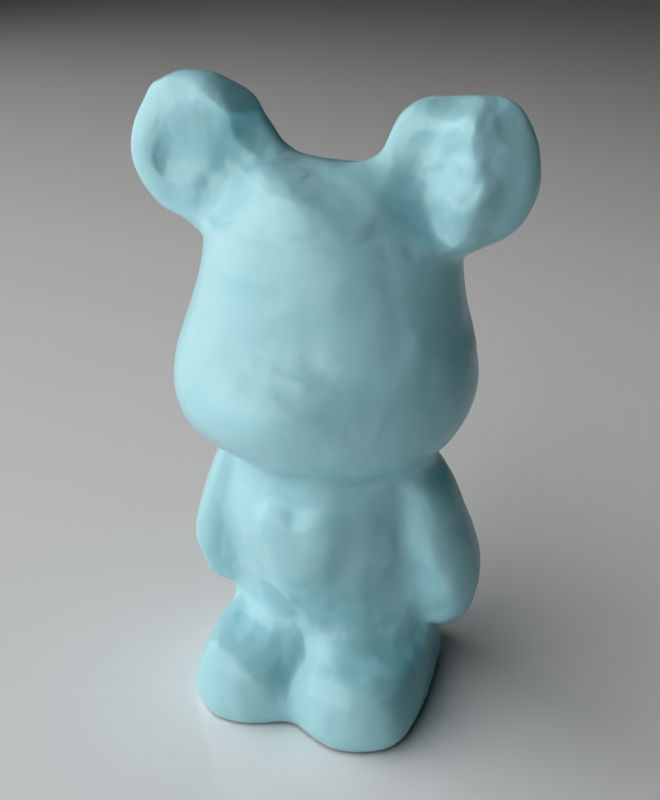}
                \put(-9.0,1.3){\scalebox{0.6}{%
                  \tightcolorbox[1pt]{black!70}{white}{5.5}%
                }}
            \end{minipage}%
            \begin{minipage}{.0832\textwidth}
                \includegraphics[width=\textwidth]{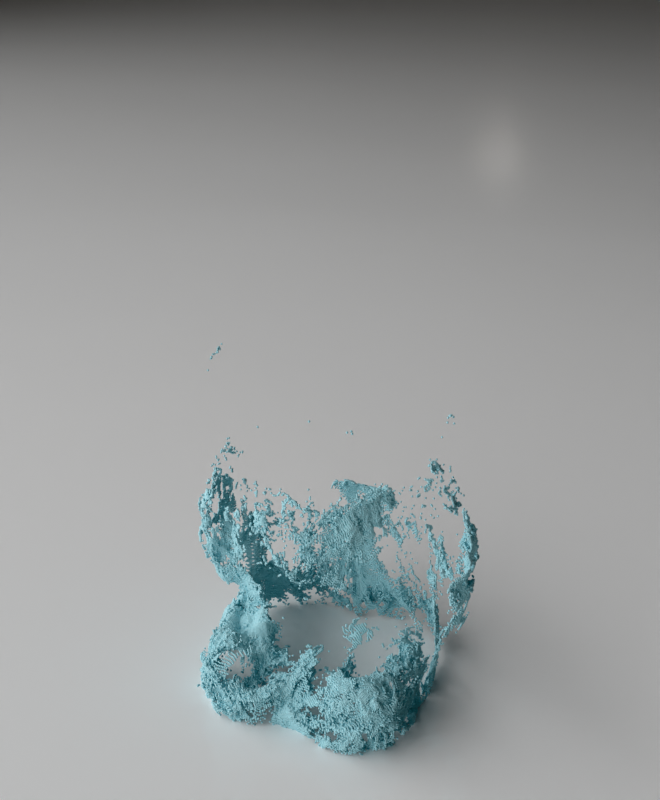}
                \put(-17.4,1.3){\scalebox{0.6}{%
                  \tightcolorbox[1pt]{black!70}{white}{1144.8}%
                }}
            \end{minipage}%
            \begin{minipage}{.0832\textwidth}
                \includegraphics[width=\textwidth]{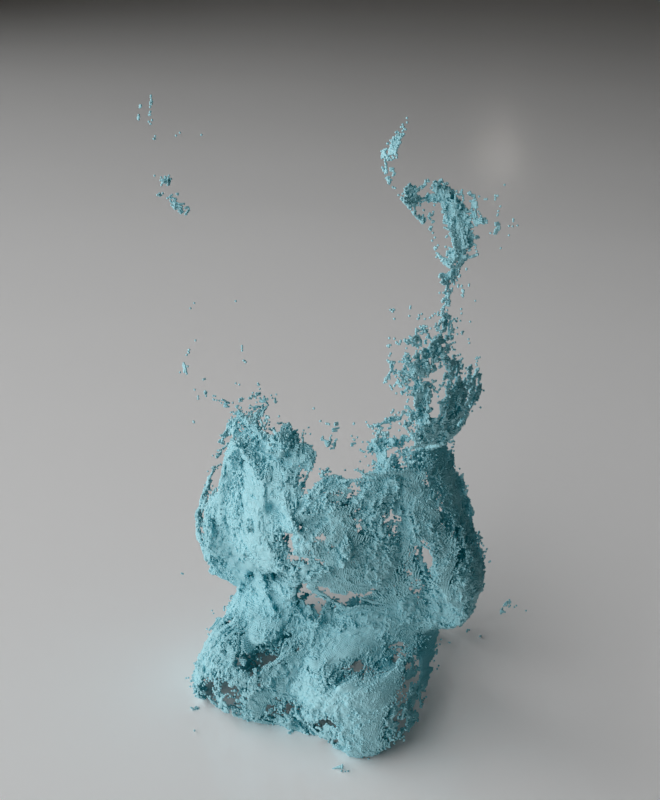}
                \put(-14.5,1.3){\scalebox{0.6}{%
                  \tightcolorbox[1pt]{black!70}{white}{223.2}%
                }}
            \end{minipage}%
            \begin{minipage}{.0832\textwidth}
                \includegraphics[width=\textwidth]{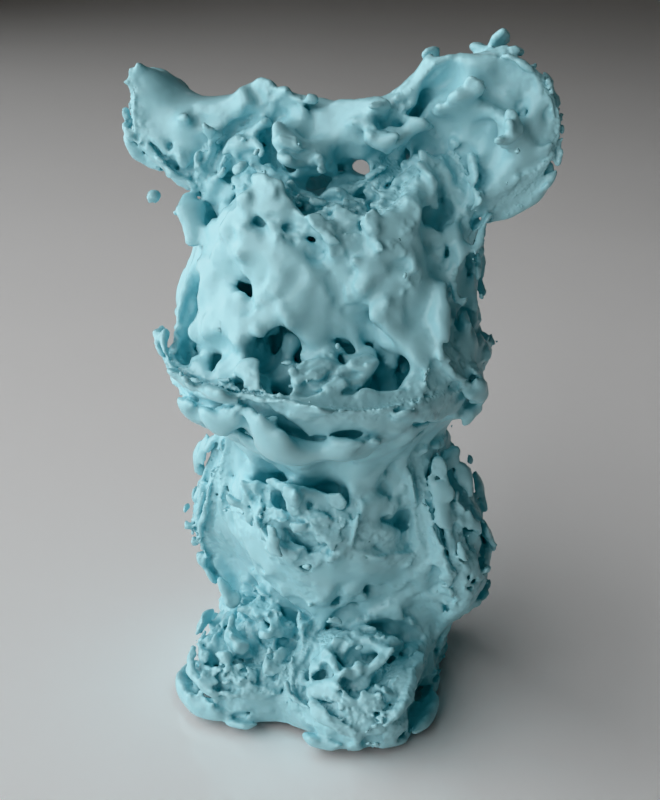}
                \put(-11.7,1.3){\scalebox{0.6}{%
                  \tightcolorbox[1pt]{black!70}{white}{34.4}%
                }}
            \end{minipage}%
        \end{minipage}
    \end{minipage}
    \begin{minipage}{\textwidth}
        \centering
        \begin{minipage}{0.065in}	
            \centering
            \rotatebox{90}{\footnotesize \textsc{Bird}}
        \end{minipage}
        \begin{minipage}{.985\textwidth}
            \centering
            \begin{minipage}{.0832\textwidth}
                \includegraphics[width=\textwidth]{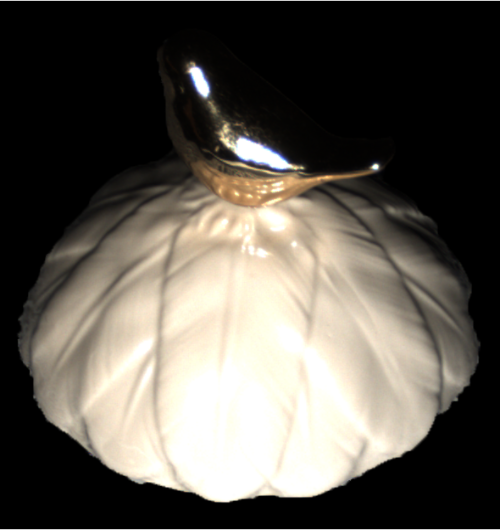}
            \end{minipage}%
            \begin{minipage}{.0832\textwidth}
                \includegraphics[width=\textwidth]{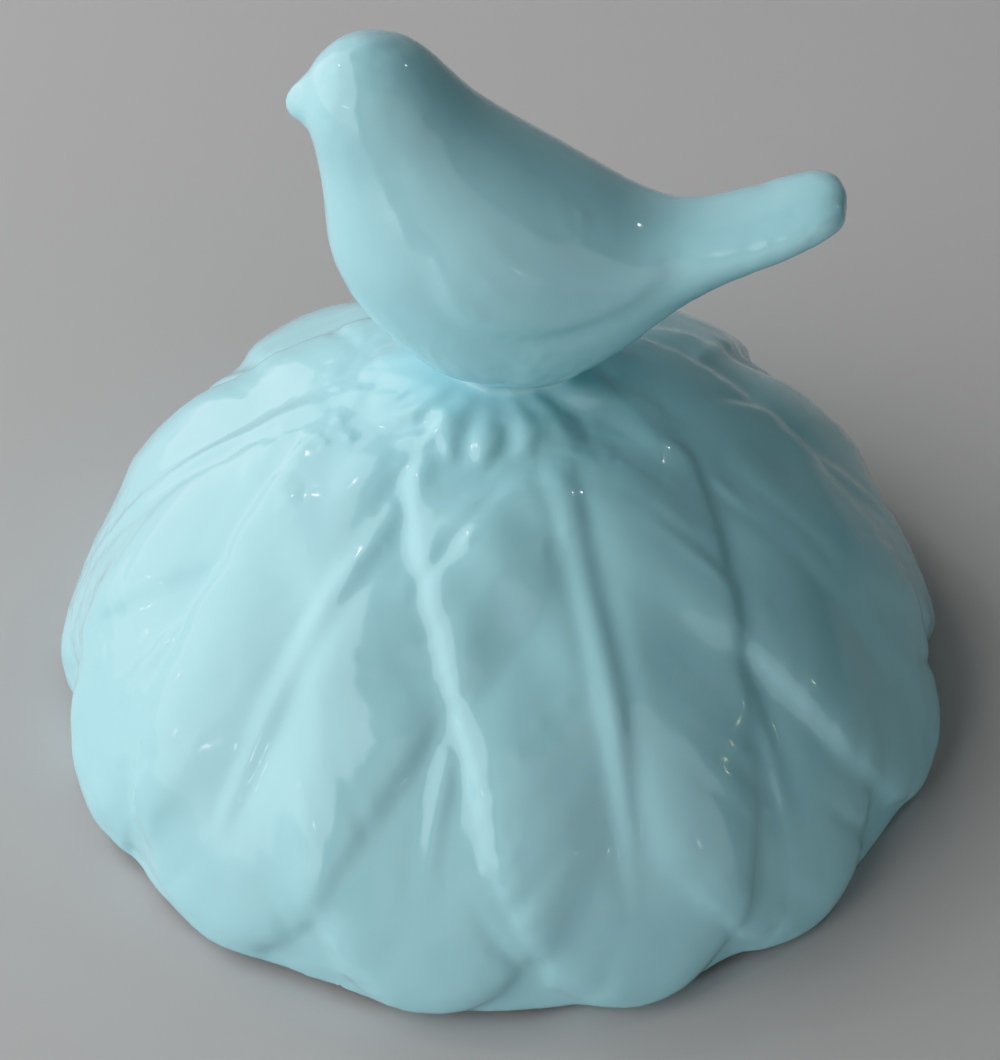}
            \end{minipage}%
            \begin{minipage}{.0832\textwidth}
                \includegraphics[width=\textwidth]{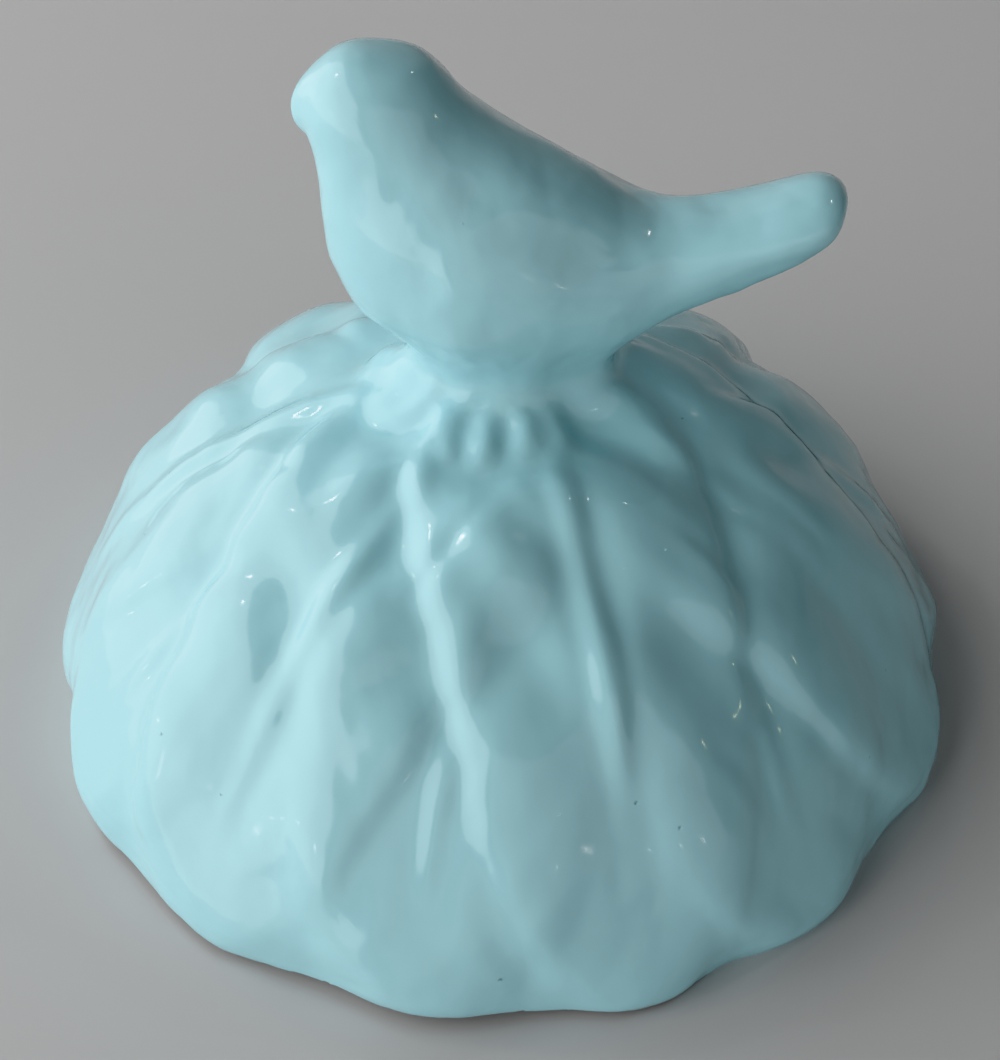}
                \put(-9.0,1.3){\scalebox{0.6}{%
                  \tightcolorbox[1pt]{black!70}{white}{0.3}%
                }}
            \end{minipage}%
            \begin{minipage}{.0832\textwidth}
                \includegraphics[width=\textwidth]{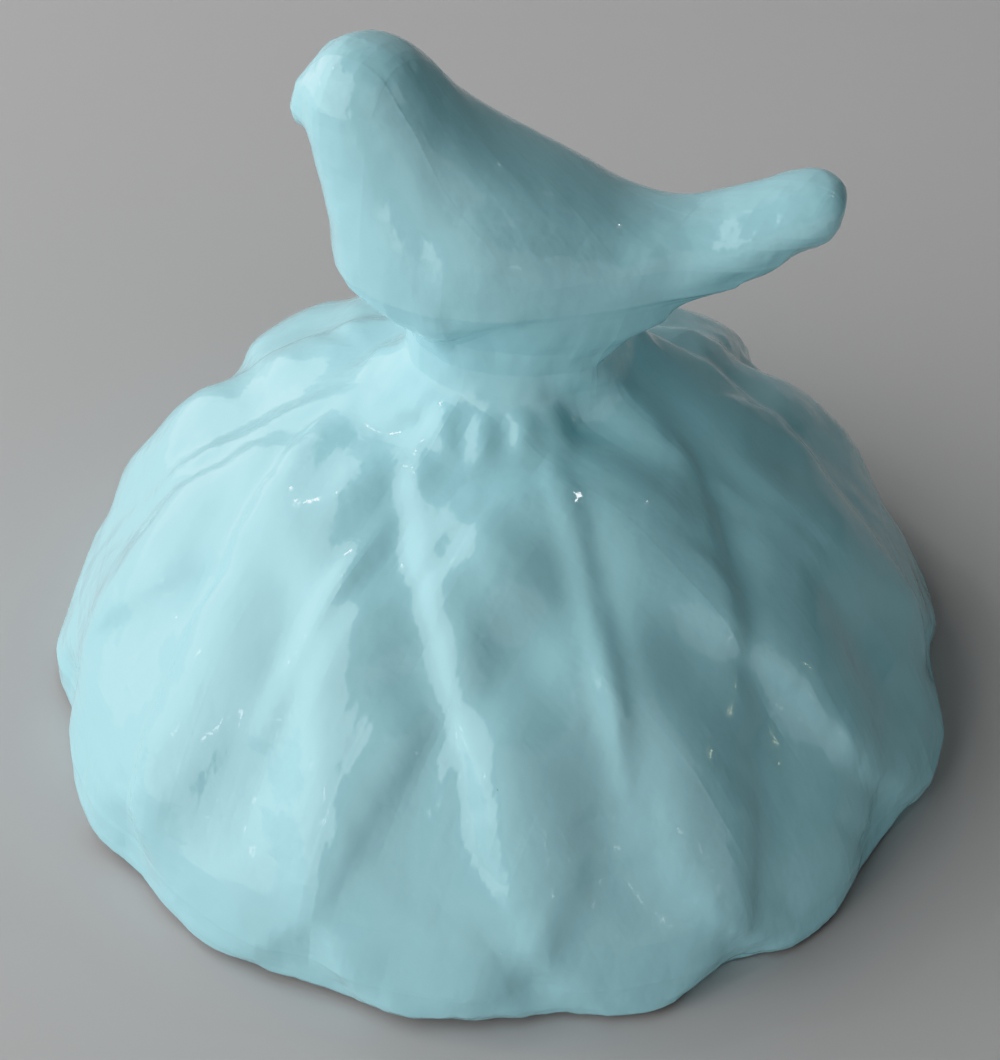}
                \put(-9.0,1.3){\scalebox{0.6}{%
                  \tightcolorbox[1pt]{black!70}{white}{0.5}%
                }}
            \end{minipage}%
            \begin{minipage}{.0832\textwidth}
                \includegraphics[width=\textwidth]{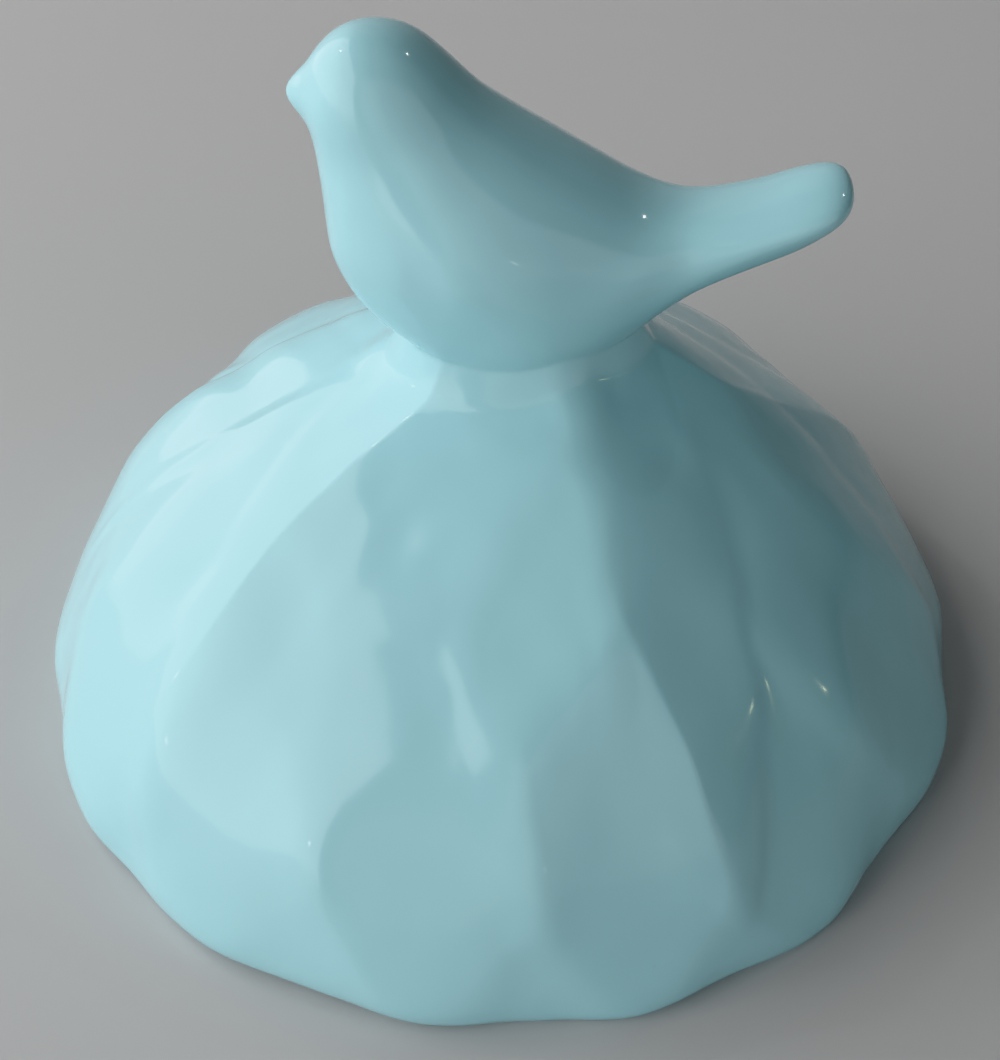}
                \put(-9.0,1.3){\scalebox{0.6}{%
                  \tightcolorbox[1pt]{black!70}{white}{1.8}%
                }}
            \end{minipage}%
            \begin{minipage}{.0832\textwidth}
                \includegraphics[width=\textwidth]{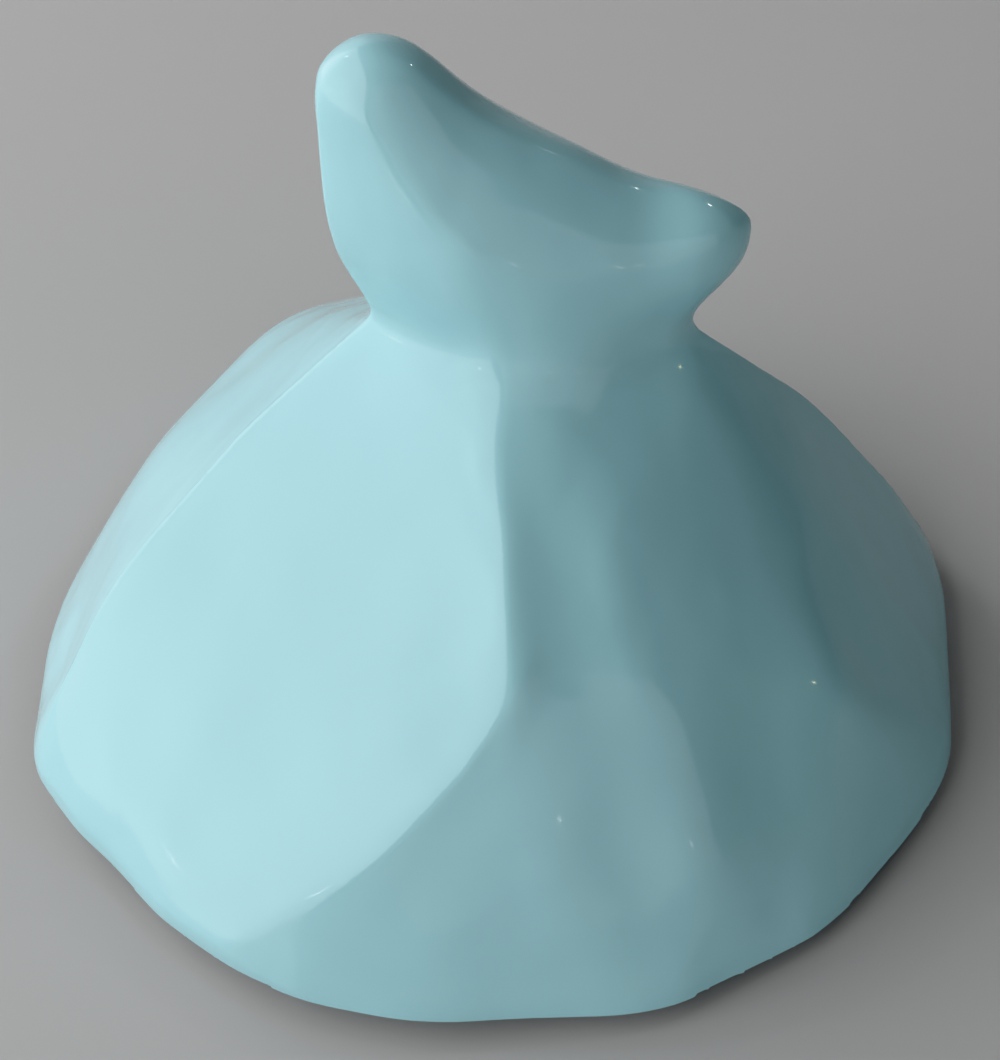}
                \put(-9.0,1.3){\scalebox{0.6}{%
                  \tightcolorbox[1pt]{black!70}{white}{5.2}%
                }}
            \end{minipage}%
            \begin{minipage}{.0832\textwidth}
                \includegraphics[width=\textwidth]{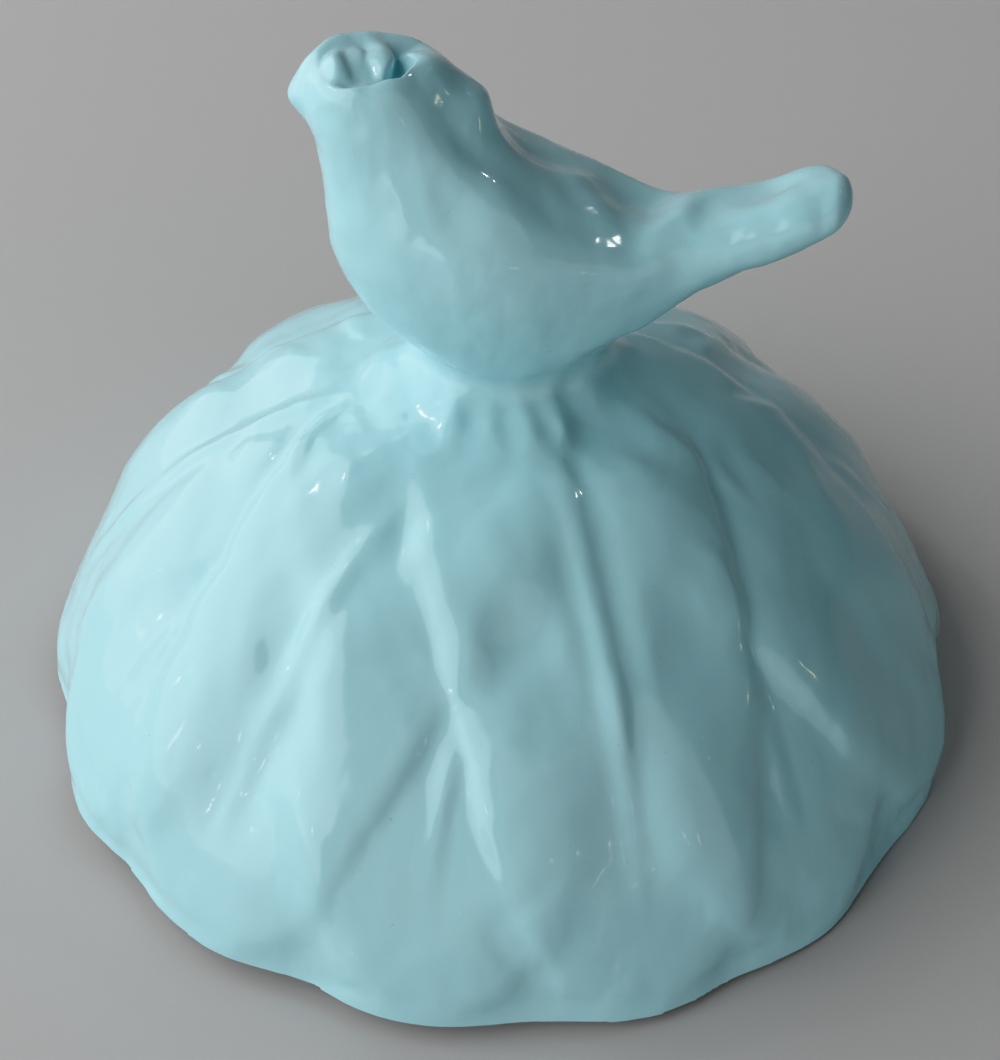}
                \put(-9.0,1.3){\scalebox{0.6}{%
                  \tightcolorbox[1pt]{black!70}{white}{0.7}%
                }}
            \end{minipage}%
            \begin{minipage}{.0832\textwidth}
                \includegraphics[width=\textwidth]{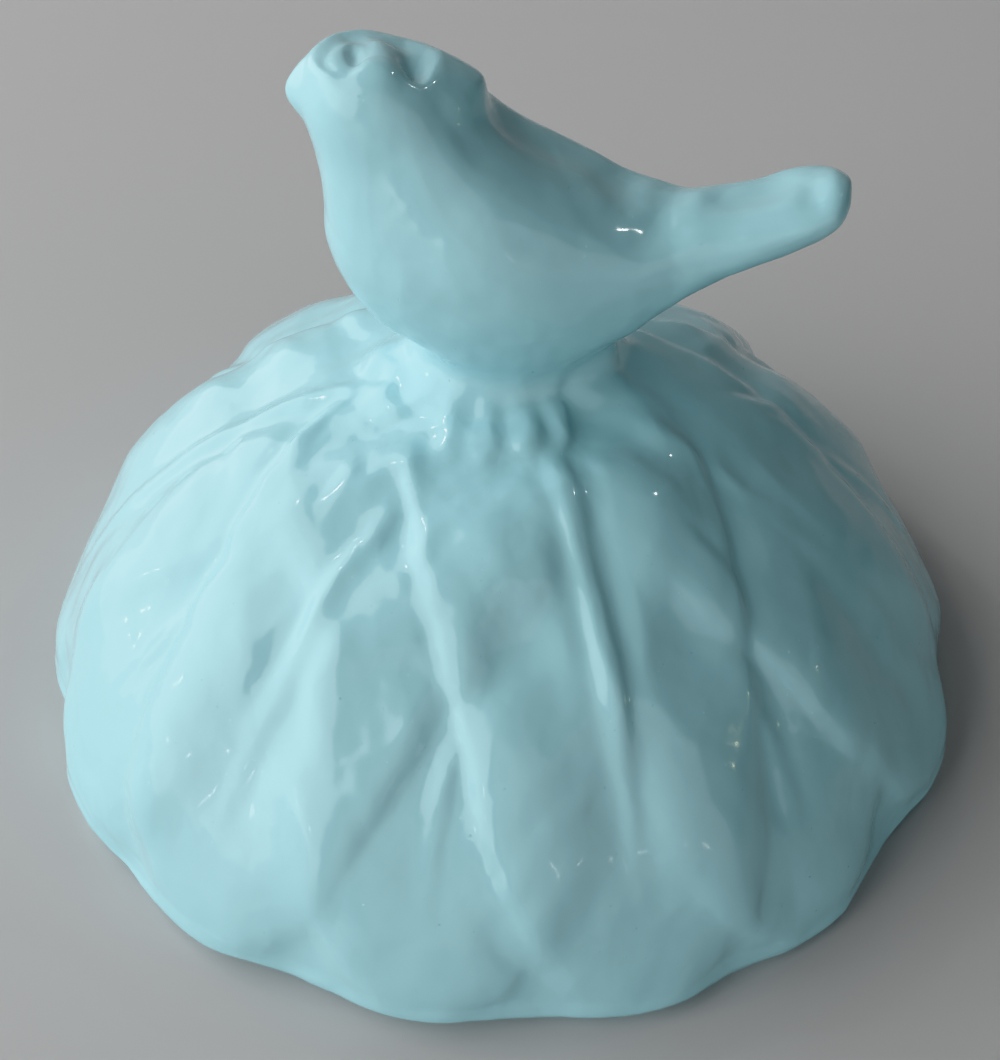}
                \put(-9.0,1.3){\scalebox{0.6}{%
                  \tightcolorbox[1pt]{black!70}{white}{0.8}%
                }}
            \end{minipage}%
            \begin{minipage}{.0832\textwidth}
                \includegraphics[width=\textwidth]{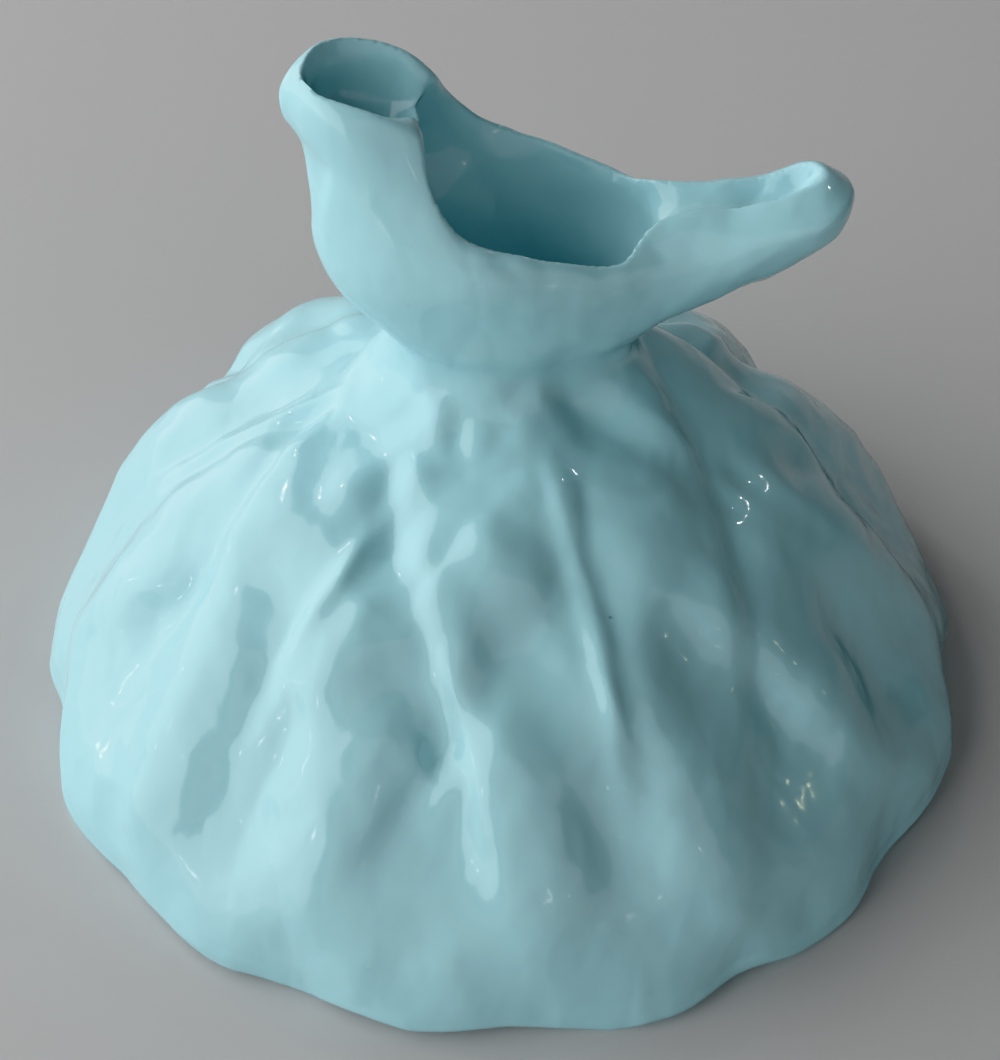}
                \put(-9.0,1.3){\scalebox{0.6}{%
                  \tightcolorbox[1pt]{black!70}{white}{2.9}%
                }}
            \end{minipage}%
            \begin{minipage}{.0832\textwidth}
                \includegraphics[width=\textwidth]{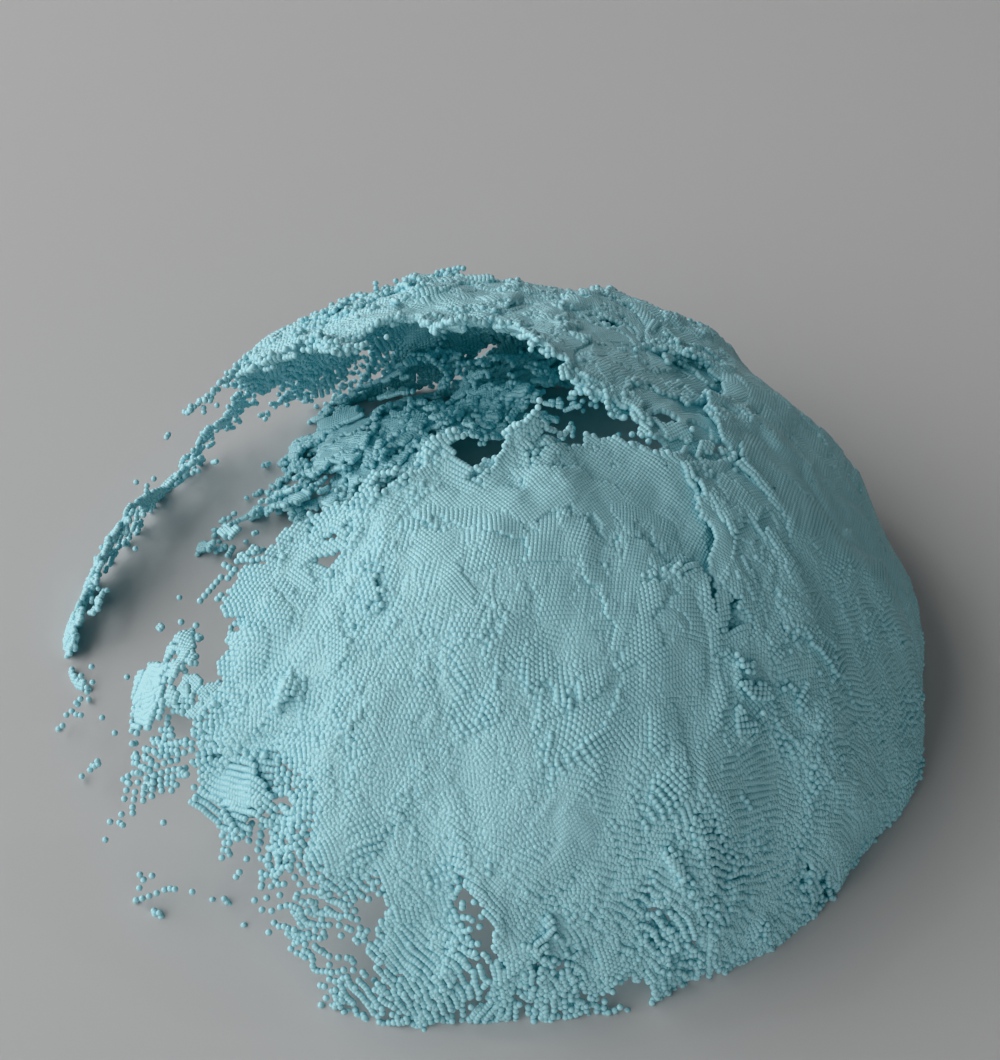}
                \put(-11.7,1.3){\scalebox{0.6}{%
                  \tightcolorbox[1pt]{black!70}{white}{83.6}%
                }}
            \end{minipage}%
            \begin{minipage}{.0832\textwidth}
                \includegraphics[width=\textwidth]{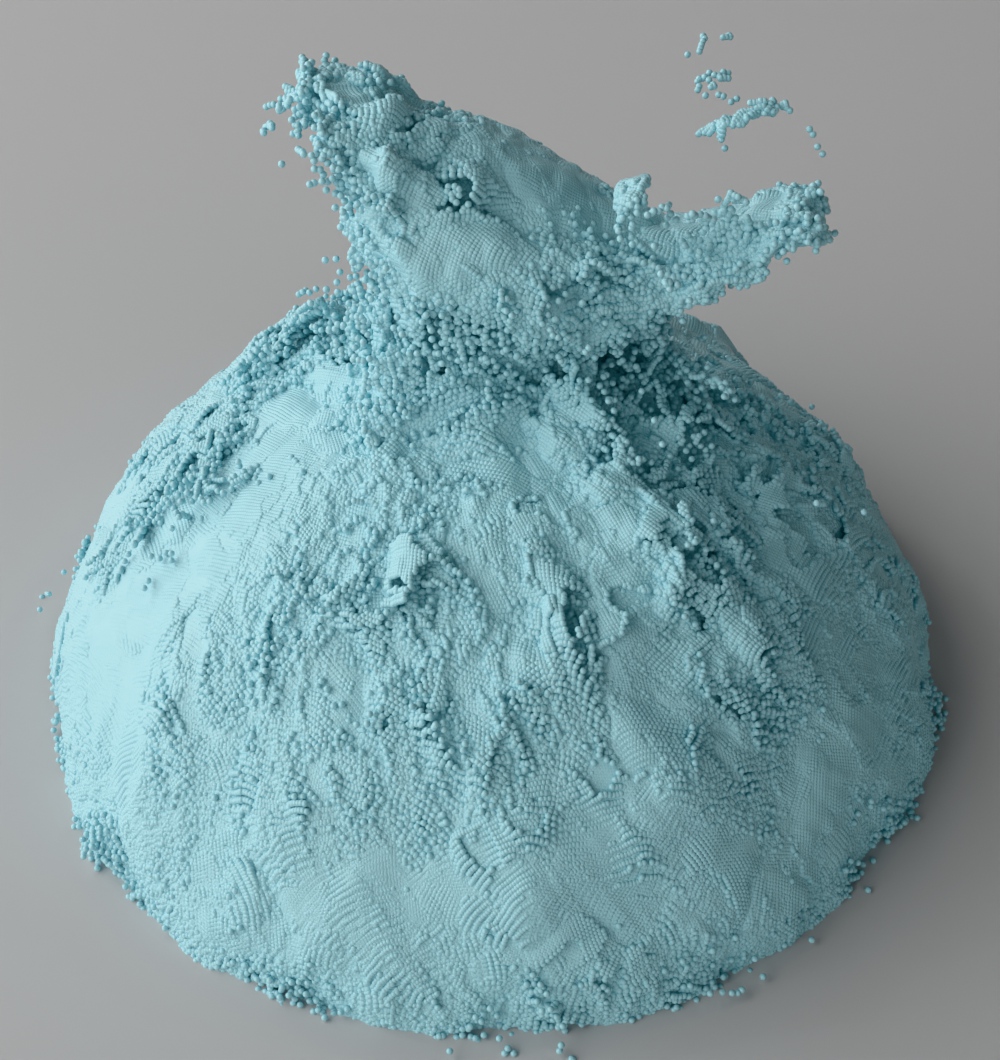}
                \put(-9.0,1.3){\scalebox{0.6}{%
                  \tightcolorbox[1pt]{black!70}{white}{7.3}%
                }}
            \end{minipage}%
            \begin{minipage}{.0832\textwidth}
                \includegraphics[width=\textwidth]{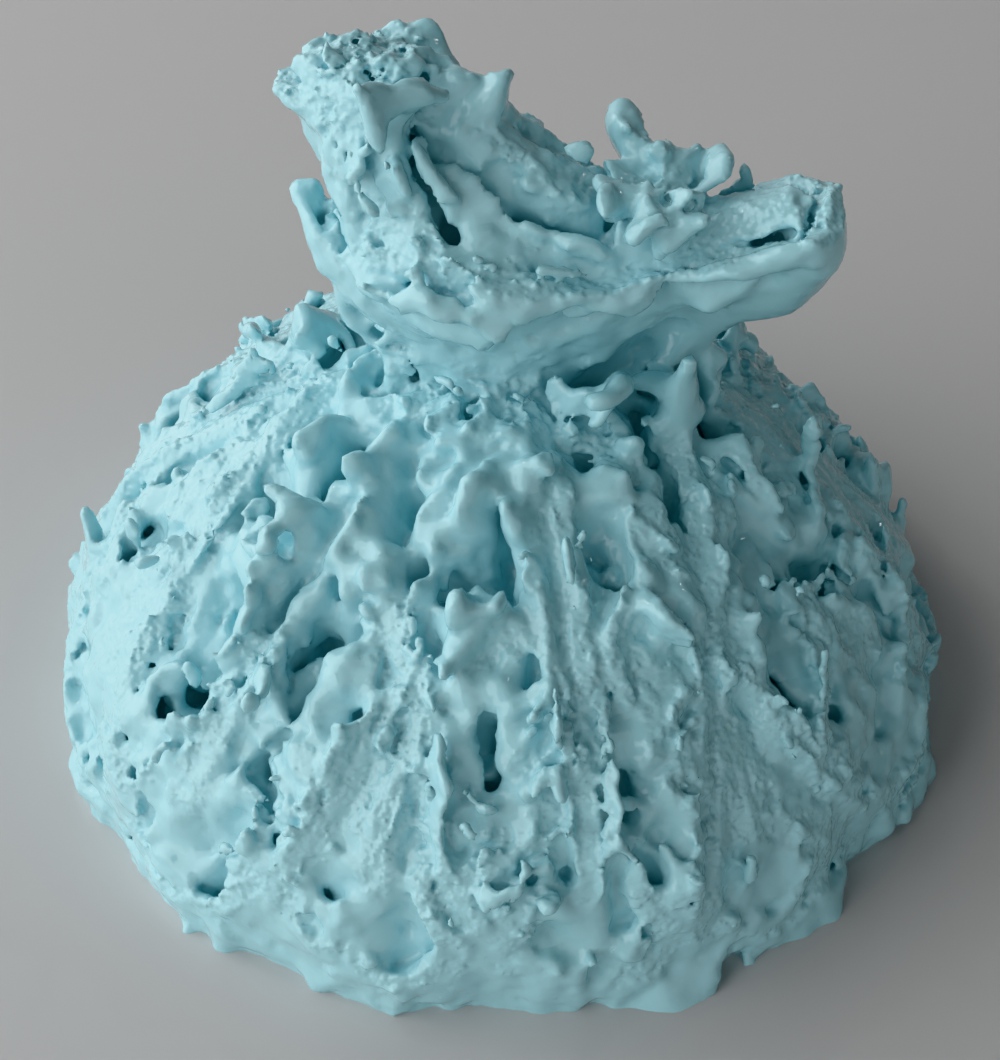}
                \put(-9.0,1.3){\scalebox{0.6}{%
                  \tightcolorbox[1pt]{black!70}{white}{4.0}%
                }}
            \end{minipage}%
        \end{minipage}
    \end{minipage}
    \begin{minipage}{\textwidth}
        \centering
        \begin{minipage}{0.065in}	
            \centering
            \rotatebox{90}{\footnotesize \textsc{Dice}}
        \end{minipage}
        \begin{minipage}{.985\textwidth}
            \centering
            \begin{minipage}{.0832\textwidth}
                \includegraphics[width=\textwidth]{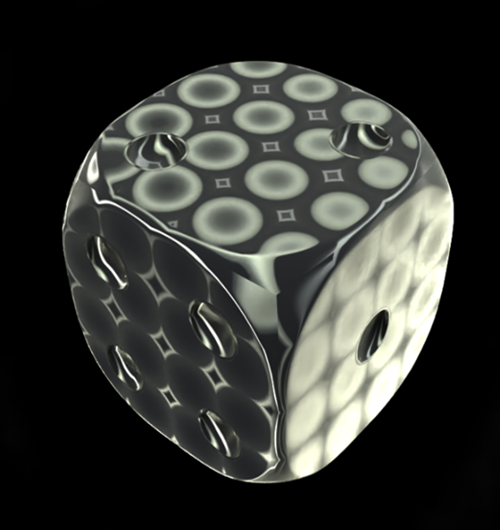}
            \end{minipage}%
            \begin{minipage}{.0832\textwidth}
                \includegraphics[width=\textwidth]{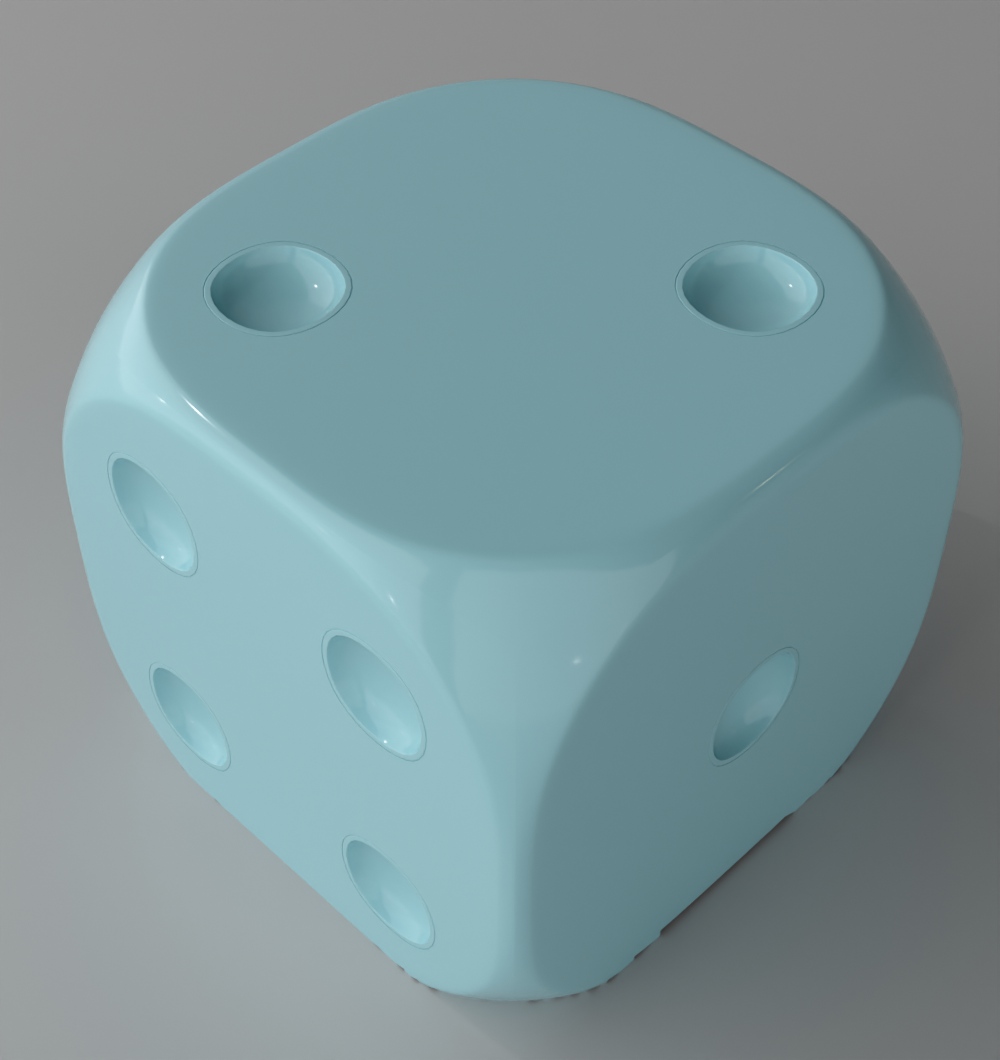}
            \end{minipage}%
            \begin{minipage}{.0832\textwidth}
                \includegraphics[width=\textwidth]{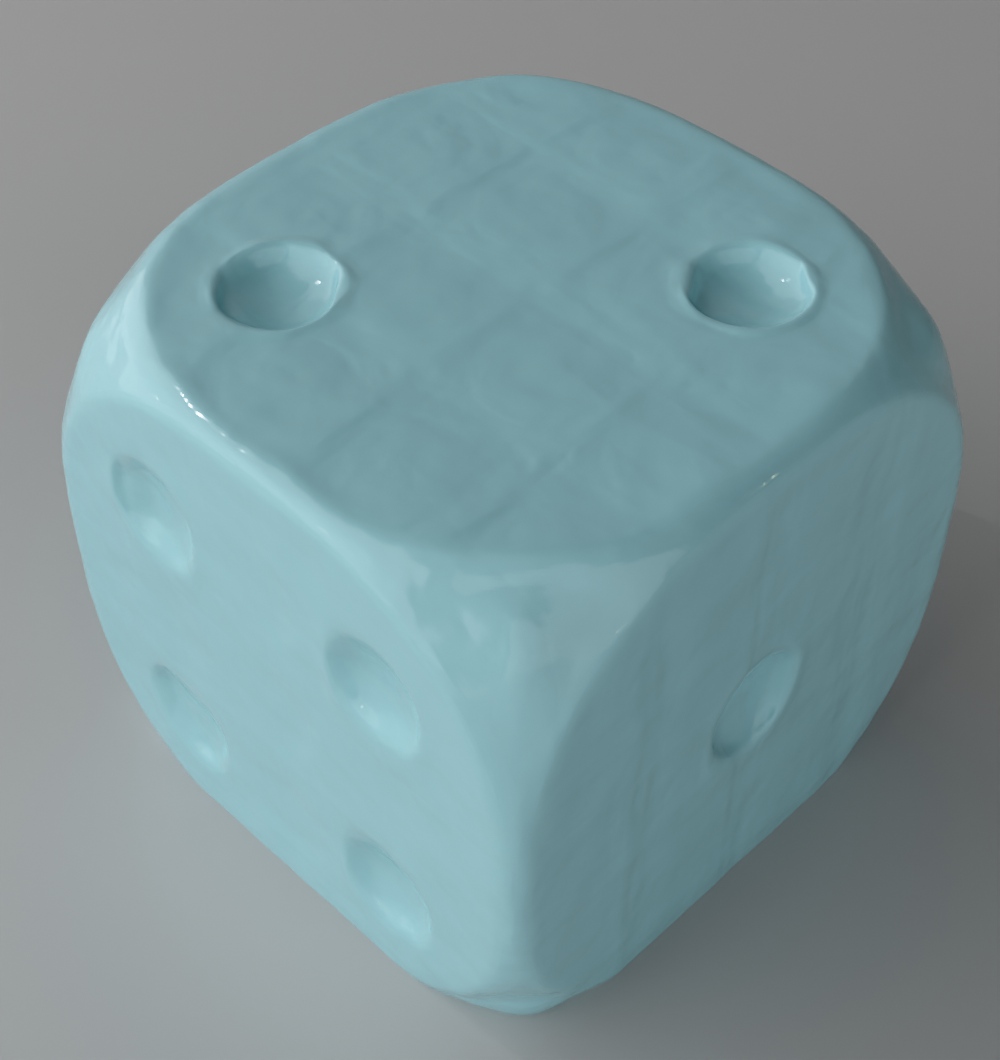}
                \put(-9.0,1.3){\scalebox{0.6}{%
                  \tightcolorbox[1pt]{black!70}{white}{0.9}%
                }}
            \end{minipage}%
            \begin{minipage}{.0832\textwidth}
                \includegraphics[width=\textwidth]{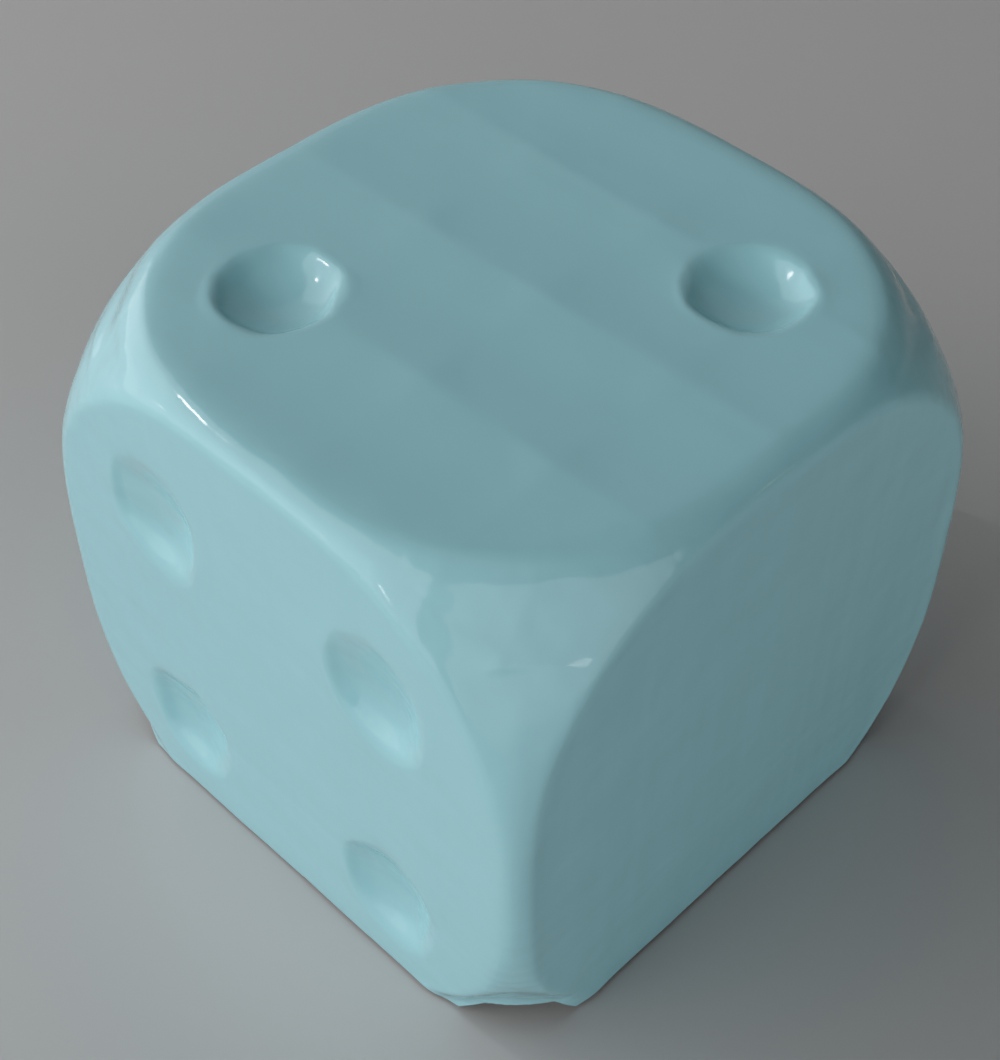}
                \put(-9.0,1.3){\scalebox{0.6}{%
                  \tightcolorbox[1pt]{black!70}{white}{1.0}%
                }}
            \end{minipage}%
            \begin{minipage}{.0832\textwidth}
                \includegraphics[width=\textwidth]{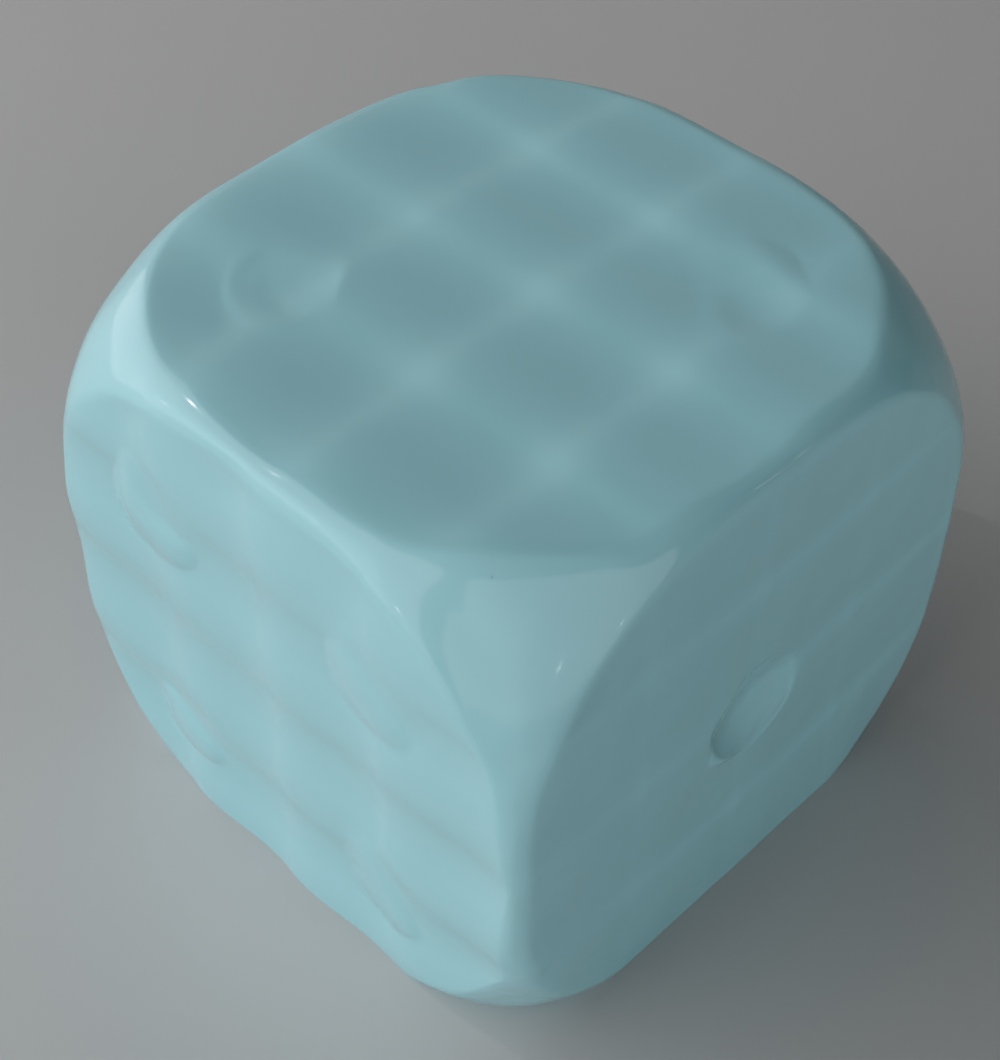}
                \put(-9.0,1.3){\scalebox{0.6}{%
                  \tightcolorbox[1pt]{black!70}{white}{1.5}%
                }}
            \end{minipage}%
            \begin{minipage}{.0832\textwidth}
                \includegraphics[width=\textwidth]{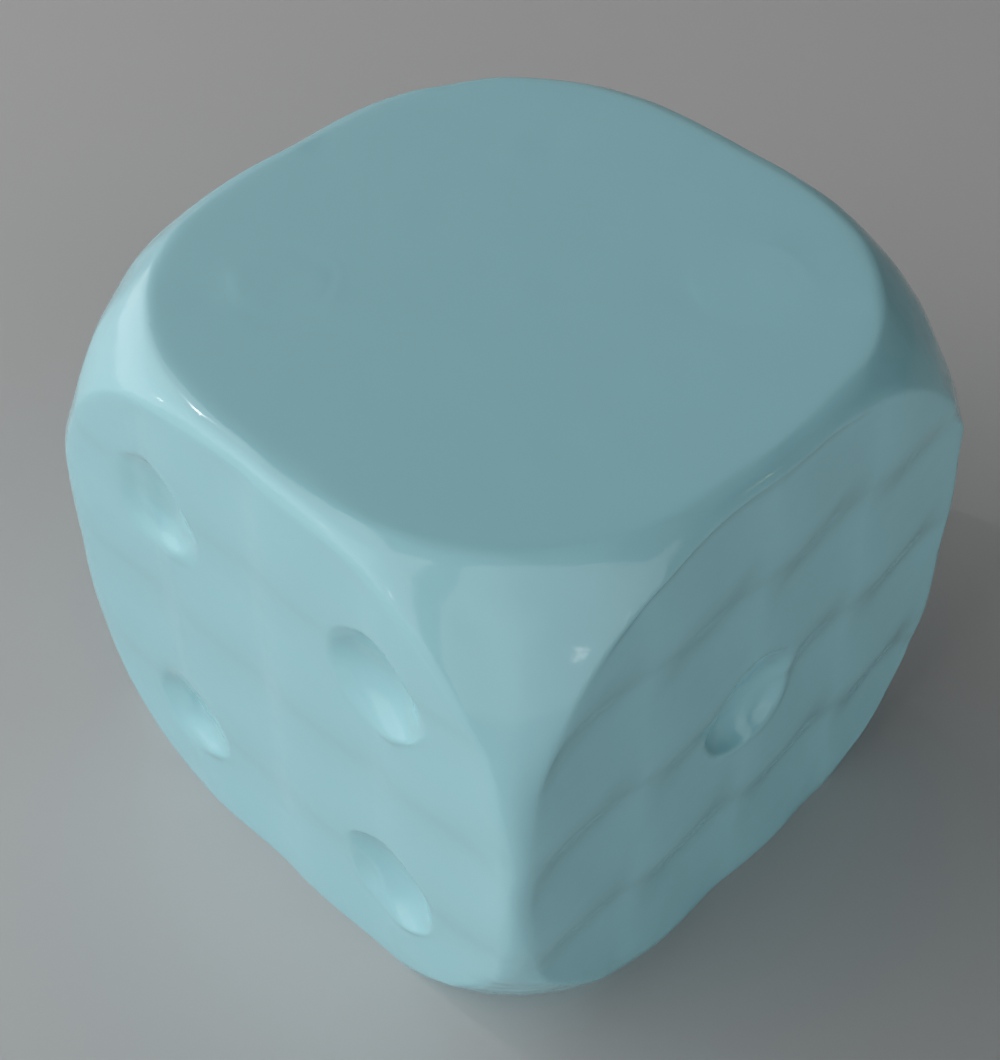}
                \put(-9.0,1.3){\scalebox{0.6}{%
                  \tightcolorbox[1pt]{black!70}{white}{1.4}%
                }}
            \end{minipage}%
            \begin{minipage}{.0832\textwidth}
                \includegraphics[width=\textwidth]{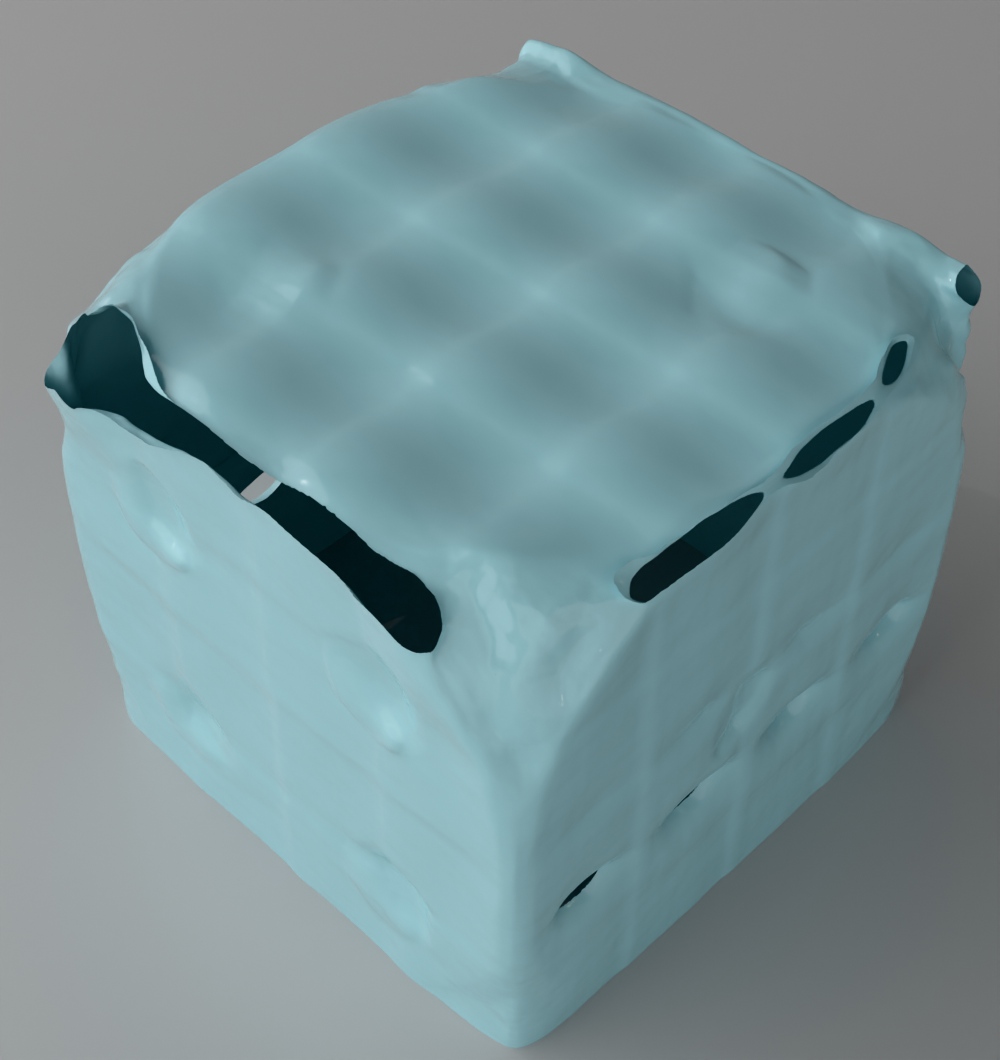}
                \put(-9.0,1.3){\scalebox{0.6}{%
                  \tightcolorbox[1pt]{black!70}{white}{2.4}%
                }}
            \end{minipage}%
            \begin{minipage}{.0832\textwidth}
                \includegraphics[width=\textwidth]{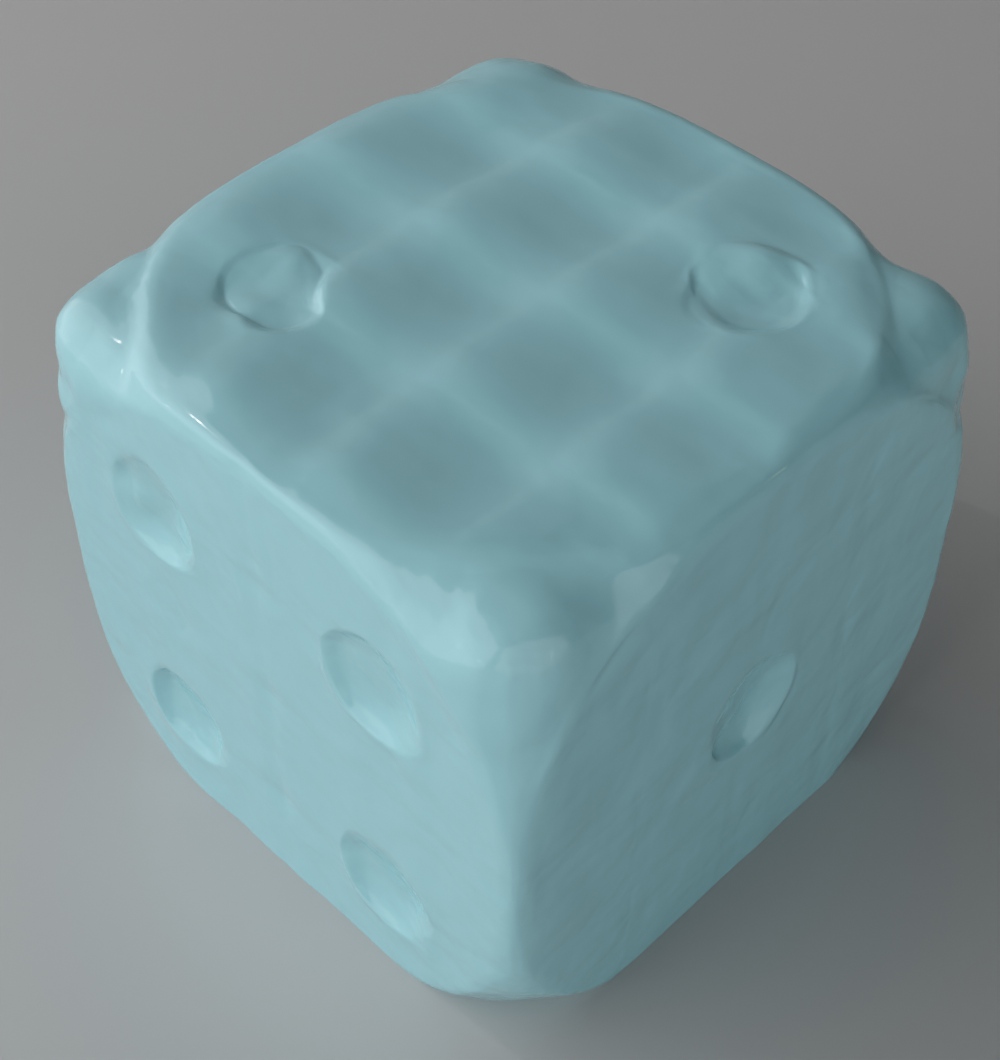}
                \put(-9.0,1.3){\scalebox{0.6}{%
                  \tightcolorbox[1pt]{black!70}{white}{1.4}%
                }}
            \end{minipage}%
            \begin{minipage}{.0832\textwidth}
                \includegraphics[width=\textwidth]{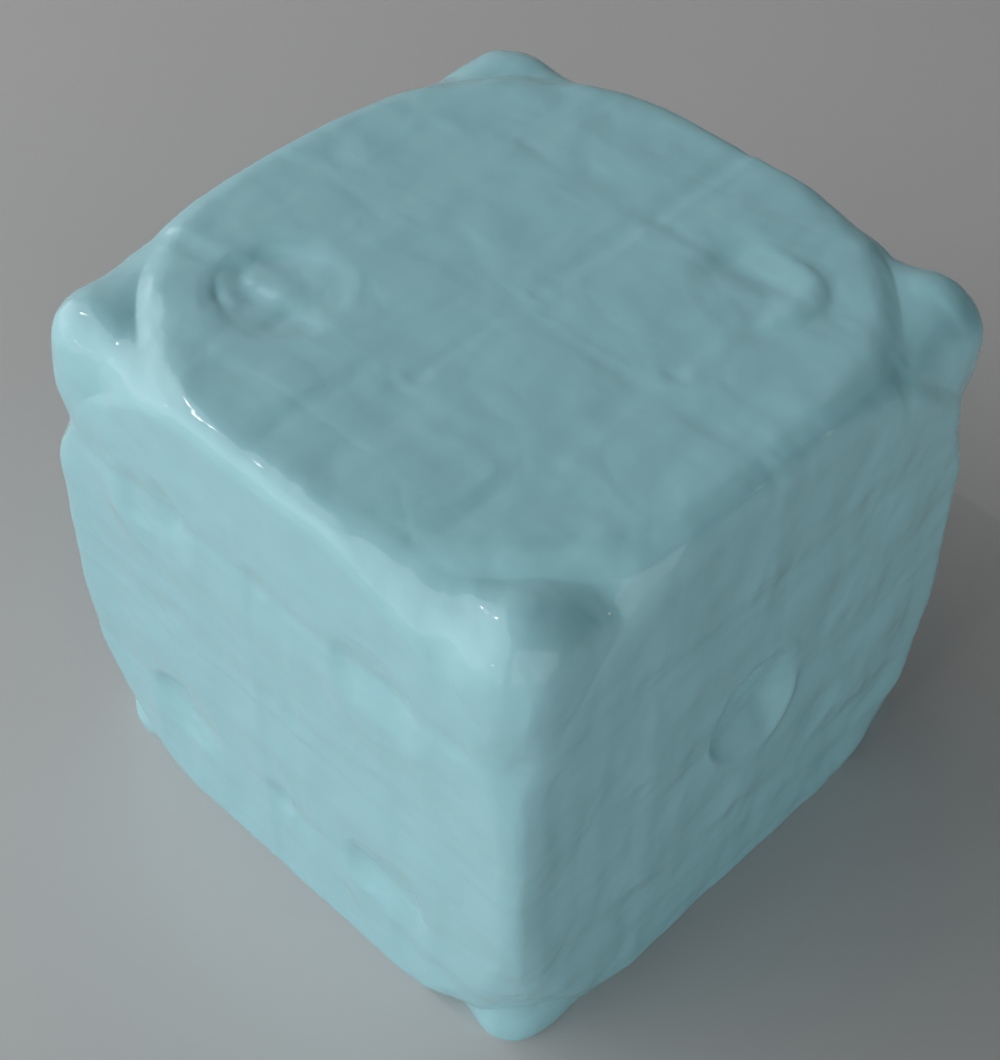}
                \put(-9.0,1.3){\scalebox{0.6}{%
                  \tightcolorbox[1pt]{black!70}{white}{2.3}%
                }}
            \end{minipage}%
            \begin{minipage}{.0832\textwidth}
                \includegraphics[width=\textwidth]{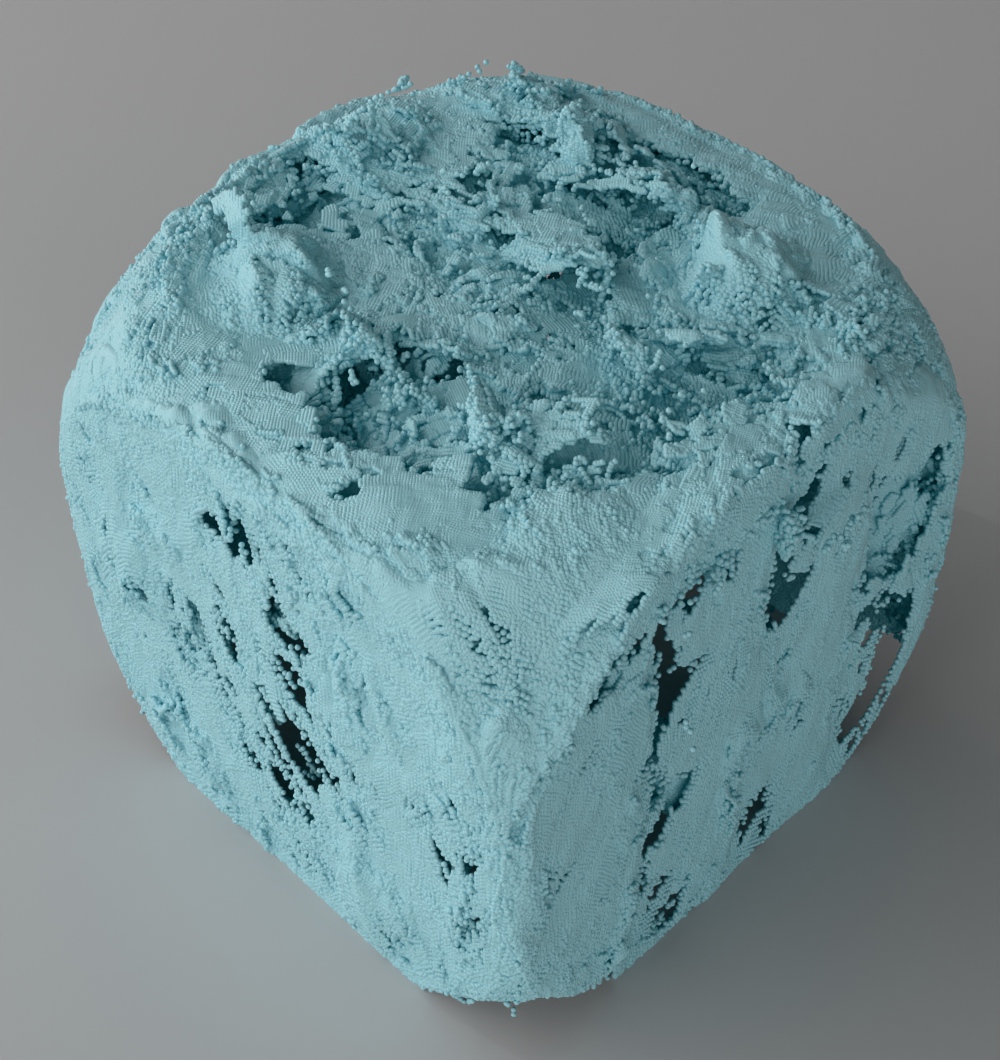}
                \put(-9.0,1.3){\scalebox{0.6}{%
                  \tightcolorbox[1pt]{black!70}{white}{5.5}%
                }}
            \end{minipage}%
            \begin{minipage}{.0832\textwidth}
                \includegraphics[width=\textwidth]{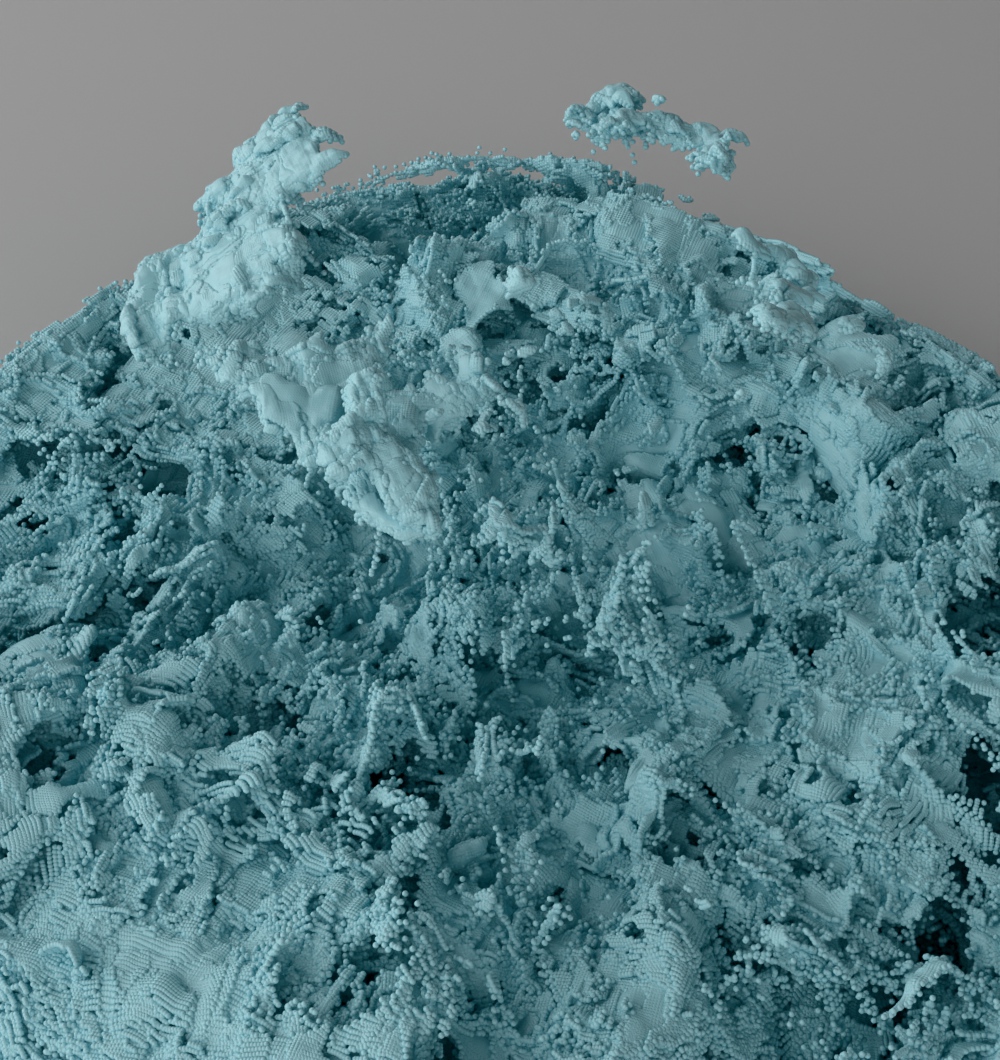}
                \put(-14.5,1.3){\scalebox{0.6}{%
                  \tightcolorbox[1pt]{black!70}{white}{437.3}%
                }}
            \end{minipage}%
            \begin{minipage}{.0832\textwidth}
                \includegraphics[width=\textwidth]{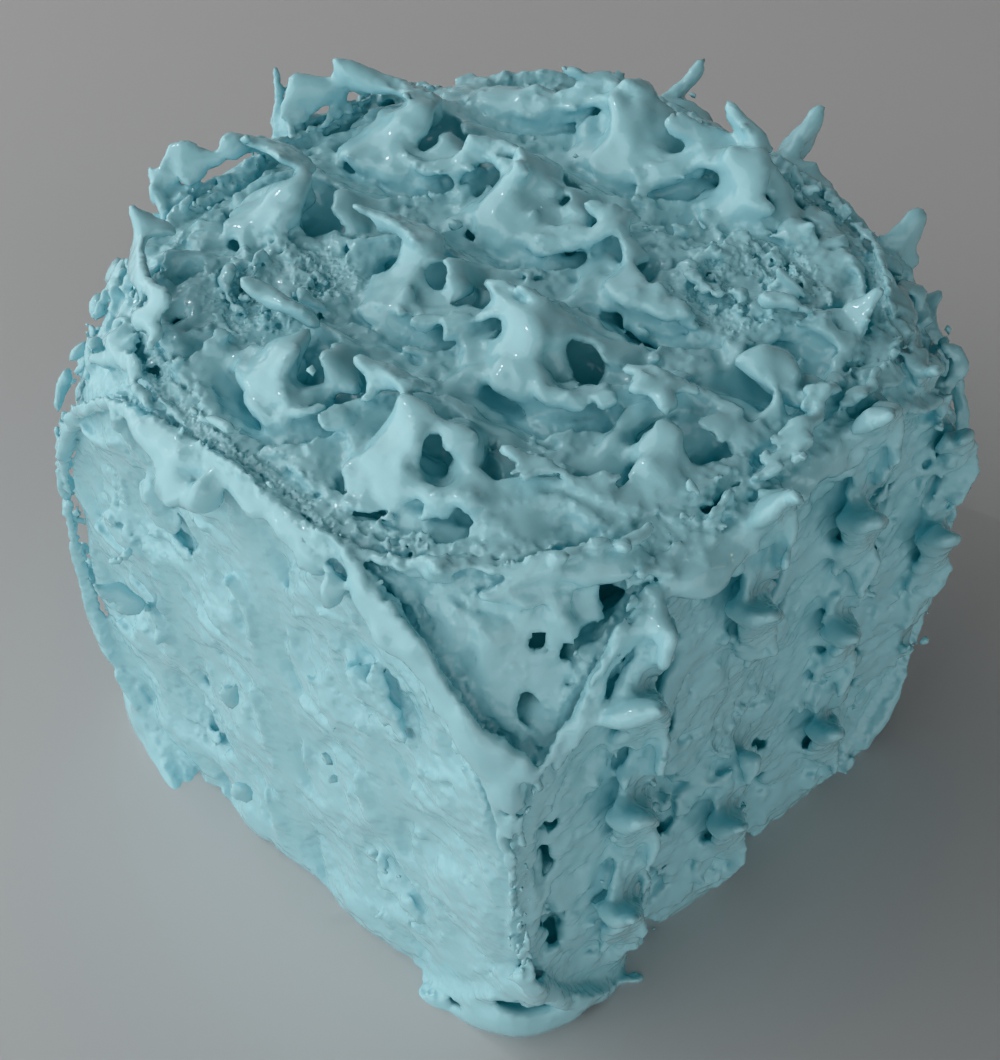}
                \put(-9.0,1.3){\scalebox{0.6}{%
                  \tightcolorbox[1pt]{black!70}{white}{5.8}%
                }}
            \end{minipage}%
        \end{minipage}
    \end{minipage}
    \caption{Comparison with different geometric reconstruction techniques. From the left to right, a photograph, ground-truth geometry, our results with NeuS\cite{wang2021neus}/Neuralangelo\cite{li2023neuralangelo} as the backend, Ref-NeuS~\cite{ge2023ref}, NeRO\cite{liu2023nero}, Neuralangelo\cite{li2023neuralangelo}, NeuS\cite{wang2021neus}, EPFT\cite{Kang_2021_ICCV}, CasMvsNet\cite{gu2020cascade} with 5/1 lighting pattern(s), and COLMAP\cite{schoenberger2016mvs}. Vanilla Neuralangelo, NeuS and COLMAP take input photos captured under an indoor office environment lighting. EPFT is re-trained with a single learnable lighting pattern to adapt to free-form scanning. CasMVSNet is supplied with input photos under 1(corresponding to our center view) or 5 of our learned patterns. \textsc{Bowl, Dog, Bear, Bird} are real objects, and \textsc{Dice} is synthetic. Neuralangelo fails to produce a mesh for \textsc{Dog}. Quantitative errors in Chamfer distance are reported in the bottom right corner of related images.}
    \label{fig:geo_comp}
\end{figure*}

\begin{table}
\caption{Correlation between transformed features and various parameters, averaged over our training dataset. From the 2nd column to the 4th, our features from unnormalized/normalized branch, and the final combined features (\sec{sec:ft}). Higher values indicate highers correlations.}\label{tab:correlation}

\begin{center}
\arrayrulecolor{black}
\begin{tabular}{m{1.8cm}>
{\centering\arraybackslash}m{1.8cm}>{\centering\arraybackslash}m{1.8cm}>{\centering\arraybackslash}m{1.8cm}}

Correlation &
Unnormalized Branch & Normalized Branch & Combined Result \\
\hline

Diffuse &
\cellcolor{corr_l3} \textcolor{black}{0.89} &
\cellcolor{corr_l0} 0.42 &
\cellcolor{corr_l3} \textcolor{black}{0.89}
\\

Specular &
\cellcolor{corr_l2} 0.68 &
\cellcolor{corr_l0} 0.38 &
\cellcolor{corr_l2} 0.68
\\

Roughness &
\cellcolor{corr_l1}0.49 &
\cellcolor{corr_l0}0.41 &
\cellcolor{corr_l1} 0.52
\\

Normal &
\cellcolor{corr_l1} 0.55 &
\cellcolor{corr_l3} \textcolor{black}{0.95} &
\cellcolor{corr_l3} \textcolor{black}{0.94}
\\

Tangent &
\cellcolor{corr_l0} 0.39 &
\cellcolor{corr_l2} 0.62 &
\cellcolor{corr_l2} 0.62
\\

Depth &
\cellcolor{corr_l1}0.46 &
\cellcolor{corr_l1}0.54 &
\cellcolor{corr_l1}0.56
\\

Position &
\cellcolor{corr_l1}0.58 &
\cellcolor{corr_l3}\textcolor{black}{0.83} &
\cellcolor{corr_l3}\textcolor{black}{0.84}
\\

\end{tabular}
\end{center}
\label{tab:correlation}
\end{table}

\begin{figure*}
    \begin{minipage}{\linewidth}
        \centering
        \begin{minipage}{\textwidth}
            \centering
            \begin{minipage}{0.08in}	
                \centering
                \hspace{0.08in}
            \end{minipage}
            \begin{minipage}{.85\textwidth}
                \centering
                \begin{minipage}{.25\textwidth}
                    \centering
                    \begin{minipage}{.5\linewidth}
                        \centering
                        \subcaption*{\small Photo}
                    \end{minipage}%
                    \begin{minipage}{.5\linewidth}
                        \centering
                        \subcaption*{\small G.T. Normal}
                    \end{minipage}
                \end{minipage}%
                \begin{minipage}{.25\textwidth}
                    \centering
                    \subcaption*{\small Ours}
                \end{minipage}%
                \begin{minipage}{.25\textwidth}
                    \centering
                    \subcaption*{\small IRON\cite{iron-2022}}
                \end{minipage}%
                \begin{minipage}{.25\textwidth}
                    \centering
                    \subcaption*{\small NVDIFFREC\cite{munkberg2022extracting}}
                \end{minipage}
            \end{minipage}
        \end{minipage}       
    \end{minipage}

    \begin{minipage}{\linewidth}
        \centering
        \begin{minipage}{0.08in}	
            \centering
            \rotatebox{90}{\small \textsc{Bowl}}
        \end{minipage}
        \begin{minipage}{.85\textwidth}
            \centering
            \begin{minipage}{.12\textwidth}
                \includegraphics[width=\textwidth]{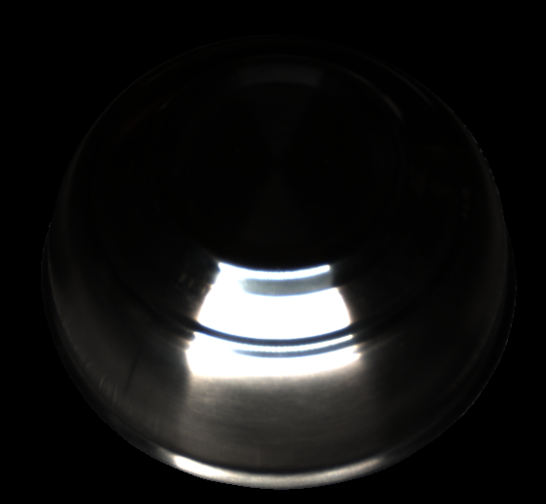}
                \put(-20,3){\scalebox{.8}{\color{white} SSIM}}
            \end{minipage}
            \begin{minipage}{.12\textwidth}
                \includegraphics[width=\textwidth]{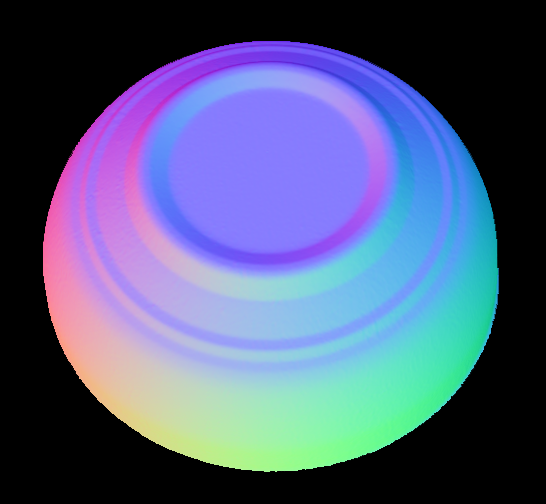}
            \end{minipage}
            \begin{minipage}{.12\textwidth}
                \includegraphics[width=\textwidth]{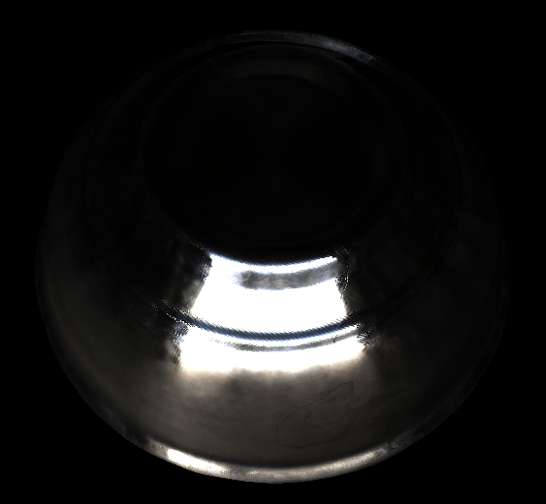}
                \put(-16,3){\scalebox{.8}{\color{white} 0.93}}
            \end{minipage}
            \begin{minipage}{.12\textwidth}
                \includegraphics[width=\textwidth]{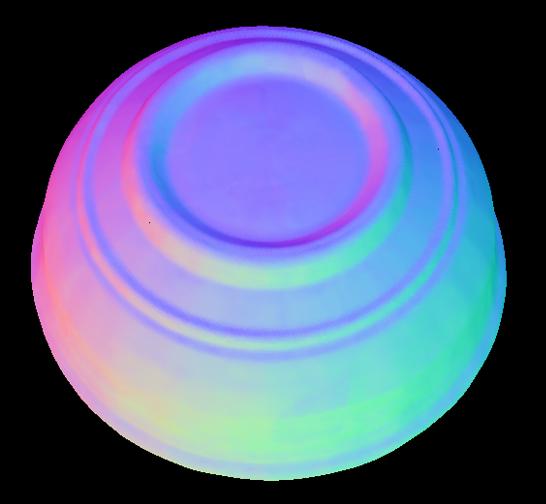}
            \end{minipage}
            \begin{minipage}{.12\textwidth}
                \includegraphics[width=\textwidth]{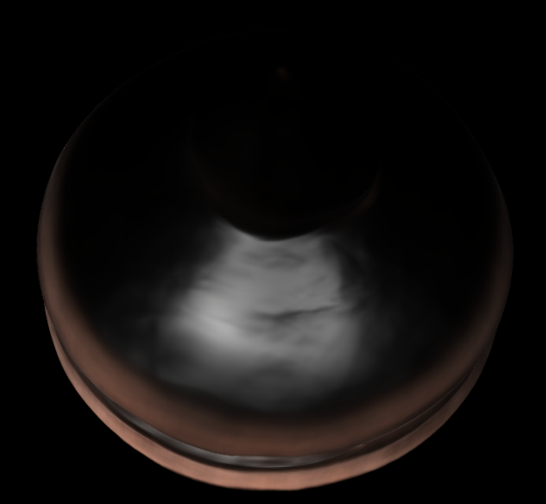}
                \put(-16,3){\scalebox{.8}{\color{white} 0.88}}
            \end{minipage}
            \begin{minipage}{.12\textwidth}
                \includegraphics[width=\textwidth]{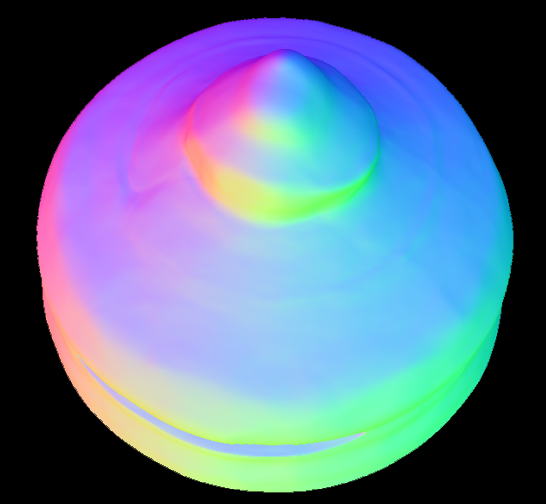}
            \end{minipage}
            \begin{minipage}{.12\textwidth}
                \includegraphics[width=\textwidth]{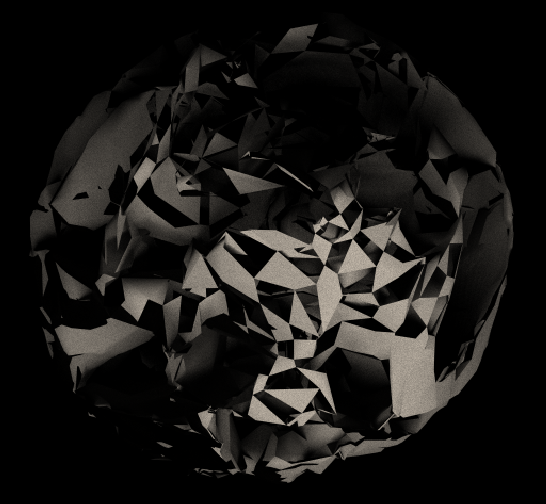}
                \put(-16,3){\scalebox{.8}{\color{white} 0.76}}
            \end{minipage}
            \begin{minipage}{.12\textwidth}
                \includegraphics[width=\textwidth]{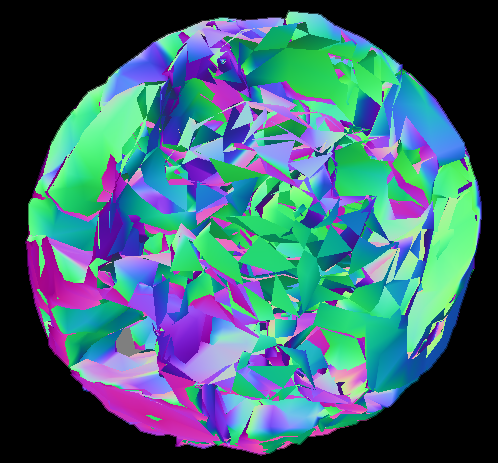}
            \end{minipage}
        \end{minipage}
    \end{minipage}

    \begin{minipage}{\linewidth}
            \centering
        \begin{minipage}{0.08in}	
            \centering
            \rotatebox{90}{\small \textsc{Bear}}
        \end{minipage}
        \begin{minipage}{.85\textwidth}
            \centering
            \begin{minipage}{.12\textwidth}
                \includegraphics[width=\textwidth]{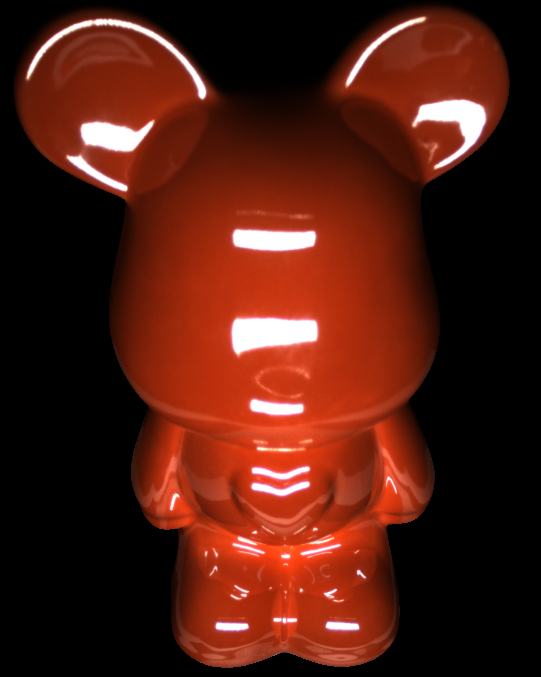}
                \put(-20,3){\scalebox{.8}{\color{white} SSIM}}
            \end{minipage}
            \begin{minipage}{.12\textwidth}
                \includegraphics[width=\textwidth]{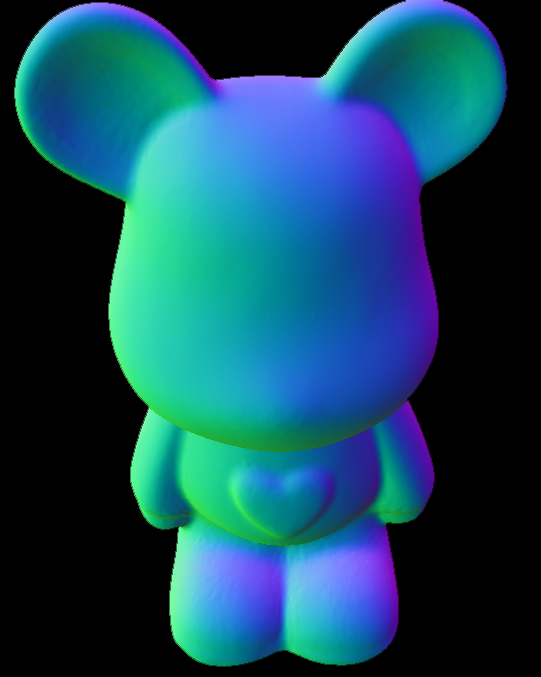}
            \end{minipage}
            \begin{minipage}{.12\textwidth}
                \includegraphics[width=\textwidth]{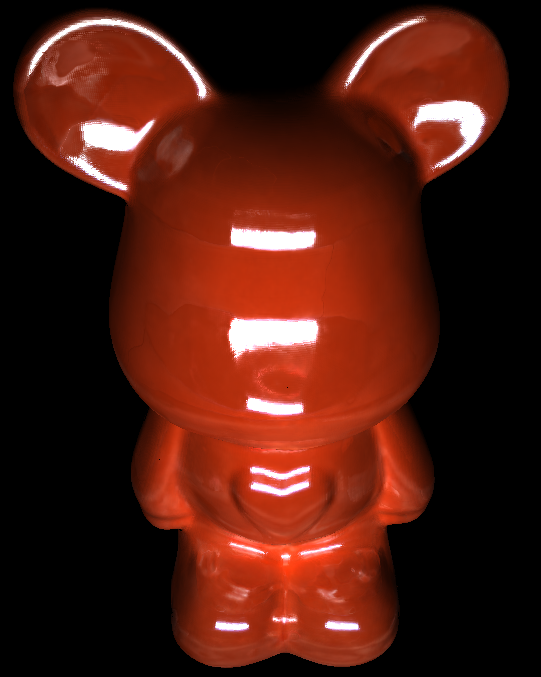}
                \put(-16,3){\scalebox{.8}{\color{white} 0.96}}
            \end{minipage}
            \begin{minipage}{.12\textwidth}
                \includegraphics[width=\textwidth]{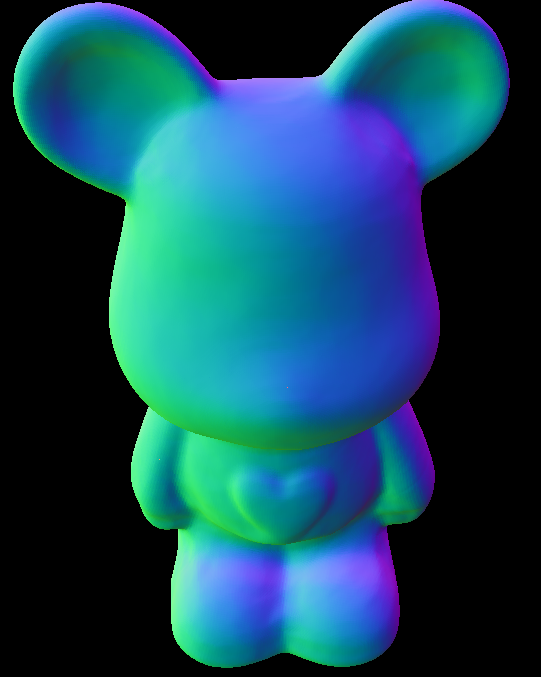}
            \end{minipage}
            \begin{minipage}{.12\textwidth}
                \includegraphics[width=\textwidth]{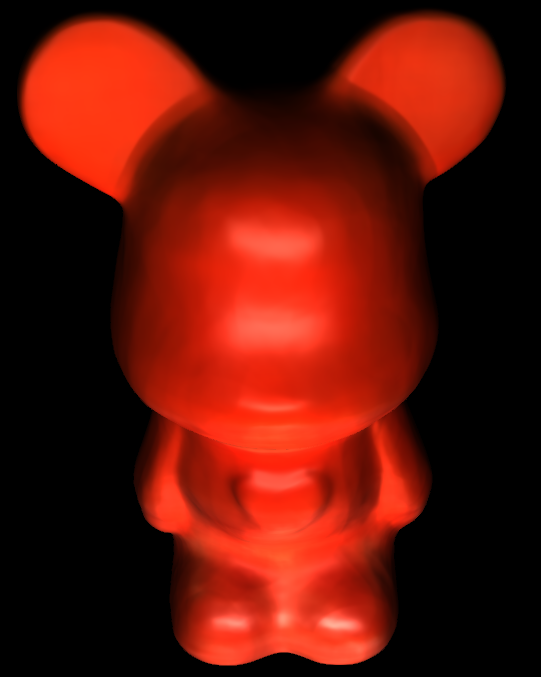}
                \put(-16,3){\scalebox{.8}{\color{white} 0.91}}
            \end{minipage}
            \begin{minipage}{.12\textwidth}
                \includegraphics[width=\textwidth]{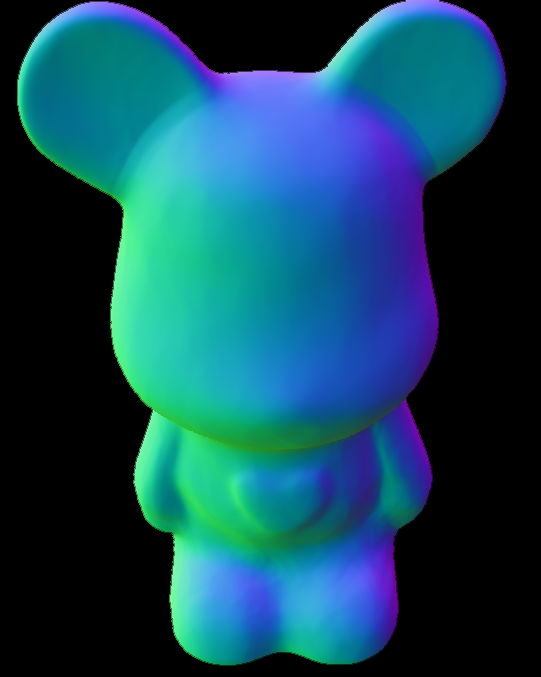}
            \end{minipage}
            \begin{minipage}{.12\textwidth}
                \includegraphics[width=\textwidth]{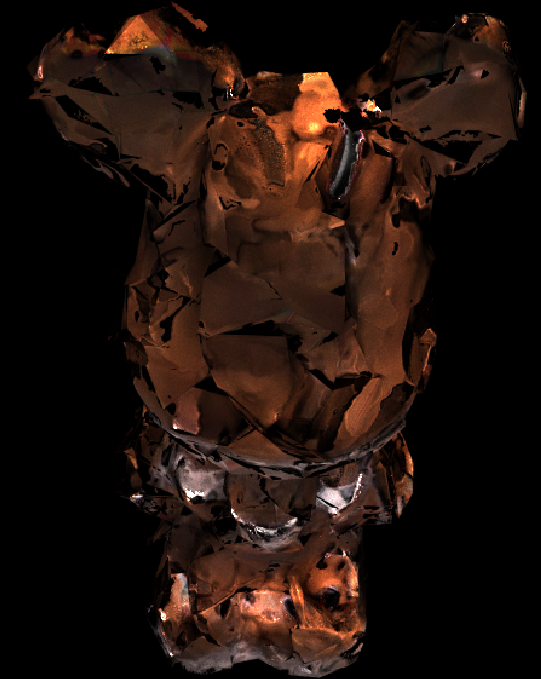}
                \put(-16,3){\scalebox{.8}{\color{white} 0.74}}
            \end{minipage}
            \begin{minipage}{.12\textwidth}
                \includegraphics[width=\textwidth]{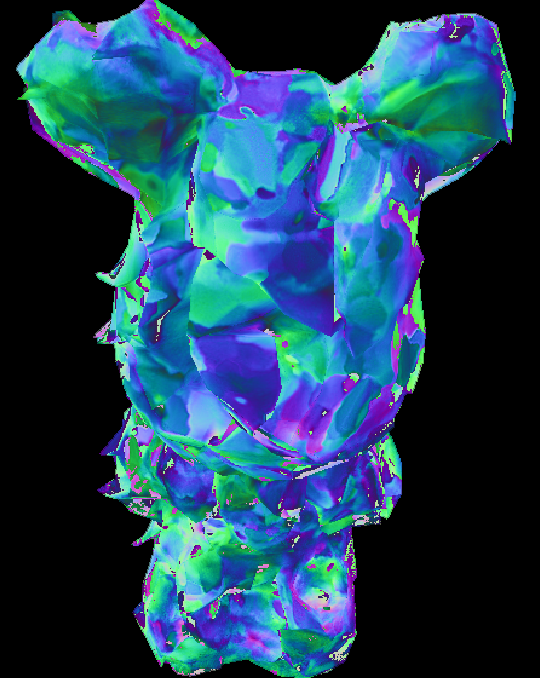}
            \end{minipage}
        \end{minipage}
    \end{minipage}

    \begin{minipage}{\linewidth}
            \centering
        \begin{minipage}{0.08in}	
            \centering
            \rotatebox{90}{\small \textsc{Matball}}
        \end{minipage}
        \begin{minipage}{.85\textwidth}
            \centering
            \begin{minipage}{.12\textwidth}
                \includegraphics[width=\textwidth]{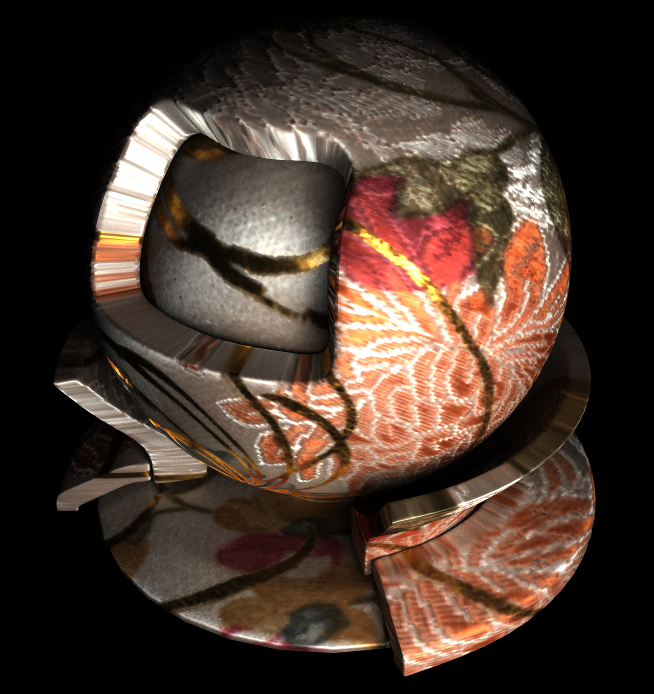}
                \put(-20,3){\scalebox{.8}{\color{white} SSIM}}
            \end{minipage}
            \begin{minipage}{.12\textwidth}
                \includegraphics[width=\textwidth]{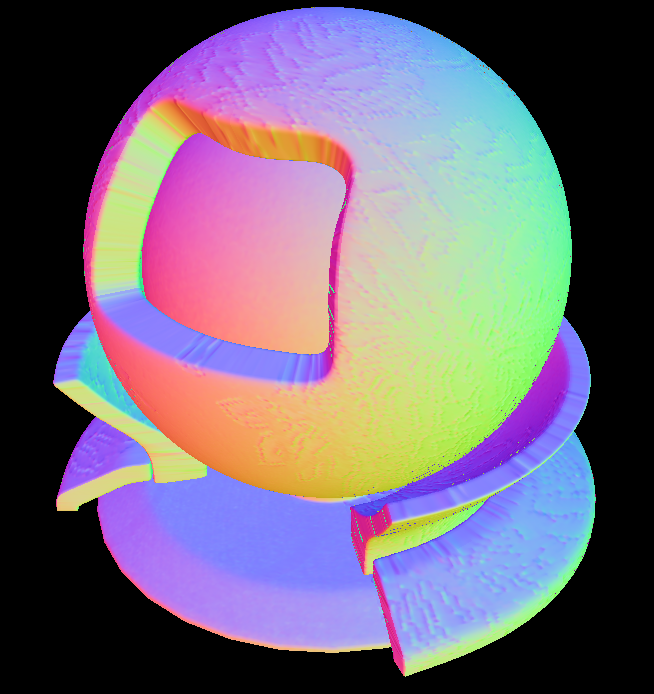}
            \end{minipage}
            \begin{minipage}{.12\textwidth}
                \includegraphics[width=\textwidth]{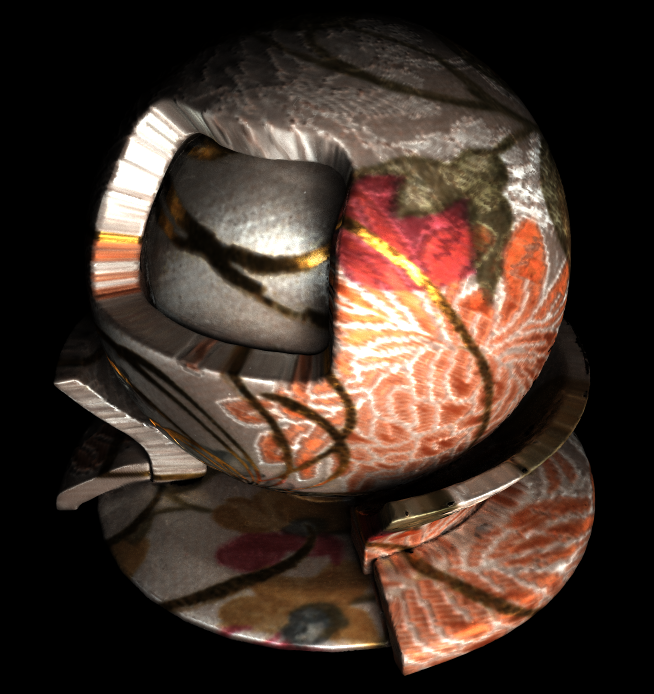}
                \put(-16,3){\scalebox{.8}{\color{white} 0.97}}
            \end{minipage}
            \begin{minipage}{.12\textwidth}
                \includegraphics[width=\textwidth]{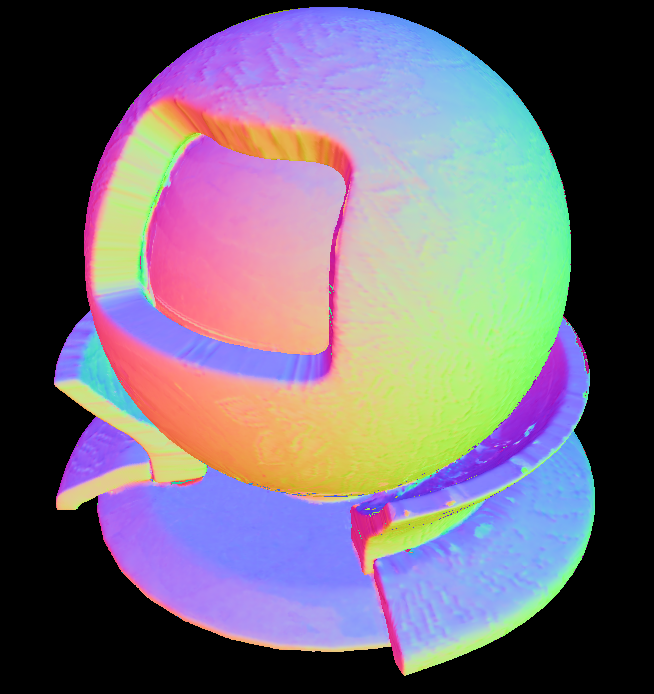}
            \end{minipage}
            \begin{minipage}{.12\textwidth}
                \includegraphics[width=\textwidth]{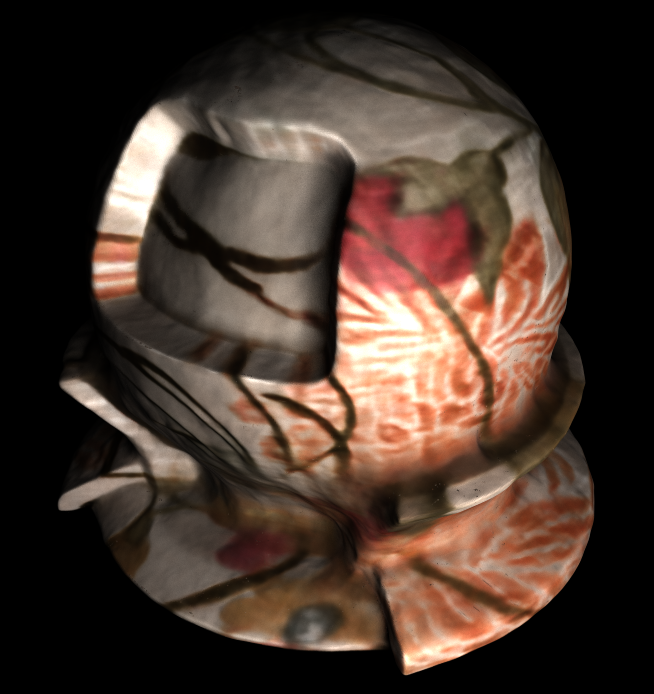}
                \put(-16,3){\scalebox{.8}{\color{white} 0.89}}
            \end{minipage}
            \begin{minipage}{.12\textwidth}
                \includegraphics[width=\textwidth]{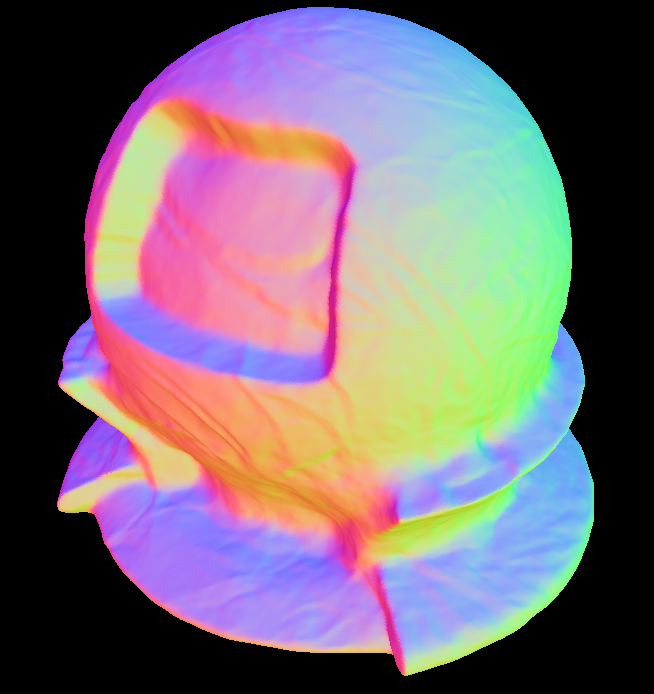}
            \end{minipage}
            \begin{minipage}{.12\textwidth}
                \includegraphics[width=\textwidth]{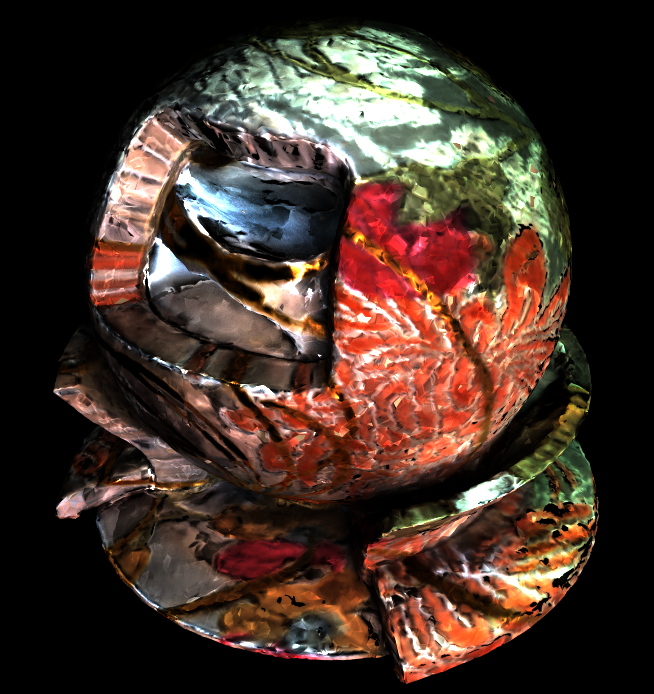}
                \put(-16,3){\scalebox{.8}{\color{white} 0.80}}
            \end{minipage}
            \begin{minipage}{.12\textwidth}
                \includegraphics[width=\textwidth]{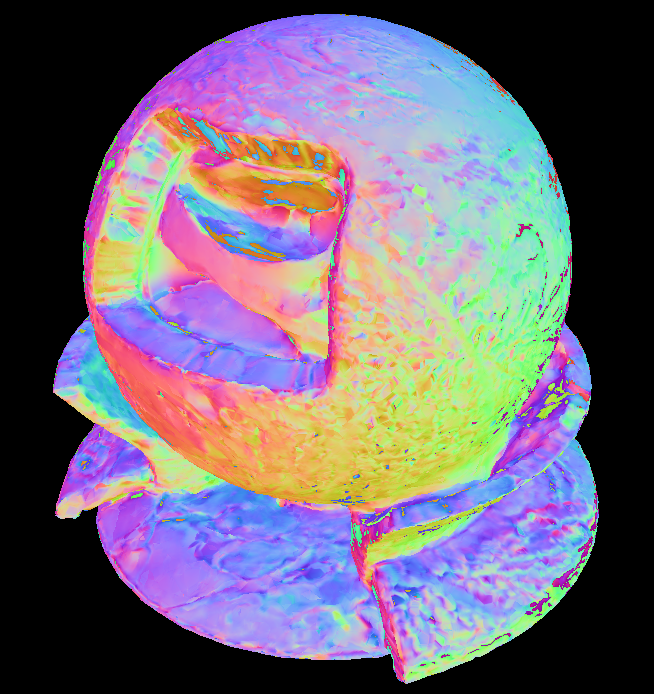}
            \end{minipage}
        \end{minipage}
    \end{minipage}

    \begin{minipage}{\linewidth}
            \centering
        \begin{minipage}{0.08in}	
            \centering
            \rotatebox{90}{\small \textsc{Bird}}
        \end{minipage}
        \begin{minipage}{.85\textwidth}
            \centering
            \begin{minipage}{.12\textwidth}
                \includegraphics[width=\textwidth]{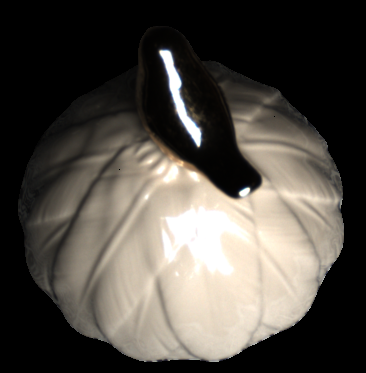}
                \put(-20,3){\scalebox{.8}{\color{white} SSIM}}
            \end{minipage}
            \begin{minipage}{.12\textwidth}
                \includegraphics[width=\textwidth]{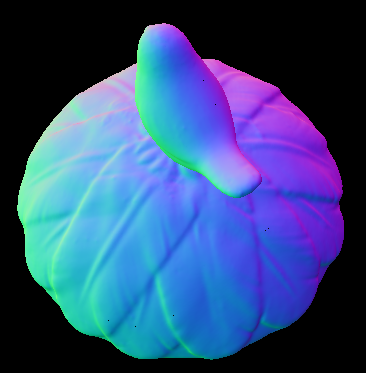}
            \end{minipage}
            \begin{minipage}{.12\textwidth}
                \includegraphics[width=\textwidth]{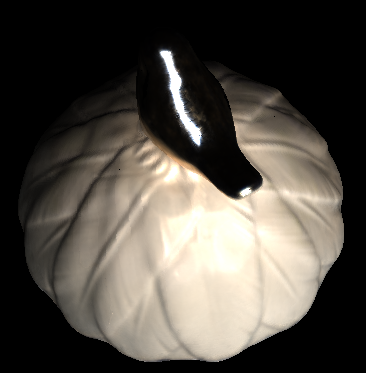}
                \put(-16,3){\scalebox{.8}{\color{white} 0.98}}
            \end{minipage}
            \begin{minipage}{.12\textwidth}
                \includegraphics[width=\textwidth]{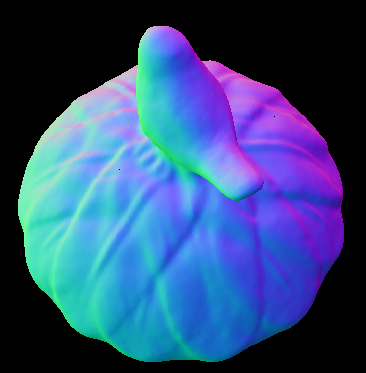}
            \end{minipage}
            \begin{minipage}{.12\textwidth}
                \includegraphics[width=\textwidth]{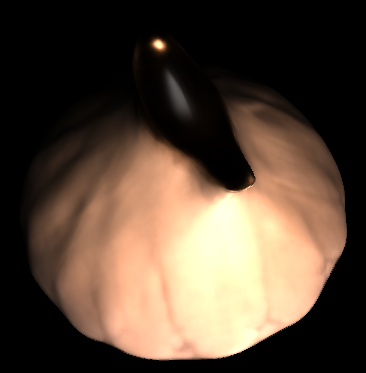}
                \put(-16,3){\scalebox{.8}{\color{white} 0.94}}
            \end{minipage}
            \begin{minipage}{.12\textwidth}
                \includegraphics[width=\textwidth]{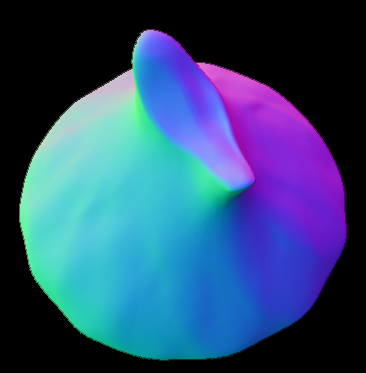}
            \end{minipage}
            \begin{minipage}{.12\textwidth}
                \includegraphics[width=\textwidth]{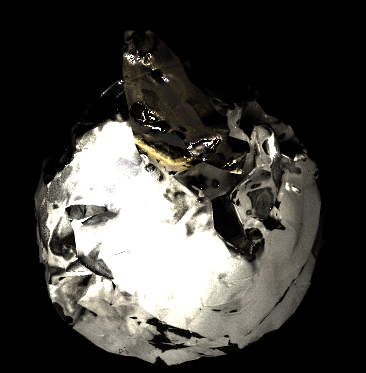}
                \put(-16,3){\scalebox{.8}{\color{white} 0.89}}
            \end{minipage}
            \begin{minipage}{.12\textwidth}
                \includegraphics[width=\textwidth]{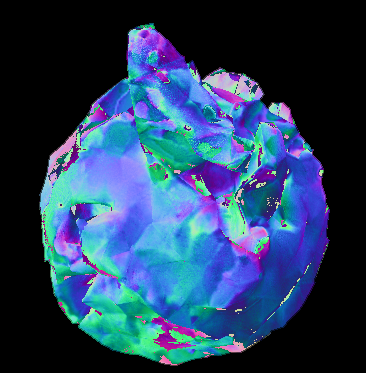}
            \end{minipage}
        \end{minipage}
    \end{minipage}

    \begin{minipage}{\linewidth}
            \centering
        \begin{minipage}{0.08in}	
            \centering
            \rotatebox{90}{\small \textsc{Dice}}
        \end{minipage}
        \begin{minipage}{.85\textwidth}
            \centering
            \begin{minipage}{.12\textwidth}
                \includegraphics[width=\textwidth]{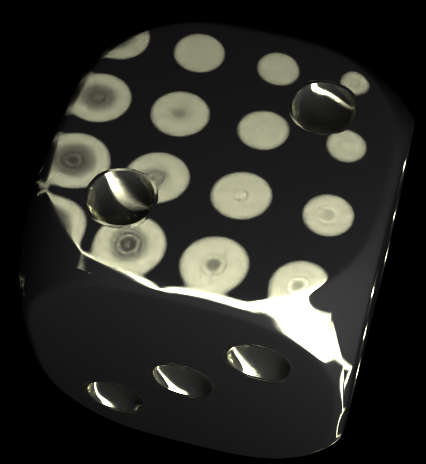}
                \put(-20,3){\scalebox{.8}{\color{white} SSIM}}
            \end{minipage}
            \begin{minipage}{.12\textwidth}
                \includegraphics[width=\textwidth]{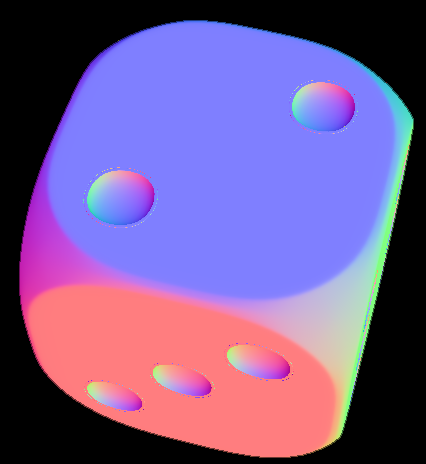}
            \end{minipage}
            \begin{minipage}{.12\textwidth}
                \includegraphics[width=\textwidth]{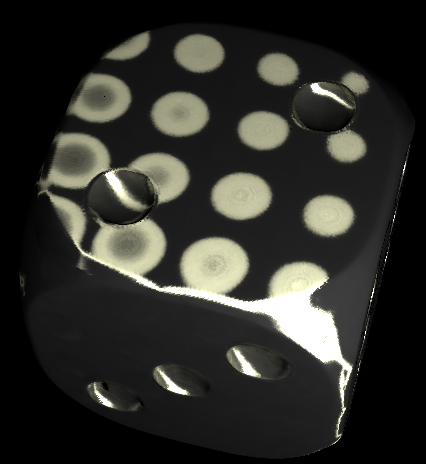}
                \put(-16,3){\scalebox{.8}{\color{white} 0.98}}
            \end{minipage}
            \begin{minipage}{.12\textwidth}
                \includegraphics[width=\textwidth]{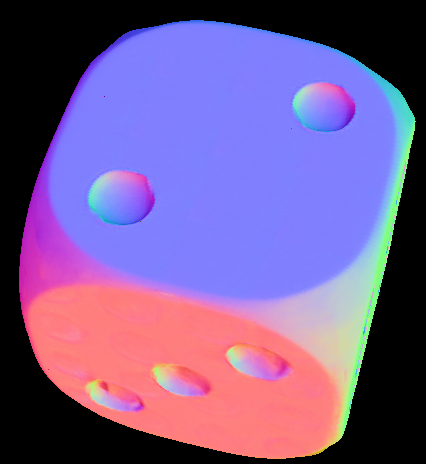}
            \end{minipage}
            \begin{minipage}{.12\textwidth}
                \includegraphics[width=\textwidth]{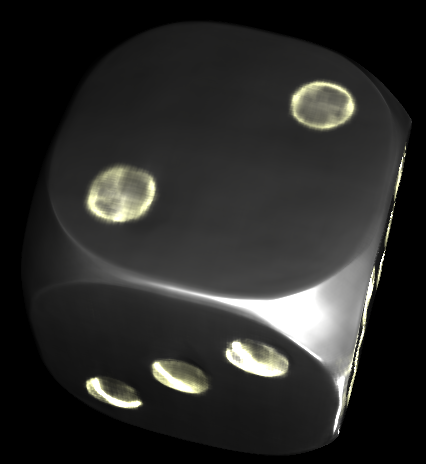}
                \put(-16,3){\scalebox{.8}{\color{white} 0.85}}
            \end{minipage}
            \begin{minipage}{.12\textwidth}
                \includegraphics[width=\textwidth]{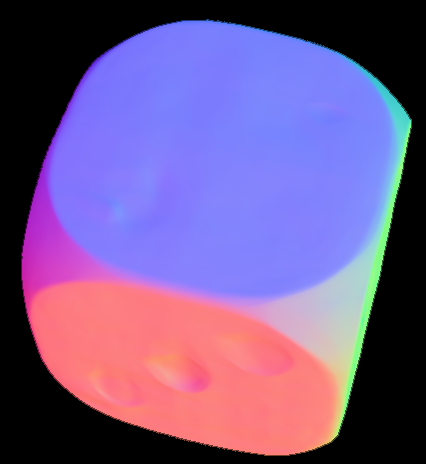}
            \end{minipage}
            \begin{minipage}{.12\textwidth}
                \includegraphics[width=\textwidth]{paper_figures/blank.png}
            \end{minipage}
            \begin{minipage}{.12\textwidth}
                \includegraphics[width=\textwidth]{paper_figures/blank.png}
            \end{minipage}
        \end{minipage}
    \end{minipage}

    \caption{Comparison with techniques on joint optimization of shape and appearance. For each pair of images, the left is a photograph/appearance rendered with novel view and lighting, and the right a normal map from the corresponding geometry. From the left to right: ground-truth, our reconstructions, the results from IRON\cite{iron-2022}, and NVDIFFREC\cite{munkberg2022extracting}. Note that NVDIFFREC fails to reconstruct \textsc{Dice} due to strong anisotropic appearance. The input images are captured with a point light for IRON, and with an indoor office environment lighting for NVDIFFREC. Quantitative errors in SSIM are reported in the bottom right corner of related images.}    
    \label{fig:geo_app_comp}
\end{figure*}

\begin{figure}[tbp]
    \begin{minipage}{\linewidth}
        \begin{minipage}{\textwidth}
            \centering
            \begin{minipage}{\textwidth}
                \centering
                \begin{minipage}{.325\textwidth}
                    \centering
                    \subcaption*{\small G.T./Photo}
                \end{minipage}
                \begin{minipage}{.325\textwidth}
                    \centering
                    \subcaption*{\small Prototype-Scanned}
                \end{minipage}
                \begin{minipage}{.325\textwidth}
                    \centering
                    \subcaption*{\small Tablet-Scanned}
                \end{minipage}
            \end{minipage}
        \end{minipage}       
    \end{minipage}

    \begin{minipage}{\linewidth}
        \begin{minipage}{\textwidth}
            \centering
            \begin{minipage}{.325\textwidth}
                \centering
                \includegraphics[width=\textwidth]{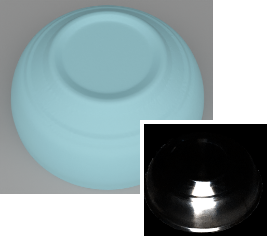}
            \end{minipage}
            \begin{minipage}{.325\textwidth}
                \centering
                \includegraphics[width=\textwidth]{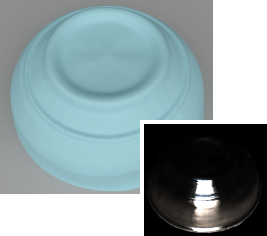}
            \end{minipage}
            \begin{minipage}{.325\textwidth}
                \centering
                \includegraphics[width=\textwidth]{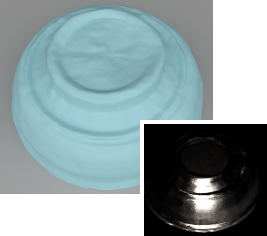}
            \end{minipage}
        \end{minipage}
    \end{minipage}
 
    \caption{Comparison of reconstructions of the same object using our prototype and tablet. The main image shows the shape, while the inset is the rendered appearance with novel view and lighting.}
    \label{fig:abl_ipad}
\end{figure}

\section{Results \& Discussions}
\label{sec:results_and_discussion}

We use our prototype to scan 4 real objects with complex appearance (\textsc{Bowl}, \textsc{Dog}, \textsc{Bear} and \textsc{Bird}). The maximum dimension of each object ranges from 6 to 18 cm. The reconstructed geometry and appearance are shown in~\figref{fig:geo_comp}~\&~\ref{fig:geo_app_comp}. The iPad is used to scan \textsc{Bowl}~(\figref{fig:abl_ipad}). It takes about 60 seconds to scan a real object. The remaining objects (\textsc{Dice}, \textsc{Cup}, \textsc{Najade}, and \textsc{Matball}) are synthetic, whose images are computed via physically based rendering with a virtual scanner. The SVBRDF of \textsc{Matball} is \textsc{satin0112} from \cite{ma2023opensvbrdf}.

All computation is performed on a server with dual AMD EPYC 7763 CPUs, 768GB DDR4 memory and 8 NVIDIA GeForce RTX 4090 GPUs. All results are rendered with NVIDIA OptiX. It takes 70 minutes to preprocess the data of an object, including blurriness computation, object segmentation and camera pose estimation. For a group of 5 images, our network needs 2.5 seconds to transform them to a feature map. On average we use 60 groups to reconstruct an object. The geometry reconstruction via NeuS/Neuralangelo takes 6/14 hours respectively, and the appearance optimization about 1 hour. Samples of the captured images under our optimized lighting patterns and corresponding feature maps are visualized in~\figref{fig:feature_visualize}.

\subsection{Comparisons} 

In the following comparisons, our approach uses on average 60 groups of images (\#images = 300) for reconstructing each object, while competing methods use around 300 images for fairness. In appearance comparisons, we randomly choose 70\% of the captured images for training and use the remaining 30\% as test images to evaluate novel view and lighting synthesis. The ground-truth geometry of a physical object is obtained with a commercial 3D scanner~\cite{shining3D}. Due to the challenging appearance (e.g., strong anisotropic/specular reflections), we have to apply fine powder to object surfaces for the scanner to work properly.

\begin{figure}[tbp]
    \begin{minipage}{\linewidth}
        \centering
        \begin{minipage}{\linewidth}
            \centering
            \begin{minipage}{\linewidth}
                \centering
                \begin{minipage}{.32\linewidth}
                    \centering
                    \subcaption*{\small Ground-truth}
                \end{minipage}
                \begin{minipage}{.32\linewidth}
                    \centering
                    \subcaption*{\small Ours}
                \end{minipage}
                \begin{minipage}{.32\linewidth}
                    \centering
                    \subcaption*{\small \cite{zeng2023nrhints}}
                \end{minipage}
            \end{minipage}
        \end{minipage}       
    \end{minipage}

    \begin{minipage}{\linewidth}    
        \begin{minipage}{\textwidth}
            \centering
            \begin{minipage}{.158\textwidth}
                \centering
                \includegraphics[width=\textwidth]{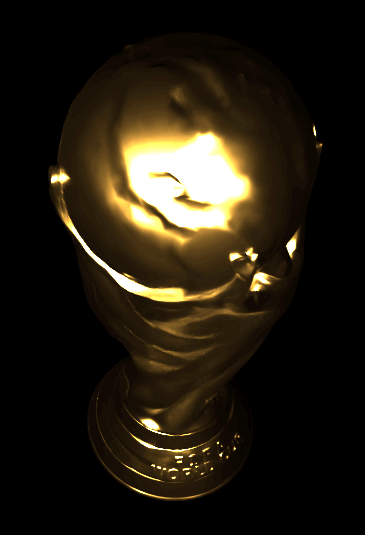}
            \end{minipage}
            \begin{minipage}{.158\textwidth}
                \centering
                \includegraphics[width=\textwidth]{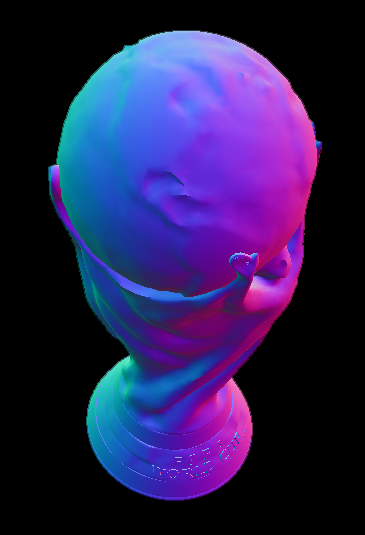}
            \end{minipage}
            \begin{minipage}{.158\textwidth}
                \centering
                \includegraphics[width=\textwidth]{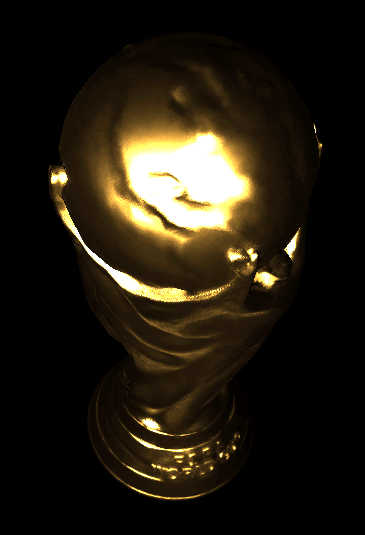}
            \end{minipage}
            \begin{minipage}{.158\textwidth}
                \centering
                \includegraphics[width=\textwidth]{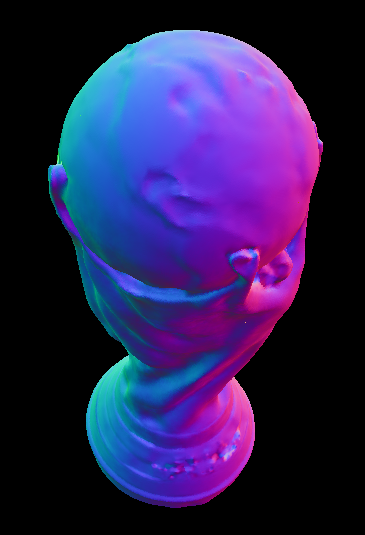}
            \end{minipage}
            \begin{minipage}{.158\textwidth}
                \centering
                \includegraphics[width=\textwidth]{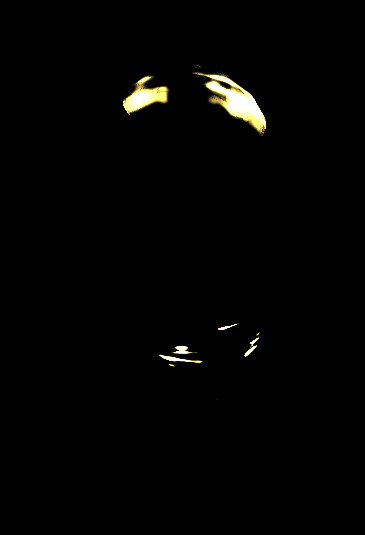}
            \end{minipage}
            \begin{minipage}{.158\textwidth}
                \centering
                \includegraphics[width=\textwidth]{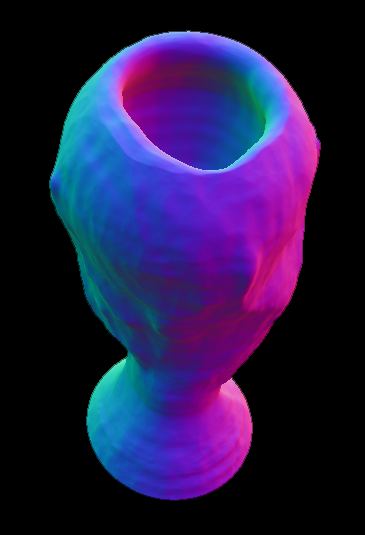}
            \end{minipage}
        \end{minipage}
    \end{minipage}
    \caption{Comparison between our reconstruction and~\cite{zeng2023nrhints}. For each pair of images, the left one is the appearance rendering, and the right a normal map from the corresponding geometry.}
    \label{fig:vs_nr2}
\end{figure}

\begin{figure}[tbp]
    \begin{minipage}{\linewidth}
        \begin{minipage}{\textwidth}
            \centering
            \begin{minipage}{.19\textwidth}
                \includegraphics[width=\textwidth]{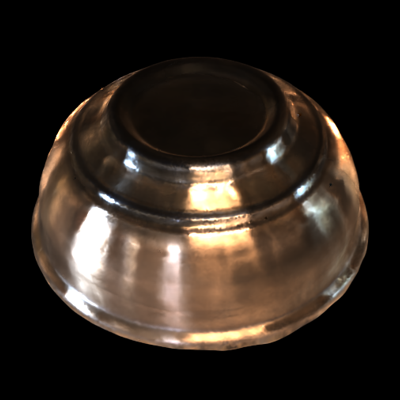}
            \end{minipage}
            \begin{minipage}{.19\textwidth}
                \includegraphics[width=\textwidth]{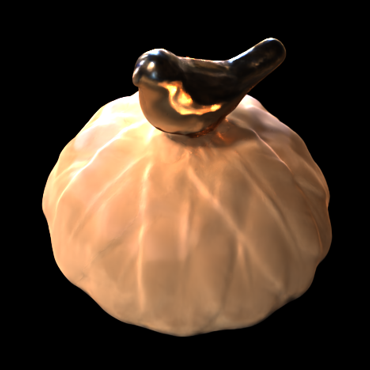}
            \end{minipage}
            \begin{minipage}{.19\textwidth}
                \includegraphics[width=\textwidth]{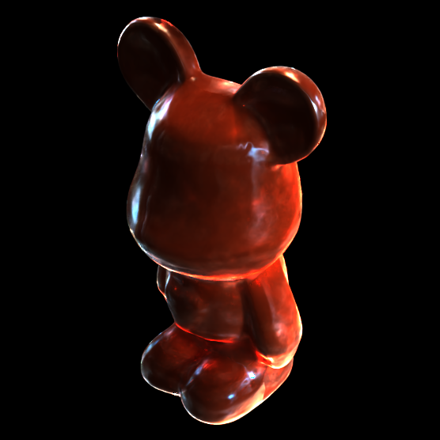}
            \end{minipage}
            \begin{minipage}{.19\textwidth}
                \includegraphics[width=\textwidth]{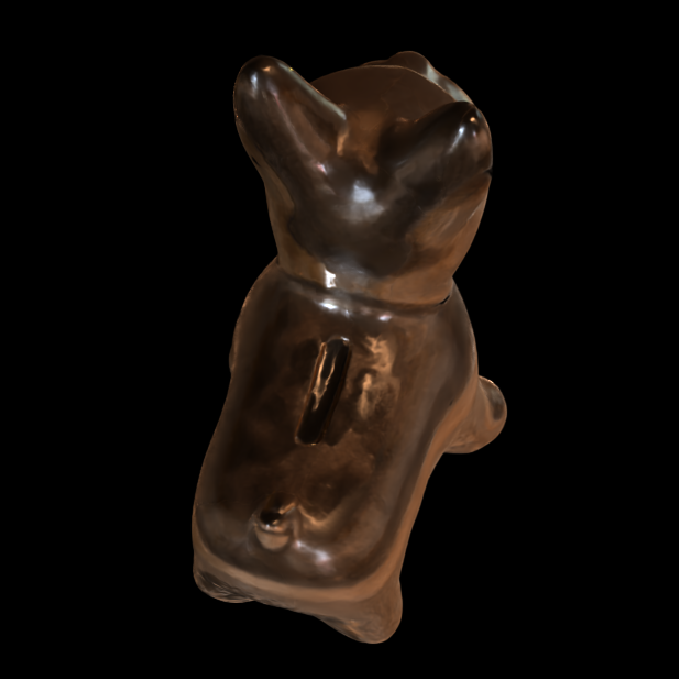}
            \end{minipage}
            \begin{minipage}{.19\textwidth}
                \includegraphics[width=\textwidth]{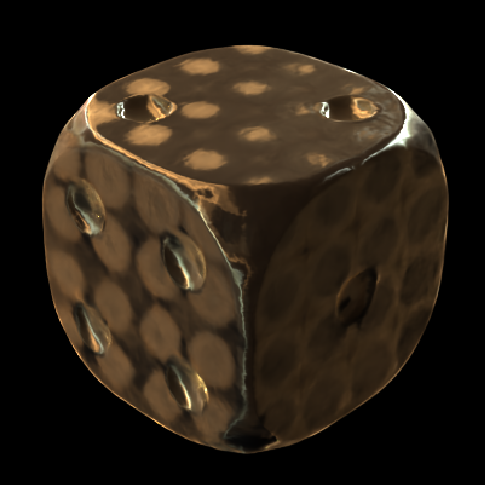}
            \end{minipage}
        \end{minipage}
    \end{minipage}
    \begin{minipage}{\linewidth}
        \begin{minipage}{\textwidth}
            \centering
            \begin{minipage}{.19\textwidth}
                \includegraphics[width=\textwidth]{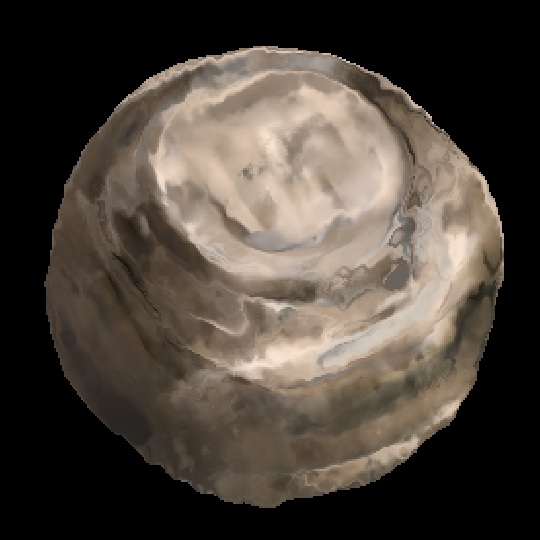}
            \end{minipage}
            \begin{minipage}{.19\textwidth}
                \includegraphics[width=\textwidth]{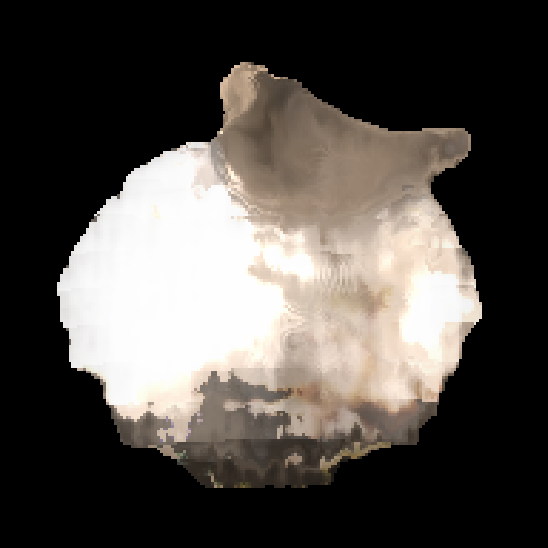}
            \end{minipage}
            \begin{minipage}{.19\textwidth}
                \includegraphics[width=\textwidth]{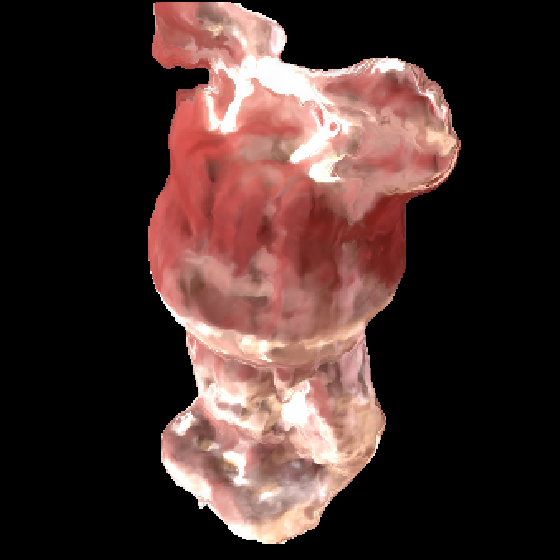}
            \end{minipage}
            \begin{minipage}{.19\textwidth}
                \includegraphics[width=\textwidth]{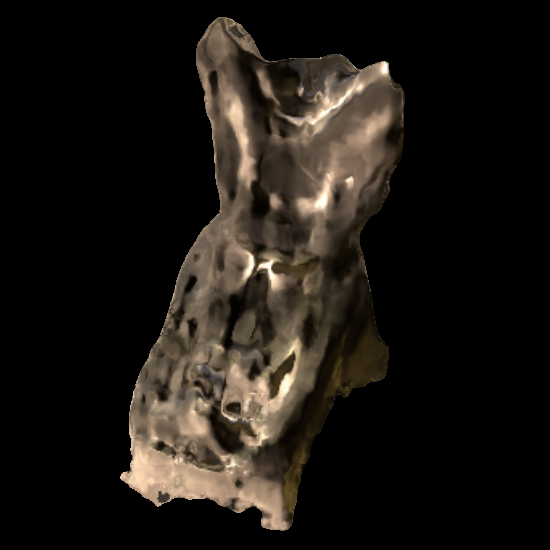}
            \end{minipage}
            \begin{minipage}{.19\textwidth}
                \includegraphics[width=\textwidth]{paper_figures/blank.png}
            \end{minipage}
        \end{minipage}
    \end{minipage} 
    \caption{Appearance reconstruction rendered with novel environment lighting. The top row are our relighting results, and the bottom row results from NeRFactor\cite{zhang2021nerfactor}, which fails to reconstruct the strong anisotropic reflectance of \textsc{Dice}.}
    \label{fig:envlight_comp}
\end{figure}

\textbf{Geometry Reconstruction.} In~\figref{fig:geo_comp}, we compare our geometry reconstruction results with state-of-the-art methods on 5 objects with challenging appearance, including specular reflections and textureless regions. The \textsc{Bowl} and \textsc{Dice} also exhibit strong anisotropic appearance. For a fair comparison, the acquisition lighting condition for Ref-NeuS, NeRO, Neuralangelo, NeuS and COLMAP is a conventional indoor office environment. This is because our pilot study shows that for these methods, using input photographs under our lighting patterns leads to lower reconstruction quality. For CasMVSNet, we test with input photographs under 1(center-view) or 5 of our learned patterns. Due to the lack of efficient handling of complex appearance variations, \figref{fig:geo_comp} shows unsatisfactory reconstructions from Ref-Neus, NeRO, Neuralangelo, NeuS, CasMVSNet or COLMAP. Moreover, the effectiveness of EPFT\cite{Kang_2021_ICCV} is limited, as it relies on photometric information captured from a fixed view. Our results (the last two columns of~\figref{fig:geo_comp}) outperform other methods both qualitatively and quantitatively.

\textbf{Joint Reconstruction of Geometry \& Appearance.} In~\figref{fig:geo_app_comp}, we compare both geometry and appearance reconstructions with related methods. For a fair comparison, IRON\cite{iron-2022} takes as input photographs with a co-located flash, and the acquisition lighting for NVDIFFREC\cite{munkberg2022extracting} is the same indoor office environment as used in~\figref{fig:geo_comp}, similar to the original papers. We also find input photographing under our patterns results in lower-quality reconstructions with their methods. Our approach outperforms competing methods qualitatively and quantitively, as we can handle challenging appearance by efficiently probing the angular domain with learned illumination multiplexing.  

In~\figref{fig:vs_nr2}, we compare with a state-of-the-art differentiable optimization technique~\cite{zeng2023nrhints}. Their method struggles to accurately reconstruct the geometry in the presence highly specular reflectance. The end-to-end, joint optimization for shape and appearance is challenging to converge to correct results. Plus, their point light does not have sufficient sampling capability, leading to an under-constrained optimization. We also compare with NeRFactor\cite{zhang2021nerfactor} on appearance reconstruction in~\figref{fig:envlight_comp}, by rendering the results under novel view and light. Their approach relies on precise estimation of the density field, which becomes inaccurate in the presence of complex appearance. This leads to unsatisfactory reconstructions, and consequently low-quality renderings.

\begin{figure*}[tbp]
    \centering
    \begin{minipage}{\linewidth}
        \centering
        \begin{minipage}{\linewidth}
            \centering
            \begin{minipage}{.138\textwidth}
                \centering
                \subcaption*{\small (a) G.T.}
            \end{minipage}
            \begin{minipage}{.138\textwidth}
                \centering
                \subcaption*{\small (b) Ours}
            \end{minipage}
            \begin{minipage}{.138\textwidth}
                \centering
                \subcaption*{\small (c) Gaussian}
            \end{minipage}
            \begin{minipage}{.138\textwidth}
                \centering
                \subcaption*{\small (d) Full-on}
            \end{minipage}
            \begin{minipage}{.138\textwidth}
                \centering
                \subcaption*{\small (e) w/o Warp}
            \end{minipage}
            \begin{minipage}{.138\textwidth}
                \centering
                \subcaption*{\small (f) 2px}
            \end{minipage}
            \begin{minipage}{.138\textwidth}
                \centering
                \subcaption*{\small (g) 4px}
            \end{minipage}
        \end{minipage}
    \end{minipage}
    \begin{minipage}{\linewidth}
        \centering
        \begin{minipage}{\linewidth}
            \centering
            \begin{minipage}{.138\textwidth}
                \includegraphics[width=\textwidth]{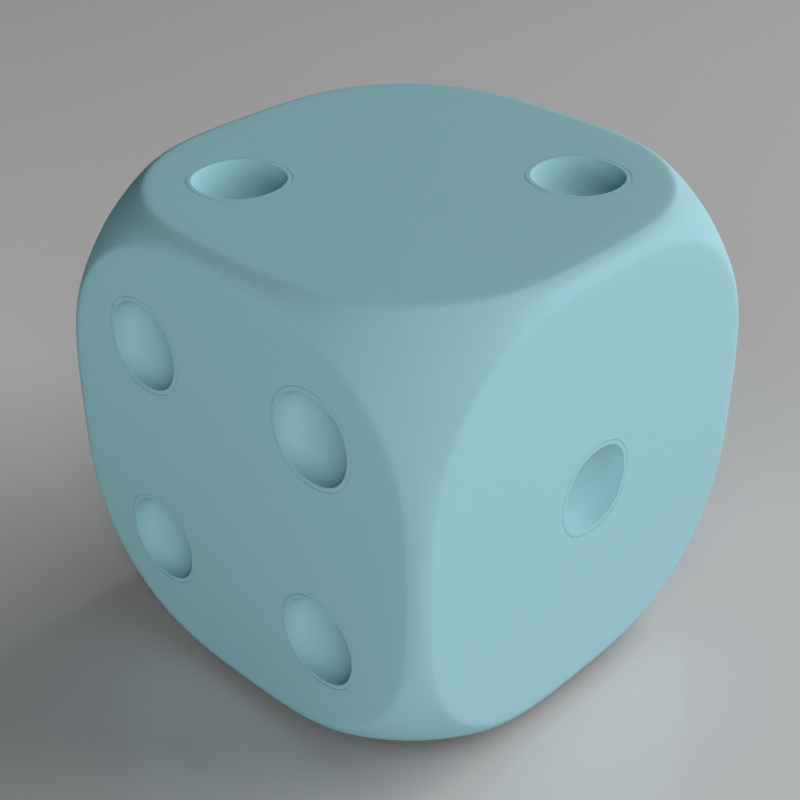}
            \end{minipage}
            \begin{minipage}{.138\textwidth}
                \includegraphics[width=\textwidth]{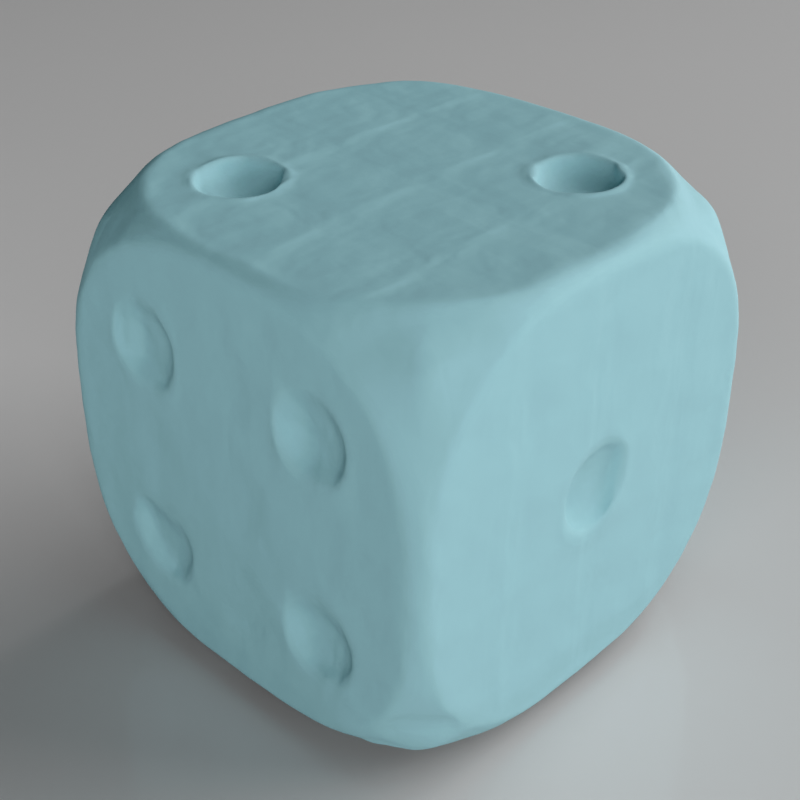}
                \put(-13,3){\scalebox{.8}{\color{black} 0.9}}
            \end{minipage}
            \begin{minipage}{.138\textwidth}
                \includegraphics[width=\textwidth]{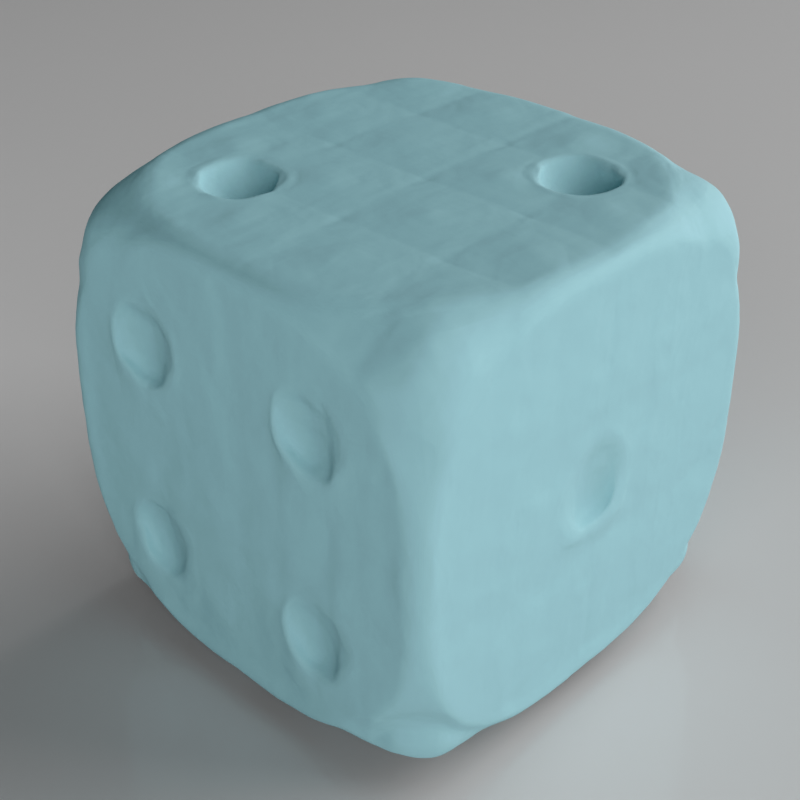}
                \put(-13,3){\scalebox{.8}{\color{black} 1.4}}
            \end{minipage}
            \begin{minipage}{.138\textwidth}
                \includegraphics[width=\textwidth]{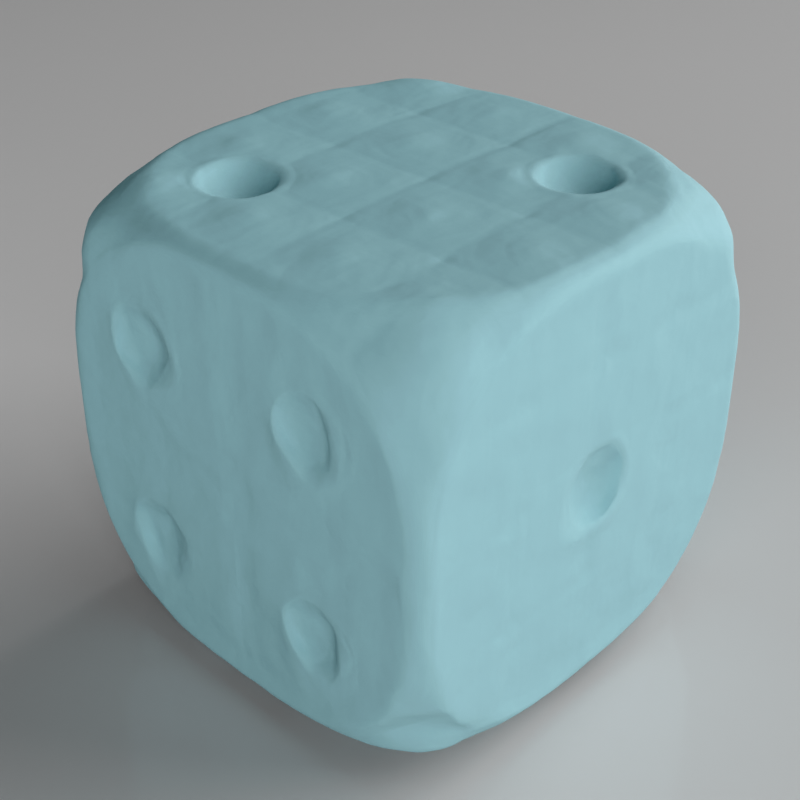}
                \put(-13,3){\scalebox{.8}{\color{black} 1.3}}
            \end{minipage}
            \begin{minipage}{.138\textwidth}
                \includegraphics[width=\textwidth]{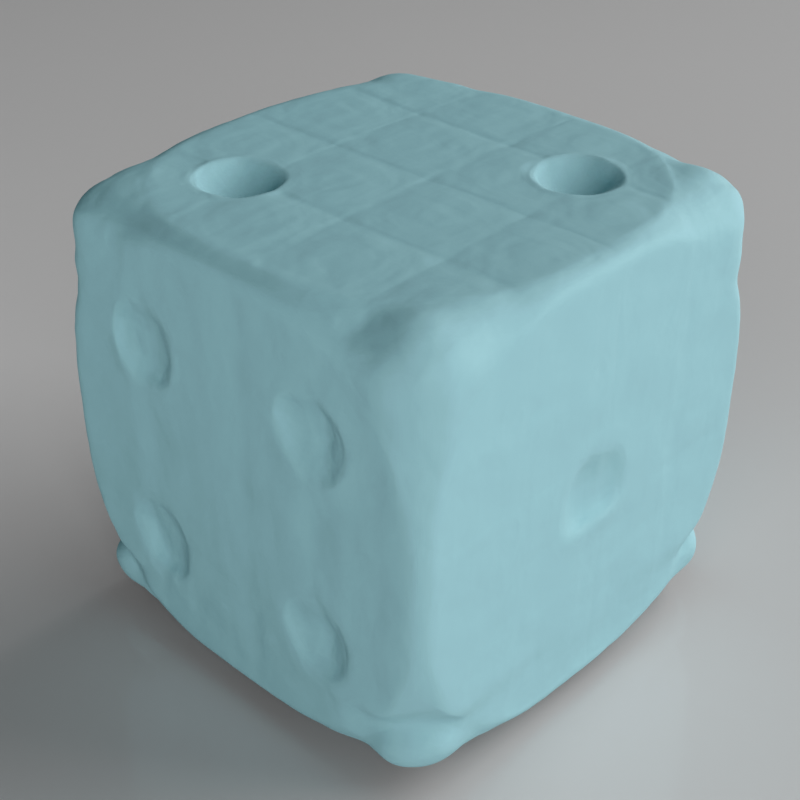}
                \put(-13,3){\scalebox{.8}{\color{black} 2.3}}
            \end{minipage}
            \begin{minipage}{.138\textwidth}
                \includegraphics[width=\textwidth]{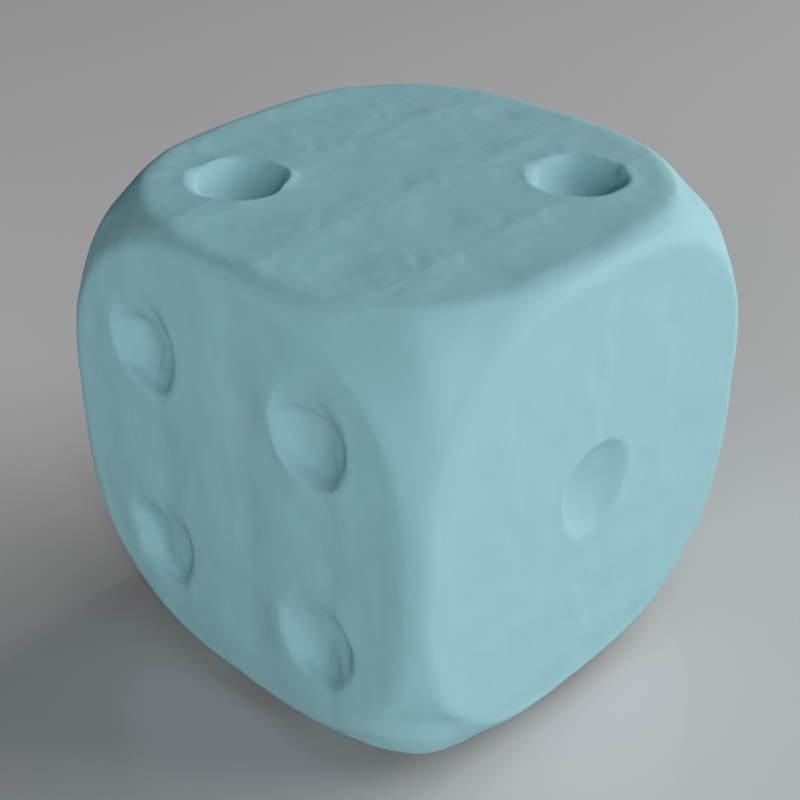}
                \put(-13,3){\scalebox{.8}{\color{black} 0.9}}
            \end{minipage}
            \begin{minipage}{.138\textwidth}
                \includegraphics[width=\textwidth]{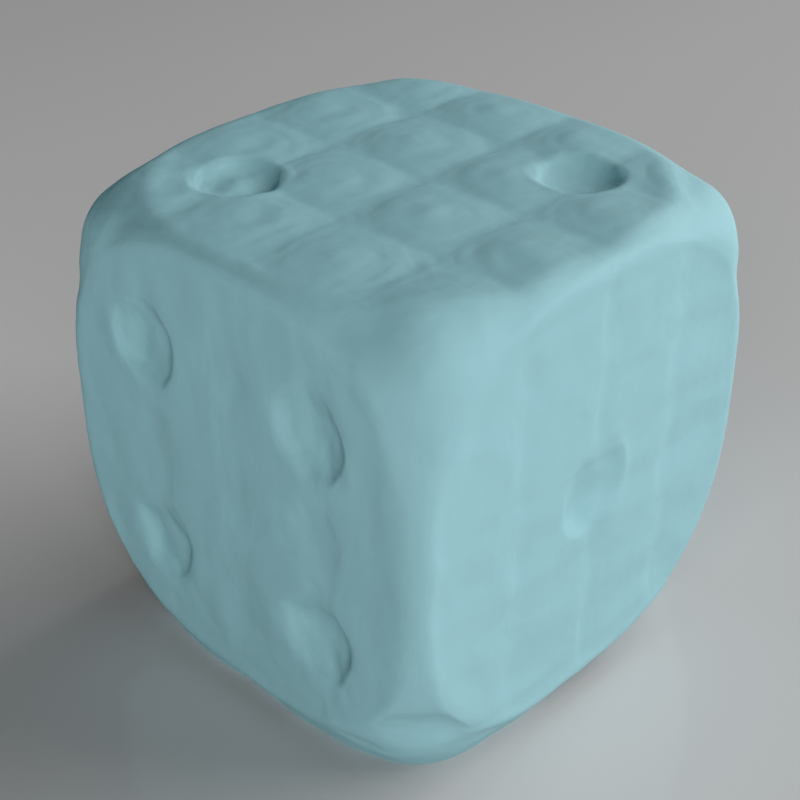}
                \put(-13,3){\scalebox{.8}{\color{black} 1.4}}
            \end{minipage}
        \end{minipage}
    \end{minipage}
    \caption{Impact of lighting patterns, warping and camera pose errors over geometric reconstruction. From the left to right: the ground-truth, the reconstruction using our network, ours with 5 Gaussian noise lighting patterns, ours with a full-on pattern, ours without the warping step and ours with perturbed camera poses (average reprojection error = 2px/4px). Quantitative errors in Chamfer distance are reported in the bottom right corner.}
    \label{fig:ablation_inone}
\end{figure*}

\begin{figure*}[tbp]
    \centering
    \begin{minipage}{\linewidth}
        \centering
        \begin{minipage}{\linewidth}
            \centering
            \begin{minipage}{.12\textwidth}
                \centering
                \subcaption*{\small (a) G.T.}
            \end{minipage}
            \begin{minipage}{.12\textwidth}
                \centering
                \subcaption*{\small (b) $\#p=3$}
            \end{minipage}
            \begin{minipage}{.12\textwidth}
                \centering
                \subcaption*{\small (c) $\#p=5$}
            \end{minipage}
            \begin{minipage}{.12\textwidth}
                \centering
                \subcaption*{\small (d) $\#p=7$}
            \end{minipage}
            \begin{minipage}{.12\textwidth}
                \centering
                \subcaption*{\small (e) Normalized}
            \end{minipage}
            \begin{minipage}{.12\textwidth}
                \centering
                \subcaption*{\small (f) Unnorm.}
            \end{minipage}
            \begin{minipage}{.12\textwidth}
                \centering
                \subcaption*{\small (g) $3\times$~Speed}
            \end{minipage}
            \begin{minipage}{.12\textwidth}
                \centering
                \subcaption*{\small (h) $6\times$~Speed}
            \end{minipage}
        \end{minipage}
    \end{minipage}
    \begin{minipage}{\textwidth}
        \centering
        \begin{minipage}{.12\linewidth}
            \centering
            \includegraphics[width=\textwidth]{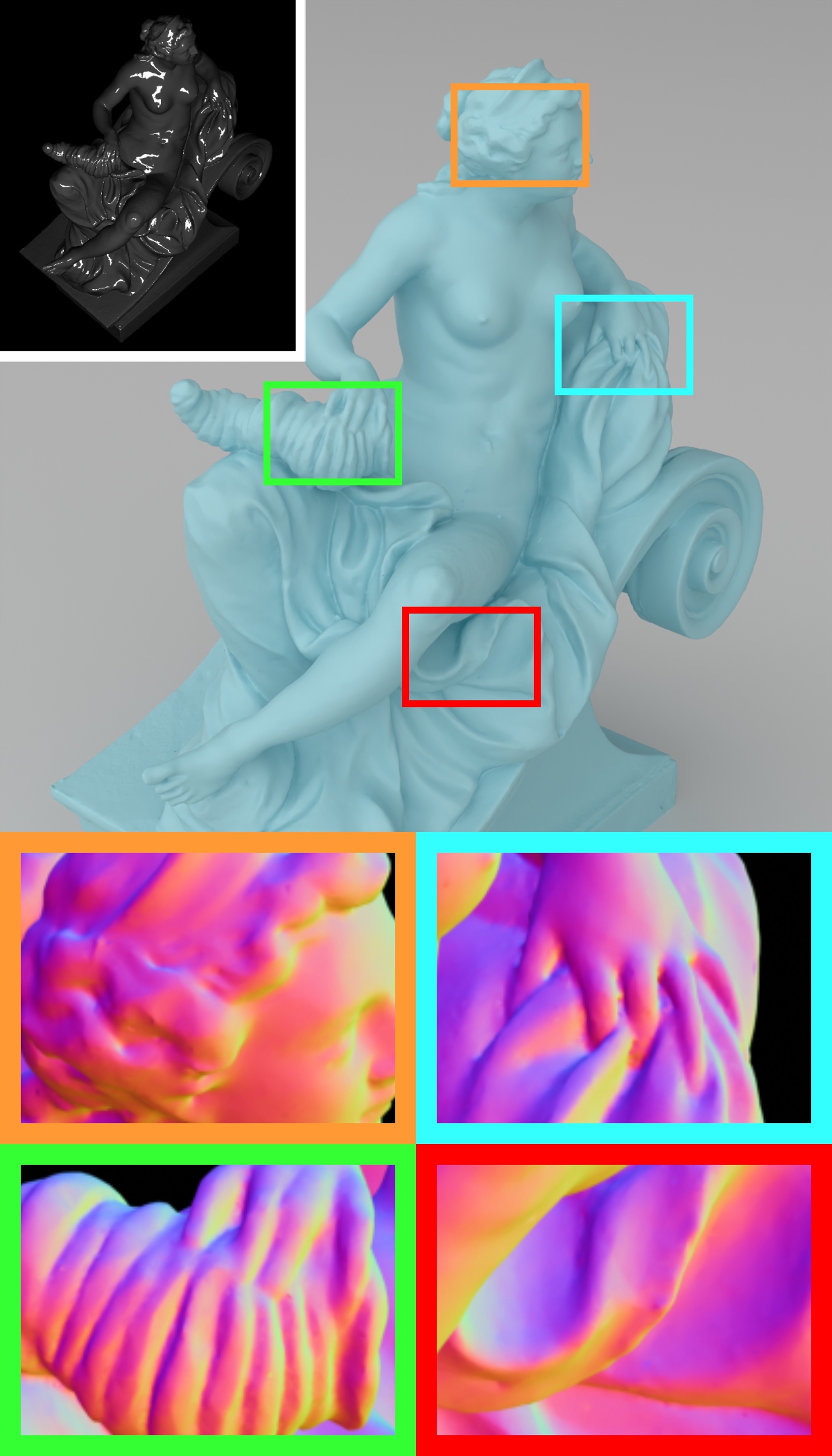}
        \end{minipage}
        \begin{minipage}{.12\linewidth}
            \centering
            \includegraphics[width=\textwidth]{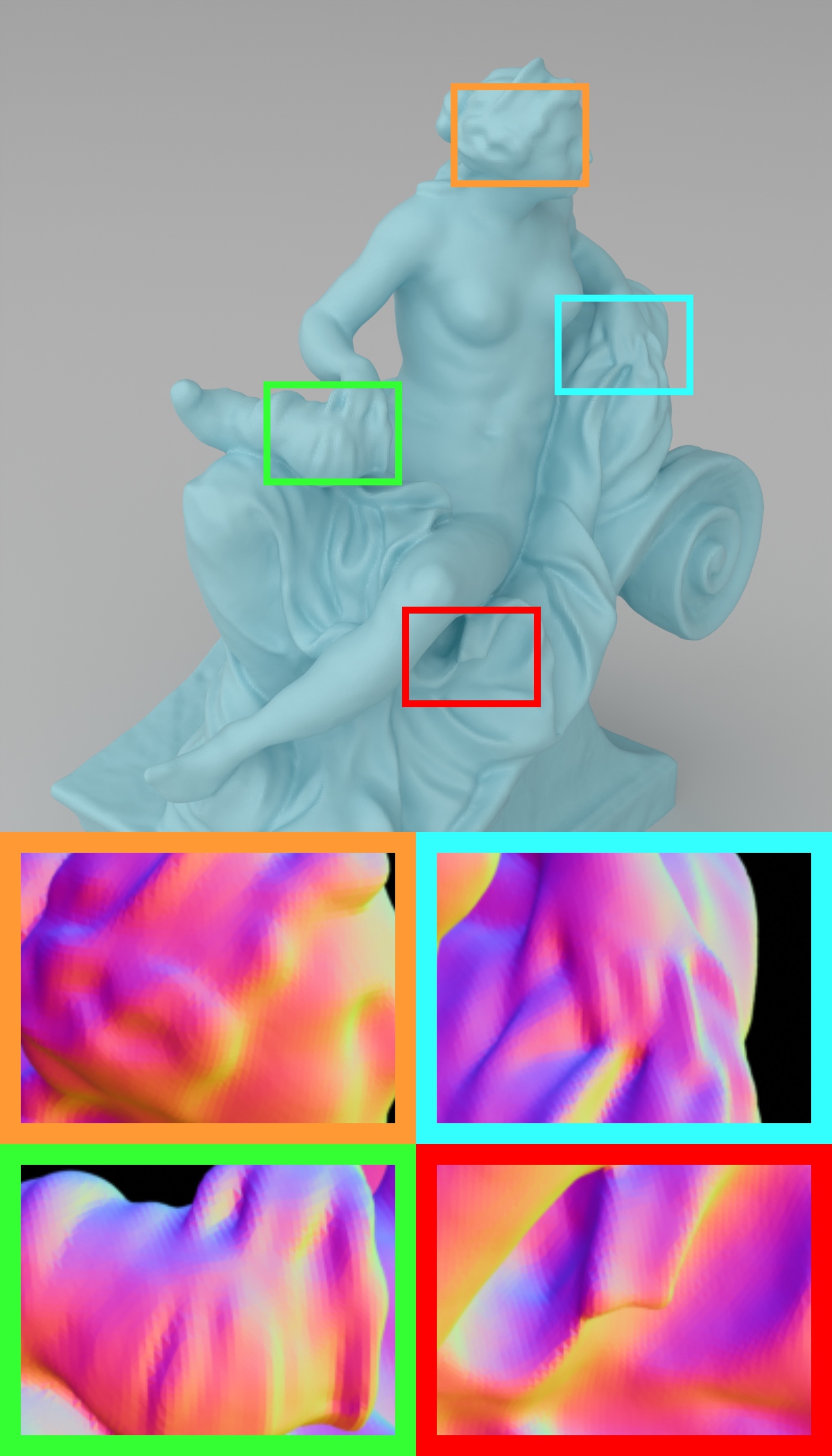}
            \put(-58,100){\scalebox{.8}{\color{black} 0.35}}
        \end{minipage}
        \begin{minipage}{.12\linewidth}
            \centering
            \includegraphics[width=\textwidth]{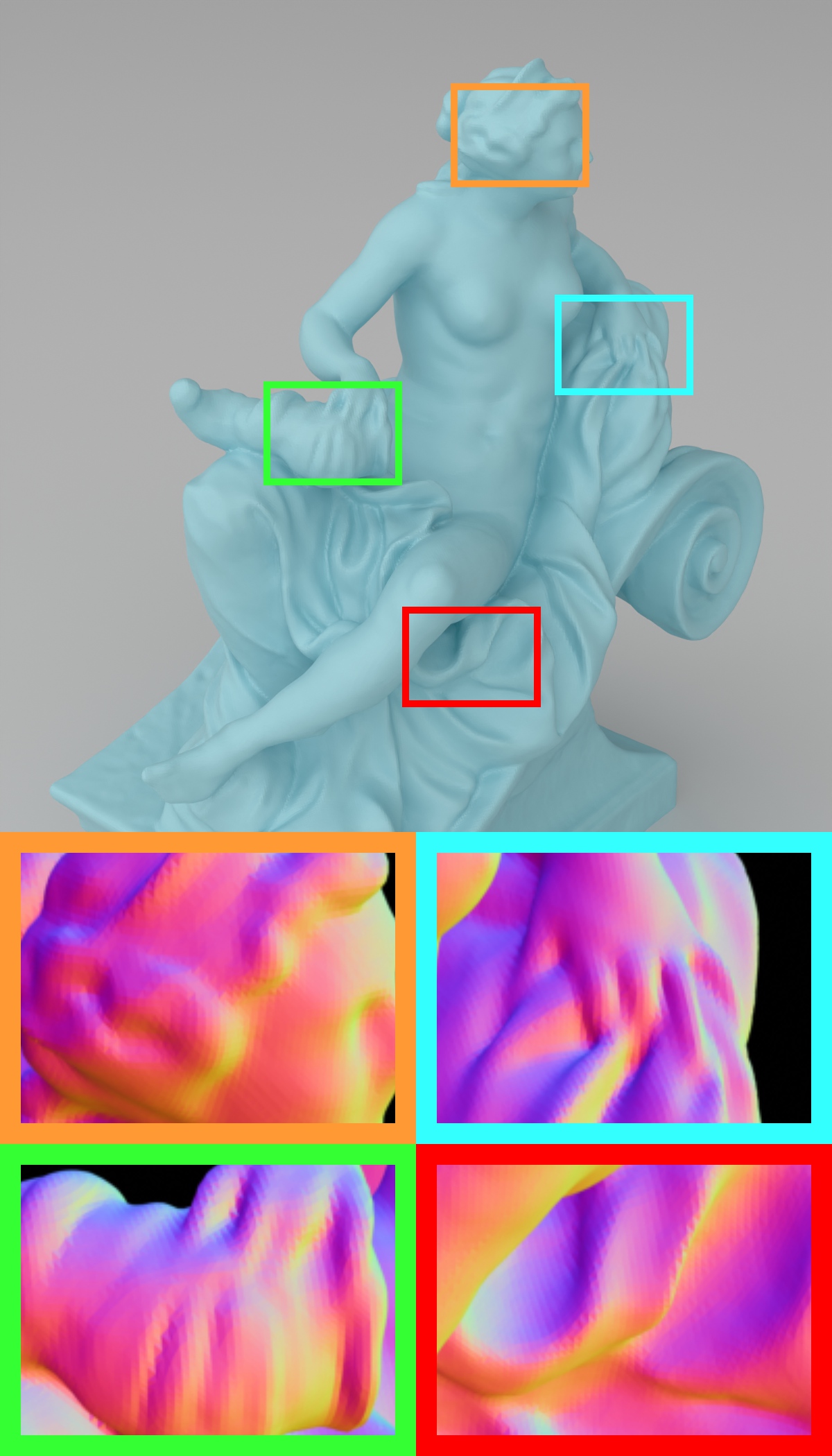}
            \put(-58,100){\scalebox{.8}{\color{black} 0.22}}
        \end{minipage}
        \begin{minipage}{.12\linewidth}
            \centering
            \includegraphics[width=\textwidth]{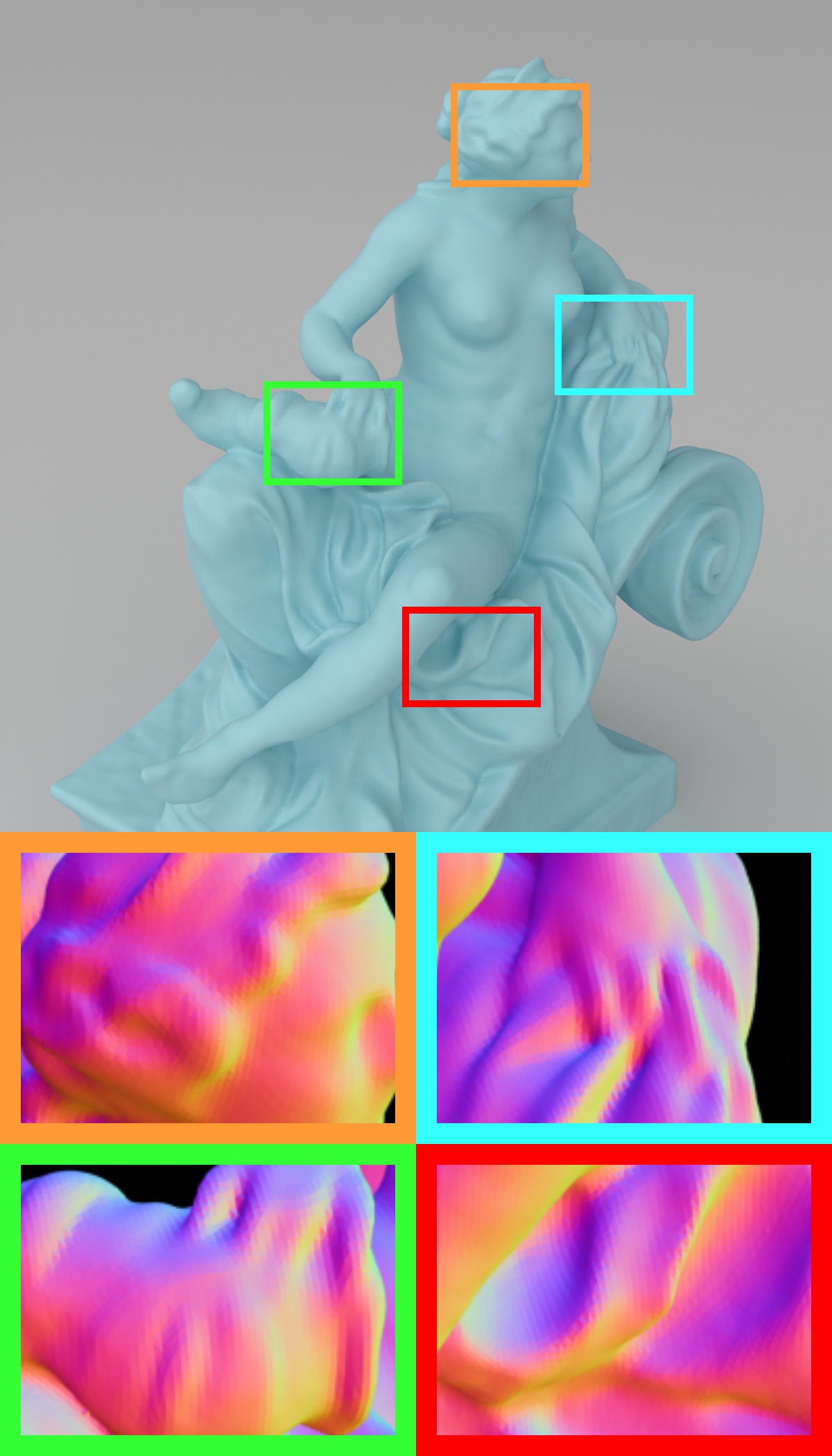}
            \put(-58,100){\scalebox{.8}{\color{black} 0.21}}
        \end{minipage}
        \begin{minipage}{.12\linewidth}
            \centering
            \includegraphics[width=\textwidth]{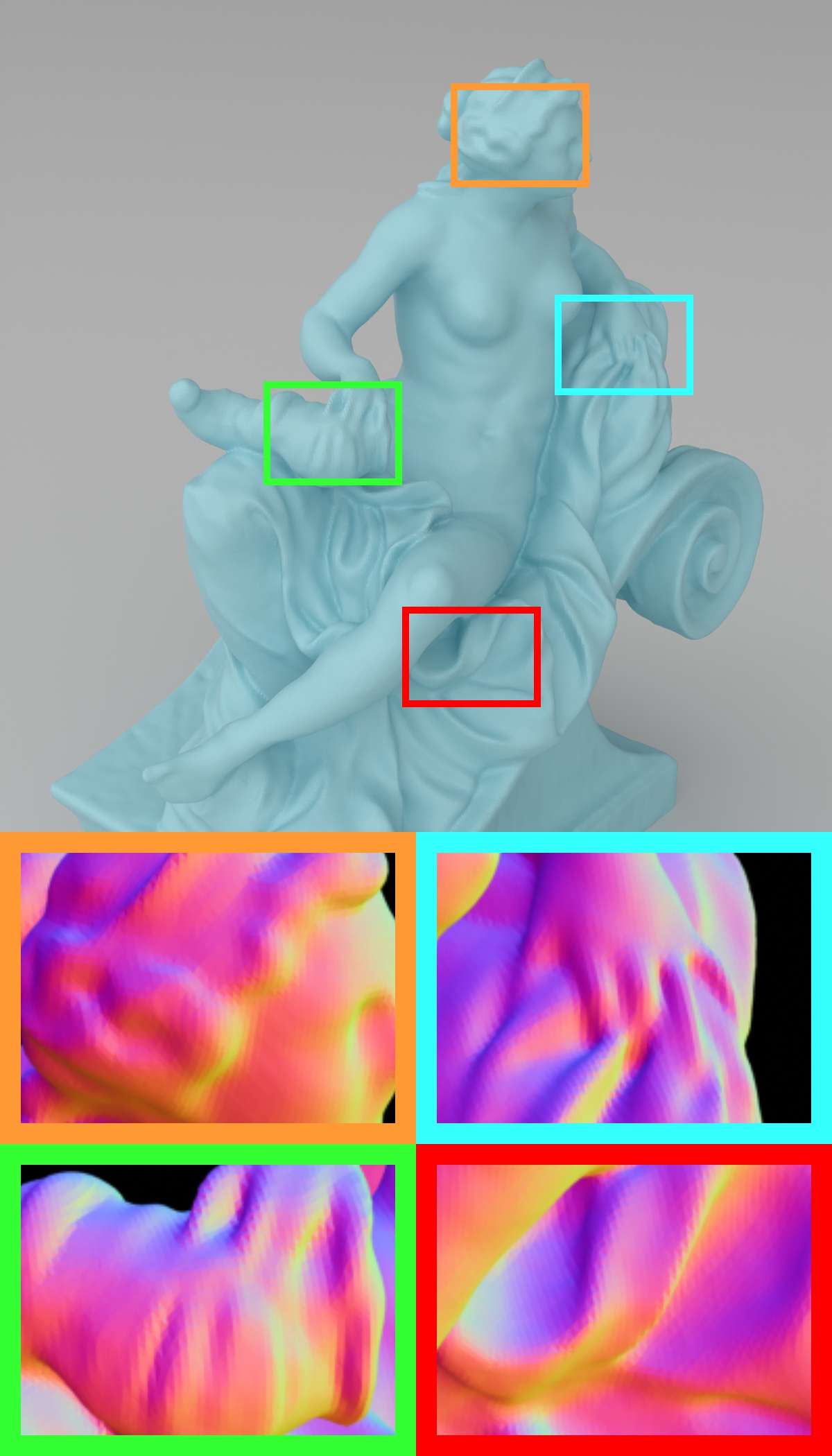}
            \put(-58,100){\scalebox{.8}{\color{black} 0.22}}
        \end{minipage}
        \begin{minipage}{.12\linewidth}
            \centering
            \includegraphics[width=\textwidth]{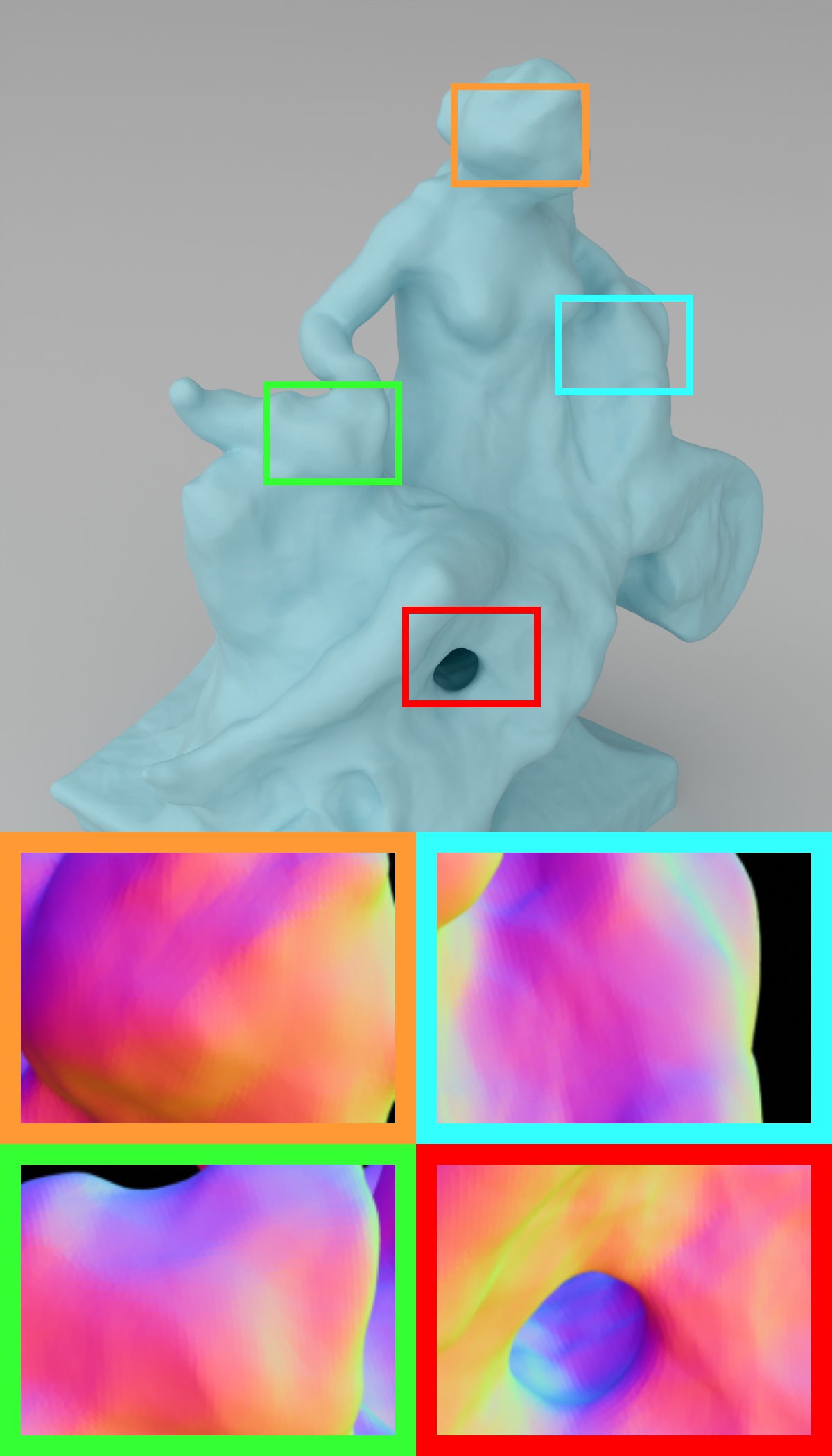}
            \put(-58,100){\scalebox{.8}{\color{black} 9.43}}
        \end{minipage}
        \begin{minipage}{.12\linewidth}
            \centering
            \includegraphics[width=\textwidth]{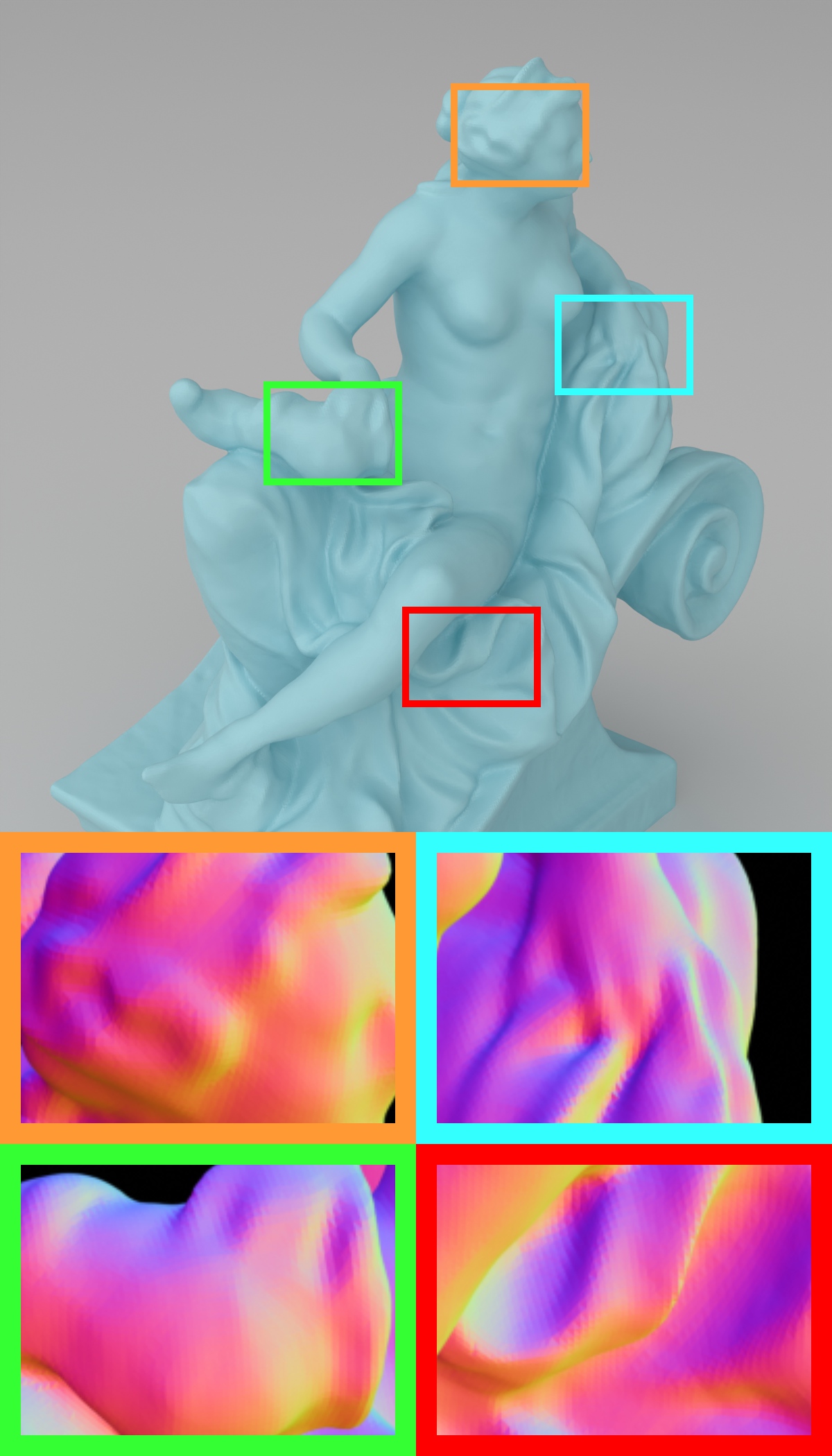}
            \put(-58,100){\scalebox{.8}{\color{black} 0.39}}
        \end{minipage}
        \begin{minipage}{.12\linewidth}
            \centering
            \includegraphics[width=\textwidth]{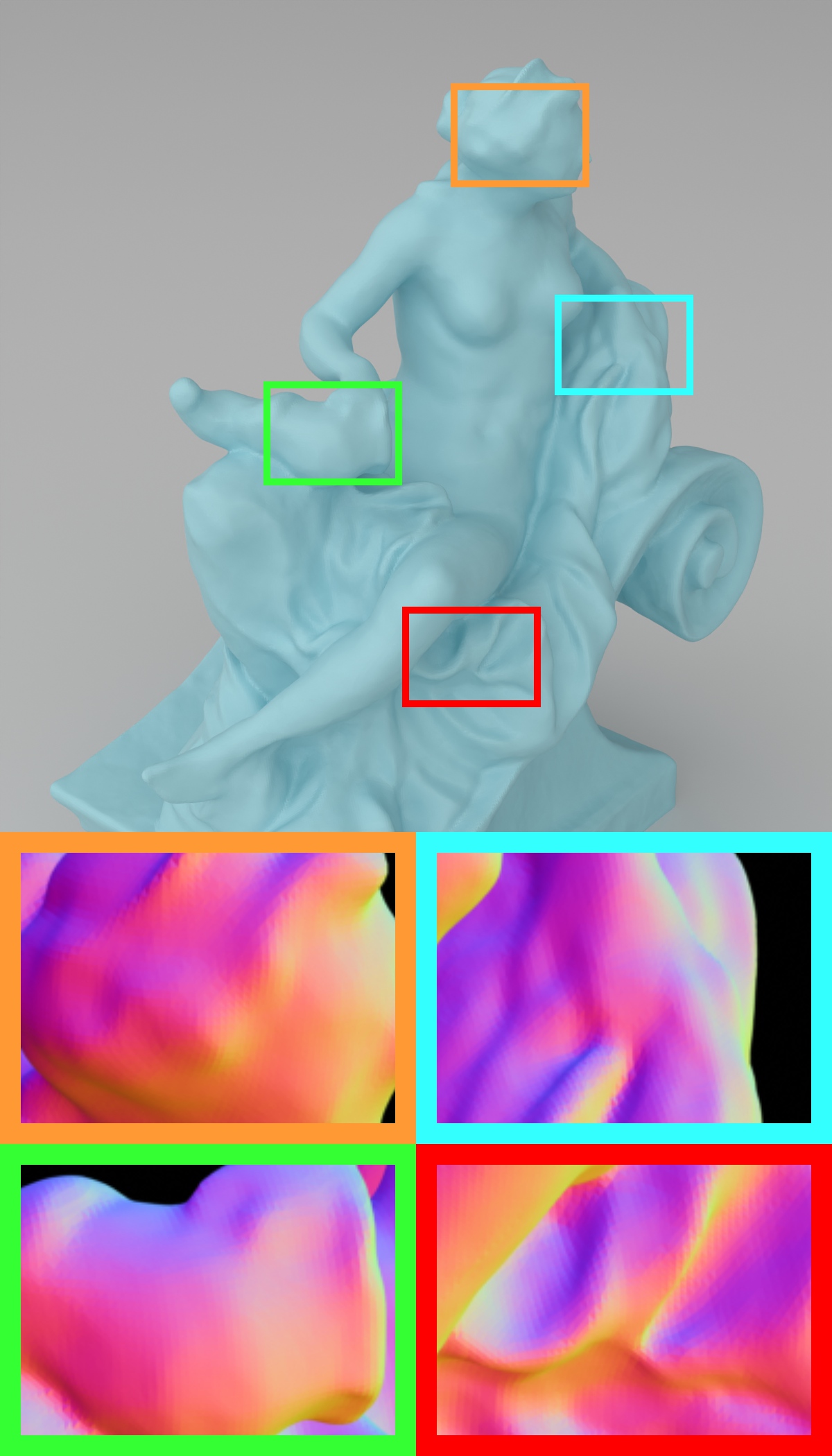}
            \put(-58,100){\scalebox{.8}{\color{black} 1.24}}
        \end{minipage}
    \end{minipage}
    \caption{Impact of the number of lighting patterns, two-branch architecture and scanning speed over geometric reconstruction. From the left to right: the ground-truth, reconstructions using our network trained with 3, 5, and 7 lighting patterns; reconstruction using features from normalized/unnormalized branch of our network; and reconstructions from $3/6\times$ the average scanning speed of the training dataset. The bottom insets visualize the reconstructed normals. Quantitative errors in Chamfer distance are reported in the top left corner. Unnorm. = Unnormalized.}
    \label{fig:ablation_inone2}
\end{figure*}

\subsection{Evaluations}

We first analyze the correlations between our learned features with various common parameters in~\tabref{tab:correlation}, by computing Canonical Correlation Analysis (CCA) between our features and individual parameters, averaged over our synthetic training dataset.

In~\figref{fig:ablation_inone}, we evaluate the impact of lighting patterns, warping and camera pose errors on reconstructing the geometry of \textsc{Dice}. We first compute the shape with our approach on images rendered with camera motions different from training, as shown in (b). Next, we evaluate the impact of different lighting patterns. (c)/(d) shows the result computed using our network trained with 5 fixed Gaussian noise patterns/a full-on pattern. The effectiveness of our learned patterns is clear, by comparing (b),(c) \& (d). Moreover, we train a network without warping, which results in considerably higher reconstruction error, as shown in (e). This demonstrates the effectiveness of our warping, which efficiently models the 3D uncertainty. The last 2 images evaluate the robustness of our approach against errors in camera poses. We perturb the camera poses with Gaussian noise of different standard deviations as in~\cite{Ma:2021:TRACE}, and report the average reprojection errors on top of the 2 images. Our approach can tolerate a reprojection error of 2 pixels, as shown in (f). However, when the error increases to 4, the 3D reconstruction quality degrades (g). For reference, the average reprojection error with standard SfM~\cite{schoenberger2016sfm} is 0.6 pixel. 

In~\figref{fig:ablation_inone2}, we evaluate the impact of the number of lighting patterns, two-branch architecture and scanning speed on reconstructing the geometry of \textsc{Najade}. Its appearance is textureless and highly specular (as shown in the first inset of the figure). We first compute the shape with our network trained on 3, 5, and 7 lighting patterns, as shown in~\figref{fig:ablation_inone2}~(b), (c) \& (d). While using $5$ patterns can improve geometric details over $3$ patterns, adopting $7$ patterns brings marginal benefits. Next, we assess the impact of normalized/unnormalized branches in ~\figref{fig:ablation_inone2}~(e)/(f). For the textureluess \textsc{Najade}, normalized features clearly outputperform unnormalized ones. Finally, we evaluate the impact of scanning speed in~\figref{fig:ablation_inone2}~(g) \& (h). Higher speed results in lower reconstruction quality, as the content in a group of patches becomes less correlated. This makes it more difficult to extract useful information from the input.

\begin{figure}[tbp]
    \begin{minipage}{\linewidth}
        \begin{minipage}{\textwidth}
            \centering
            \begin{minipage}{\textwidth}
                \centering
                \begin{minipage}{.325\textwidth}
                    \centering
                    \subcaption*{\small Ground-truth}
                \end{minipage}
                \begin{minipage}{.325\textwidth}
                    \centering
                    \subcaption*{\small COLMAP}
                \end{minipage}
                \begin{minipage}{.325\textwidth}
                    \centering
                    \subcaption*{\scriptsize COLMAP+Our Features}
                \end{minipage}
            \end{minipage}
        \end{minipage}       
    \end{minipage}

    \begin{minipage}{\linewidth}
        \begin{minipage}{\textwidth}
            \centering
            \begin{minipage}{.325\textwidth}
                \includegraphics[width=\textwidth]{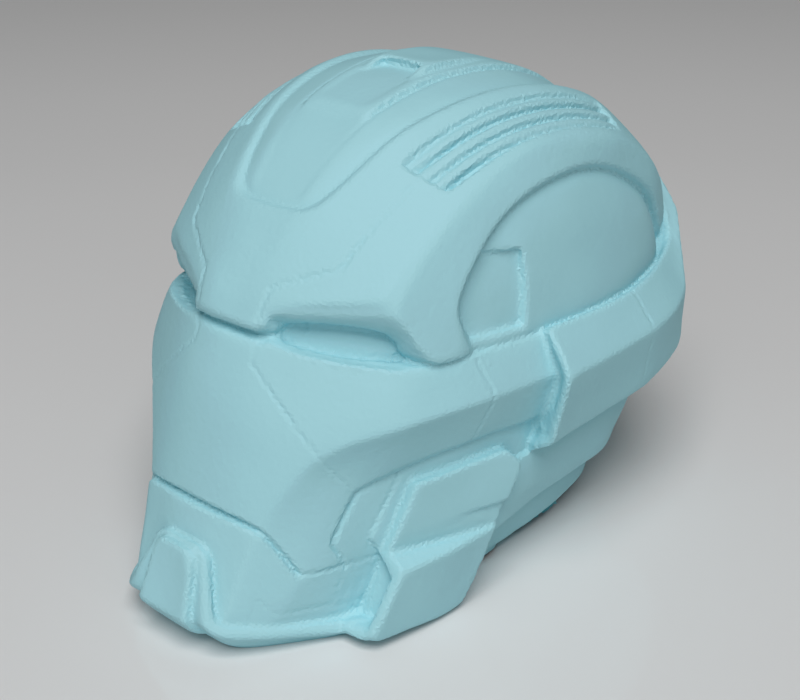}
                 \put(-18,2){\scalebox{.8}{\color{black} A/C}}
            \end{minipage}
            \begin{minipage}{.325\textwidth}
                \includegraphics[width=\textwidth]{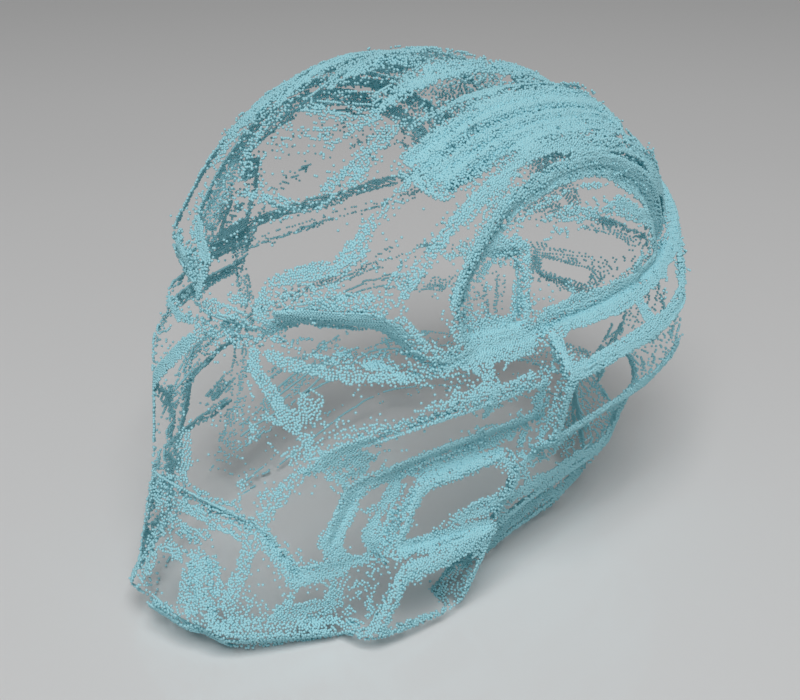}
                \put(-33,2){\scalebox{.8}{\color{black} 60.9/43.6}}
            \end{minipage}
            \begin{minipage}{.325\textwidth}
                \includegraphics[width=\textwidth]{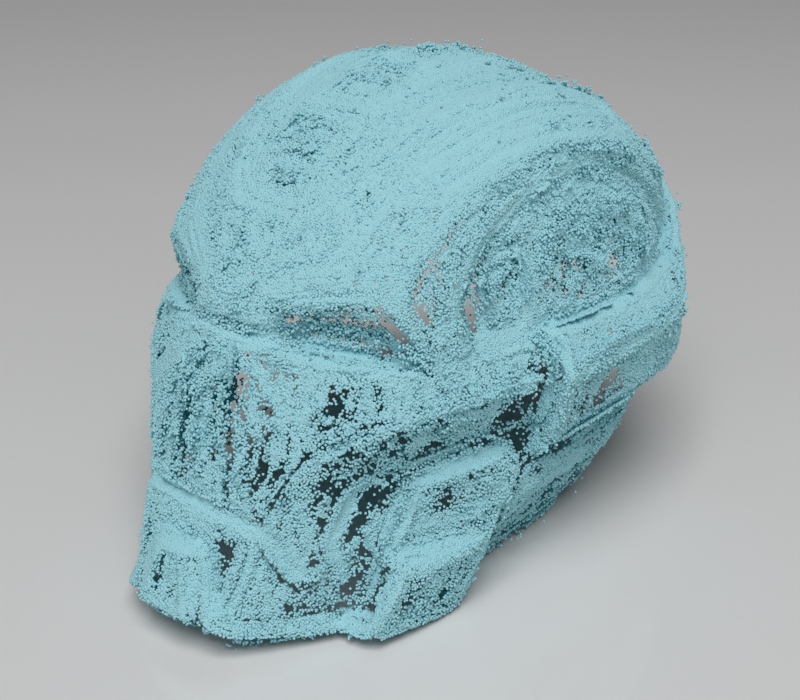}
                \put(-33,2){\scalebox{.8}{\color{black} 69.7/64.9}}
            \end{minipage}
        \end{minipage}
    \end{minipage}
 
    \caption{Boosting another MVS method (COLMAP) with our features. From the left to right: the ground-truth, geometric reconstruction by COLMAP from environment-lit images, and the result by sending our feature maps to COLMAP. Quantitative errors in accuracy/completeness percentage are indicated in the bottom right corner.}
    \label{fig:abl_backend}
\end{figure}

\begin{figure}[h]
    \begin{minipage}{\linewidth}
        \begin{minipage}{0.03in}
            \hspace{0.03in}
        \end{minipage}	
        \begin{minipage}{\textwidth}
            \centering
            \begin{minipage}{\textwidth}
                \centering
                \begin{minipage}{.325\textwidth}
                    \centering
                    \subcaption*{\small Ground-truth}
                \end{minipage}
                \begin{minipage}{.325\textwidth}
                    \centering
                    \subcaption*{\small Scan\#1}
                \end{minipage}
                \begin{minipage}{.325\textwidth}
                    \centering
                    \subcaption*{\small Scan\#2}
                \end{minipage}
            \end{minipage}
        \end{minipage}       
    \end{minipage}

    \begin{minipage}{\linewidth}    
        \begin{minipage}{\textwidth}
            \centering
            \begin{minipage}{.325\textwidth}
                \includegraphics[width=\textwidth]{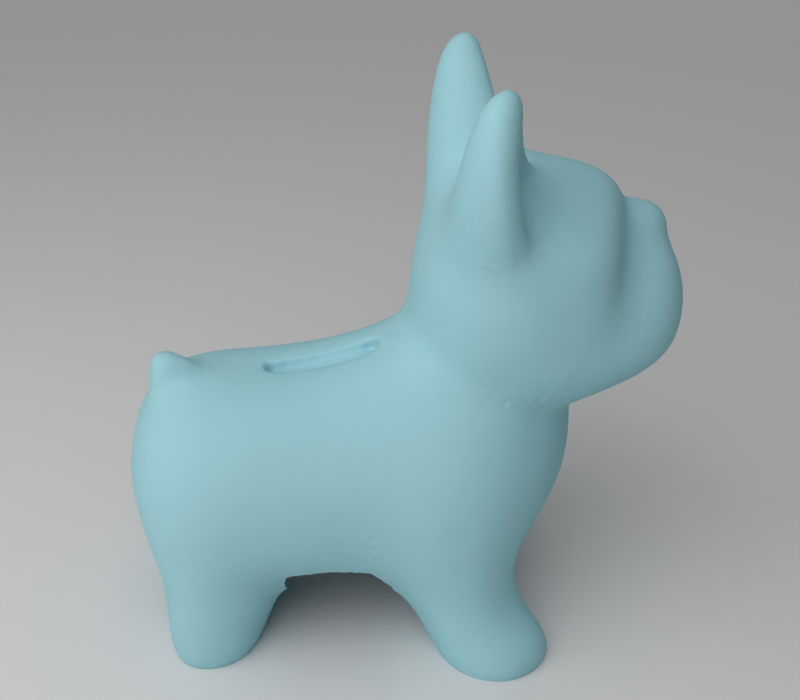}
            \end{minipage}
            \begin{minipage}{.325\textwidth}
                \includegraphics[width=\textwidth]{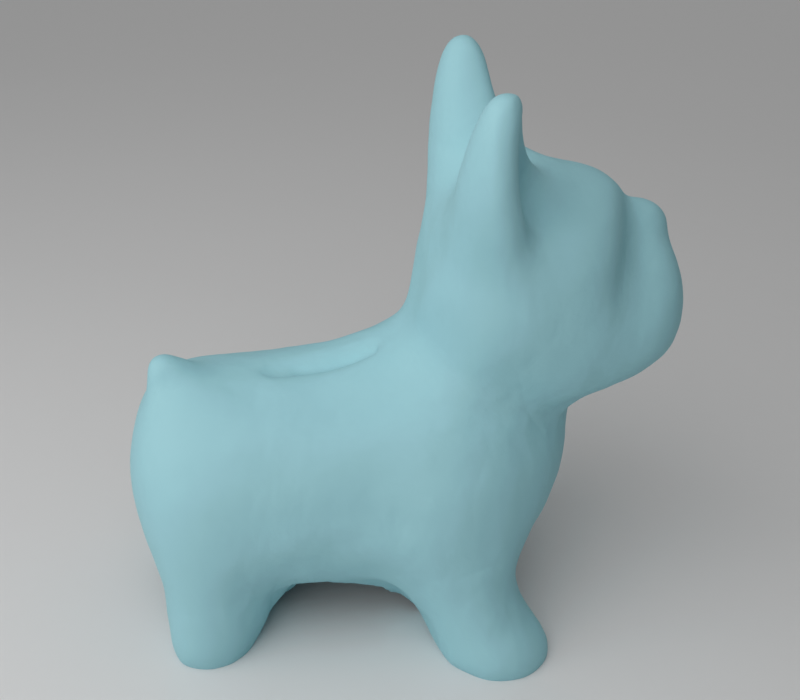}
                \put(-13,3){\scalebox{.8}{\color{black} 2.9}}
            \end{minipage}
            \begin{minipage}{.325\textwidth}
                \includegraphics[width=\textwidth]{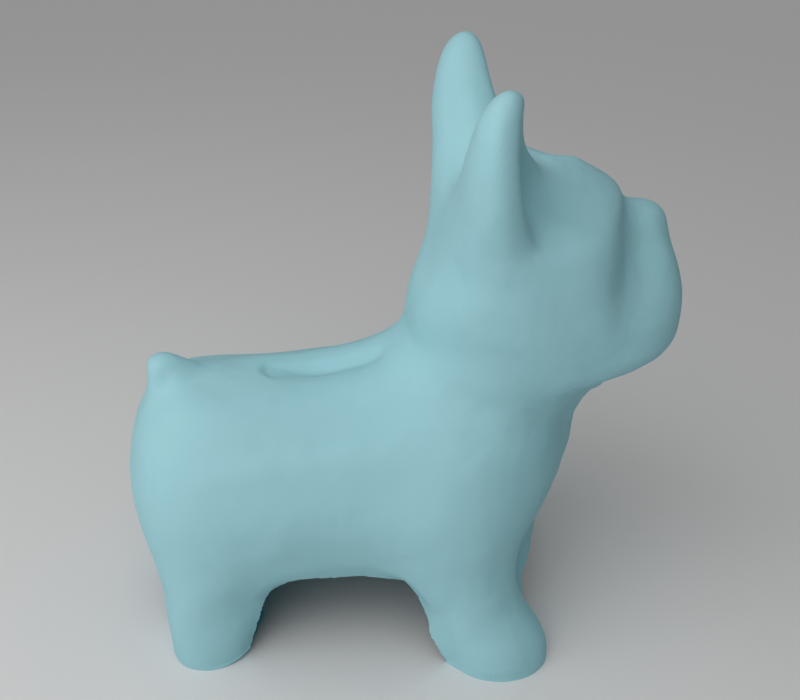}
                \put(-13,3){\scalebox{.8}{\color{black} 3.5}}
            \end{minipage}
        \end{minipage}
    \end{minipage}
    \caption{Repeatability experiment on our geometric reconstruction from 2 scans by 2 different students. Quantitative errors in Chamfer distance are reported in the bottom right corner.}
    \label{fig:repeat}
\end{figure}

We further demonstrate the modular property of our features, by applying them to boost 3D reconstruction with a different backend, COLMAP. We test on a synthetic object with a homogeneous shiny material. In~\figref{fig:abl_backend}, we employ COLMAP to reconstruct a 3D shape from images of the object rendered under environment lighting, as well as feature maps computed with our network from the same number of images under learned lighting patterns. Considerable quality improvement is shown with the help of our learned features. Finally, we test the repeatability of our approach. Two students, who are not involved in this project, are asked to independently capture about the same number of photographs of the same object with our prototype. The reconstructed shapes, shown in~\figref{fig:repeat}, are visually similar.

\section{Limitations \& Future Work}
Our work is subject to a number of limitations. First, our current imaging pipeline does not account for global illumination effects like inter-reflections. Also our system cannot capture transparent/translucent objects, which require special processing. In addition, we need a dark room for high signal-to-noise ratio acquisition, as the uncontrolled environment illumination is not modeled.

It will be interesting future work to address the above limitations. We are also interested in combing our system with a differentiable appearance scanner~\cite{Ma:2021:TRACE}, to optimize lighting patterns for the acquisition of both shape and reflectance. It is also promising to combine our acquisition framework with an efficient and expressive neural representation~\cite{bi2024rgs}. Finally, we expect that the reconstruction quality could be further improved, if future tablets could offer APIs that enable hardware camera-screen synchronization.

 \ifCLASSOPTIONcompsoc
   \section*{Acknowledgments}
 \else
   \section*{Acknowledgment}
 \fi
 
The authors would like to thank Chong Zeng, Xiaohe Ma and Yizhong Zhang for their generous help and support. This work is partially supported by NSF China (62332015, 62227806 \& 62421003), the Fundamental Research Funds for the Central Universities (226-2023-00145), the XPLORER PRIZE and Information Technology Center and State Key Lab of CAD\&CG, Zhejiang University.

\ifCLASSOPTIONcaptionsoff
  \newpage
\fi

\bibliographystyle{IEEEtran}
\bibliography{IEEEabrv,paper}

\begin{thebibliography}{10}
\providecommand{\url}[1]{#1}
\csname url@samestyle\endcsname
\providecommand{\newblock}{\relax}
\providecommand{\bibinfo}[2]{#2}
\providecommand{\BIBentrySTDinterwordspacing}{\spaceskip=0pt\relax}
\providecommand{\BIBentryALTinterwordstretchfactor}{4}
\providecommand{\BIBentryALTinterwordspacing}{\spaceskip=\fontdimen2\font plus
\BIBentryALTinterwordstretchfactor\fontdimen3\font minus \fontdimen4\font\relax}
\providecommand{\BIBforeignlanguage}[2]{{%
\expandafter\ifx\csname l@#1\endcsname\relax
\typeout{** WARNING: IEEEtran.bst: No hyphenation pattern has been}%
\typeout{** loaded for the language `#1'. Using the pattern for}%
\typeout{** the default language instead.}%
\else
\language=\csname l@#1\endcsname
\fi
#2}}
\providecommand{\BIBdecl}{\relax}
\BIBdecl

\bibitem{furukawa2015multi}
Y.~Furukawa, C.~Hern{\'a}ndez \emph{et~al.}, ``Multi-view stereo: A tutorial,'' \emph{Foundations and Trends{\textregistered} in Computer Graphics and Vision}, vol.~9, no. 1-2, pp. 1--148, 2015.

\bibitem{chen2018ps}
G.~Chen, K.~Han, and K.-Y.~K. Wong, ``Ps-fcn: A flexible learning framework for photometric stereo,'' in \emph{Proceedings of the European conference on computer vision (ECCV)}, 2018, pp. 3--18.

\bibitem{woodham1980photometric}
R.~J. Woodham, ``Photometric method for determining surface orientation from multiple images,'' \emph{Optical engineering}, vol.~19, no.~1, p. 191139, 1980.

\bibitem{wang2021neus}
P.~Wang, L.~Liu, Y.~Liu, C.~Theobalt, T.~Komura, and W.~Wang, ``Neus: Learning neural implicit surfaces by volume rendering for multi-view reconstruction,'' \emph{NeurIPS}, 2021.

\bibitem{zeng2023nrhints}
C.~Zeng, G.~Chen, Y.~Dong, P.~Peers, H.~Wu, and X.~Tong, ``Relighting neural radiance fields with shadow and highlight hints,'' in \emph{ACM SIGGRAPH 2023 Conference Proceedings}, 2023.

\bibitem{li2023neuralangelo}
Z.~Li, T.~M\"uller, A.~Evans, R.~H. Taylor, M.~Unberath, M.-Y. Liu, and C.-H. Lin, ``Neuralangelo: High-fidelity neural surface reconstruction,'' in \emph{IEEE Conference on Computer Vision and Pattern Recognition ({CVPR})}, 2023.

\bibitem{zhang2021nerfactor}
X.~Zhang, P.~P. Srinivasan, B.~Deng, P.~Debevec, W.~T. Freeman, and J.~T. Barron, ``Nerfactor: Neural factorization of shape and reflectance under an unknown illumination,'' \emph{ACM Transactions on Graphics (TOG)}, vol.~40, no.~6, pp. 1--18, 2021.

\bibitem{wenger2005performance}
A.~Wenger, A.~Gardner, C.~Tchou, J.~Unger, T.~Hawkins, and P.~Debevec, ``Performance relighting and reflectance transformation with time-multiplexed illumination,'' \emph{ACM Transactions on Graphics (TOG)}, vol.~24, no.~3, pp. 756--764, 2005.

\bibitem{Ma:2021:TRACE}
\BIBentryALTinterwordspacing
X.~Ma, K.~Kang, R.~Zhu, H.~Wu, and K.~Zhou, ``Free-form scanning of non-planar appearance with neural trace photography,'' \emph{ACM Trans. Graph.}, vol.~40, no.~4, Jul. 2021. [Online]. Available: \url{https://doi.org/10.1145/3450626.3459679}
\BIBentrySTDinterwordspacing

\bibitem{shi2016benchmark}
B.~Shi, Z.~Wu, Z.~Mo, D.~Duan, S.-K. Yeung, and P.~Tan, ``A benchmark dataset and evaluation for non-lambertian and uncalibrated photometric stereo,'' in \emph{CVPR}, 2016, pp. 3707--3716.

\bibitem{alldrin2008photometric}
N.~Alldrin, T.~Zickler, and D.~Kriegman, ``Photometric stereo with non-parametric and spatially-varying reflectance,'' in \emph{CVPR}, 2008, pp. 1--8.

\bibitem{goldman2009shape}
D.~B. Goldman, B.~Curless, A.~Hertzmann, and S.~M. Seitz, ``Shape and spatially-varying brdfs from photometric stereo,'' \emph{TPAMI}, vol.~32, no.~6, pp. 1060--1071, 2009.

\bibitem{shi2012elevation}
B.~Shi, P.~Tan, Y.~Matsushita, and K.~Ikeuchi, ``Elevation angle from reflectance monotonicity: Photometric stereo for general isotropic reflectances,'' in \emph{ECCV}.\hskip 1em plus 0.5em minus 0.4em\relax Springer, 2012, pp. 455--468.

\bibitem{alldrin2007resolving}
N.~G. Alldrin, S.~P. Mallick, and D.~J. Kriegman, ``Resolving the generalized bas-relief ambiguity by entropy minimization,'' in \emph{CVPR}, 2007, pp. 1--7.

\bibitem{basri2007photometric}
R.~Basri, D.~Jacobs, and I.~Kemelmacher, ``Photometric stereo with general, unknown lighting,'' \emph{IJCV}, vol.~72, no.~3, pp. 239--257, 2007.

\bibitem{lu2013uncalibrated}
F.~Lu, Y.~Matsushita, I.~Sato, T.~Okabe, and Y.~Sato, ``Uncalibrated photometric stereo for unknown isotropic reflectances,'' in \emph{CVPR}, 2013, pp. 1490--1497.

\bibitem{hernandez2008multiview}
C.~Hernandez, G.~Vogiatzis, and R.~Cipolla, ``Multiview photometric stereo,'' \emph{TPAMI}, vol.~30, no.~3, pp. 548--554, 2008.

\bibitem{li2020multi}
M.~Li, Z.~Zhou, Z.~Wu, B.~Shi, C.~Diao, and P.~Tan, ``Multi-view photometric stereo: a robust solution and benchmark dataset for spatially varying isotropic materials,'' \emph{IEEE Transactions on Image Processing}, vol.~29, pp. 4159--4173, 2020.

\bibitem{zhou2013multi}
Z.~Zhou, Z.~Wu, and P.~Tan, ``Multi-view photometric stereo with spatially varying isotropic materials,'' in \emph{CVPR}, 2013, pp. 1482--1489.

\bibitem{vlasic2009dynamic}
D.~Vlasic, P.~Peers, I.~Baran, P.~Debevec, J.~Popovi{\'c}, S.~Rusinkiewicz, and W.~Matusik, ``Dynamic shape capture using multi-view photometric stereo,'' in \emph{ACM SIGGRAPH Asia 2009 Papers}, 2009, pp. 1--11.

\bibitem{logothetis2019differential}
F.~Logothetis, R.~Mecca, and R.~Cipolla, ``A differential volumetric approach to multi-view photometric stereo,'' in \emph{Proceedings of the IEEE/CVF International Conference on Computer Vision}, 2019, pp. 1052--1061.

\bibitem{bi2020deep}
S.~Bi, Z.~Xu, K.~Sunkavalli, M.~Ha{\v{s}}an, Y.~Hold-Geoffroy, D.~Kriegman, and R.~Ramamoorthi, ``Deep reflectance volumes: Relightable reconstructions from multi-view photometric images,'' in \emph{Computer Vision--ECCV 2020: 16th European Conference, Glasgow, UK, August 23--28, 2020, Proceedings, Part III 16}.\hskip 1em plus 0.5em minus 0.4em\relax Springer, 2020, pp. 294--311.

\bibitem{yang2022ps}
W.~Yang, G.~Chen, C.~Chen, Z.~Chen, and K.-Y.~K. Wong, ``Ps-nerf: Neural inverse rendering for multi-view photometric stereo,'' in \emph{European Conference on Computer Vision}.\hskip 1em plus 0.5em minus 0.4em\relax Springer, 2022, pp. 266--284.

\bibitem{Kang_2021_ICCV}
K.~Kang, C.~Xie, R.~Zhu, X.~Ma, P.~Tan, H.~Wu, and K.~Zhou, ``Learning efficient photometric feature transform for multi-view stereo,'' in \emph{Proceedings of the IEEE/CVF International Conference on Computer Vision (ICCV)}, October 2021, pp. 5956--5965.

\bibitem{galliani2015massively}
S.~Galliani, K.~Lasinger, and K.~Schindler, ``Massively parallel multiview stereopsis by surface normal diffusion,'' in \emph{ICCV}, 2015, pp. 873--881.

\bibitem{schoenberger2016mvs}
J.~L. Sch\"{o}nberger, E.~Zheng, M.~Pollefeys, and J.-M. Frahm, ``Pixelwise view selection for unstructured multi-view stereo,'' in \emph{ECCV}, 2016.

\bibitem{levoy2000digital}
M.~Levoy, K.~Pulli, B.~Curless, S.~Rusinkiewicz, D.~Koller, L.~Pereira, M.~Ginzton, S.~Anderson, J.~Davis, J.~Ginsberg \emph{et~al.}, ``The digital michelangelo project: 3d scanning of large statues,'' in \emph{Proc. SIGGRAPH}, 2000, pp. 131--144.

\bibitem{salvi2004pattern}
J.~Salvi, J.~Pages, and J.~Batlle, ``Pattern codification strategies in structured light systems,'' \emph{Pattern recognition}, vol.~37, no.~4, pp. 827--849, 2004.

\bibitem{simonyan2014learning}
K.~Simonyan, A.~Vedaldi, and A.~Zisserman, ``Learning local feature descriptors using convex optimisation,'' \emph{TPAMI}, vol.~36, no.~8, pp. 1573--1585, 2014.

\bibitem{zagoruyko2015learning}
S.~Zagoruyko and N.~Komodakis, ``Learning to compare image patches via convolutional neural networks,'' in \emph{CVPR}, 2015, pp. 4353--4361.

\bibitem{luo2016efficient}
W.~Luo, A.~G. Schwing, and R.~Urtasun, ``Efficient deep learning for stereo matching,'' in \emph{Proceedings of the IEEE conference on computer vision and pattern recognition}, 2016, pp. 5695--5703.

\bibitem{yao2018mvsnet}
Y.~Yao, Z.~Luo, S.~Li, T.~Fang, and L.~Quan, ``Mvsnet: Depth inference for unstructured multi-view stereo,'' in \emph{Proceedings of the European conference on computer vision (ECCV)}, 2018, pp. 767--783.

\bibitem{gu2020cascade}
X.~Gu, Z.~Fan, S.~Zhu, Z.~Dai, F.~Tan, and P.~Tan, ``Cascade cost volume for high-resolution multi-view stereo and stereo matching,'' in \emph{Proceedings of the IEEE/CVF Conference on Computer Vision and Pattern Recognition}, 2020, pp. 2495--2504.

\bibitem{munkberg2022extracting}
J.~Munkberg, J.~Hasselgren, T.~Shen, J.~Gao, W.~Chen, A.~Evans, T.~M{\"u}ller, and S.~Fidler, ``Extracting triangular 3d models, materials, and lighting from images,'' in \emph{Proceedings of the IEEE/CVF Conference on Computer Vision and Pattern Recognition}, 2022, pp. 8280--8290.

\bibitem{yariv2020multiview}
L.~Yariv, Y.~Kasten, D.~Moran, M.~Galun, M.~Atzmon, B.~Ronen, and Y.~Lipman, ``Multiview neural surface reconstruction by disentangling geometry and appearance,'' \emph{Advances in Neural Information Processing Systems}, vol.~33, pp. 2492--2502, 2020.

\bibitem{yariv2021volume}
L.~Yariv, J.~Gu, Y.~Kasten, and Y.~Lipman, ``Volume rendering of neural implicit surfaces,'' in \emph{Thirty-Fifth Conference on Neural Information Processing Systems}, 2021.

\bibitem{luan2021unified}
F.~Luan, S.~Zhao, K.~Bala, and Z.~Dong, ``Unified shape and svbrdf recovery using differentiable monte carlo rendering,'' in \emph{Computer Graphics Forum}, vol.~40, no.~4.\hskip 1em plus 0.5em minus 0.4em\relax Wiley Online Library, 2021, pp. 101--113.

\bibitem{nam2018practical}
G.~Nam, J.~H. Lee, D.~Gutierrez, and M.~H. Kim, ``Practical svbrdf acquisition of 3d objects with unstructured flash photography,'' \emph{ACM Transactions on Graphics (TOG)}, vol.~37, no.~6, pp. 1--12, 2018.

\bibitem{iron-2022}
K.~Zhang, F.~Luan, Z.~Li, and N.~Snavely, ``Iron: Inverse rendering by optimizing neural sdfs and materials from photometric images,'' in \emph{IEEE Conf. Comput. Vis. Pattern Recog.}, 2022.

\bibitem{srinivasan2021nerv}
P.~P. Srinivasan, B.~Deng, X.~Zhang, M.~Tancik, B.~Mildenhall, and J.~T. Barron, ``Nerv: Neural reflectance and visibility fields for relighting and view synthesis,'' in \emph{Proceedings of the IEEE/CVF Conference on Computer Vision and Pattern Recognition}, 2021, pp. 7495--7504.

\bibitem{zhang2021physg}
K.~Zhang, F.~Luan, Q.~Wang, K.~Bala, and N.~Snavely, ``Physg: Inverse rendering with spherical gaussians for physics-based material editing and relighting,'' in \emph{Proceedings of the IEEE/CVF Conference on Computer Vision and Pattern Recognition}, 2021, pp. 5453--5462.

\bibitem{boss2021nerd}
M.~Boss, R.~Braun, V.~Jampani, J.~T. Barron, C.~Liu, and H.~Lensch, ``Nerd: Neural reflectance decomposition from image collections,'' in \emph{Proceedings of the IEEE/CVF International Conference on Computer Vision}, 2021, pp. 12\,684--12\,694.

\bibitem{liu2023nero}
Y.~Liu, P.~Wang, C.~Lin, X.~Long, J.~Wang, L.~Liu, T.~Komura, and W.~Wang, ``Nero: Neural geometry and brdf reconstruction of reflective objects from multiview images,'' in \emph{SIGGRAPH}, 2023.

\bibitem{wu2015simultaneous}
H.~Wu, Z.~Wang, and K.~Zhou, ``Simultaneous localization and appearance estimation with a consumer rgb-d camera,'' \emph{IEEE transactions on visualization and computer graphics}, vol.~22, no.~8, pp. 2012--2023, 2015.

\bibitem{Walter:2007:GGX}
B.~Walter, S.~R. Marschner, H.~Li, and K.~E. Torrance, ``{Microfacet Models for Refraction through Rough Surfaces},'' in \emph{Rendering Techniques (Proc. EGWR)}, 2007.

\bibitem{lensch2003image}
H.~P. Lensch, J.~Kautz, M.~Goesele, W.~Heidrich, and H.-P. Seidel, ``Image-based reconstruction of spatial appearance and geometric detail,'' \emph{ACM Transactions on Graphics (TOG)}, vol.~22, no.~2, pp. 234--257, 2003.

\bibitem{tian2017l2}
Y.~Tian, B.~Fan, and F.~Wu, ``L2-net: Deep learning of discriminative patch descriptor in euclidean space,'' in \emph{Proceedings of the IEEE conference on computer vision and pattern recognition}, 2017, pp. 661--669.

\bibitem{Kang:2019:JOINT}
\BIBentryALTinterwordspacing
K.~Kang, C.~Xie, C.~He, M.~Yi, M.~Gu, Z.~Chen, K.~Zhou, and H.~Wu, ``Learning efficient illumination multiplexing for joint capture of reflectance and shape,'' \emph{ACM Trans. Graph.}, vol.~38, no.~6, pp. 165:1--165:12, Nov. 2019. [Online]. Available: \url{http://doi.acm.org/10.1145/3355089.3356492}
\BIBentrySTDinterwordspacing

\bibitem{Xu:2023:StructuredLight}
X.~X. Xu, L.~Yuxin, H.~Zhou, C.~Zeng, Y.~Yu, K.~Zhou, and H.~Wu, ``A unified spatial-angular structured light for single-view acquisition of shape and reflectance,'' in \emph{CVPR}, 2023.

\bibitem{crete2007blur}
F.~Cr{\'e}t{\'e}-Roffet, T.~Dolmiere, P.~Ladret, and M.~Nicolas, ``The blur effect: Perception and estimation with a new no-reference perceptual blur metric,'' in \emph{SPIE Electronic Imaging Symposium Conf Human Vision and Electronic Imaging}, vol.~12, 2007, pp. EI--6492.

\bibitem{schoenberger2016sfm}
J.~L. Sch\"{o}nberger and J.-M. Frahm, ``Structure-from-motion revisited,'' in \emph{Conference on Computer Vision and Pattern Recognition (CVPR)}, 2016.

\bibitem{fiala2005artag}
M.~Fiala, ``Artag, a fiducial marker system using digital techniques,'' in \emph{2005 IEEE Computer Society Conference on Computer Vision and Pattern Recognition (CVPR'05)}, vol.~2.\hskip 1em plus 0.5em minus 0.4em\relax IEEE, 2005, pp. 590--596.

\bibitem{kirillov2023segany}
A.~Kirillov, E.~Mintun, N.~Ravi, H.~Mao, C.~Rolland, L.~Gustafson, T.~Xiao, S.~Whitehead, A.~C. Berg, W.-Y. Lo, P.~Doll{\'a}r, and R.~Girshick, ``Segment anything,'' \emph{arXiv:2304.02643}, 2023.

\bibitem{ge2023ref}
W.~Ge, T.~Hu, H.~Zhao, S.~Liu, and Y.-C. Chen, ``Ref-neus: Ambiguity-reduced neural implicit surface learning for multi-view reconstruction with reflection,'' \emph{arXiv preprint arXiv:2303.10840}, 2023.

\bibitem{ma2023opensvbrdf}
X.~Ma, X.~Xu, L.~Zhang, K.~Zhou, and H.~Wu, ``Opensvbrdf: A database of measured spatially-varying reflectance,'' \emph{ACM Transactions on Graphics (TOG)}, vol.~42, no.~6, pp. 1--14, 2023.

\bibitem{shining3D}
Shining3D, ``Einscan pro 2x plus handheld industrial scanner,'' \url{https://www.einscan.com/handheld-3d-scanner/2x-plus/}, 2023, [Online; accessed May-2023].

\bibitem{bi2024rgs}
Z.~Bi, Y.~Zeng, C.~Zeng, F.~Pei, X.~Feng, K.~Zhou, and H.~Wu, ``Gs\textsuperscript{3}: Efficient relighting with triple gaussian splatting,'' in \emph{SIGGRAPH Asia 2024 Conference Papers}, 2024.

\end{thebibliography}

\begin{IEEEbiography}[{\includegraphics[width=1in,height=1.25in,clip,keepaspectratio]{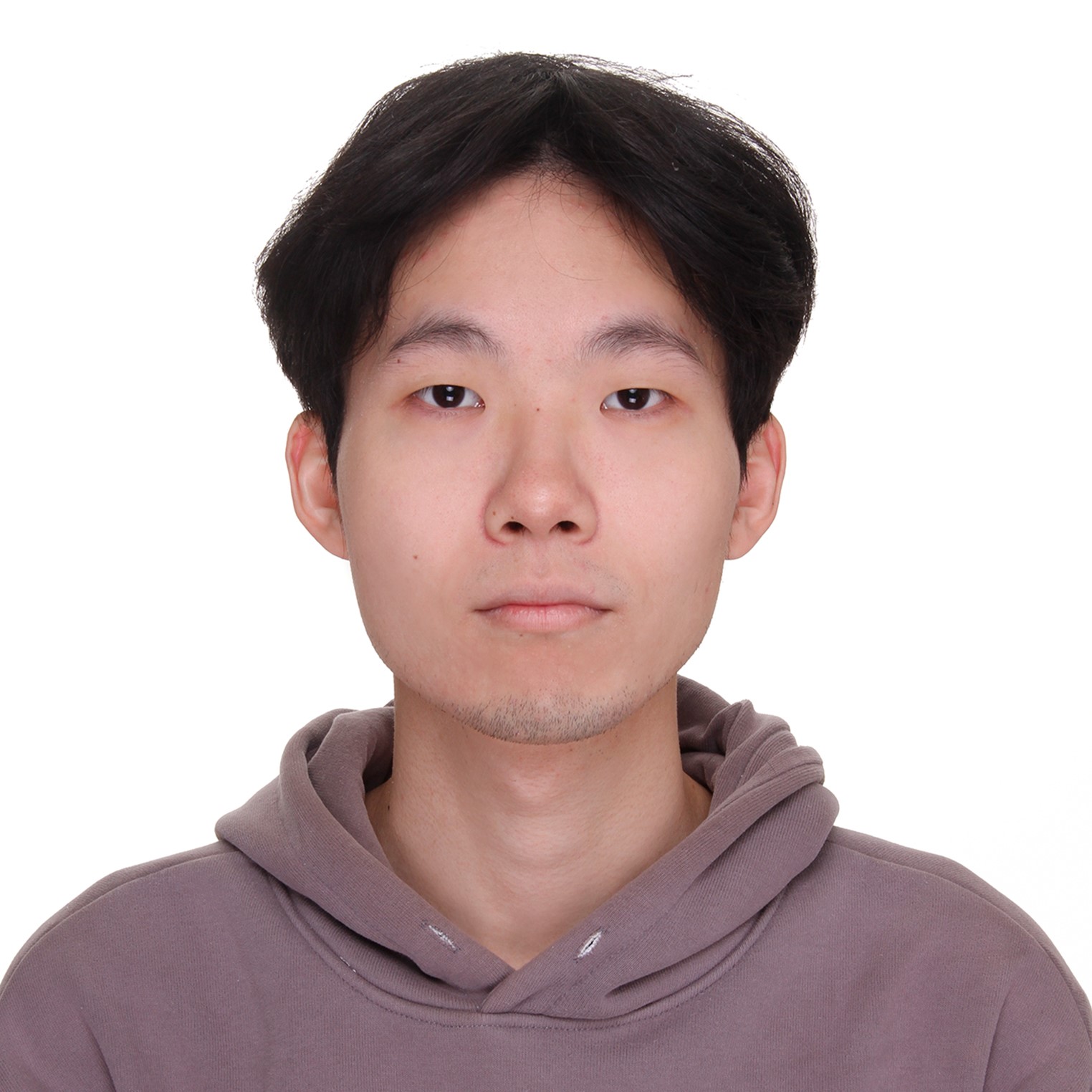}}]{Xiang Feng} is a master student in the State Key Lab of CAD~\&~CG, Zhejiang University. He received his B.Eng. in computer science from the same university in 2022. His research interests include generation/reconstruction of appearance and shape.
\end{IEEEbiography}

\begin{IEEEbiography}[{\includegraphics[width=1in,height=1.25in,clip,keepaspectratio]{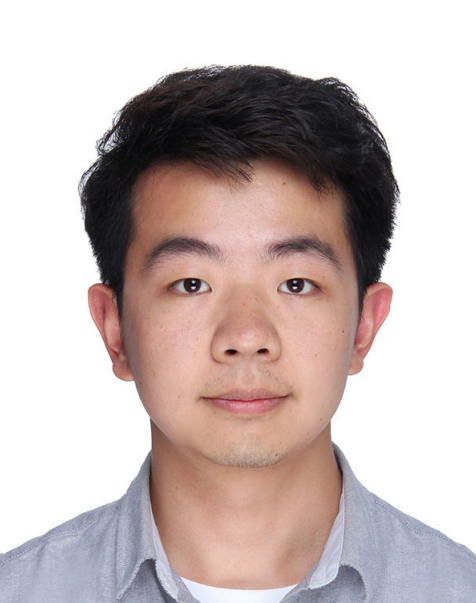}}]{Kaizhang Kang} is a postdoctoral researcher at King Abdullah University of Science and Technology (KAUST). He received Ph.D. degree from Zhejiang University. His research focuses on the acquisition and reconstruction of physical information, including high-dimensional appearance, 3D geometry and volumes. He received Microsoft Asia Fellowship in 2021.
\end{IEEEbiography}

\begin{IEEEbiography}[{\includegraphics[width=1in,height=1.25in,clip,keepaspectratio]{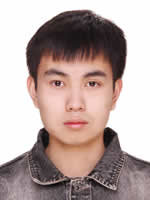}}]{Fan Pei} is a master student in the State Key Lab of CAD~\&~CG, Zhejiang University. He received his B.Eng. from Zhejiang university in 2022. His research interests include
3D feature learning and appearance acquisition.
\end{IEEEbiography}

\begin{IEEEbiography}[{\includegraphics[width=1in,height=1.25in,clip,keepaspectratio]{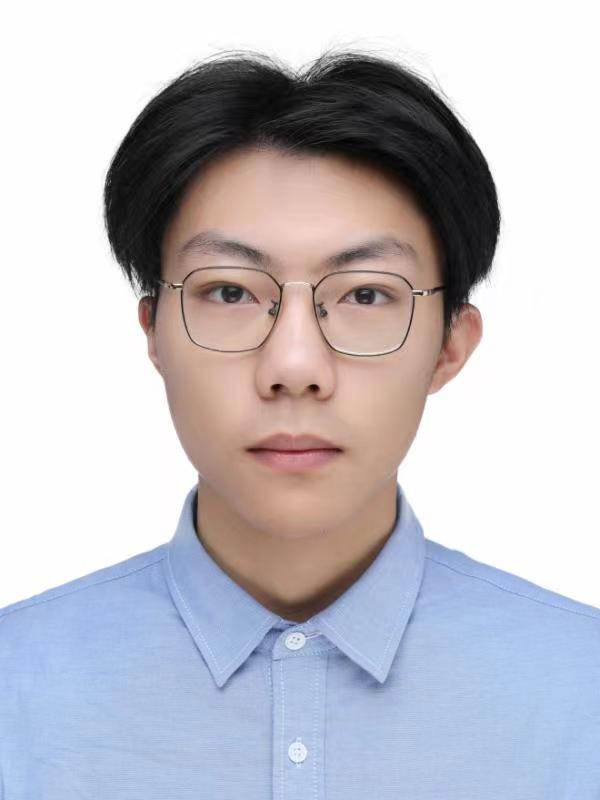}}]{Huakeng Ding} is an undergraduate student in the State Key Lab of CAD~\&~CG, Zhejiang University. His research interests include appearance acquisition and shape reconstruction.
\end{IEEEbiography}

\begin{IEEEbiography}[{\includegraphics[width=1in,keepaspectratio,clip,trim={0cm 10.0cm 0cm 0.25cm}]{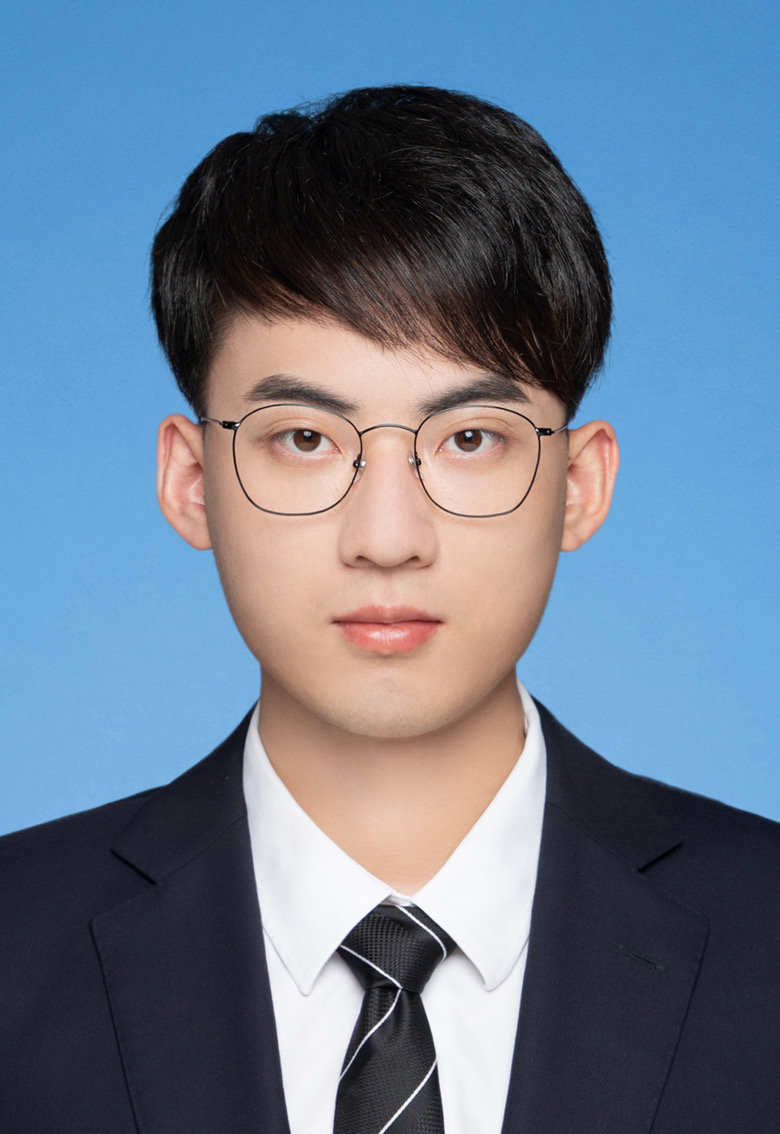}}]{Jinjiang You} is a master student in the Carnegie Mellon University. He received his B.Eng. in computer science from Zhejiang University in 2023. His research interests include appearance acquisition and 3d vision.
\end{IEEEbiography}

\begin{IEEEbiography}[{\includegraphics[width=1in,height=1.25in,clip,keepaspectratio]{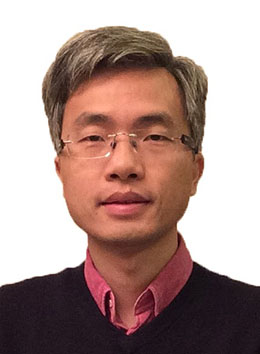}}]{Ping Tan} is a Professor in Department of Electronic and Computer Engineering at  Hong Kong University of Science and Technology. He has previously served as the head of XR Lab at Alibaba's DAMO Academy, Chief Scientist for Computer Vision at the Artificial Intelligence Lab, and as an Associate Professor at Simon Fraser University and National University of Singapore. His research areas include computer vision and computer graphics.
\end{IEEEbiography}

\begin{IEEEbiography}[{\includegraphics[width=1in,height=1.25in,clip,keepaspectratio]{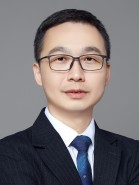}}]{Kun Zhou} is a Cheung Kong Professor in the Computer Science Department of Zhejiang University, and the Director of the State Key Lab of CAD~\&~CG. Prior to that, he was a Leader Researcher of the Internet Graphics Group at Microsoft Research Asia. He received B.S. and Ph.D. in computer science from Zhejiang University. His research interests are in visual computing, parallel computing, human computer interaction, and virtual reality. 
\end{IEEEbiography}

\begin{IEEEbiography}[{\includegraphics[width=1in,height=1.25in,clip,keepaspectratio]{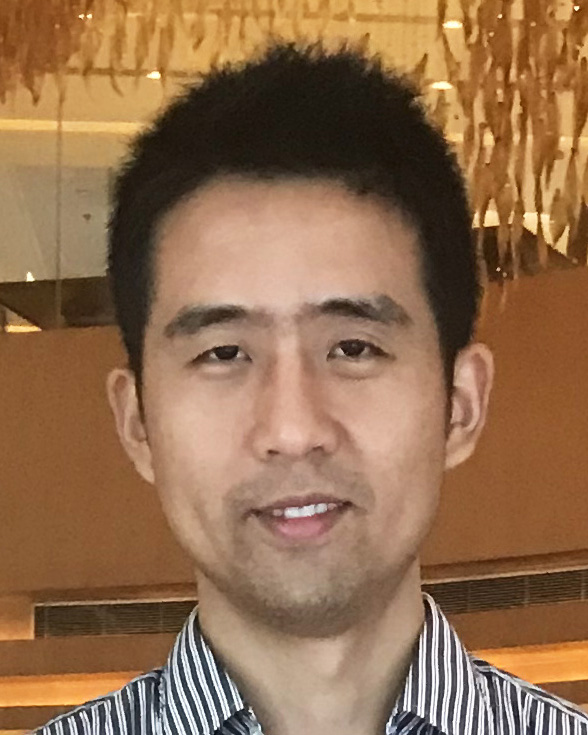}}]{Hongzhi Wu} is a professor in the State Key Lab of CAD~\&~CG, Zhejiang University. He received B.Sc. in computer science from Fudan University, and Ph.D. in computer science from Yale University. His current research interests include high-density illumination multiplexing devices and differentiable acquisition. Hongzhi is a recipient of Excellent Young Scholars, NSF China. He is on the editorial board of IEEE TVCG.
\end{IEEEbiography}

\clearpage
\thispagestyle{empty}

\section{Supplementary Material}

\subsection{Details on Geometric Reconstruction}

Once our approach convert the groups of input images into feature maps, they are directly fed as input to existing multi-view stereo techniques (be it COLMAP, NeuS or Neuralangelo), with only minor modifications: the number of channels of an input image is changed from 3 (RGB) to 12, to match the dimension of our features. The objective function stays the same as in respective reconstruction approaches. Moreover, for methods based on inverse rendering such as NeuS or Neuralangelo, we further modify their rendering process to produce 12-channel output images, in order to compute the loss against our input feature maps. This is done by changing the output dimension of the final fully connected layer in NeuS/Neuralangelo from 3 to 12.

\subsection{Details on Appearance Reconstruction}

After geometric reconstruction, we establish a uv-parame-terization over object surfaces, and compute BRDF parameters at each valid texel via differentiable optimization. While not being tied to any specific model, we adopt the anisotropic GGX BRDF in this paper:
\begin{align}
f(\mathbf{\omega_{i}};& \mathbf{\omega_{o}}, \mathbf{p})=\frac{\rho_d}{\pi}+ \notag\\
&\rho_s\frac{D_{GGX}(\mathbf{\omega_{h}};\alpha_x,\alpha_y)F(\mathbf{\omega_{i}},\mathbf{\omega_{h}})G_{GGX}(\mathbf{\omega_{i}}, \mathbf{\omega_{o}};\alpha_x,\alpha_y)}{4(\mathbf{\omega_{i}}\cdot\mathbf{n_p})(\mathbf{\omega_{o}}\cdot\mathbf{n_p})}.\notag
\end{align}
Here $\rho_d$/$\rho_s$ are the diffuse/specular albedo, $\alpha_x$/$\alpha_y$ are the roughness parameters, and $\mathbf{\omega_{h}}$ is the half vector. $D_{GGX}$ is the microfacet distribution function, $F$ is the Fresnel term and $G_{GGX}$ accounts for shadowing/masking effects. The BRDF model is defined in the local frame $\mathbf{n_p}$/$\mathbf{t_p}$ of $\mathbf{p}$, where $\mathbf{n_p}$/$\mathbf{t_p}$ are the normal and tangent, respectively.

\begin{figure}[htbp]
    \centering
    \begin{minipage}{\linewidth}
        \begin{minipage}{0.08in}
            \hspace{0.08in}
        \end{minipage}	
        \begin{minipage}{.983\linewidth}
            \centering
            \begin{minipage}{\textwidth}
                \centering
                \begin{minipage}{.24\textwidth}
                    \centering
                    \subcaption*{\small Diffuse}
                \end{minipage}
                \begin{minipage}{.24\textwidth}
                    \centering
                    \subcaption*{\small Specular}
                \end{minipage}
                \begin{minipage}{.24\textwidth}
                    \centering
                    \subcaption*{\small Roughness}
                \end{minipage}
                \begin{minipage}{.24\textwidth}
                    \centering
                    \subcaption*{\small Tangent}
                \end{minipage}
            \end{minipage}
        \end{minipage}       
    \end{minipage}

    \begin{minipage}{\linewidth}
        \begin{minipage}{0.08in}	
            \centering
            \rotatebox{90}{\small \textsc{Bowl}}
        \end{minipage}
        \begin{minipage}{.983\linewidth}
            \centering
            \begin{minipage}{.24\textwidth}
                \includegraphics[width=\textwidth]{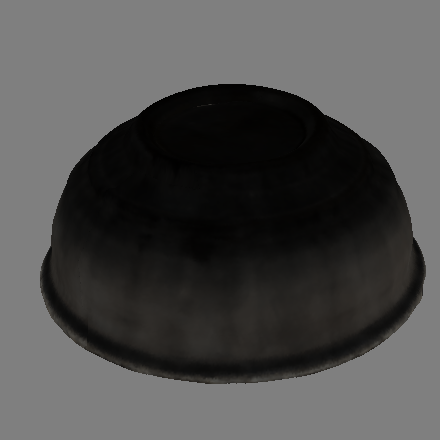 }
            \end{minipage}
            \begin{minipage}{.24\textwidth}
                \includegraphics[width=\textwidth]{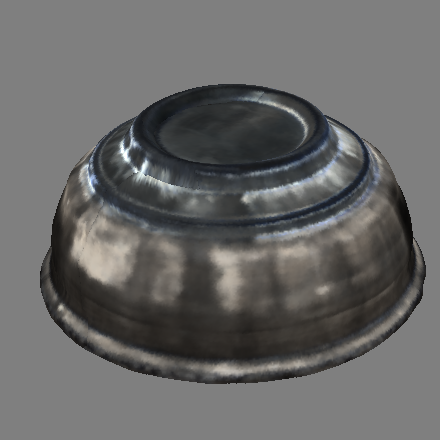}
            \end{minipage}
            \begin{minipage}{.24\textwidth}
                \includegraphics[width=\textwidth]{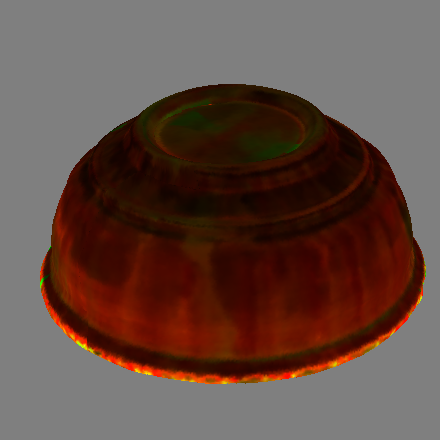}
            \end{minipage}
            \begin{minipage}{.24\textwidth}
                \includegraphics[width=\textwidth]{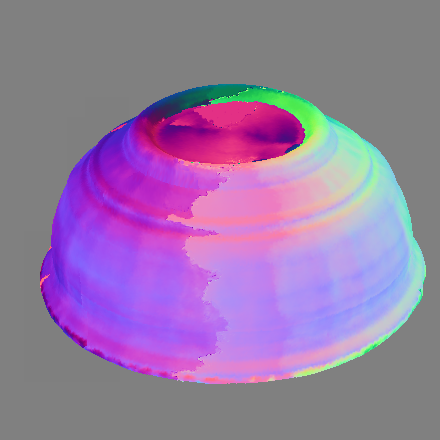}
            \end{minipage}
        \end{minipage}
    \end{minipage}

    \begin{minipage}{\linewidth}
        \begin{minipage}{0.08in}	
            \centering
            \rotatebox{90}{\small \textsc{Dog}}
        \end{minipage}
        \begin{minipage}{.983\linewidth}
            \centering
            \begin{minipage}{.24\textwidth}
                \includegraphics[width=\textwidth]{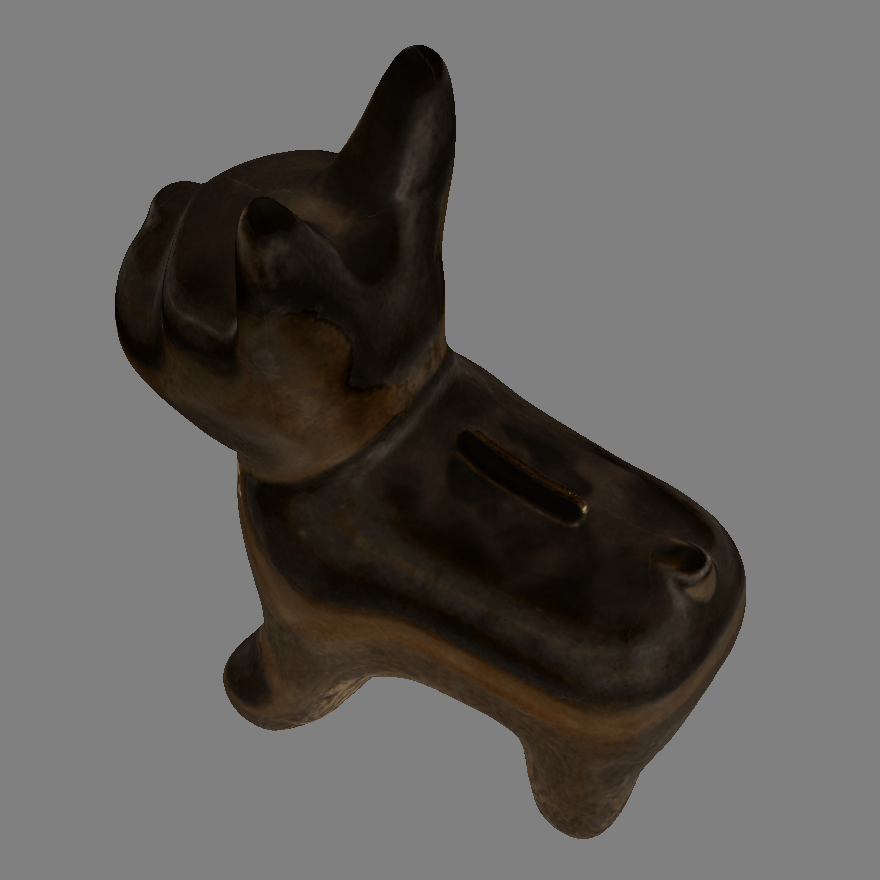 }
            \end{minipage}
            \begin{minipage}{.24\textwidth}
                \includegraphics[width=\textwidth]{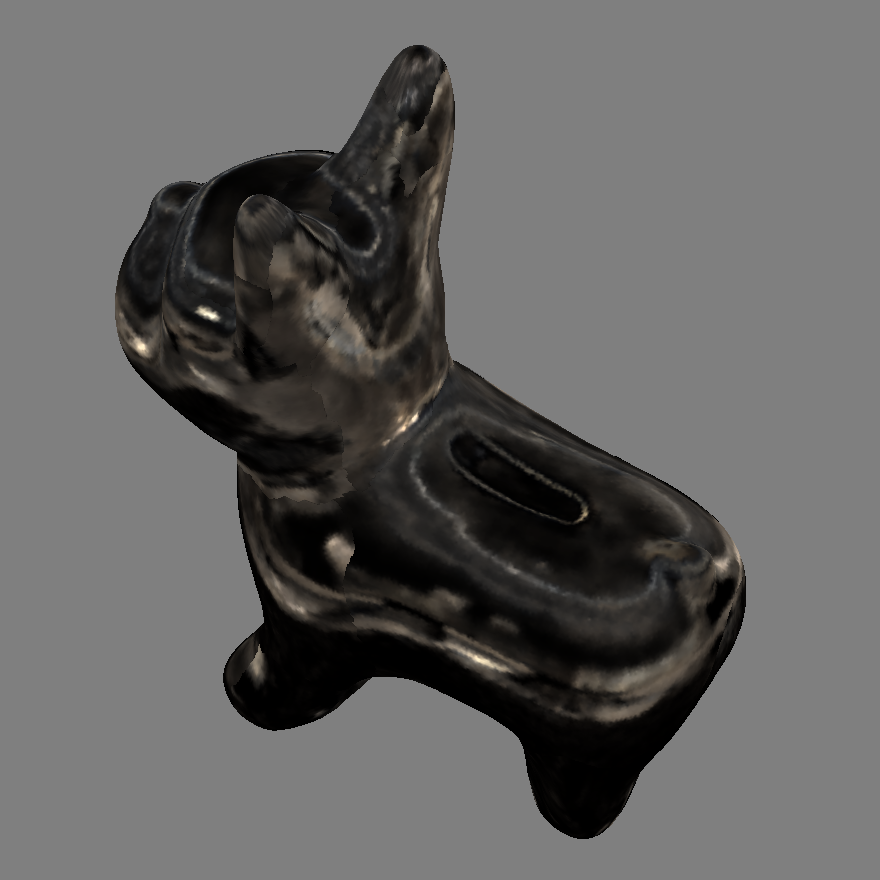}
            \end{minipage}
            \begin{minipage}{.24\textwidth}
                \includegraphics[width=\textwidth]{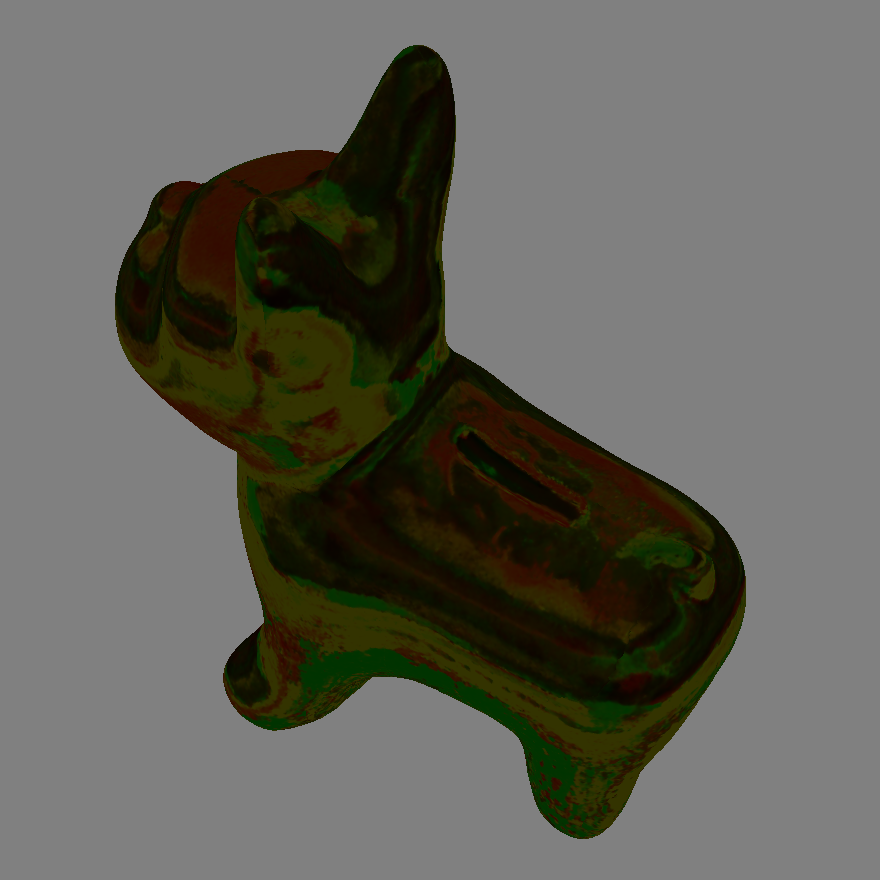}
            \end{minipage}
            \begin{minipage}{.24\textwidth}
                \includegraphics[width=\textwidth]{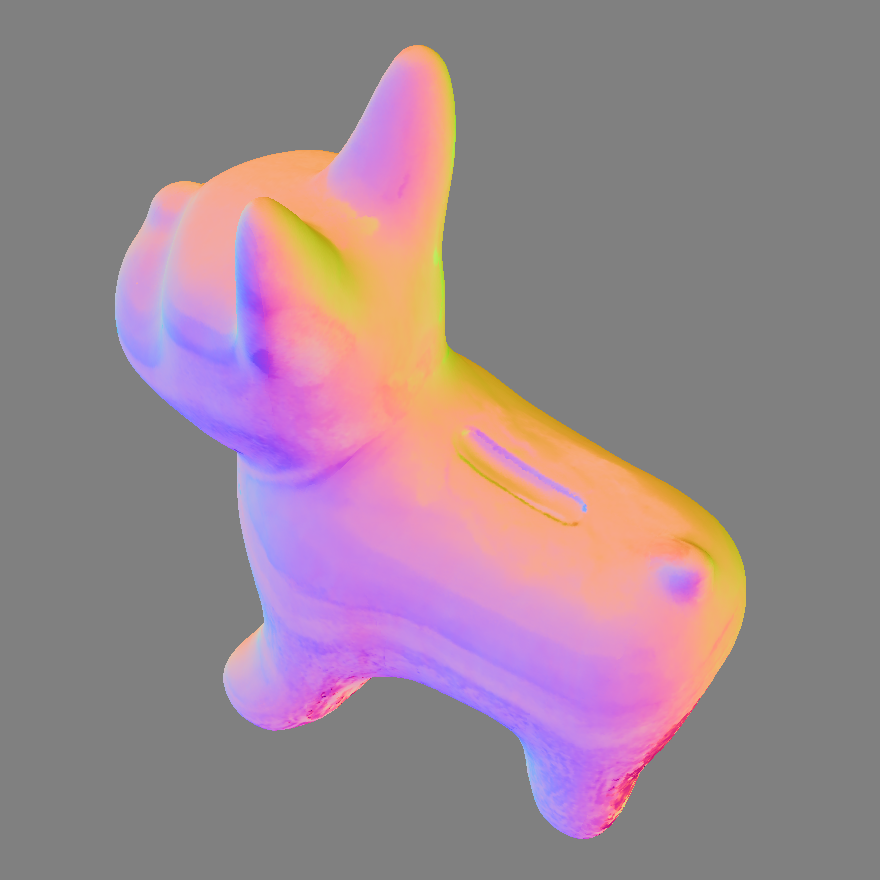}
            \end{minipage}
        \end{minipage}
    \end{minipage}

    \begin{minipage}{\linewidth}
        \begin{minipage}{0.08in}	
            \centering
            \rotatebox{90}{\small \textsc{Bear}}
        \end{minipage}
        \begin{minipage}{.983\linewidth}
            \centering
            \begin{minipage}{.24\textwidth}
                \includegraphics[width=\textwidth]{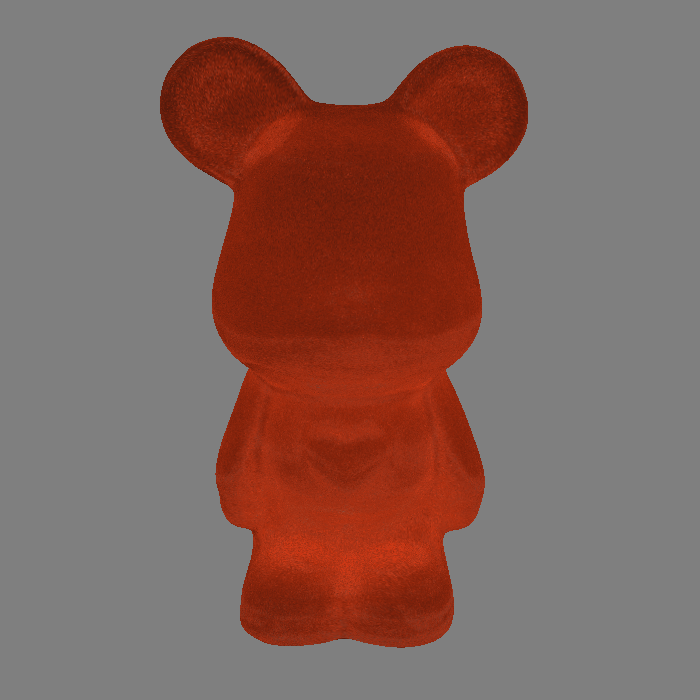}
            \end{minipage}
            \begin{minipage}{.24\textwidth}
                \includegraphics[width=\textwidth]{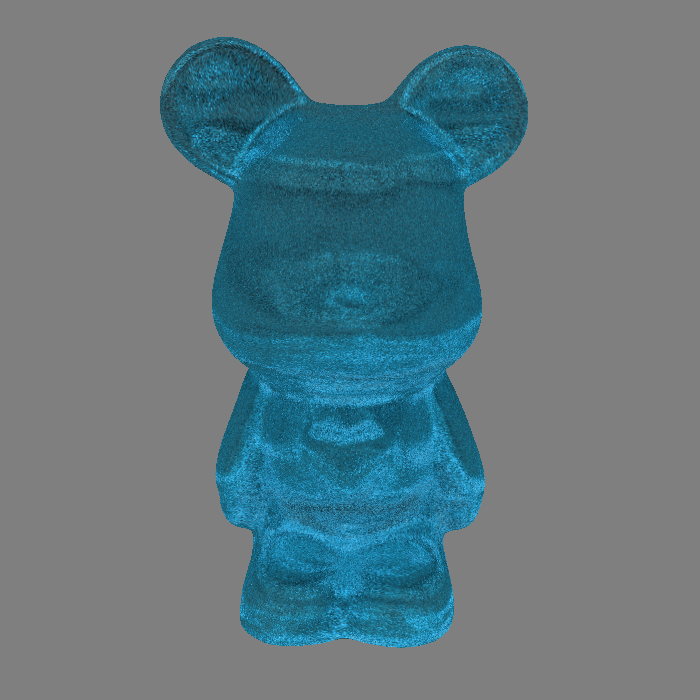}
            \end{minipage}
            \begin{minipage}{.24\textwidth}
                \includegraphics[width=\textwidth]{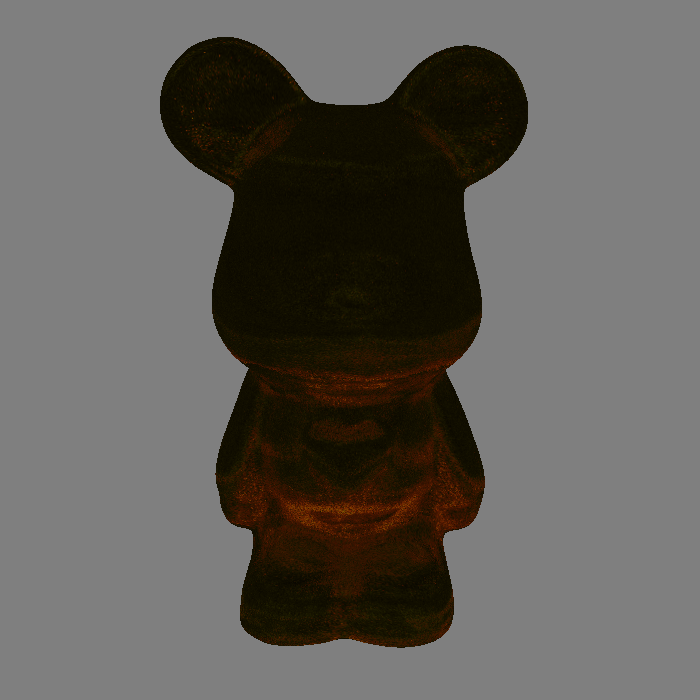}
            \end{minipage}
            \begin{minipage}{.24\textwidth}
                \includegraphics[width=\textwidth]{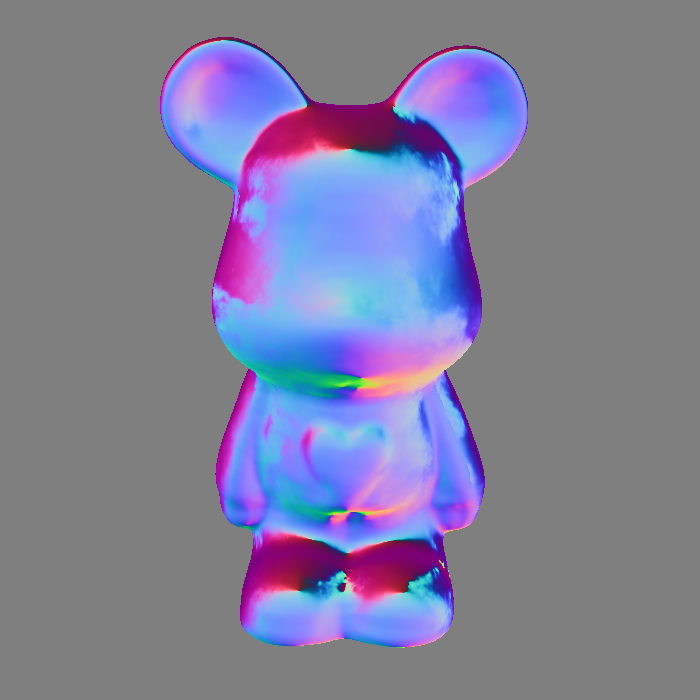}
            \end{minipage}
        \end{minipage}
    \end{minipage}

    \begin{minipage}{\linewidth}
        \begin{minipage}{0.08in}	
            \centering
            \rotatebox{90}{\small \textsc{Bird}}
        \end{minipage}
        \begin{minipage}{.983\linewidth}
            \centering
            \begin{minipage}{.24\textwidth}
                \includegraphics[width=\textwidth]{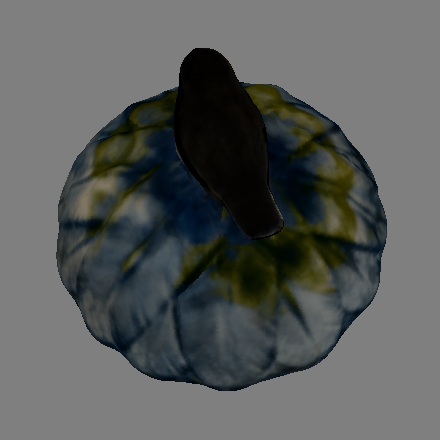 }
            \end{minipage}
            \begin{minipage}{.24\textwidth}
                \includegraphics[width=\textwidth]{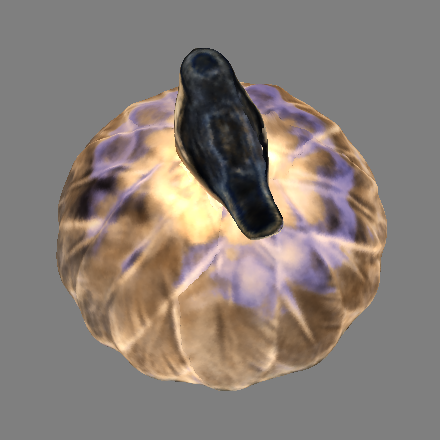}
            \end{minipage}
            \begin{minipage}{.24\textwidth}
                \includegraphics[width=\textwidth]{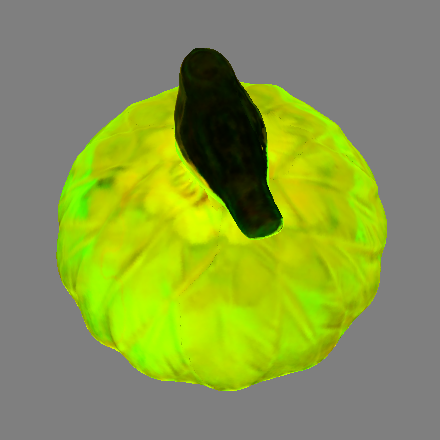}
            \end{minipage}
            \begin{minipage}{.24\textwidth}
                \includegraphics[width=\textwidth]{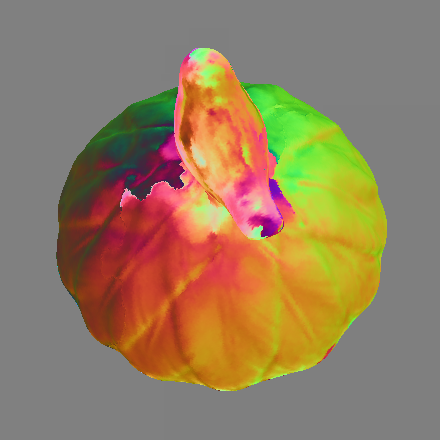}
            \end{minipage}
        \end{minipage}
    \end{minipage}

    \begin{minipage}{\linewidth}
        \begin{minipage}{0.08in}	
            \centering
            \rotatebox{90}{\small \textsc{Dice}}
        \end{minipage}
        \begin{minipage}{.983\linewidth}
            \centering
            \begin{minipage}{.24\textwidth}
                \includegraphics[width=\textwidth]{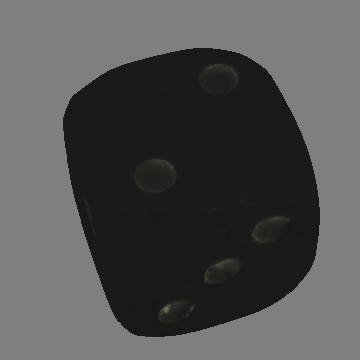}
            \end{minipage}
            \begin{minipage}{.24\textwidth}
                \includegraphics[width=\textwidth]{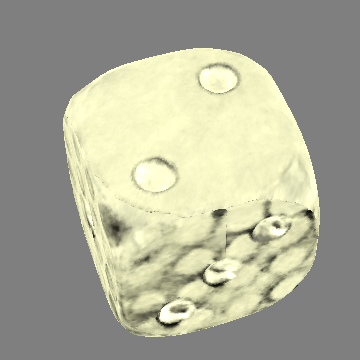}
            \end{minipage}
            \begin{minipage}{.24\textwidth}
                \includegraphics[width=\textwidth]{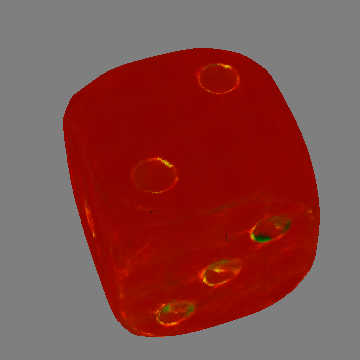}
            \end{minipage}
            \begin{minipage}{.24\textwidth}
                \includegraphics[width=\textwidth]{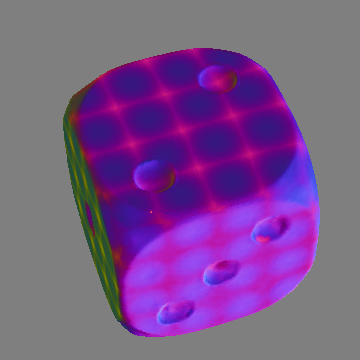}
            \end{minipage}
        \end{minipage}
    \end{minipage}

    \begin{minipage}{\linewidth}
        \begin{minipage}{0.08in}	
            \centering
            \rotatebox{90}{\small \textsc{Cup}}
        \end{minipage}
        \begin{minipage}{.983\linewidth}
            \centering
            \begin{minipage}{.24\textwidth}
                \includegraphics[width=\textwidth]{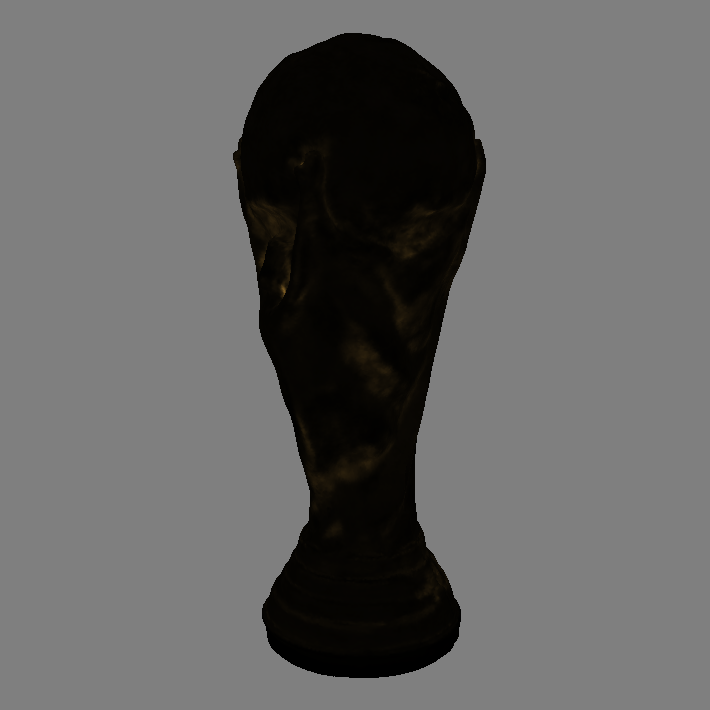 }
            \end{minipage}
            \begin{minipage}{.24\textwidth}
                \includegraphics[width=\textwidth]{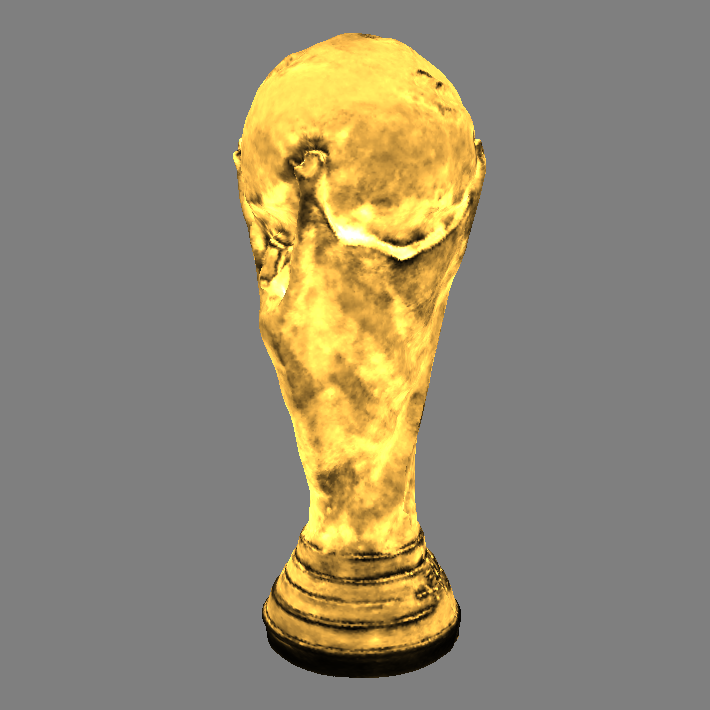}
            \end{minipage}
            \begin{minipage}{.24\textwidth}
                \includegraphics[width=\textwidth]{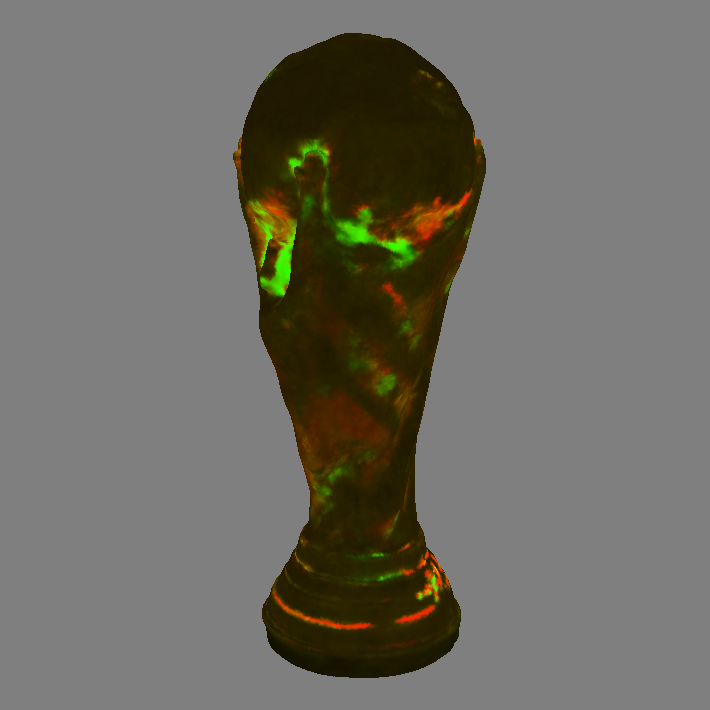}
            \end{minipage}
            \begin{minipage}{.24\textwidth}
                \includegraphics[width=\textwidth]{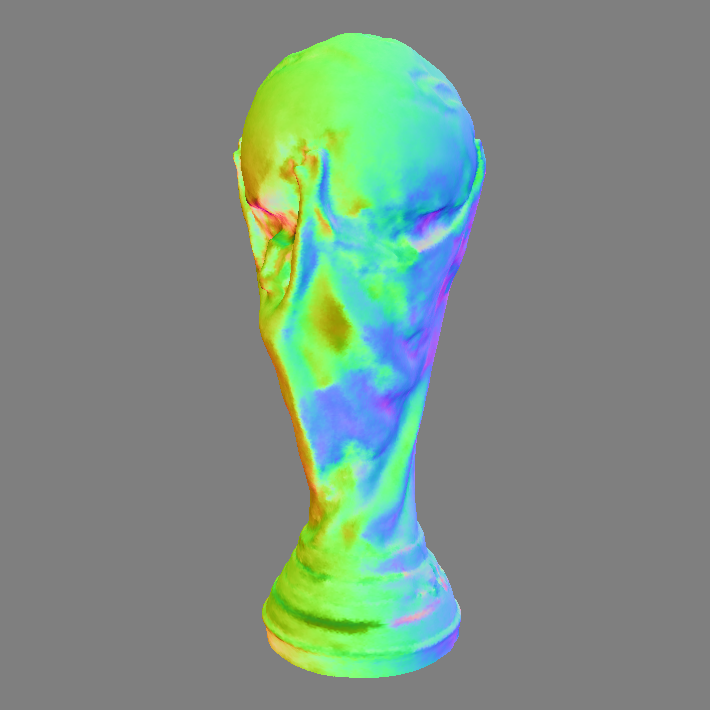}
            \end{minipage}
        \end{minipage}
    \end{minipage}

    \begin{minipage}{\linewidth}
        \begin{minipage}{0.08in}	
            \centering
            \rotatebox{90}{\small \textsc{Matball}}
        \end{minipage}
        \begin{minipage}{.983\linewidth}
            \centering
            \begin{minipage}{.24\textwidth}
                \includegraphics[width=\textwidth]{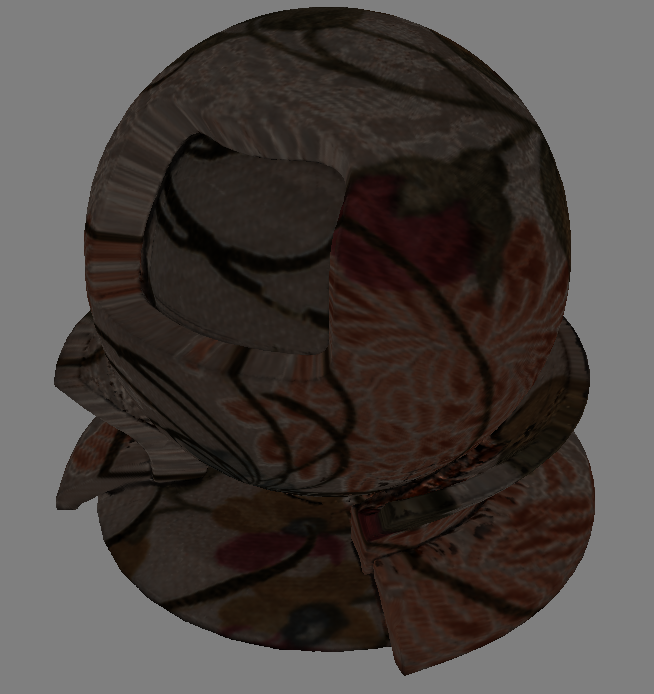 }
            \end{minipage}
            \begin{minipage}{.24\textwidth}
                \includegraphics[width=\textwidth]{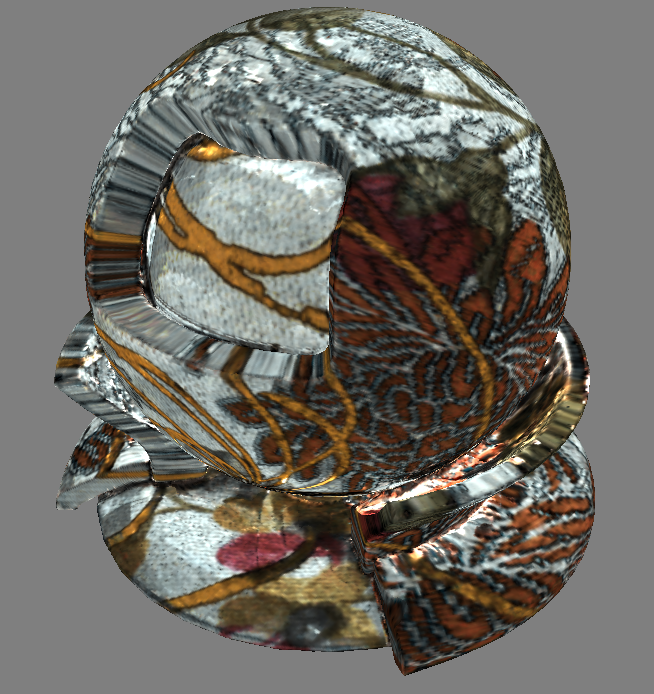}
            \end{minipage}
            \begin{minipage}{.24\textwidth}
                \includegraphics[width=\textwidth]{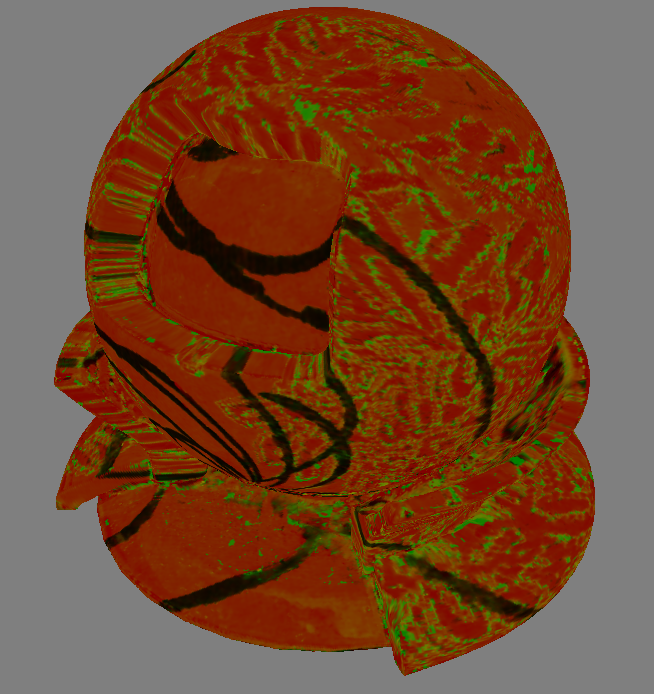}
            \end{minipage}
            \begin{minipage}{.24\textwidth}
                \includegraphics[width=\textwidth]{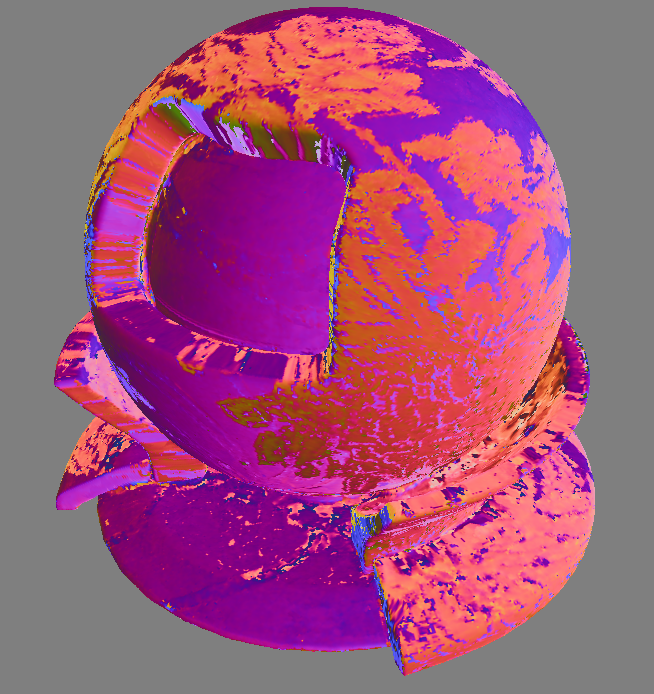}
            \end{minipage}
        \end{minipage}
    \end{minipage}

    \caption{Reconstructed SVBRDF parameters. For visualization purpose, each tangent is added with $\left(1, 1, 1\right)$ and then divided by 2 to fit to the range of $[0, 1]$; the specular albedo is re-scaled; and $\alpha_x$/$\alpha_y$ are visualized in the red/green channel.}
    \label{fig:vis_svbrdf_maps}
\end{figure}

To fit BRDF parameters for a particular texel, we first project its corresponding 3D position to all visible views to gather its image measurements. Next, we employ a 16D latent vector to represent the BRDF parameters: a decoder network is also trained to transform the latent vector to the parameters ($\rho_d,\rho_s,\alpha_x,\alpha_y,\mathbf{n_p}, \mathbf{t_p}$). These parameters will be used to produce rendering results, whose difference with the aforementioned image measurements is minimized. All latent vectors and the corresponding decoder are jointly optimized. Finally, we convert the latent vector at each texel to anisotropic GGX BRDF parameters, and store them in texture maps as the appearance result (as visualized in~\figref{fig:vis_svbrdf_maps}).

\subsection{Features Incorporating Correlated Factors}
According to~\tabref{tab:correlation}, diffuse albedos and normals are mostly correlated with our learned features. Here we test the impact of replacing part of our learned features with the predictions of these highly correlated factors. Specifically, we encourage our network to learn to explicitly predict the first 6D of the output feature as diffuse albedo and normal, with the following modified loss:
$$
L=\lambda_{0}L_0+\lambda_{1}L_1+\lambda_{2}L_2+\lambda_p L_p+\lambda_{\rm reg}L_{\rm reg},
$$
where
$$
L_{\rm reg}=L_{\rm diffuse} + L_{\rm normal}.
$$
We reserve the first 6 dimensions of the final feature for diffuse and normal predictions, and leave the remaining dimensions for data-learned features. Here $L_{\rm diffuse}$ represents the mean squared error (MSE) between the first three dimensions of the final feature and the ground-truth diffuse albedo, while $L_{\rm normal}$ is the MSE between the next three dimensions of the final feature and the ground-truth normal. We set $\lambda_{\rm reg} = 5$ in our experiment.

We test the new features on reconstructing the geometry of \textsc{Matball}. Its Chamfer distance increases from 5.13 (our features) to 5.28 (new features). We find that while it is faster to train the new features due to the extra regularization term, the reconstruction quality is reduced, as the features are not completely learned from data.

\end{document}